\documentclass{article}

\usepackage[final]{neurips_2023}

\usepackage[utf8]{inputenc} %
\usepackage[T1]{fontenc}    %
\usepackage{hyperref}       %
\usepackage{url}            %
\usepackage{booktabs}       %
\usepackage{amsfonts}       %
\usepackage{nicefrac}       %
\usepackage{microtype}      %
\usepackage{xcolor}         %
\usepackage{subfigure}
\usepackage[textsize=tiny]{todonotes}
\usepackage{amsmath}
\usepackage{amssymb}
\usepackage{mathtools}
\usepackage{amsthm}
\usepackage{bm}
\usepackage{float}
\hypersetup{
   colorlinks=true,
   linkcolor=[RGB]{30, 30, 180},
   citecolor=[RGB]{30, 30, 180},
   urlcolor=magenta,
   pdfborder=0 0 0,
   pdftitle={},
   pdfsubject={}, 
   pdfkeywords={},
   pdfauthor={},%
   pdfstartview=FitH
}
\usepackage{graphicx,subfigure}
\usepackage{caption}
\usepackage[labelfont=bf]{caption}
\usepackage[capitalize,noabbrev]{cleveref}
\usepackage{wrapfig}
\usepackage{adjustbox}
\usepackage{natbib}
\usepackage{multirow}
\usepackage{selectp}

\title{Learning to Modulate pre-trained Models in RL}

\author{
\hspace{-0.67cm}
  \textbf{Thomas Schmied}$^{1}$,
  \textbf{Markus Hofmarcher}$^{2}$,
  \textbf{Fabian Paischer}$^{1}$,\\
  \textbf{Razvan Pascanu}$^{3,4}$,
  \textbf{Sepp Hochreiter}$^{1}$ \\
\hspace{-0.67cm}
  $^{1}$~ELLIS Unit Linz and LIT AI Lab, Institute for Machine Learning,\\
\hspace{-0.67cm}
  $^{2}$~JKU LIT SAL eSPML Lab, Institute for Machine Learning,\\
\hspace{-0.67cm}
                  ~~~~Johannes Kepler University, Linz, Austria\\
\hspace{-0.67cm}
  $^{3}$~Google DeepMind, $^{4}$~UCL\\
\hspace{-0.67cm}
  \texttt{schmied@ml.jku.at}
}

\begin{document}

\maketitle

\begin{abstract}
Reinforcement Learning (RL) has been successful in various domains like robotics, game playing, and simulation.
While RL agents have shown impressive capabilities in their specific tasks, they insufficiently adapt to new tasks.
In supervised learning, this adaptation problem is addressed by large-scale pre-training followed by fine-tuning to new down-stream tasks.
Recently, pre-training on multiple tasks has been gaining traction in RL.
However, fine-tuning a pre-trained model often suffers from catastrophic forgetting.
That is, the performance on the pre-training tasks deteriorates when fine-tuning on new tasks.
To investigate the catastrophic forgetting phenomenon, we first jointly pre-train a model on datasets from two benchmark suites, namely Meta-World and DMControl.
Then, we evaluate and compare a variety of fine-tuning methods prevalent in natural language processing, both in terms of performance on new tasks, and how well performance on pre-training tasks is retained.
Our study shows that with most fine-tuning approaches, the performance on pre-training tasks deteriorates significantly.
Therefore, we propose a novel method, Learning-to-Modulate (L2M), that avoids the degradation of learned skills by modulating the information flow of the frozen pre-trained model via a learnable modulation pool. 
Our method achieves state-of-the-art performance on the Continual-World benchmark, while retaining performance on the pre-training tasks.
Finally, to aid future research in this area, we release a dataset encompassing $50$ Meta-World and $16$ DMControl tasks.
\end{abstract}

\section{Introduction}
Reinforcement Learning (RL) has been instrumental in training agents capable of achieving notable successes, both in simulation, and in the real-world \citep{Silver:16, Vinyals:19,Berner:19,Arjona:19,Bellemare:20,Degrave:2022}. 
However, such agents are usually highly specialized and incapable of performing well outside of a narrowly-defined task.
Furthermore, adapting a pre-trained agent to a new task by fine-tuning usually results in decreased performance on prior tasks.
This effect is well-known in the literature as \textit{catastrophic forgetting} \citep{McCloskey:1989}.

A common paradigm to learn multiple tasks concurrently is multi-task learning \citep{caruana:97}.
However, typically, not all tasks we want an agent to learn are available at training time.
In this case, new tasks must be learned in a sequential manner.
Learning a new task ideally exploits knowledge from previously learned tasks and does not adversely affect the performance on these prior tasks.
Recent works have demonstrated that models based on the Transformer architecture \citep{Vaswani:17} excel at learning multiple tasks concurrently from large offline datasets \citep{Lee:22,Reed:22,Jiang:22,Brohan:22,Gupta:22,Shridar:22}. 
We want to utilize this capability in RL, thereby obtaining agents that can quickly learn new skills and adapt to new tasks.
However, it is still unclear how an agent can learn new tasks and skills without negatively impacting its performance on the tasks it has already learned.
 \begin{figure}
        \centering
        \includegraphics[width=1\textwidth]{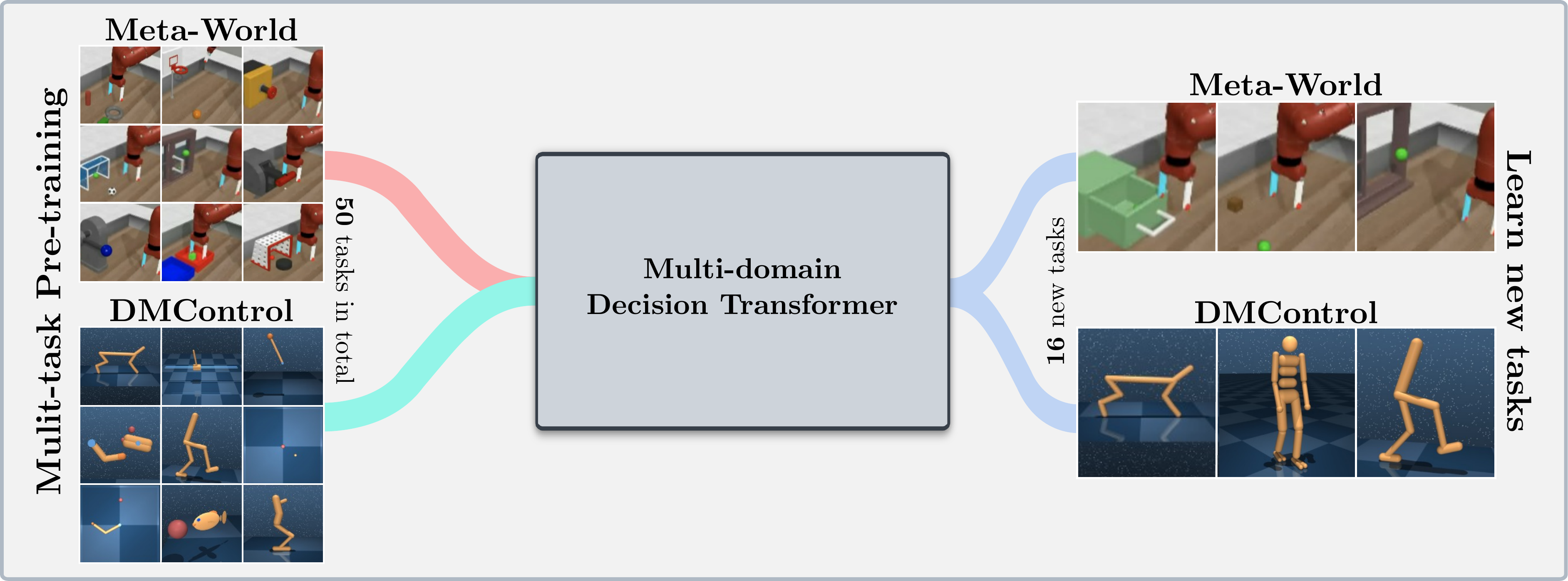}
        \caption{General setup of the experiments. We pre-train a Multi-Domain Decision Transformer (MDDT) model on a large set of tasks from multiple domains, namely Meta-World and DMControl. Then we fine-tune this model to learn additional tasks from these domains.}
        \label{fig:mddt}
\vspace{-0.5cm}
\end{figure}

Our aim is to find methods that enable an agent to efficiently learn new tasks without compromising its proficiency in previously acquired tasks. 
In this regard, we draw inspiration from fine-tuning (FT) techniques prevalent in supervised learning, such as parameter-efficient fine-tuning (PEFT,~\citep{Houlsby:19,Hu:21,Liu:22}) and prompt-based tuning (PBT,~\citep{Lester:21,Xiang:21}). 
Both PBT, and PEFT, incorporate a small set of new parameters to adapt a pre-trained model to new tasks at low cost. 
Thus, they intrinsically avoid catastrophic forgetting. 
However, it is unclear how well these methods can be adopted for training RL agents from offline datasets. 
Therefore, we first conduct a comprehensive evaluation for fine-tuning a pretrained Decision Transformer (DT, \citealp{Chen:21})
on two established RL benchmarks, namely Meta-World~\citep{Yu:22,Wolczyk:21} and DMControl~\citep{Tassa:18}. 
We first pre-train a DT jointly on tasks from both domains, then we transfer the pre-trained model to new tasks using various FT, PEFT and PBT methods (Figure \ref{fig:mddt}). 
We find that FT methods adjust well to new tasks, but performance on previous tasks generally decreases. 
PBT, in contrast, retains performance on previous tasks but does not adapt well to new tasks.

\begin{wrapfigure}{r}{0.51\textwidth}
    \centering
    \vspace{-0.6cm}
    \subfigure{\includegraphics[width=0.875\linewidth]{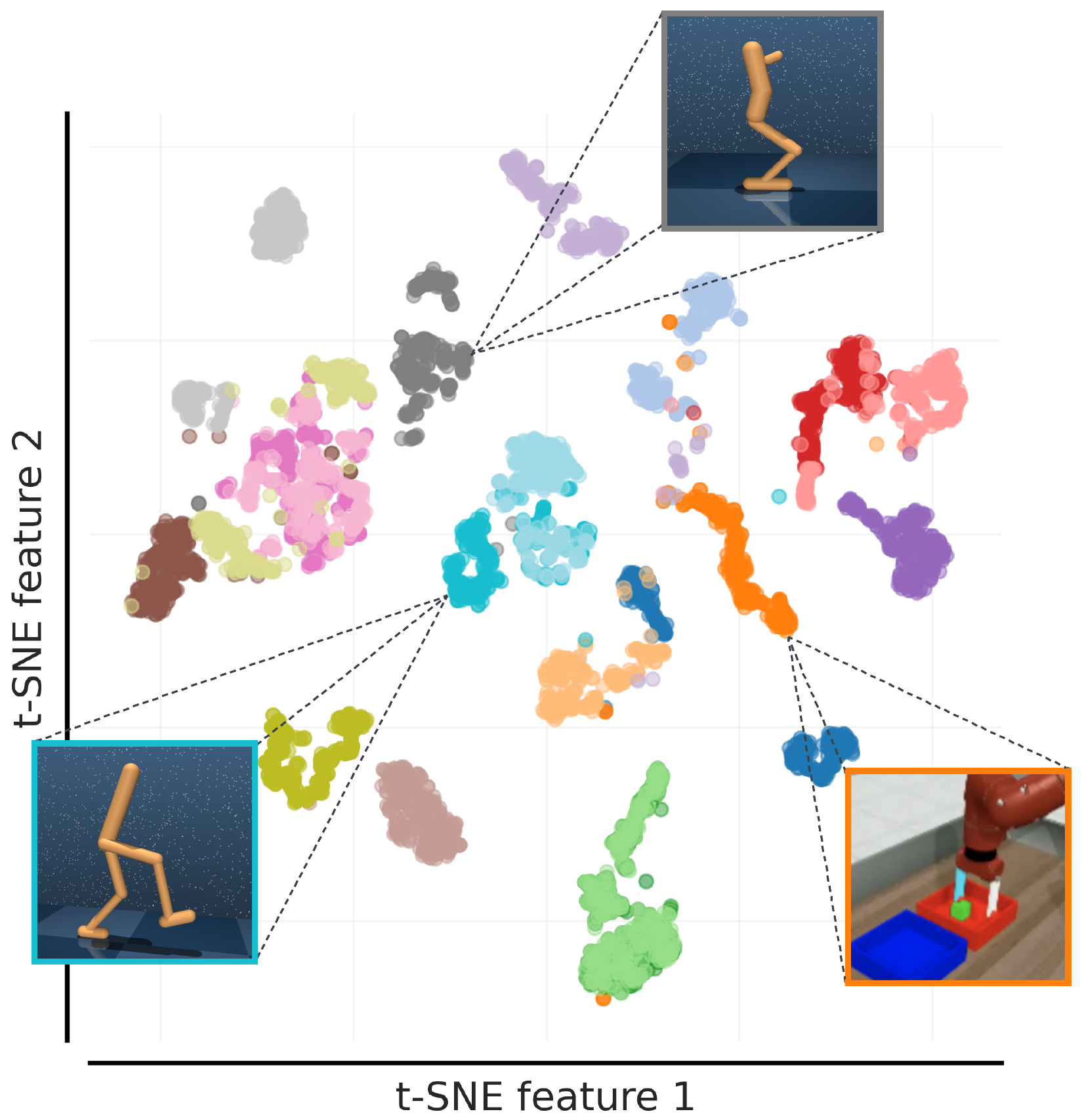}}
    \caption{t-SNE clustering of state embeddings for the first ten MT40 and DMControl tasks. Similar tasks are clustered together, while dissimilar tasks are apart (legend is available in Appendix \ref{appendix:task-sep}).}
    \label{fig:pretrain-tsne-l0-s}
\end{wrapfigure}

To combine the advantages of both approaches, we propose Learning-to-Modulate (L2M), 
which is based on two recent works from NLP and computer vision, 
namely Learning-to-prompt (L2P, \citealp{Wang:22}) and Low Rank Adaptation (LoRA, \citealp{Hu:21}). 
L2M operates in a task-agnostic manner and efficiently adapts a pre-trained model to new tasks via a learnable modulation pool.
The modulation pool consists of learnable keys that are associated with learnable modulation matrices.
Based on the current context, L2M selects a set of modulators which are learned during fine-tuning on the new task. 
This assumes that tasks can be well discriminated, such that suitable modulators are selected.
Indeed, we observe an emergent clustering of tasks in the embedding layer of our model after pre-training, as illustrated in Figure \ref{fig:pretrain-tsne-l0-s}. 
In turn, L2M is capable of quickly adapting to new tasks, while retaining performance on prior tasks, and introduces only minimal additional parameters per task relative to the full model size. 

To summarize, we make the following \textbf{contributions}: 
\begin{itemize}
\setlength\partopsep{0pt}
\setlength\itemsep{0pt}
\item We conduct an extensive evaluation of fine-tuning, parameter-efficient fine-tuning, and prompting methods for Transformers in RL.
\item We propose the novel L2M for efficient fine-tuning of a frozen pre-trained model by modulating the information flow via learnable modulation matrices.
\item We release a dataset of trajectories for the Meta-World and DMControl benchmark suites.
\end{itemize}

 \begin{figure}[t]
        \centering
        \includegraphics[width=1\textwidth]{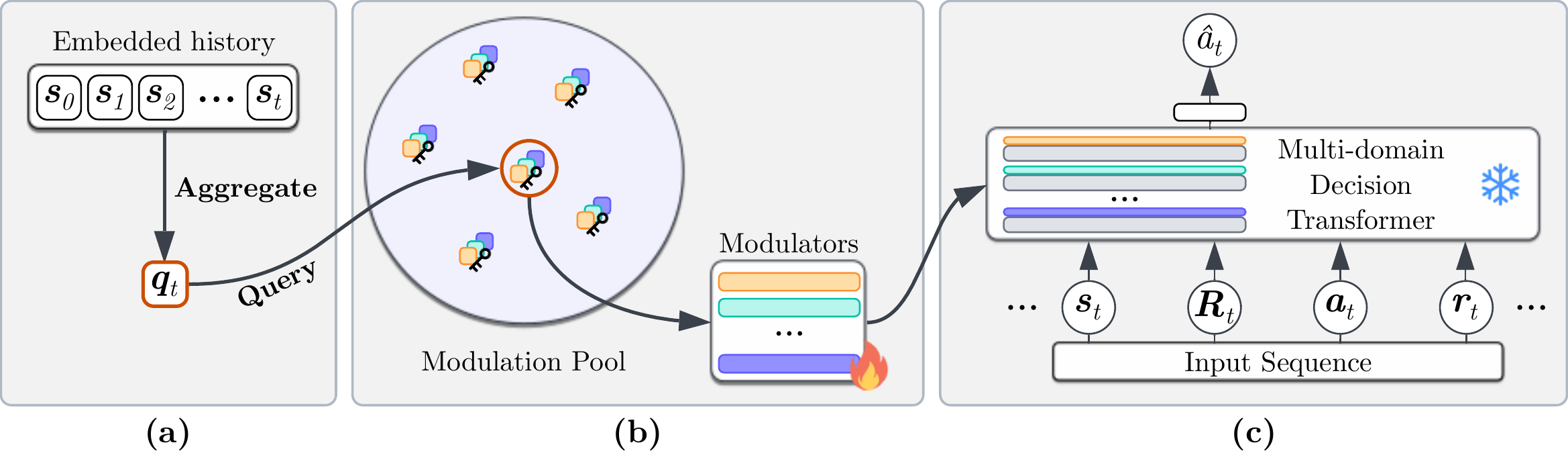}
        \caption{Illustration of L2M. \textbf{(a)} We construct a query $\mathbf{q}_t$ by aggregating the embedded history of state-tokens. \textbf{(b)} The query $\mathbf{q}_t$ is matched with learnable keys in a modulation pool, which map to learnable modulation matrices. We retrieve the modulation matrices with the highest similarity between $\mathbf{q}_t$ and every key in the modulation pool. \textbf{(c)} Retrieved modulation matrices modify the pre-trained and frozen multi-domain DT.}
        \label{fig:l2m-architecture}
\end{figure}

\section{Method}\label{sec:method}

We propose L2M, a method to adapt a pre-trained RL agent such that new tasks can be learned efficiently without compromising performance on pre-training tasks.
To this end, we combine approaches from prompt-based tuning and parameter-efficient fine-tuning.

\subsection{Background} 

\textbf{Reinforcement Learning.} We consider a Markov decision process (MDP) represented by the tuple  $(\mathcal{S},\mathcal{A},\mathcal{P},\mathcal{R})$. 
$\mathcal{S}$ and $\mathcal{A}$ denote state and action spaces, respectively, with states $s_t \in \mathcal{S}$ and actions $a_t \in \mathcal{A}$ at timestep $t$. 
The transition function $\mathcal{P}(s_{t+1}\mid s_t, a_t)$ takes a state-action pair and yields a probability distribution over next states. 
The reward function $\mathcal{R}(r_t \mid s_t, a_t)$ maps a given state-action pair to a scalar reward $r_t$.
The objective is to learn a policy $\pi(a_t\mid s_t)$ that predicts an action $a_t$ given $s_t$ that maximizes the reward $r_t$.

\textbf{Decision Transformer.} Decision Transformer (DT, \citealp{Chen:21}) rephrases RL as a sequence modelling problem.
This assumes access to a dataset $\mathcal{D}=\{\tau_i \mid 1 \leq i \leq N \}$, where $\tau_i=(s_0, a_0, r_0, \ldots, s_T, a_T, r_T)$ represents a trajectory of length $T$ consisting of state-action-reward triplets and $N$ defines the number of trajectories in $\mathcal{D}$. 
Further, we augment each trajectory $\tau_i$ with \emph{returns-to-go} $\hat R_t=\sum_{t'=t}^T r_{t'}$.
The policy $\pi_{\theta} (a_t \mid s_{t-C: t}, \hat R_{t-C:t}, a_{t-C:t-1}, r_{t-C:t-1})$, with parameters $\theta$ and context length $C$, is then trained in a supervised manner via upside-down-RL \citep{Schmidhuber:19}, minimising the cross-entropy $\mathcal{L}_{CE}(\hat{a}_t, a_t)$ between predicted action $\hat{a}_t$ and ground-truth action $a_t$.  
During inference, the policy produces a sequence of actions conditioned on a high return-to-go.

\textbf{Low Rank Adaptation (LoRA).} 
LoRA \citep{Hu:21} is a parameter efficient method for fine-tuning pre-trained models.
It performs a low-rank decomposition of a weight matrix that is utilized to modulate the information flow in the pre-trained model.
Particularly, we assume a pretrained layer $l$ with weight matrix $\bm{W}_l \in \mathbb{R}^{d_{\text{in}}\times d_{\text{out}}}$ that receives $\bm{x}_{l-1} \in \mathbb{R} ^ {1\times d_{\text{in}}}$ as input.
LoRA adds learnable low rank modulation matrices $\bm{A}_l \in \mathbb{R}^{d_{\text{in}}\times r}$ and $\bm{B}_l \in \mathbb{R}^{r \times d_{\text{out}}}$, where $r \ll d_{\text{in}}$.
The output $\bm{h}_l$ of layer $l$ is then computed as:
\begin{equation}\label{eq:lora}
    \bm{h}_l = \bm{W}_l\bm{x}_{l-1} +  \bm{B}_l\bm{A}_l\bm{x}_{l-1}.
\end{equation}
During the fine-tuning stage, only the parameters of $\bm{A}_l$ and $\bm{B}_l$ are updated.

\subsection{Learning-to-Modulate (L2M)}
\label{sec:l2m}
We propose L2M, a method that combines the advantages of PEFT and PBT approaches.
In particular, we rely on LoRA (\citealp{Hu:21}), and refer to the trainable parameters it induces  as modulators of the pre-trained model. 
Further, we draw inspiration from Learning-to-prompt (L2P, \citealp{Wang:22}), and maintain a pool of modulators associated with learnable keys. 
For a given input sequence, we retrieve the best modulator weights from the pool, and leverage them to alter the behaviour of the pretrained model (Figure \ref{fig:l2m-architecture}).

We define a \emph{modulation pool} that contains a set of $M$ learnable keys, $\mathbf{K}_{\text{pool}} = \{\mathbf{k}_i \mid 1 \leq i \leq M \}$.
All keys in $\mathbf{K}_{\text{pool}}$ are initialized uniformly from $[-1, 1]$. 
Each $\mathbf{k}_i$ is associated with a set of modulation matrices $\{ \mathbf{A}_{b}^i, \mathbf{B}_{b}^i \mid 1 \leq b \leq B\}$, for each layer block $b$ of a DT with $B$ layer blocks. As in \citet{Hu:21}, we initialize $\mathbf{A}$ according to a normal distribution around zero and $\mathbf{B}$ with zeros.
Thus, no modulation can occur at the beginning of training.

At time $t$, we generate a query vector $\mathbf{q}_t \in \mathbb{R}^{d}$, where $d$ is the embedding dimension of the pre-trained model, in the following way.
First, we construct a matrix $\mathbf{S}_{t-C:t} \in \mathbb{R}^{C\times d}$ from the states of a trajectory $\tau$ with the context length $C$, after they are processed via the embedding layer of the frozen pre-trained model.
Then, we reduce the matrix to a query vector $\mathbf{q}_t$ by an aggregation function $g(\cdot)$:
{
\setlength{\abovedisplayskip}{3pt}
\setlength{\belowdisplayskip}{3pt}
\begin{equation}
    \mathbf{q}_t = g(\mathbf{S}_{t-C:t})
\end{equation}
}

where we use mean-pooling for $g(\cdot)$.
We retrieve a set of modulation matrices $\{ \mathbf{A}_{b}^j, \mathbf{B}_{b}^j \mid 1 \leq b \leq B\}$ by the maximum similarity between each $\mathbf{k}_i \in \mathbf{K}_{\text{pool}}$ and the query $\mathbf{q}_t$ at timestep $t$:
{
\setlength{\abovedisplayskip}{4pt}
\setlength{\belowdisplayskip}{3pt}
\begin{gather}
    j \ = \ \operatorname*{argmax}_{\mathbf{k}_i \in \mathbf{K}_\text{pool}} \ \operatorname{sim} (\mathbf{q}_t, \mathbf{k}_i) \ n(\mathbf{k}_i)^{-1}
\end{gather}
}

In our case, $\operatorname{sim}(\cdot, \cdot)$ corresponds to the cosine similarity and $n(\mathbf{k}_i)$ represents the selection count for key $\mathbf{k}_i$. 
Every time a key $\mathbf{k}_i$ is selected, we increase $n(\mathbf{k}_i)$ by one.
This way, we discourage that different queries always attend to the same key. 

We use $\{ \mathbf{A}_{b}^j, \mathbf{B}_{b}^j \mid 1 \leq b \leq B\}$ to modulate the pre-trained model according to \cref{eq:lora}. 
More specifically, we apply LoRA on the queries and values in the self-attention mechanism, as well as on the activation in the feed-forward block of the pre-trained model. 
Only the modulators are learned via gradient descent, while the pre-trained model remains frozen.
Following \citet{Wang:22}, the key is updated by maximizing the cosine similarity between $\mathbf{q}_t$ and $\mathbf{k}_j$ via an additional term in the end-to-end objective:
{
\setlength{\abovedisplayskip}{4pt}
\setlength{\belowdisplayskip}{3pt}
\begin{gather}
    \min_{\theta, \mathbf{k}_i} \mathcal{L}_{CE}(\hat{a}_t, a_t) - \lambda \;\operatorname{sim} (\operatorname{stopgrad}(\mathbf{q}_t), \mathbf{k}_j)
\end{gather}
}

where $\lambda$ is a hyperparameter.

L2M unifies the benefits of LoRA and L2P, ensuring high-performance and few additional learnable parameters, while also avoiding forgetting on the pre-trained tasks.

\subsection{Multi-Domain Decision Transformer (MDDT)}
We extend the Decision Transformer architecture proposed by \citet{Chen:21} to handle inputs from multiple domains with varying state/action spaces.
Since the dimensionality of states differ between Meta-World and DMControl we construct a unified state space, that comprises all dimension of DMControl and Meta-World environments. 
This state-space consists of 204 dimensions in total.
Dimensions that are unused by a certain task are padded with zeros.
Finally, the states are embedded with a linear layer before serving as input to the MDDT.
Similar to \citet{Reed:22} and \citet{Brohan:22}, we tokenise each action dimension and autoregressively predict action tokens. 
In the environments we consider, all actions are bounded by [-1, 1]. 
We use a min-max tokenisation, which comprises min-max normalisation and subsequent uniform discretisation into 64 bins. 
These changes enable processing observations and actions originating from different environments at training and test time. 
In line with prior work \citep{Chen:21, Lee:22}, we train the DT via return-conditioned upside-down RL using a cross-entropy loss to predict next actions autoregressively.
We set the target return to the maximum observed return in the respective dataset, but also found that constant proxies per domain work well.
We provide further implementation details on our MDDT as well as a discussion on limitations and implications of our design choices in Appendix \ref{appendix:mddt}.

\section{Experiments}\label{sec:experiments}
We consider two different benchmark suites that comprise 66 different tasks in total, namely Meta-World~\citep{Yu:22,Wolczyk:21} and DMControl~\citep{Tassa:18}.
Meta-World consists of 50 diverse tasks for robotic manipulation, such as grasping, manipulating objects, opening/closing a window, pushing buttons, locking/unlocking a door, and throwing a basketball. 
We follow \citet{Wolczyk:21} and split them into $40$ pre-training (MT40), and $10$ fine-tuning tasks (CW10). 
We use Meta-World v2, instead of v1 used by \citet{Wolczyk:21}, but will refer to it as Meta-World throughout this work.
Further, we select $16$ tasks from DMControl and assign $10$ tasks to the pre-training set (DMC10) and $6$ tasks to the fine-tuning set (DMC6).
We elaborate on the different environments and on the data collection procedure in Appendix \ref{appendix:environments}, and Appendix \ref{appendix:data}, respectively.
Unlike Meta-World, which shares the same state and action spaces across tasks ($|\mathcal{S}|=39, |\mathcal{A}|=6 $), the state and action spaces of DMControl vary for each task %
(e.g., for \texttt{hopper} $|\mathcal{S}|=15, |\mathcal{A}|=4$, for \texttt{cheetah} $|\mathcal{S}|=17, |\mathcal{A}|=6$), as listed in Appendix \ref{appendix:data}.

We collect a dataset\footnote{Source code and datasets are available at: \url{https://github.com/ml-jku/L2M}} by training task-specific agents on each task using an agent based on Soft Actor Critic (SAC)  \citep{Haarnoja:18}.
The average performance for SAC across the Meta-World and DMControl task splits is shown in Table \ref{tab:mt40-cw10}. 
For each of the $50$ tasks in Meta-World, we include $10$K trajectories of length $200$ in the dataset, amounting to 100M transitions in total. 
For DMControl, we include $1000$ trajectories of length $1000$ for each task, amounting to $16$M transitions in total.
The datasets contain the entire replay buffer of the task-specific agents, and consequently behaviours ranging from random to expert.
We choose this collection scheme since prior work has illustrated the benefits of training agents on data that comprises mixed behaviour \citep{Lee:22}.
We give further details on the training procedure, including hyperparameters and learning curves in Appendix \ref{appendix:data}.
\begin{table}[h]
\centering
    \caption{Performance measures of task specific SAC-based agents used for data collection averaged over the different task splits. Mean and standard deviation are shown.}
    \subtable[Meta-World]{
        \centering
        \begin{tabular}{l c r}
            \toprule
             \textbf{Dataset} & \textbf{Success Rate} &      \textbf{Mean Reward} \\
            \midrule
             MT40 &  0.84 ± 0.34 & 1414.62 ± 439.39 \\
             CW10 &    1.0 ± 0.0 & 1540.49 ± 184.43 \\
            \bottomrule
        \end{tabular}
    }
    \hspace{1cm}
    \subtable[DMControl]{
        \centering
        \begin{tabular}{l r}
            \toprule
            \textbf{Dataset} &  \textbf{Mean Reward} \\
            \midrule
            DMC10 & 788.36 ± 219.11 \\
            DMC6 & 840.37 ± 216.63 \\
            \bottomrule
        \end{tabular}
     }
    \label{tab:mt40-cw10}
\end{table}

Figure \ref{fig:mddt} illustrates our experimental design.
First, we pre-train a MDDT on all pre-training datasets (MT40 and DMC10) simultaneously (see Section \ref{sec:exp-pretrain}). 
Next, we conduct a broad evaluation of different FT, PEFT and PBT methods to adapt the pre-trained model to each of the held-out fine-tuning tasks (CW10 and DMC6).
In Section~\ref{sec:exp-adaptation}, we show the results for fine-tuning on all CW10 and DMC6 tasks individually.
Finally, we present results for training all methods in a continual RL setup, in which tasks are introduced sequentially (Section \ref{sec:exp-cl}). 
All our models are trained offline on the generated datasets and evaluated online in the respective environments.

Unless mentioned otherwise, we report IQM and 95\% bootstrapped confidence intervals, as proposed by \citet{Agarwal:21} ($3$ seeds per method and 50K bootstrap samples). 
We report success rates and data-normalized scores (normalized by mean score in dataset and random agent performance as suggested by \citet{Fu:20}) for Meta-World and DMControl, respectively.

\subsection{Pre-training} \label{sec:exp-pretrain}
We pre-train a MDDT on our datasets collected for MT40 and DMC10. 
This already results in strong performance on the pre-training tasks (60\% for MT40, 94\% for DMC10). 
We experiment with different model sizes (40M to 200M parameters), varying number of Transformer layers, number of heads per layer and embedding dimension. 
Generally, there is a trend that more complex models perform better. 
For all our subsequent experiments, we use a 40M parameter model, as it achieves a good trade-off between performance and training time. 
Performance scores, learning curves, implementation details and hyperparameters for our pre-training experiments are provided in Appendix \ref{appendix:pre-training}.

After pre-training, we analyse the learned representations of the model via t-SNE \citep{Van:08}. 
Figure \ref{fig:pretrain-tsne-l0-s} illustrates emerging clusters of learned state embeddings for ten tasks of MT40 and DMC10. 
We observe that similar tasks cluster together (e.g., \textit{button-press-v2} and \textit{button-press-wall-v2}), whereas dissimilar tasks are well separated (e.g., \textit{assembly-v2} and \textit{bin-picking-v2}). 
We repeat this analysis for rewards, RTG and action token embeddings and explain the clustering procedure in Appendix \ref{appendix:task-sep}. 

\begin{figure}[h]
     \centering
     \subfigure[CW10]{\includegraphics[width=0.45\textwidth]{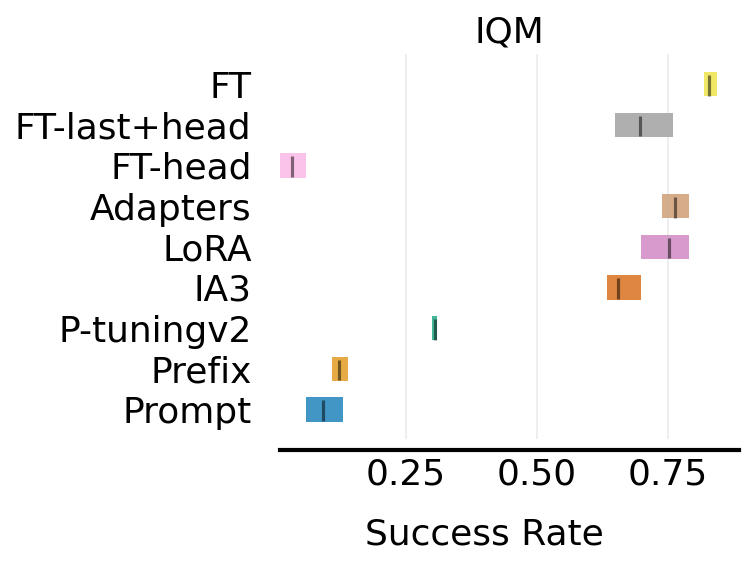}}
    \hspace{0.3cm}
     ~
     \subfigure[DMC6]{\includegraphics[width=0.46\textwidth]{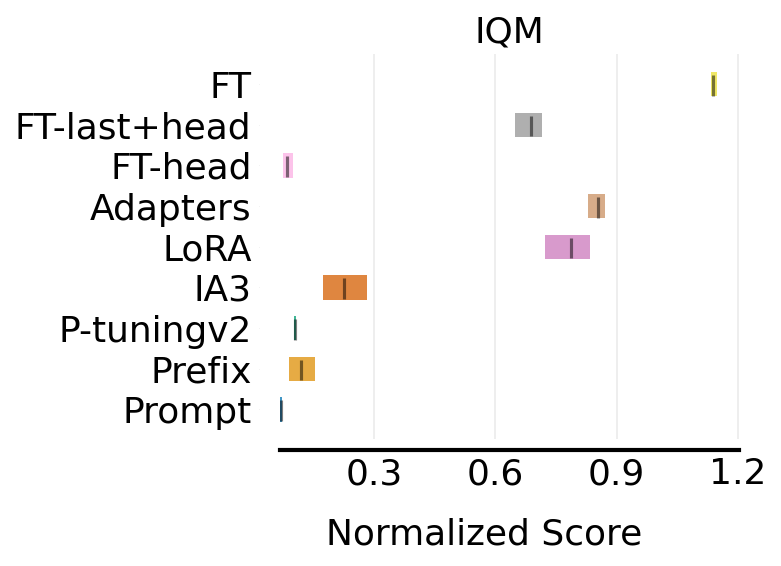}}
    \caption{IQM and 95\% CIs for single-task fine-tuning on \textbf{(a)} CW10 and \textbf{(b)} DMC6. Models are pre-trained on MT40/DMC6 and are then optimized for each CW10/DMC6 task individually.}
    \label{fig:off-main1}
\end{figure}

\subsection{Single-Task Fine-Tuning}
\label{sec:exp-adaptation}
We evaluate the performance of various FT, PEFT and PBT strategies for fine-tuning on the held-out tasks.
Overall, we compare a total of 9 methods: Full fine-tuning (FT), FT of action head (FT-head), FT of last Transformer layer and action head (FT-last+head), Adapters \citep{Houlsby:19}, LoRA \citep{Hu:21}, (IA$\text{)}^3$ \citep{Liu:22}, Prompt-tuning \citep{Lester:21}, Prefix-tuning \citep{Xiang:21}, and P-tuning v2 \citep{Liu:21a}. 
We do not compare against meta-RL algorithms, as \citet{Mandi:22} showed that FT approaches perform on-par or better on several tasks.
A detailed list of hyperparameters, training details and description for each method are provided in Appendix \ref{appendix:adaptation}. 
We fine-tune the pre-trained model for each task individually and aggregate the performance over all tasks within each domain.

\begin{wrapfigure}{r}{0.5\textwidth}
     \centering
     \includegraphics[width=0.90\linewidth]{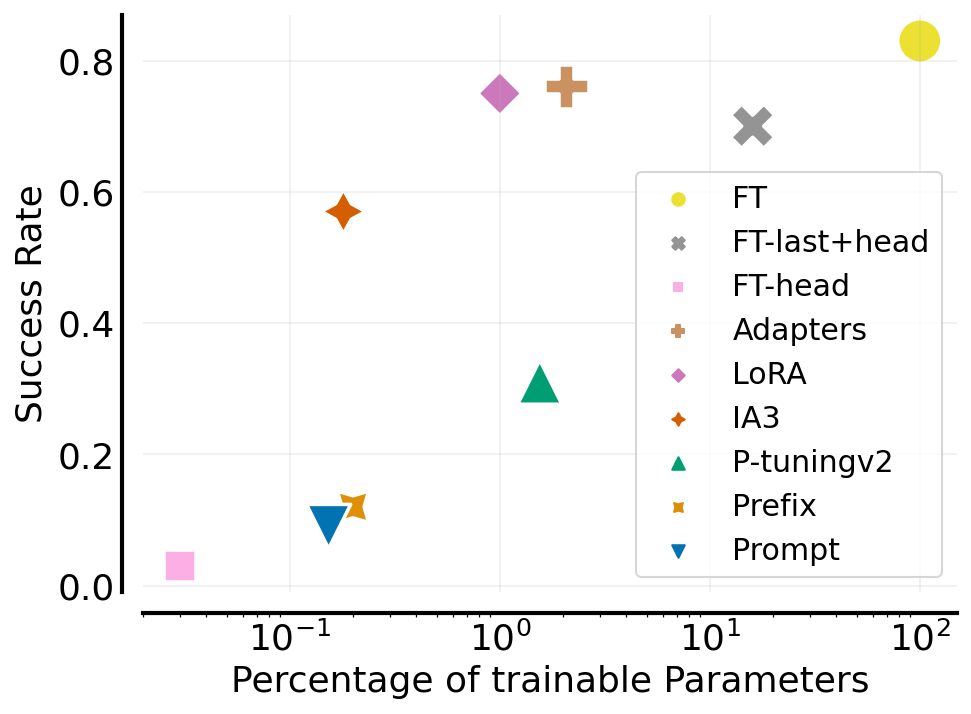}
    \caption{Success rate vs. fraction of parameters trained for various fine-tuning techniques on single-task experiments for CW10.}
    \label{fig:off-main2}
\end{wrapfigure}

In Figure \ref{fig:off-main1}, we show IQM over all tasks for all methods on CW10 and DMC6.
Overall, FT attains the highest scores, since it can utilize the entire model capacity.
Thus, it serves as an upper bound in terms of performance. 
Adapters achieve the second-highest scores on average, followed by LoRA, (IA$\text{)}^3$ and FT-last+head. 
Notably, there is a large gap between PBT and PEFT methods, particularly on CW10.
Fine-tuning the action head for each task separately, results in the worst performance.
While the absolute scores differ between the two domains, the relative performance ranking among methods remains equivalent. 
On DMC6, we observe a larger performance gap between full FT and PEFT methods. 
While in Meta-World only the locomotion differs between tasks, in DMControl the entire robot morphology may change. 
Therefore, more capacity of the model is required to sufficiently adapt to the new tasks.

We highlight the efficacy of the considered methods by comparing the fraction of trainable parameters against attained performance on CW10 in Figure \ref{fig:off-main2} (see Appendix \ref{appendix:adaptation} for same comparison on DMC6).
FT updates all parameters, while Adapters update two orders of magnitude less parameters (2.1\%). 
In contrast, LoRA requires less than half the amount of parameters of Adapters (1\%), while reaching a similar level of performance. 
Notably, (IA$\text{)}^3$, trains approximately the same amount of parameters as Prompt-tuning (0.21\%), but attains significantly higher performance.
This result indicates that we can achieve decent performance with a small amount of trainable parameters if the pre-training and fine-tuning tasks are similar.
Overall, LoRA compares favourably in both dimensions.

\subsection{Continual Fine-Tuning}\label{sec:exp-cl}

Ultimately, our goal is to adapt the pre-trained model to multiple novel tasks, while alleviating forgetting of tasks learned during pre-training.
In this regard, we adapt the pre-trained model to all fine-tuning tasks in a sequential manner, as proposed by \citet{Wolczyk:21} for CW10.
Moreover, we evaluate forgetting by measuring the performance on tasks acquired during pre-training after fine-tuning. 

\begin{figure}[h]
     \centering
     \subfigure[CW10]{\includegraphics[width=0.44\textwidth]{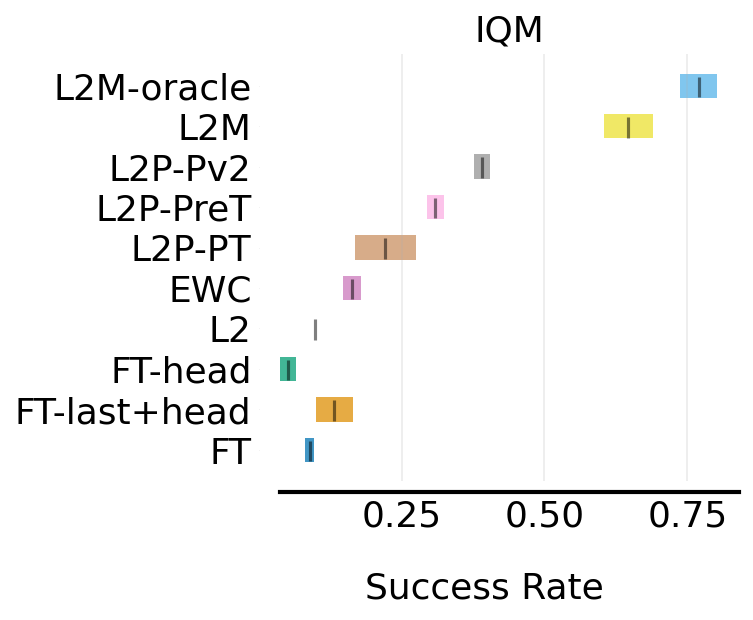}\label{fig:cl-main-ie-1}}
    \hspace{0.3cm}
     ~
     \subfigure[DMC6]{\includegraphics[width=0.44\textwidth]{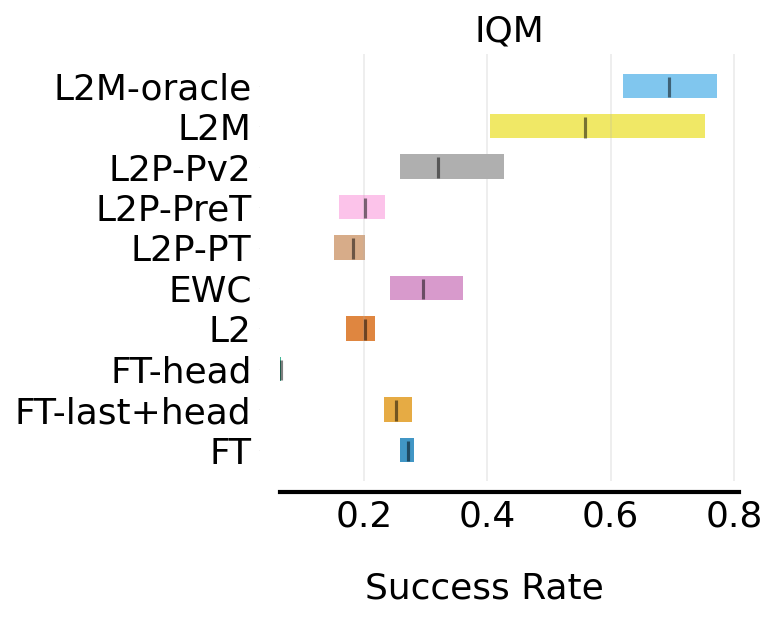}\label{fig:cl-main-ie-2}}
    \caption{IQM and 95\% CIs of success rates for CRL experiments on \textbf{(a)} CW10 and \textbf{(b)} DMC6. Models are pre-trained on MT40+DMC10 and are then trained on the tasks from CW10/DMC6 sequentially. On each task, we train for 100K steps and then move to the next task in the sequence.}
    \label{fig:cl-main-ie}
\end{figure}

Again, we compare the following fine-tuning variations: FT, FT-head, and FT-last+head.
Additionally, we augment the PBT methods of the previous section with a prompt pool as in L2P, which enables learning of task-specific prompts. 
We refer to these methods as L2P + Prompt-tuning (L2P-PT), L2P + Prefix-tuning (L2P-PreT), L2P + P-tuning v2 (L2P-Pv2).
We note that L2P-PT reflects the original version of L2P proposed by \citet{Wang:22} which was originally applied in computer vision. 
Additionally, we compare L2M to two established methods from Continual RL (CRL), namely Elastic Weight Consolidation (EWC), and L2, both propposed by \citet{Kirkpatrick:17}. 
EWC constrains task-specific weights to not change too much when adapting to a new task by estimating their individual importances. 
L2 applies a uniform constraint to all weights without considering their relevance.
Moreover, we add another implementation of L2M, which is equipped with an oracle that provides information on what task is currently being observed in terms of a task index (L2M-Oracle).
For L2M-oracle, the modulation pool contains as many modulators as there are tasks. 
At training time, the task index refers to the dataset the batches are sampled from. At inference time, the task index refers to the environment the DT is currently evaluated in.
In contrast to \citet{Wolczyk:21}, we do not use separate action heads per task for L2M to remain task agnostic.
To illustrate an upper bound on performance, we provide scores for two multi-task baselines, which train on all tasks simultaneously, either from scratch, or after pre-training (Table \ref{tab:crl_full} in Appendix \ref{appendix:cl}).
We train each method for 100K steps per task and evaluate on all tasks every 50K steps. 
We provide further training details and hyperparameters in Appendix \ref{appendix:cl}.

In Figures \ref{fig:cl-main-ie-1} and \ref{fig:cl-main-ie-2}, we show the performance scores of all methods on CW10 and DMC6, respectively. 
In addition, we report forgetting scores (as calculated by \citet{Wolczyk:21}), and rewards obtained at the end of training, in Appendix \ref{appendix:cl}.
L2M outperforms all other approaches in both domains, attaining an average success rate of 65\% across CW10 tasks and a normalized score of 56\% across DMC6 tasks. 
Adding a task oracle to L2M increases the success rate to 76\% and 70\% on CW10 and DMC6, respectively, and closely matches the single-task performance of LoRA.
L2P combined with different prompting approaches performs considerably worse than L2M.
Interestingly, the established CRL method EWC does not mitigate forgetting sufficiently. 
Similar results for EWC were reported by \citet{Ben:22} in an online RL setting. 
To the best of our knowledge, the results of our method, L2M, are the highest reported results on Continual-World v2 to date. 
Prior work reports average success rates of roughly 40\% \citep{Caccia:22,Ben:22}. 
Even though L2M outperforms other approaches, there is a considerable gap between L2M and L2M-oracle. 
We surmise this is due to conflation effects in the prompt pool and aim to investigate this in more detail in future work.
\begin{figure}
     \centering
     \includegraphics[width=0.42\linewidth]{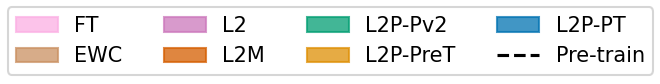}
     \\
    \begin{minipage}[b]{0.44\linewidth}
        \centering
        \includegraphics[width=1\linewidth]{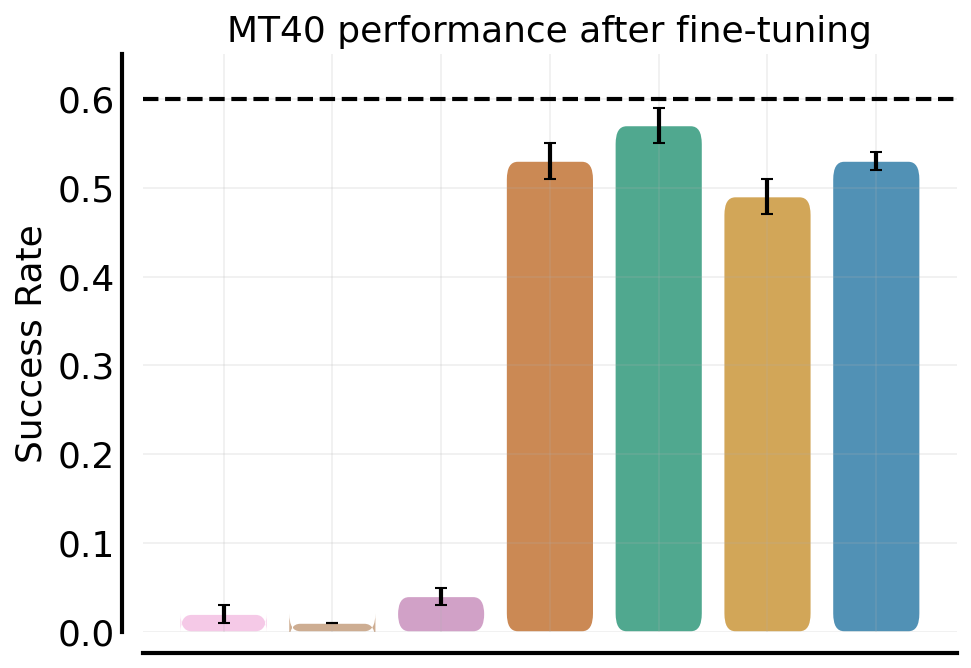}
    \end{minipage}
    \hspace{0.25cm}
     ~
    \begin{minipage}[b]{0.44\linewidth}
        \centering
        \includegraphics[width=1\linewidth]{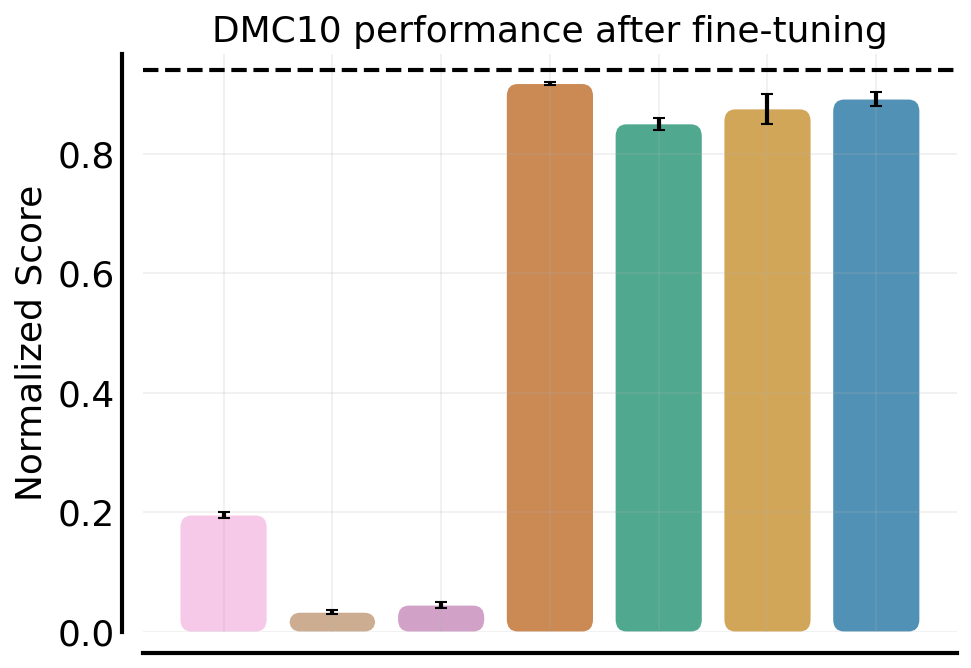}
    \end{minipage}
    \caption{Performance on the MT40 and DMC10 pre-training tasks after fine-tuning.}
    \label{fig:cl-pretrain-perf}
\end{figure}

\textbf{Performance on pre-training tasks.} Finally, we evaluate performance on the pre-training tasks after fine-tuning, as shown in Figure \ref{fig:cl-pretrain-perf}. 
To remain task-agnostic, we train a set of 100 keys for the pre-training tasks, which we concatenate to the set of keys introduced during the fine-tuning stage.
While FT, L2 and EWC experience a severe drop in performance, L2M and L2P-based approaches maintain a similar performance level as prior to fine-tuning. 
Thus, L2M preserves performance on the pre-training tasks, while it effectively adapts to new tasks.

\subsection{Ablation Studies}\label{sec:exp-ablation}
To gain a better understanding of our method and its limitations, we conduct a number of additional ablations. 

\begin{wrapfigure}{r}{0.5\textwidth}
    \vspace{-0.2cm}
    \centering
    \includegraphics[width=0.5\textwidth]{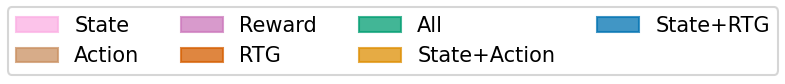}\\
    \includegraphics[width=0.45\textwidth]{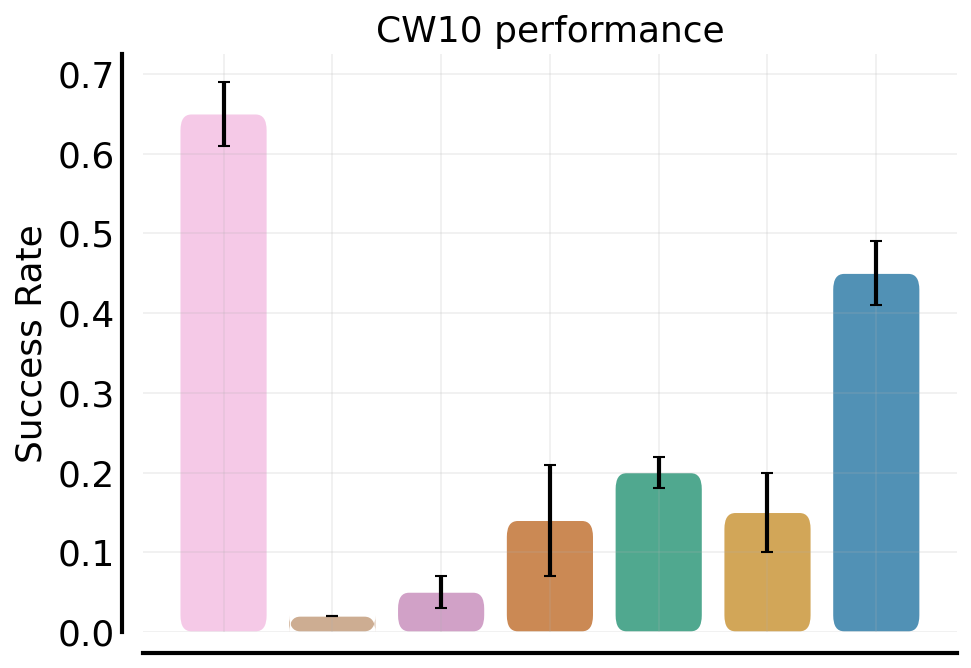}
    \caption{Aggregation token ablation on CW10.}
    \label{fig:query-rep-token}
\end{wrapfigure}

\textbf{Modulation Targets.} 
One important design choice in L2M is which weights of the pre-trained model to modulate. In principle, LoRA can be applied to any weight matrix in the pre-trained model. 
By default, we employ LoRA on the queries and values in the self-attention mechanism, and on the position-wise feedforward layer. 
In Appendix \ref{appendix:lora-rank}, we conduct an ablation study in which we vary the modulation targets. 
Our analysis suggests that modulating the attention mechanism is not as important as modulating the position-wise feed-forward layer.
However, modulating both, the attention mechanism, and the position-wise feed-forward layer yields the best results. 
Nevertheless, this performance gain comes at the cost of more trainable parameters.

\textbf{Query Representation.} 
As specified in Section \ref{sec:l2m}, we use embedded state tokens aggregated over a specified context window as input query to the prompt pool in L2M.
This design choice is inspired by the observed task separation in the embedding layer after pre-training (see Figure \ref{fig:pretrain-tsne-l0-s}). 
We investigate other choices of individual tokens and token combinations (state, action, reward, RTG, state-action, state-RTG) to represent the query (see Figure \ref{fig:query-rep-token}). 
Indeed, we observe that using the state embeddings results in the best overall performance, as it aids task separation. 
In contrast, using RTG or action embeddings deteriorates performance.
Furthermore, we also investigate the effect of using different layer representations and context windows to construct the query in Appendix \ref{appendix:queryrep}.

\textbf{Alternative Modulators.} %
We conduct an ablation study in which we compare L2M against L2M in combination with (IA$\text{)}^3$ in Appendix \ref{appendix:altmod}.
Instead of the low-rank modulation matrices, (IA$\text{)}^3$ employs elementwise multiplication with learnable modulation vectors. 
While performance decreases with L2M-(IA$\text{)}^3$, it compares favourably in terms of parameter-efficiency. Depending on the task, this may be preferred. 

\textbf{Single-domain DT on Meta-World.}
To better understand the potential drawbacks of using a multi-domain model, we pre-train and fine-tune a single-domain model only on Meta-World (see Appendix \ref{sec:appendix-singledomain}). 
Due to the common state and action spaces in Meta-World, we use a simpler, non-discretized, action space and training objective (MSE) for this experiment.
This specialised single-domain model obtains considerably higher performance scores. 
However, these performance gains come at the cost of the loss of generality, as the specialised model can only handle the particular state/action space it was trained on. 
Thus, fine-tuning it to tasks with new state/action spaces is not possible. 
Nevertheless, this experiment highlights the drawbacks of our current mechanisms, such as action discretisation and autoregressive action-prediction, that enable multi-domain training. 

\textbf{Cross-domain FT}. 
Finally, to better understand the potential upsides of using a multi-domain model, we pre-train a DT (with unified state space and action discretisation) on Meta-World only (MT40) and then fine-tune it on DMControl (DMC6).
We observe that the fine-tuning performance on DMC6 (different domain) is considerably worse than for the MDDT (see Appendix \ref{sec:appendix-crossdomain}). 
In addition, we also fine-tune the pre-trained single-domain model on CW10 (same domain). 
Interestingly, the final performance on CW10 is also lower compared to the MDDT that was pre-trained on both domains.
This experiment indicates, that multi-domain pre-training can, indeed, have a positive effect on the fine-tuning performance.

We provide additional details on our ablation studies in Appendix \ref{appendix:adaptation} and \ref{appendix:cl}.

\section{Related Work}\label{sec:relatedwork}

\textbf{Transformers in RL.} Since the inception of Transformers \citep{Vaswani:17} there has been a widespread adoption of the underlying architecture in various areas such as NLP \citep{Devlin:18,Radford:19,Brown:20}, computer vision \citep{Dosovitskiy:21,He:22,Radford:21,Fürst:21,Ramesh:21,Rombach:22}, speech recognition \citep{Radford:22,Baevski:20} or video generation \citep{Ho:22,Singer:22}. 
More recently, Transformers have found their way into RL. 
Similar to the DT \citet{Chen:21}, Trajectory Transformer \citep{Janner:21} is based on the GPT architecture \citep{Radford:18}, but relies on dynamics modelling. \citet{Lee:22} extended DT to a multi-game setup to learn to play 46 Atari games.
Meanwhile, a variety of DT-variants have been proposed \citep{Zheng:22,Wang_bootstrapped:22,Shang:22,Meng:21}. \citet{Siebenborn:22} replace the Transformer in DT with an LSTM \citep{Hochreiter:97}. 
PromptDT \citep{Xu:22} demonstrated that prompting a pre-trained DT model with expert trajectories can improve the agent's ability to generalize to new tasks. 
\citet{Jiang:22} presented a prompt-based Transformer for robot manipulation, that integrates multi-modal prompts via cross-attention. \citet{XuMengdi:22} propose to augment a DT with a hyper-network for parameter-efficient policy adaptation.
A number of other works (which are largely orthogonal to our approach) instead aim to improve the pre-training stage, for example by predicting masked-out parts of the sequence \citep{Carroll:22,LiuFangchen:22,Sun:22}. 
\citet{Reed:22} trained a Transformer that scaled to over 600 tasks. 
Most recently, \citet{Brohan:22} presented a scalable Transformer for real-world robotics manipulation.
\citet{Laskin:22} and \citet{Liu:23} make use of a multi-episodic context to elicit in-context learning capabilities.
Other works use a Transformer backbone for history compression in online RL \citep{Parisotto:20,Paischer:22,Paischer:23}.
\citet{Li:23} and \citet{Yang:23} cover the landscape of Transformers in RL in more detail.

\textbf{Continual and multi-task RL.}
A plethora of works have considered learning multiple tasks concurrently via RL \citep{Tanaka:03,Rusu:policy:16,Borsa:16,Rajeswaran:17,Bsat:17,Andreas:17,Deramo:20,Yu:20,Igl:20,Sodhani:21}.
In CRL, however, tasks are considered to not be readily available at the same time.
Many continual learning methods were proposed for computer vision, but can also be applied to a CRL setting. 
Early works include regularization approaches, such as EWC \citep{Kirkpatrick:17}, Synaptic Intelligence \citep{Zenke:17}, Memory-aware Synapses \citep{Aljundi:18}, Gradient Episodic Memory (GEM, \citealp{Lopez:17}) and A-GEM \citep{Chaudhry:18}. 
Other approaches rely on adding new modules to existing architecture \citep{Rusu:16}, iterative pruning \citep{Mallya:18}, or improved exploration \citep{Steinparz:22}. 
A number of approaches have been tested on the Continual-World benchmark \citep{Wolczyk:21}, including 3RL \citep{Caccia:22}, ClonExSAC \citep{Wolczyk:22} and Modulating masks \citep{Ben:22}. 
L2P \citep{Wang:22} learns to prompt a frozen Vision Transformer \citep{Dosovitskiy:21} and consistently outperforms prior methods on a number of continual learning benchmarks. 
Other follow-up works that rely on a prompting mechanism for CL have been proposed \citep{Wang_dual:22, Smith:22, Razdaibiedina:23}. 
Recent works provide a comprehensive overview of the field of CRL \citep{Hadsell:20,Lesort:20,Khetarpal:22,Baker:23}. 

\textbf{Parameter-efficient fine-tuning and Prompting.} Large-language models (LLMs) are usually pre-trained on vast amounts of data \citep{Devlin:18,Radford:19,Brown:20}. 
After pre-training, it is desirable to specialize or fine-tune the foundation model \citep{Bommasani:21} to a down-stream task. 
A common way is fine-tuning all network weights or a fraction thereof. 
Fine-tuning the entire network is costly and suffers from catastrophic forgetting.
Parameter-efficient fine-tuning methods and prompt-based tuning offer attractive alternatives. 
\citet{Houlsby:19} repurposed Adapter modules \citep{Rebuffi:17} to interleave pretrained Transformer layers. 
Variations thereof have been proposed \citep{Bapna:19,Pfeiffer:20,Pfeiffer:20a,Karimi:21}. 
Low Rank Adaptation injects trainable low-rank decomposition matrices into every layer of the model \citep{Hu:21}.
(IA$\text{)}^3$ modulates the inner activation flow of the Transformer layers by elementwise multiplication with learned modulation vectors \citep{Liu:22}.
Prompt tuning conditions the pre-trained model on new tasks by prepending learnable prompts to the embedded input sequence \citep{Lester:21}. 
Similarly, prefix-tuning adds learnable prefix vectors to the keys and values of each attention head input \citep{Xiang:21}.
P-tuning v2, applies learnable prompts at each Transformer layer, instead of merely the input layer \citep{Liu:21a}.
UniPELT combines different PEFT methods in a unified framework \citep{Mao:22}
\citet{Liu:21} give a comprehensive overview of prompt-based learning for NLP.

\section{Conclusion}\label{sec:conclusion}
Adapting agents to new tasks, while preserving performance on previously learned tasks, remains a major challenge towards more general RL agents.
Consequently, we conduct a comprehensive evaluation of established fine-tuning, PEFT and PBT methods for Transformers in RL. 
We evaluate both how well new tasks are learned and how strongly performance deteriorates on the pre-training tasks.
While full fine-tuning of a pre-trained model adapts well to new tasks, it suffers from catastrophic forgetting of pre-training tasks. 
Prompt-based tuning techniques preserve performance on pre-training tasks, but cannot compete on new tasks.
We propose a novel method, L2M, which performs well in both dimensions. 
L2M enables efficient fine-tuning of pre-trained models via learnable modulation matrices, resulting in high performance both on new tasks and on pre-training tasks.
Finally, we release a large dataset of trajectories for the Meta-World and DMControl benchmark suites to facilitate future research in the domain of offline RL.

For future work, we envision an investigation on the compositionality of learned modulation matrices.
While currently a distinct set of modulators is selected for a given input sequence, we believe that by combining modulators across tasks, our method can improve in terms of performance and parameter-efficiency. 
Such an approach could improve forward-transfer, where new tasks benefit even more from previously learned tasks by combining existing skills. 
Furthermore, the context length of the models we consider is currently limited. Extending the context to leverage information from multiple episodes and tasks may improve the model's ability to generalize to new tasks or domains. 
Another promising avenue of research is to investigate more diverse domains and explore how different fine-tuning methods perform on more distinct fine-tuning domains.

\begin{ack}
The ELLIS Unit Linz, the LIT AI Lab, the Institute for Machine Learning, are supported by the Federal State Upper Austria. We thank the projects AI-MOTION (LIT-2018-6-YOU-212), DeepFlood (LIT-2019-8-YOU-213), Medical Cognitive Computing Center (MC3), INCONTROL-RL (FFG-881064), PRIMAL (FFG-873979), S3AI (FFG-872172), DL for GranularFlow (FFG-871302), EPILEPSIA (FFG-892171), AIRI FG 9-N (FWF-36284, FWF-36235), AI4GreenHeatingGrids(FFG- 899943), INTEGRATE (FFG-892418), ELISE (H2020-ICT-2019-3 ID: 951847), Stars4Waters (HORIZON-CL6-2021-CLIMATE-01-01). We thank Audi.JKU Deep Learning Center, TGW LOGISTICS GROUP GMBH, Silicon Austria Labs (SAL), University SAL Labs initiative, FILL Gesellschaft mbH, Anyline GmbH, Google, ZF Friedrichshafen AG, Robert Bosch GmbH, UCB Biopharma SRL, Merck Healthcare KGaA, Verbund AG, GLS (Univ. Waterloo) Software Competence Center Hagenberg GmbH, T\"{U}V Austria, Frauscher Sensonic, TRUMPF and the NVIDIA Corporation.
\end{ack}

\bibliography{refs}

\begin{thebibliography}{}

\bibitem[Agarwal et~al., 2021]{Agarwal:21}
Agarwal, R., Schwarzer, M., Castro, P.~S., Courville, A.~C., and Bellemare, M.
  (2021).
\newblock Deep reinforcement learning at the edge of the statistical precipice.
\newblock {\em Advances in neural information processing systems},
  34:29304--29320.

\bibitem[Aljundi et~al., 2018]{Aljundi:18}
Aljundi, R., Babiloni, F., Elhoseiny, M., Rohrbach, M., and Tuytelaars, T.
  (2018).
\newblock Memory aware synapses: Learning what (not) to forget.
\newblock In {\em Proceedings of the European Conference on Computer Vision
  (ECCV)}, pages 139--154.

\bibitem[Andreas et~al., 2017]{Andreas:17}
Andreas, J., Klein, D., and Levine, S. (2017).
\newblock Modular multitask reinforcement learning with policy sketches.
\newblock In Precup, D. and Teh, Y.~W., editors, {\em Proceedings of the 34th
  International Conference on Machine Learning, {ICML} 2017, Sydney, NSW,
  Australia, 6-11 August 2017}, volume~70 of {\em Proceedings of Machine
  Learning Research}, pages 166--175. {PMLR}.

\bibitem[Arjona-Medina et~al., 2019]{Arjona:19}
Arjona-Medina, J.~A., Gillhofer, M., Widrich, M., Unterthiner, T.,
  Brandstetter, J., and Hochreiter, S. (2019).
\newblock Rudder: Return decomposition for delayed rewards.
\newblock {\em Advances in Neural Information Processing Systems}, 32.

\bibitem[Ba et~al., 2016]{Ba:16}
Ba, J.~L., Kiros, J.~R., and Hinton, G.~E. (2016).
\newblock Layer normalization.
\newblock {\em arXiv preprint arXiv:1607.06450}.

\bibitem[Baevski et~al., 2020]{Baevski:20}
Baevski, A., Zhou, Y., Mohamed, A., and Auli, M. (2020).
\newblock wav2vec 2.0: A framework for self-supervised learning of speech
  representations.
\newblock {\em Advances in Neural Information Processing Systems},
  33:12449--12460.

\bibitem[Baker et~al., 2023]{Baker:23}
Baker, M.~M., New, A., Aguilar-Simon, M., Al-Halah, Z., Arnold, S.~M.,
  Ben-Iwhiwhu, E., Brna, A.~P., Brooks, E., Brown, R.~C., Daniels, Z., et~al.
  (2023).
\newblock A domain-agnostic approach for characterization of lifelong learning
  systems.
\newblock {\em Neural Networks}.

\bibitem[Bapna et~al., 2019]{Bapna:19}
Bapna, A., Arivazhagan, N., and Firat, O. (2019).
\newblock Simple, scalable adaptation for neural machine translation.
\newblock {\em arXiv preprint arXiv:1909.08478}.

\bibitem[Bellemare et~al., 2020]{Bellemare:20}
Bellemare, M.~G., Candido, S., Castro, P.~S., Gong, J., Machado, M.~C., Moitra,
  S., Ponda, S.~S., and Wang, Z. (2020).
\newblock Autonomous navigation of stratospheric balloons using reinforcement
  learning.
\newblock {\em Nature}, 588(7836):77--82.

\bibitem[Ben-Iwhiwhu et~al., 2022]{Ben:22}
Ben-Iwhiwhu, E., Nath, S., Pilly, P.~K., Kolouri, S., and Soltoggio, A. (2022).
\newblock Lifelong reinforcement learning with modulating masks.
\newblock {\em arXiv preprint arXiv:2212.11110}.

\bibitem[Berner et~al., 2019]{Berner:19}
Berner, C., Brockman, G., Chan, B., Cheung, V., D{k{e}}biak, P., Dennison, C.,
  Farhi, D., Fischer, Q., Hashme, S., Hesse, C., et~al. (2019).
\newblock Dota 2 with large scale deep reinforcement learning.
\newblock {\em arXiv preprint arXiv:1912.06680}.

\bibitem[Bommasani et~al., 2021]{Bommasani:21}
Bommasani, R., Hudson, D.~A., Adeli, E., Altman, R., Arora, S., von Arx, S.,
  Bernstein, M.~S., Bohg, J., Bosselut, A., Brunskill, E., et~al. (2021).
\newblock On the opportunities and risks of foundation models.
\newblock {\em arXiv preprint arXiv:2108.07258}.

\bibitem[Borsa et~al., 2016]{Borsa:16}
Borsa, D., Graepel, T., and Shawe{-}Taylor, J. (2016).
\newblock Learning shared representations in multi-task reinforcement learning.
\newblock {\em CoRR}, abs/1603.02041.

\bibitem[Brohan et~al., 2022]{Brohan:22}
Brohan, A., Brown, N., Carbajal, J., Chebotar, Y., Dabis, J., Finn, C.,
  Gopalakrishnan, K., Hausman, K., Herzog, A., Hsu, J., et~al. (2022).
\newblock Rt-1: Robotics transformer for real-world control at scale.
\newblock {\em arXiv preprint arXiv:2212.06817}.

\bibitem[Brown et~al., 2020]{Brown:20}
Brown, T., Mann, B., Ryder, N., Subbiah, M., Kaplan, J.~D., Dhariwal, P.,
  Neelakantan, A., Shyam, P., Sastry, G., Askell, A., et~al. (2020).
\newblock Language models are few-shot learners.
\newblock {\em Advances in neural information processing systems},
  33:1877--1901.

\bibitem[Caccia et~al., 2022]{Caccia:22}
Caccia, M., Mueller, J., Kim, T., Charlin, L., and Fakoor, R. (2022).
\newblock Task-agnostic continual reinforcement learning: In praise of a simple
  baseline.
\newblock {\em arXiv preprint arXiv:2205.14495}.

\bibitem[Carroll et~al., 2022]{Carroll:22}
Carroll, M., Paradise, O., Lin, J., Georgescu, R., Sun, M., Bignell, D.,
  Milani, S., Hofmann, K., Hausknecht, M., Dragan, A., et~al. (2022).
\newblock Uni [mask]: Unified inference in sequential decision problems.
\newblock {\em Advances in neural information processing systems},
  35:35365--35378.

\bibitem[Caruana, 1997]{caruana:97}
Caruana, R. (1997).
\newblock Multitask learning.
\newblock {\em Machine Learning}, 28.

\bibitem[Chaudhry et~al., 2019]{Chaudhry:18}
Chaudhry, A., Ranzato, M., Rohrbach, M., and Elhoseiny, M. (2019).
\newblock Efficient lifelong learning with {A-GEM}.
\newblock In {\em 7th International Conference on Learning Representations,
  {ICLR} 2019, New Orleans, LA, USA, May 6-9, 2019}. OpenReview.net.

\bibitem[Chen et~al., 2021]{Chen:21}
Chen, L., Lu, K., Rajeswaran, A., Lee, K., Grover, A., Laskin, M., Abbeel, P.,
  Srinivas, A., and Mordatch, I. (2021).
\newblock Decision transformer: Reinforcement learning via sequence modeling.
\newblock {\em Advances in neural information processing systems},
  34:15084--15097.

\bibitem[Degrave et~al., 2022]{Degrave:2022}
Degrave, J., Felici, F., Buchli, J., Neunert, M., Tracey, B., Carpanese, F.,
  Ewalds, T., Hafner, R., Abdolmaleki, A., de~Las~Casas, D., et~al. (2022).
\newblock Magnetic control of tokamak plasmas through deep reinforcement
  learning.
\newblock {\em Nature}, 602(7897):414--419.

\bibitem[D'Eramo et~al., 2020]{Deramo:20}
D'Eramo, C., Tateo, D., Bonarini, A., Restelli, M., and Peters, J. (2020).
\newblock Sharing knowledge in multi-task deep reinforcement learning.
\newblock In {\em International Conference on Learning Representations}.

\bibitem[Devlin et~al., 2019]{Devlin:18}
Devlin, J., Chang, M., Lee, K., and Toutanova, K. (2019).
\newblock {BERT:} pre-training of deep bidirectional transformers for language
  understanding.
\newblock In Burstein, J., Doran, C., and Solorio, T., editors, {\em
  Proceedings of the 2019 Conference of the North American Chapter of the
  Association for Computational Linguistics: Human Language Technologies,
  {NAACL-HLT} 2019, Minneapolis, MN, USA, June 2-7, 2019, Volume 1 (Long and
  Short Papers)}, pages 4171--4186. Association for Computational Linguistics.

\bibitem[Dosovitskiy et~al., 2021]{Dosovitskiy:21}
Dosovitskiy, A., Beyer, L., Kolesnikov, A., Weissenborn, D., Zhai, X.,
  Unterthiner, T., Dehghani, M., Minderer, M., Heigold, G., Gelly, S.,
  Uszkoreit, J., and Houlsby, N. (2021).
\newblock An image is worth 16x16 words: Transformers for image recognition at
  scale.
\newblock In {\em 9th International Conference on Learning Representations,
  {ICLR} 2021, Virtual Event, Austria, May 3-7, 2021}. OpenReview.net.

\bibitem[El~Bsat et~al., 2017]{Bsat:17}
El~Bsat, S., Bou~Ammar, H., and Taylor, M. (2017).
\newblock Scalable multitask policy gradient reinforcement learning.
\newblock {\em Proceedings of the AAAI Conference on Artificial Intelligence},
  31(1).

\bibitem[Fu et~al., 2020]{Fu:20}
Fu, J., Kumar, A., Nachum, O., Tucker, G., and Levine, S. (2020).
\newblock D4rl: Datasets for deep data-driven reinforcement learning.
\newblock {\em arXiv preprint arXiv:2004.07219}.

\bibitem[Fürst et~al., 2022]{Fürst:21}
Fürst, A., Rumetshofer, E., Lehner, J., Tran, V., Tang, F., Ramsauer, H.,
  Kreil, D., Kopp, M., Klambauer, G., Bitto-Nemling, A., and Hochreiter, S.
  (2022).
\newblock Cloob: Modern hopfield networks with infoloob outperform clip.

\bibitem[Gupta et~al., 2022]{Gupta:22}
Gupta, A., Fan, L., Ganguli, S., and Fei{-}Fei, L. (2022).
\newblock Metamorph: Learning universal controllers with transformers.
\newblock In {\em The Tenth International Conference on Learning
  Representations, {ICLR} 2022, Virtual Event, April 25-29, 2022}.
  OpenReview.net.

\bibitem[Haarnoja et~al., 2018]{Haarnoja:18}
Haarnoja, T., Zhou, A., Abbeel, P., and Levine, S. (2018).
\newblock Soft actor-critic: Off-policy maximum entropy deep reinforcement
  learning with a stochastic actor.
\newblock In Dy, J.~G. and Krause, A., editors, {\em Proceedings of the 35th
  International Conference on Machine Learning, {ICML} 2018,
  Stockholmsm{\"{a}}ssan, Stockholm, Sweden, July 10-15, 2018}, volume~80 of
  {\em Proceedings of Machine Learning Research}, pages 1856--1865. {PMLR}.

\bibitem[Hadsell et~al., 2020]{Hadsell:20}
Hadsell, R., Rao, D., Rusu, A.~A., and Pascanu, R. (2020).
\newblock Embracing change: Continual learning in deep neural networks.
\newblock {\em Trends in cognitive sciences}, 24(12):1028--1040.

\bibitem[Hafner et~al., 2019]{Hafner:19}
Hafner, D., Lillicrap, T., Fischer, I., Villegas, R., Ha, D., Lee, H., and
  Davidson, J. (2019).
\newblock Learning latent dynamics for planning from pixels.
\newblock In {\em International conference on machine learning}, pages
  2555--2565. PMLR.

\bibitem[He et~al., 2022]{He:22}
He, K., Chen, X., Xie, S., Li, Y., Doll{\'{a}}r, P., and Girshick, R.~B.
  (2022).
\newblock Masked autoencoders are scalable vision learners.
\newblock In {\em {IEEE/CVF} Conference on Computer Vision and Pattern
  Recognition, {CVPR} 2022, New Orleans, LA, USA, June 18-24, 2022}, pages
  15979--15988. {IEEE}.

\bibitem[Ho et~al., 2022]{Ho:22}
Ho, J., Chan, W., Saharia, C., Whang, J., Gao, R., Gritsenko, A.~A., Kingma,
  D.~P., Poole, B., Norouzi, M., Fleet, D.~J., and Salimans, T. (2022).
\newblock Imagen video: High definition video generation with diffusion models.
\newblock {\em CoRR}, abs/2210.02303.

\bibitem[Hochreiter and Schmidhuber, 1997]{Hochreiter:97}
Hochreiter, S. and Schmidhuber, J. (1997).
\newblock Long short-term memory.
\newblock {\em Neural computation}, 9(8):1735--1780.

\bibitem[Houlsby et~al., 2019]{Houlsby:19}
Houlsby, N., Giurgiu, A., Jastrzebski, S., Morrone, B., De~Laroussilhe, Q.,
  Gesmundo, A., Attariyan, M., and Gelly, S. (2019).
\newblock Parameter-efficient transfer learning for nlp.
\newblock In {\em International Conference on Machine Learning}, pages
  2790--2799. PMLR.

\bibitem[Hu et~al., 2022]{Hu:21}
Hu, E.~J., Shen, Y., Wallis, P., Allen{-}Zhu, Z., Li, Y., Wang, S., Wang, L.,
  and Chen, W. (2022).
\newblock Lora: Low-rank adaptation of large language models.
\newblock In {\em The Tenth International Conference on Learning
  Representations, {ICLR} 2022, Virtual Event, April 25-29, 2022}.
  OpenReview.net.

\bibitem[Igl et~al., 2020]{Igl:20}
Igl, M., Gambardella, A., He, J., Nardelli, N., Siddharth, N., Boehmer, W., and
  Whiteson, S. (2020).
\newblock Multitask soft option learning.
\newblock In Adams, R.~P. and Gogate, V., editors, {\em Proceedings of the
  Thirty-Sixth Conference on Uncertainty in Artificial Intelligence, {UAI}
  2020, virtual online, August 3-6, 2020}, volume 124 of {\em Proceedings of
  Machine Learning Research}, pages 969--978. {AUAI} Press.

\bibitem[Janner et~al., 2021]{Janner:21}
Janner, M., Li, Q., and Levine, S. (2021).
\newblock Offline reinforcement learning as one big sequence modeling problem.
\newblock {\em Advances in neural information processing systems},
  34:1273--1286.

\bibitem[Jiang et~al., 2022]{Jiang:22}
Jiang, Y., Gupta, A., Zhang, Z., Wang, G., Dou, Y., Chen, Y., Fei-Fei, L.,
  Anandkumar, A., Zhu, Y., and Fan, L. (2022).
\newblock Vima: General robot manipulation with multimodal prompts.
\newblock {\em arXiv preprint arXiv:2210.03094}.

\bibitem[Karimi~Mahabadi et~al., 2021]{Karimi:21}
Karimi~Mahabadi, R., Henderson, J., and Ruder, S. (2021).
\newblock Compacter: Efficient low-rank hypercomplex adapter layers.
\newblock {\em Advances in Neural Information Processing Systems},
  34:1022--1035.

\bibitem[Khetarpal et~al., 2022]{Khetarpal:22}
Khetarpal, K., Riemer, M., Rish, I., and Precup, D. (2022).
\newblock Towards continual reinforcement learning: A review and perspectives.
\newblock {\em Journal of Artificial Intelligence Research}, 75:1401--1476.

\bibitem[Kingma and Ba, 2015]{Kingma:14}
Kingma, D.~P. and Ba, J. (2015).
\newblock Adam: {A} method for stochastic optimization.
\newblock In Bengio, Y. and LeCun, Y., editors, {\em 3rd International
  Conference on Learning Representations, {ICLR} 2015, San Diego, CA, USA, May
  7-9, 2015, Conference Track Proceedings}.

\bibitem[Kirkpatrick et~al., 2017]{Kirkpatrick:17}
Kirkpatrick, J., Pascanu, R., Rabinowitz, N., Veness, J., Desjardins, G., Rusu,
  A.~A., Milan, K., Quan, J., Ramalho, T., Grabska-Barwinska, A., et~al.
  (2017).
\newblock Overcoming catastrophic forgetting in neural networks.
\newblock {\em Proceedings of the national academy of sciences},
  114(13):3521--3526.

\bibitem[Laskin et~al., 2022]{Laskin:22}
Laskin, M., Wang, L., Oh, J., Parisotto, E., Spencer, S., Steigerwald, R.,
  Strouse, D., Hansen, S., Filos, A., Brooks, E., et~al. (2022).
\newblock In-context reinforcement learning with algorithm distillation.
\newblock {\em arXiv preprint arXiv:2210.14215}.

\bibitem[Lee et~al., 2022]{Lee:22}
Lee, K.-H., Nachum, O., Yang, M., Lee, L., Freeman, D., Xu, W., Guadarrama, S.,
  Fischer, I., Jang, E., Michalewski, H., et~al. (2022).
\newblock Multi-game decision transformers.
\newblock {\em arXiv preprint arXiv:2205.15241}.

\bibitem[Lesort et~al., 2020]{Lesort:20}
Lesort, T., Lomonaco, V., Stoian, A., Maltoni, D., Filliat, D., and
  D{\'\i}az-Rodr{\'\i}guez, N. (2020).
\newblock Continual learning for robotics: Definition, framework, learning
  strategies, opportunities and challenges.
\newblock {\em Information fusion}, 58:52--68.

\bibitem[Lester et~al., 2021]{Lester:21}
Lester, B., Al-Rfou, R., and Constant, N. (2021).
\newblock The power of scale for parameter-efficient prompt tuning.
\newblock In {\em Proceedings of the 2021 Conference on Empirical Methods in
  Natural Language Processing}, pages 3045--3059, Online and Punta Cana,
  Dominican Republic. Association for Computational Linguistics.

\bibitem[Li et~al., 2023]{Li:23}
Li, W., Luo, H., Lin, Z., Zhang, C., Lu, Z., and Ye, D. (2023).
\newblock A survey on transformers in reinforcement learning.
\newblock {\em arXiv preprint arXiv:2301.03044}.

\bibitem[Li and Liang, 2021]{Xiang:21}
Li, X.~L. and Liang, P. (2021).
\newblock Prefix-tuning: Optimizing continuous prompts for generation.
\newblock In Zong, C., Xia, F., Li, W., and Navigli, R., editors, {\em
  Proceedings of the 59th Annual Meeting of the Association for Computational
  Linguistics and the 11th International Joint Conference on Natural Language
  Processing, {ACL/IJCNLP} 2021, (Volume 1: Long Papers), Virtual Event, August
  1-6, 2021}, pages 4582--4597. Association for Computational Linguistics.

\bibitem[Liu et~al., 2022a]{LiuFangchen:22}
Liu, F., Liu, H., Grover, A., and Abbeel, P. (2022a).
\newblock Masked autoencoding for scalable and generalizable decision making.
\newblock {\em Advances in Neural Information Processing Systems},
  35:12608--12618.

\bibitem[Liu and Abbeel, 2023]{Liu:23}
Liu, H. and Abbeel, P. (2023).
\newblock Emergent agentic transformer from chain of hindsight experience.
\newblock {\em arXiv preprint arXiv:2305.16554}.

\bibitem[Liu et~al., 2022b]{Liu:22}
Liu, H., Tam, D., Muqeeth, M., Mohta, J., Huang, T., Bansal, M., and Raffel, C.
  (2022b).
\newblock Few-shot parameter-efficient fine-tuning is better and cheaper than
  in-context learning.
\newblock {\em arXiv preprint arXiv:2205.05638}.

\bibitem[Liu et~al., 2021a]{Liu:21}
Liu, P., Yuan, W., Fu, J., Jiang, Z., Hayashi, H., and Neubig, G. (2021a).
\newblock Pre-train, prompt, and predict: A systematic survey of prompting
  methods in natural language processing.
\newblock {\em arXiv preprint arXiv:2107.13586}.

\bibitem[Liu et~al., 2021b]{Liu:21a}
Liu, X., Ji, K., Fu, Y., Du, Z., Yang, Z., and Tang, J. (2021b).
\newblock P-tuning v2: Prompt tuning can be comparable to fine-tuning
  universally across scales and tasks.
\newblock {\em arXiv preprint arXiv:2110.07602}.

\bibitem[Lopez-Paz and Ranzato, 2017]{Lopez:17}
Lopez-Paz, D. and Ranzato, M. (2017).
\newblock Gradient episodic memory for continual learning.
\newblock {\em Advances in neural information processing systems}, 30.

\bibitem[Loshchilov and Hutter, 2018]{Loshchilov:18}
Loshchilov, I. and Hutter, F. (2018).
\newblock Decoupled weight decay regularization.
\newblock In {\em International Conference on Learning Representations}.

\bibitem[Maas et~al., 2013]{Maas:13}
Maas, A.~L., Hannun, A.~Y., Ng, A.~Y., et~al. (2013).
\newblock Rectifier nonlinearities improve neural network acoustic models.
\newblock In {\em Proc. icml}, volume~30, page~3. Atlanta, Georgia, USA.

\bibitem[Mallya and Lazebnik, 2018]{Mallya:18}
Mallya, A. and Lazebnik, S. (2018).
\newblock Packnet: Adding multiple tasks to a single network by iterative
  pruning.
\newblock In {\em Proceedings of the IEEE conference on Computer Vision and
  Pattern Recognition}, pages 7765--7773.

\bibitem[Mandi et~al., 2022]{Mandi:22}
Mandi, Z., Abbeel, P., and James, S. (2022).
\newblock On the effectiveness of fine-tuning versus meta-reinforcement
  learning.
\newblock {\em arXiv preprint arXiv:2206.03271}.

\bibitem[Mao et~al., 2022]{Mao:22}
Mao, Y., Mathias, L., Hou, R., Almahairi, A., Ma, H., Han, J., Yih, S., and
  Khabsa, M. (2022).
\newblock Unipelt: A unified framework for parameter-efficient language model
  tuning.
\newblock In {\em Proceedings of the 60th Annual Meeting of the Association for
  Computational Linguistics (Volume 1: Long Papers)}, pages 6253--6264.

\bibitem[McCloskey and Cohen, 1989]{McCloskey:1989}
McCloskey, M. and Cohen, N. (1989).
\newblock Catastrophic interference in connectionist networks: The sequential
  learning problem.
\newblock In {\em Psychology of learning and motivation}, volume~24, pages
  109--165. Elsevier.

\bibitem[Meng et~al., 2021]{Meng:21}
Meng, L., Wen, M., Yang, Y., Le, C., Li, X., Zhang, W., Wen, Y., Zhang, H.,
  Wang, J., and Xu, B. (2021).
\newblock Offline pre-trained multi-agent decision transformer: One big
  sequence model conquers all starcraftii tasks.
\newblock {\em arXiv preprint arXiv:2112.02845}.

\bibitem[Micikevicius et~al., 2017]{micikevicius2017mixed}
Micikevicius, P., Narang, S., Alben, J., Diamos, G., Elsen, E., Garcia, D.,
  Ginsburg, B., Houston, M., Kuchaiev, O., Venkatesh, G., et~al. (2017).
\newblock Mixed precision training.
\newblock {\em arXiv preprint arXiv:1710.03740}.

\bibitem[Paischer et~al., 2023]{Paischer:23}
Paischer, F., Adler, T., Hofmarcher, M., and Hochreiter, S. (2023).
\newblock Semantic {HELM:} an interpretable memory for reinforcement learning.
\newblock {\em CoRR}, abs/2306.09312.

\bibitem[Paischer et~al., 2022]{Paischer:22}
Paischer, F., Adler, T., Patil, V., Bitto-Nemling, A., Holzleitner, M., Lehner,
  S., Eghbal-Zadeh, H., and Hochreiter, S. (2022).
\newblock History compression via language models in reinforcement learning.
\newblock In {\em International Conference on Machine Learning}, pages
  17156--17185. PMLR.

\bibitem[Parisotto et~al., 2020]{Parisotto:20}
Parisotto, E., Song, H.~F., Rae, J.~W., Pascanu, R., G{\"{u}}l{\c{c}}ehre,
  {\c{C}}., Jayakumar, S.~M., Jaderberg, M., Kaufman, R.~L., Clark, A., Noury,
  S., Botvinick, M.~M., Heess, N., and Hadsell, R. (2020).
\newblock Stabilizing transformers for reinforcement learning.
\newblock In {\em Proceedings of the 37th International Conference on Machine
  Learning, {ICML} 2020, 13-18 July 2020, Virtual Event}, volume 119 of {\em
  Proceedings of Machine Learning Research}, pages 7487--7498. {PMLR}.

\bibitem[Pfeiffer et~al., 2021]{Pfeiffer:20}
Pfeiffer, J., Kamath, A., R{\"{u}}ckl{\'{e}}, A., Cho, K., and Gurevych, I.
  (2021).
\newblock Adapterfusion: Non-destructive task composition for transfer
  learning.
\newblock In Merlo, P., Tiedemann, J., and Tsarfaty, R., editors, {\em
  Proceedings of the 16th Conference of the European Chapter of the Association
  for Computational Linguistics: Main Volume, {EACL} 2021, Online, April 19 -
  23, 2021}, pages 487--503. Association for Computational Linguistics.

\bibitem[Pfeiffer et~al., 2020]{Pfeiffer:20a}
Pfeiffer, J., Vulic, I., Gurevych, I., and Ruder, S. (2020).
\newblock {MAD-X:} an adapter-based framework for multi-task cross-lingual
  transfer.
\newblock In Webber, B., Cohn, T., He, Y., and Liu, Y., editors, {\em
  Proceedings of the 2020 Conference on Empirical Methods in Natural Language
  Processing, {EMNLP} 2020, Online, November 16-20, 2020}, pages 7654--7673.
  Association for Computational Linguistics.

\bibitem[Radford et~al., 2021]{Radford:21}
Radford, A., Kim, J.~W., Hallacy, C., Ramesh, A., Goh, G., Agarwal, S., Sastry,
  G., Askell, A., Mishkin, P., Clark, J., Krueger, G., and Sutskever, I.
  (2021).
\newblock Learning transferable visual models from natural language
  supervision.
\newblock In Meila, M. and Zhang, T., editors, {\em Proceedings of the 38th
  International Conference on Machine Learning, {ICML} 2021, 18-24 July 2021,
  Virtual Event}, volume 139 of {\em Proceedings of Machine Learning Research},
  pages 8748--8763. {PMLR}.

\bibitem[Radford et~al., 2022]{Radford:22}
Radford, A., Kim, J.~W., Xu, T., Brockman, G., McLeavey, C., and Sutskever, I.
  (2022).
\newblock Robust speech recognition via large-scale weak supervision.
\newblock {\em arXiv preprint arXiv:2212.04356}.

\bibitem[Radford et~al., 2018]{Radford:18}
Radford, A., Narasimhan, K., Salimans, T., Sutskever, I., et~al. (2018).
\newblock Improving language understanding by generative pre-training.

\bibitem[Radford et~al., 2019]{Radford:19}
Radford, A., Wu, J., Child, R., Luan, D., Amodei, D., Sutskever, I., et~al.
  (2019).
\newblock Language models are unsupervised multitask learners.
\newblock {\em OpenAI blog}, 1(8):9.

\bibitem[Raffin et~al., 2021]{Raffin:21}
Raffin, A., Hill, A., Gleave, A., Kanervisto, A., Ernestus, M., and Dormann, N.
  (2021).
\newblock Stable-baselines3: Reliable reinforcement learning implementations.
\newblock {\em Journal of Machine Learning Research}, 22(268):1--8.

\bibitem[Rajeswaran et~al., 2017]{Rajeswaran:17}
Rajeswaran, A., Ghotra, S., Ravindran, B., and Levine, S. (2017).
\newblock {EPO}pt: Learning robust neural network policies using model
  ensembles.
\newblock In {\em International Conference on Learning Representations}.

\bibitem[Ramesh et~al., 2021]{Ramesh:21}
Ramesh, A., Pavlov, M., Goh, G., Gray, S., Voss, C., Radford, A., Chen, M., and
  Sutskever, I. (2021).
\newblock Zero-shot text-to-image generation.
\newblock In {\em International Conference on Machine Learning}, pages
  8821--8831. PMLR.

\bibitem[Razdaibiedina et~al., 2023]{Razdaibiedina:23}
Razdaibiedina, A., Mao, Y., Hou, R., Khabsa, M., Lewis, M., and Almahairi, A.
  (2023).
\newblock Progressive prompts: Continual learning for language models.
\newblock {\em arXiv preprint arXiv:2301.12314}.

\bibitem[Rebuffi et~al., 2017]{Rebuffi:17}
Rebuffi, S.-A., Bilen, H., and Vedaldi, A. (2017).
\newblock Learning multiple visual domains with residual adapters.
\newblock {\em Advances in neural information processing systems}, 30.

\bibitem[Reed et~al., 2022]{Reed:22}
Reed, S.~E., Zolna, K., Parisotto, E., Colmenarejo, S.~G., Novikov, A.,
  Barth{-}Maron, G., Gimenez, M., Sulsky, Y., Kay, J., Springenberg, J.~T.,
  Eccles, T., Bruce, J., Razavi, A., Edwards, A., Heess, N., Chen, Y., Hadsell,
  R., Vinyals, O., Bordbar, M., and de~Freitas, N. (2022).
\newblock A generalist agent.
\newblock {\em CoRR}, abs/2205.06175.

\bibitem[Rombach et~al., 2022]{Rombach:22}
Rombach, R., Blattmann, A., Lorenz, D., Esser, P., and Ommer, B. (2022).
\newblock High-resolution image synthesis with latent diffusion models.
\newblock In {\em Proceedings of the IEEE/CVF Conference on Computer Vision and
  Pattern Recognition}, pages 10684--10695.

\bibitem[Rusu et~al., 2016a]{Rusu:policy:16}
Rusu, A.~A., Colmenarejo, S.~G., G{\"{u}}l{\c{c}}ehre, {\c{C}}., Desjardins,
  G., Kirkpatrick, J., Pascanu, R., Mnih, V., Kavukcuoglu, K., and Hadsell, R.
  (2016a).
\newblock Policy distillation.
\newblock In Bengio, Y. and LeCun, Y., editors, {\em 4th International
  Conference on Learning Representations, {ICLR} 2016, San Juan, Puerto Rico,
  May 2-4, 2016, Conference Track Proceedings}.

\bibitem[Rusu et~al., 2016b]{Rusu:16}
Rusu, A.~A., Rabinowitz, N.~C., Desjardins, G., Soyer, H., Kirkpatrick, J.,
  Kavukcuoglu, K., Pascanu, R., and Hadsell, R. (2016b).
\newblock Progressive neural networks.
\newblock {\em arXiv preprint arXiv:1606.04671}.

\bibitem[Schmidhuber, 2019]{Schmidhuber:19}
Schmidhuber, J. (2019).
\newblock Reinforcement learning upside down: Don't predict rewards--just map
  them to actions.
\newblock {\em arXiv preprint arXiv:1912.02875}.

\bibitem[Schweighofer et~al., 2022]{Schweighofer:22}
Schweighofer, K., Dinu, M.-c., Radler, A., Hofmarcher, M., Patil, V.~P.,
  Bitto-Nemling, A., Eghbal-zadeh, H., and Hochreiter, S. (2022).
\newblock A dataset perspective on offline reinforcement learning.
\newblock In {\em Conference on Lifelong Learning Agents}, pages 470--517.
  PMLR.

\bibitem[Shang et~al., 2022]{Shang:22}
Shang, J., Kahatapitiya, K., Li, X., and Ryoo, M.~S. (2022).
\newblock Starformer: Transformer with state-action-reward representations for
  visual reinforcement learning.
\newblock In {\em European Conference on Computer Vision}, pages 462--479.
  Springer.

\bibitem[Shridhar et~al., 2022]{Shridar:22}
Shridhar, M., Manuelli, L., and Fox, D. (2022).
\newblock Perceiver-actor: {A} multi-task transformer for robotic manipulation.
\newblock In Liu, K., Kulic, D., and Ichnowski, J., editors, {\em Conference on
  Robot Learning, CoRL 2022, 14-18 December 2022, Auckland, New Zealand},
  volume 205 of {\em Proceedings of Machine Learning Research}, pages 785--799.
  {PMLR}.

\bibitem[Siebenborn et~al., 2022]{Siebenborn:22}
Siebenborn, M., Belousov, B., Huang, J., and Peters, J. (2022).
\newblock How crucial is transformer in decision transformer?
\newblock {\em arXiv preprint arXiv:2211.14655}.

\bibitem[Silver et~al., 2016]{Silver:16}
Silver, D., Huang, A., Maddison, C.~J., Guez, A., Sifre, L., van~den Driessche,
  G., Schrittwieser, J., Antonoglou, I., Panneershelvam, V., Lanctot, M.,
  Dieleman, S., Grewe, D., Nham, J., Kalchbrenner, N., Sutskever, I.,
  Lillicrap, T.~P., Leach, M., Kavukcuoglu, K., Graepel, T., and Hassabis, D.
  (2016).
\newblock Mastering the game of go with deep neural networks and tree search.
\newblock {\em Nat.}, 529(7587):484--489.

\bibitem[Singer et~al., 2022]{Singer:22}
Singer, U., Polyak, A., Hayes, T., Yin, X., An, J., Zhang, S., Hu, Q., Yang,
  H., Ashual, O., Gafni, O., et~al. (2022).
\newblock Make-a-video: Text-to-video generation without text-video data.
\newblock {\em arXiv preprint arXiv:2209.14792}.

\bibitem[Smith et~al., 2022]{Smith:22}
Smith, J.~S., Karlinsky, L., Gutta, V., Cascante-Bonilla, P., Kim, D., Arbelle,
  A., Panda, R., Feris, R., and Kira, Z. (2022).
\newblock Coda-prompt: Continual decomposed attention-based prompting for
  rehearsal-free continual learning.
\newblock {\em arXiv preprint arXiv:2211.13218}.

\bibitem[Sodhani et~al., 2021]{Sodhani:21}
Sodhani, S., Zhang, A., and Pineau, J. (2021).
\newblock Multi-task reinforcement learning with context-based representations.
\newblock In Meila, M. and Zhang, T., editors, {\em Proceedings of the 38th
  International Conference on Machine Learning, {ICML} 2021, 18-24 July 2021,
  Virtual Event}, volume 139 of {\em Proceedings of Machine Learning Research},
  pages 9767--9779. {PMLR}.

\bibitem[Srivastava et~al., 2014]{Srivastava:14}
Srivastava, N., Hinton, G., Krizhevsky, A., Sutskever, I., and Salakhutdinov,
  R. (2014).
\newblock Dropout: a simple way to prevent neural networks from overfitting.
\newblock {\em The journal of machine learning research}, 15(1):1929--1958.

\bibitem[Steinparz et~al., 2022]{Steinparz:22}
Steinparz, C.~A., Schmied, T., Paischer, F., Dinu, M.-C., Patil, V.~P.,
  Bitto-Nemling, A., Eghbal-zadeh, H., and Hochreiter, S. (2022).
\newblock Reactive exploration to cope with non-stationarity in lifelong
  reinforcement learning.
\newblock In {\em Conference on Lifelong Learning Agents}, pages 441--469.
  PMLR.

\bibitem[Sun et~al., 2022]{Sun:22}
Sun, Y., Ma, S., Madaan, R., Bonatti, R., Huang, F., and Kapoor, A. (2022).
\newblock Smart: Self-supervised multi-task pretraining with control
  transformers.
\newblock In {\em The Eleventh International Conference on Learning
  Representations}.

\bibitem[Tanaka and Yamamura, 2003]{Tanaka:03}
Tanaka, F. and Yamamura, M. (2003).
\newblock Multitask reinforcement learning on the distribution of mdps.
\newblock In {\em Proceedings 2003 IEEE International Symposium on
  Computational Intelligence in Robotics and Automation. Computational
  Intelligence in Robotics and Automation for the New Millennium (Cat.
  No.03EX694)}, volume~3, pages 1108--1113 vol.3.

\bibitem[Tassa et~al., 2018]{Tassa:18}
Tassa, Y., Doron, Y., Muldal, A., Erez, T., Li, Y., de~Las~Casas, D., Budden,
  D., Abdolmaleki, A., Merel, J., Lefrancq, A., Lillicrap, T.~P., and
  Riedmiller, M.~A. (2018).
\newblock Deepmind control suite.
\newblock {\em CoRR}, abs/1801.00690.

\bibitem[Todorov et~al., 2012]{Todorov:12}
Todorov, E., Erez, T., and Tassa, Y. (2012).
\newblock {{MuJoCo}}: {{A}} physics engine for model-based control.
\newblock In {\em 2012 {{IEEE}}/{{RSJ International Conference}} on
  {{Intelligent Robots}} and {{Systems}}}, pages 5026--5033.

\bibitem[Van~der Maaten and Hinton, 2008]{Van:08}
Van~der Maaten, L. and Hinton, G. (2008).
\newblock Visualizing data using t-sne.
\newblock {\em Journal of machine learning research}, 9(11).

\bibitem[Vaswani et~al., 2017]{Vaswani:17}
Vaswani, A., Shazeer, N., Parmar, N., Uszkoreit, J., Jones, L., Gomez, A.~N.,
  Kaiser, l., and Polosukhin, I. (2017).
\newblock Attention is all you need.
\newblock {\em Advances in neural information processing systems}, 30.

\bibitem[Vinyals et~al., 2019]{Vinyals:19}
Vinyals, O., Babuschkin, I., Czarnecki, W.~M., Mathieu, M., Dudzik, A., Chung,
  J., Choi, D.~H., Powell, R., Ewalds, T., Georgiev, P., Oh, J., Horgan, D.,
  Kroiss, M., Danihelka, I., Huang, A., Sifre, L., Cai, T., Agapiou, J.~P.,
  Jaderberg, M., Vezhnevets, A.~S., Leblond, R., Pohlen, T., Dalibard, V.,
  Budden, D., Sulsky, Y., Molloy, J., Paine, T.~L., G{\"{u}}l{\c{c}}ehre,
  {\c{C}}., Wang, Z., Pfaff, T., Wu, Y., Ring, R., Yogatama, D., W{\"{u}}nsch,
  D., McKinney, K., Smith, O., Schaul, T., Lillicrap, T.~P., Kavukcuoglu, K.,
  Hassabis, D., Apps, C., and Silver, D. (2019).
\newblock Grandmaster level in starcraft {II} using multi-agent reinforcement
  learning.
\newblock {\em Nat.}, 575(7782):350--354.

\bibitem[Wang et~al., 2022a]{Wang_bootstrapped:22}
Wang, K., Zhao, H., Luo, X., Ren, K., Zhang, W., and Li, D. (2022a).
\newblock Bootstrapped transformer for offline reinforcement learning.
\newblock {\em arXiv preprint arXiv:2206.08569}.

\bibitem[Wang et~al., 2022b]{Wang_dual:22}
Wang, Z., Zhang, Z., Ebrahimi, S., Sun, R., Zhang, H., Lee, C., Ren, X., Su,
  G., Perot, V., Dy, J.~G., and Pfister, T. (2022b).
\newblock Dualprompt: Complementary prompting for rehearsal-free continual
  learning.
\newblock In Avidan, S., Brostow, G.~J., Ciss{\'{e}}, M., Farinella, G.~M., and
  Hassner, T., editors, {\em Computer Vision - {ECCV} 2022 - 17th European
  Conference, Tel Aviv, Israel, October 23-27, 2022, Proceedings, Part {XXVI}},
  volume 13686 of {\em Lecture Notes in Computer Science}, pages 631--648.
  Springer.

\bibitem[Wang et~al., 2022c]{Wang:22}
Wang, Z., Zhang, Z., Lee, C.-Y., Zhang, H., Sun, R., Ren, X., Su, G., Perot,
  V., Dy, J., and Pfister, T. (2022c).
\newblock Learning to prompt for continual learning.
\newblock In {\em Proceedings of the IEEE/CVF Conference on Computer Vision and
  Pattern Recognition}, pages 139--149.

\bibitem[Wo{l}czyk et~al., 2021]{Wolczyk:21}
Wo{l}czyk, M., Zaj{k{a}}c, M., Pascanu, R., Kuci{\'n}ski, L., and Mi{l}o{\'s},
  P. (2021).
\newblock Continual world: A robotic benchmark for continual reinforcement
  learning.
\newblock {\em Advances in Neural Information Processing Systems},
  34:28496--28510.

\bibitem[Wo{l}czyk et~al., 2022]{Wolczyk:22}
Wo{l}czyk, M., Zaj{k{a}}c, M., Pascanu, R., Kuci{\'n}ski, L., and Mi{l}o{\'s},
  P. (2022).
\newblock Disentangling transfer in continual reinforcement learning.
\newblock {\em arXiv preprint arXiv:2209.13900}.

\bibitem[Wolf et~al., 2020]{Wolf:20}
Wolf, T., Debut, L., Sanh, V., Chaumond, J., Delangue, C., Moi, A., Cistac, P.,
  Rault, T., Louf, R., Funtowicz, M., Davison, J., Shleifer, S., von Platen,
  P., Ma, C., Jernite, Y., Plu, J., Xu, C., Scao, T.~L., Gugger, S., Drame, M.,
  Lhoest, Q., and Rush, A.~M. (2020).
\newblock Transformers: State-of-the-art natural language processing.
\newblock In {\em Proceedings of the 2020 Conference on Empirical Methods in
  Natural Language Processing: System Demonstrations}, pages 38--45, Online.
  Association for Computational Linguistics.

\bibitem[Xu et~al., 2022a]{XuMengdi:22}
Xu, M., Lu, Y., Shen, Y., Zhang, S., Zhao, D., and Gan, C. (2022a).
\newblock Hyper-decision transformer for efficient online policy adaptation.
\newblock In {\em The Eleventh International Conference on Learning
  Representations}.

\bibitem[Xu et~al., 2022b]{Xu:22}
Xu, M., Shen, Y., Zhang, S., Lu, Y., Zhao, D., Tenenbaum, J., and Gan, C.
  (2022b).
\newblock Prompting decision transformer for few-shot policy generalization.
\newblock In {\em International Conference on Machine Learning}, pages
  24631--24645. PMLR.

\bibitem[Yang et~al., 2023]{Yang:23}
Yang, S., Nachum, O., Du, Y., Wei, J., Abbeel, P., and Schuurmans, D. (2023).
\newblock Foundation models for decision making: Problems, methods, and
  opportunities.
\newblock {\em arXiv preprint arXiv:2303.04129}.

\bibitem[Yu et~al., 2020a]{Yu:20}
Yu, T., Kumar, S., Gupta, A., Levine, S., Hausman, K., and Finn, C. (2020a).
\newblock Gradient surgery for multi-task learning.
\newblock In Larochelle, H., Ranzato, M., Hadsell, R., Balcan, M., and Lin, H.,
  editors, {\em Advances in Neural Information Processing Systems 33: Annual
  Conference on Neural Information Processing Systems 2020, NeurIPS 2020,
  December 6-12, 2020, virtual}.

\bibitem[Yu et~al., 2020b]{Yu:22}
Yu, T., Quillen, D., He, Z., Julian, R., Hausman, K., Finn, C., and Levine, S.
  (2020b).
\newblock Meta-world: A benchmark and evaluation for multi-task and meta
  reinforcement learning.
\newblock In {\em Conference on robot learning}, pages 1094--1100. PMLR.

\bibitem[Zenke et~al., 2017]{Zenke:17}
Zenke, F., Poole, B., and Ganguli, S. (2017).
\newblock Continual learning through synaptic intelligence.
\newblock In {\em International Conference on Machine Learning}, pages
  3987--3995. PMLR.

\bibitem[Zheng et~al., 2022]{Zheng:22}
Zheng, Q., Zhang, A., and Grover, A. (2022).
\newblock Online decision transformer.
\newblock In Chaudhuri, K., Jegelka, S., Song, L., Szepesv{\'{a}}ri, C., Niu,
  G., and Sabato, S., editors, {\em International Conference on Machine
  Learning, {ICML} 2022, 17-23 July 2022, Baltimore, Maryland, {USA}}, volume
  162 of {\em Proceedings of Machine Learning Research}, pages 27042--27059.
  {PMLR}.

\end{thebibliography}
\bibliographystyle{apalike}

\appendix
\section{Environments} \label{appendix:environments}
\subsection{Meta-World \& Continual-World}
\textbf{Meta-World}. For our experiments, we use the Meta-World benchmark \citep{Yu:22}. Meta-World tasks consist of a Sawyer robotic arm simulated using the MuJoCo physics engine \citep{Todorov:12}. 
The observation space in Meta-World is a 39-dimensional vector.
The action space is 4-dimensional, with all actions in range $[-1, 1]$.
For each task, the reward functions are scaled to have a maximum value of 10 and a minimum of 0. The exact reward-function definitions are provided in \citet{Yu:22}. Episodes last for 200 timesteps for each task. For all our experiments on Meta-World, we report success rates and mean rewards obtained. 

\textbf{Difference between Meta-World v1 and v2 environments.} The Continual-World benchmark was published using v1 of Meta-World. However, in the meantime v2 has been released. While the v1 environments had a 9 dimensional observation space, the observation space is 39-dimensional in v2. Another major change in v2, was the introduction of dense and bounded reward functions. 

\textbf{Continual-World.}
The Continual-World benchmark was proposed by \citet{Wolczyk:21} and is built on top of Meta-World.
Continual-World is a challenging CRL benchmark, consisting of 10 of the 50 tasks contained in Meta-World, denoted as CW10. We use the tasks contained in CW10 as fine-tuning set and select the remaining 40 tasks as pre-training set (MT40). The CW10 task sequence is:

\textit{hammer-v2}, \textit{push-wall-v2}, \textit{faucet-close-v2}, \textit{push-back-v2}, \textit{stick-pull-v2}, \textit{stick-pull-v2}, \textit{handle-press-side-v2}, \textit{push-v2}, \textit{shelf-place-v2}, 
\textit{window-close-v2}, and \textit{peg-unplug-side-v2}. 

\subsection{DeepMind Control Suite (DMControl)}
The DMControl Suite benchmark \citep{Tassa:18} consists of 30 simulated continuous control tasks built on the MuJoCo physics engine. DMControl encompasses a wide variety of robot morphologies, ranging from a simple cartpole to a complex humanoid. Therefore, the observation and action spaces vary across environments. The action spaces vary between 1 (\texttt{cartpole}) and 21 continuous dimensions (\texttt{humanoid}), where each action dimension is bounded by $[-1, 1]$. Agents can be trained based on continuous vector-based representations (state-based) or directly from pixels (pixel-based). In this work, we focus on the state-based representations, which vary between 3 (\texttt{pendulum}) and 67 continuous dimensions (\texttt{humanoid}). As described in Section \ref{sec:experiments}, we select 16 of the 30 DMControl tasks and assign 10 to the pre-training set (DMC10) and 6 to the fine-tuning set (DMC6) (similar to \citet{Hafner:19}). We list the dimensionalities of their state and action spaces in Table \ref{tab:dmc10-dmc6}.

The \textbf{DMC10} tasks include: 

\textit{cartpole-balance}, \textit{finger-turn\_easy}, \textit{finger-turn\_hard}, \textit{fish-upright}, \textit{hopper-stand}, \textit{pendulum-swingup}, \textit{point\_mass-easy}, \textit{reacher-hard}, \textit{walker-run}, \textit{walker-stand}

Furthermore, the \textbf{DMC6} tasks include: 

\textit{ball\_in\_cup-catch}, \textit{cartpole-swingup}, \textit{cheetah-run}, \textit{finger-spin}, \textit{reacher-easy}, \textit{walker-walk}

For all environments in DMControl, episodes last for 1000 timesteps and the maximum achievable return is 1000. In addition to the mean reward obtained, we report data-normalized scores, as suggested by \citet{Fu:20}. For each environment, the scores are normalized by the average scores achieved by the expert agent used to collect the dataset, and the performance attained by a random agent, $\texttt{normalized\_score} = \frac{\texttt{score} - \texttt{random\_score}}{\texttt{expert\_score} - \texttt{random\_score}}$. Therefore, the normalized scores are roughly between 0 and 1. The expert and random scores used for score normalization are available in the accompanying source code\footnote{Available at: \url{https://github.com/ml-jku/L2M}}. 

\section{Methods}\label{appendix:methods}

\textbf{Fine-tuning.} In full fine-tuning, the entire pre-trained model is trained on the new task. We also try other common variations thereof, such as fine-tuning the action head, the last layer, or both.

\textbf{Adapters.} Adapters are a parameter-efficient fine-tuning approach proposed by \citet{Houlsby:19}. An adapter layer consists of a down-projection, a non-linearity and an up-projection along with a skip connection. Each Transformer block is interleaved with two Adapter modules. During training, only the Adapters (and optionally LayerNorms) are updated, while the remaining parameters remain frozen. This significantly reduces the number of trainable parameters, while preserving the ability to adapt to new tasks. \citet{Houlsby:19} adapt a BERT \citep{Devlin:18}  model to 26 different text classification tasks, while training only roughly 4\% of the parameters and attaining performance close to full fine-tuning.

\textbf{(IA$\text{)}^3$: Infused Adapter by Inhibiting and Amplifying Inner Activations.} Similar to LoRA, (IA$\text{)}^3$ \citep{Liu:22} modulates the inner activation flow of the Transformer layers and can be applied to every component of the pre-trained model. Instead of low-rank modulation matrices as used in LoRA, (IA$\text{)}^3$ performs elementwise multiplication with learnable modulation vectors. Thus, (IA$\text{)}^3$ typically incurs less additional parameters compared to other methods. 

\textbf{Prompt-tuning.} \citet{Lester:21} learn soft prompts to enable a frozen pre-trained GPT-3 \citep{Brown:20} model to generalize to downstream tasks.
In prompt-tuning, the learnable prompts are prepended to the input sequence, while the rest of the pre-trained model is kept frozen.

\textbf{Prefix-tuning.} Prefix-tuning \citep{Xiang:21} is another PBT-approach. Similar to prompt-tuning, prefix-tuning prepends learnable vectors (i.e., soft-prompts) to the input. However, unlike in prompt-tuning, the learnable prompts are prepended to the keys and values of the attention head input. Therefore, prefix-tuning also incurs more parameters than prompt-tuning, but typically results in better performance. 

\textbf{P-tuning v2.} \citet{Liu:21a} proposed P-tuning v2 as a successor of prefix-tuning. Instead of prepending continuous prompts only on the input layer, P-tuning v2 prepends them for every layer of the pre-trained model. This simple modification results in more parameters, but better performance. In particular, P-tuning v2 matches the performance of full fine-tuning on a number of NLP benchmarks and is effective across model sizes (330M to 10B parameters). 

\textbf{EWC and L2.} EWC \citep{Kirkpatrick:17} is an established technique for continual learning based on regularization. EWC helps to prevent forgetting of previous tasks when learning a new task by protecting parameters that are important for previously learned tasks. EWC uses the Fisher information matrix as a regularization term, which measures the sensitivity of each parameter with respect to each task and, thus, indicates which parameters require protection. L2 is used as a baseline in the original EWC publication \citep{Kirkpatrick:17} and utilizes the L2 penalty to protect previously learned weights from being changed. 

\section{Multi-Domain Decision Transformer (MDDT)}\label{appendix:mddt}
In this section, we describe our Multi-Domain Decision Transformer (MDDT) architecture in more detail.
We pre-train a MDDT on the 50 collected pre-training datasets simultaneously (Section \ref{sec:exp-pretrain}). Then we fine-tune the pre-trained model to 16 new tasks from two domains. 
To achieve this, we extend the DT architecture proposed by \citet{Chen:21} to handle inputs from multiple domains with varying state and/or action spaces.

\textbf{Unified state-space.} 
While all Meta-World tasks share a 39-dimensional state-space, the dimensionalities of the state-spaces in DMControl vary between 3 and 67 continuous values (See Appendix \ref{appendix:environments}). This variation reflects the diverse robot morphologies across DMControl tasks. 
To deal with these heterogeneous state spaces, we construct a unified state-space that spans all dimension of DMControl and Meta-World tasks. While the dimensionalities of state-spaces differ between tasks, they may share common observation types, such as orientation, current velocity, or arm positions of the robot. In total, DMControl encompasses 24 different observation types. Our constructed state-space unifies all of these observation-type attributes, amounting to 204 continuous dimensions in total. For example, the first 13 dimensions correspond to the current orientations of the robot and the last dimension corresponds to the current height. Dimensions that are unused by a certain task are zero-padded.
Thus, the constructed state-space covers all environments considered in this work. Finally, as in the original DT publication \citep{Chen:21}, the states are embedded with a linear layer before serving as input to the MDDT.

We choose this scheme for handling varying state-spaces instead of discretisation followed by padding, as was used in prior work \citep{Reed:22}. Discretisation of states in combination with padding drastically increases the context length and, thus, the computational requirements. For example, when tokenising a trajectory of five timesteps in Meta-World, the context would expand to 195 tokens (39 dimensions * 5 timesteps) for states alone. In contrast, our scheme results in only 5 state tokens. 

\textbf{Action tokenisation \& autoregressive prediction.} 
To handle action spaces of varying dimensionalities, we tokenize each action dimension and autoregressively predict action tokens, similar to \citet{Reed:22} and \citet{Brohan:22}. 
In the environments we consider, the action-spaces vary between 1 and 6 dimensions and all actions are bounded by [-1, 1]. We use a min-max tokenisation to discretise every continuous action dimensions into 64 bins and then pad to the maximum action dimension. Thus, we obtain 6 discrete action tokens from every action. During evaluation, we autoregressively predict the action tokens until the dimensionality of action space in the particular task is reached, and then execute the composed action. Note that autoregressive prediction of actions make the prediction task harder, compared to simply predicting the ground truth action, as is commonly done for continuous control tasks.

\textbf{Context representation.} 
Similar to \citet{Lee:22}, the policy $\pi_{\theta}$ with trainable parameters $\theta$, predicts the next action $a_t$ from a given context of the $K$ most recent states, RTGs, actions and rewards.
States and actions are pre-processed as described earlier, and then embedded using separate linear layers. Similarly, we embed the continuous scalars for RTGs and rewards using linear layers. Unlike \citet{Lee:22}, we do not discretise rewards or RTGs. However, similar to \citet{Chen:21}, we scale the rewards and RTGs such that all RTGs lie roughly within the range [0, 10]. To achieve this, we use separate reward scales of 200 and 100, for Meta-World and DMControl, respectively. Thus, each timestep is made up of 9 embedded tokens (1$\times$ state, 1$\times$ RTG, 6 $\times$ action, 1 $\times$ reward). We use a context-length of 5 timesteps for all our experiments, and therefore, the final sequence length expands to 45 embedded tokens (5 timesteps * 9 tokens per timestep). 

\textbf{Positional encodings.} Once the sequence is embedded, we add positional information to every embedded token. 
As \citet{Chen:21}, we learn an embedding for each timestep and add this embedding to every embedded token (i.e., states/action/reward/rtg tokens) in a timestep.
This provides the DT with vital information about the absolute position within the current episode.
This is important since the choice for an action may differ at the beginning vs. at the end of an episode. 
We also explored other positional encodings in preliminary experiments, but found learned time-step based embeddings to work best. 

\textbf{Training \& Evaluation.} 
In line with prior work \citep{Chen:21, Lee:22}, we train the DT via return-conditioned upside-down RL using a cross-entropy loss to predict next actions. 
During evaluation, we set the target return to the maximum observed return in the respective dataset, instead of utilizing the expert-action inference mechanism proposed by \citet{Lee:22}. However, we also found that constant proxies per domain (e.g., 2000  and 1000) work well.

These simple yet effective modifications enable processing observations and actions that originate from different environments with varying state and/or action spaces at training and test time. However, a few notable \textbf{limitations} remain: 
\begin{itemize}
    \item The constructed \textbf{state-space} is specific to the environments we consider in our work and, thus, not directly transferable to other environments. Therefore, a new unified state-space may have to be constructed (same procedure), when using a different environment. 
    \item The \textbf{discretised actions} are padded to the maximum action dimension. Therefore, an input sequence for an environment with a smaller than the maximum action dimension, may contain multiple redundant tokens. However, this design decision simplified our implementation considerably.
    \item During evaluation, we use \textbf{autoregressive prediction} of action dimensions. This makes the prediction task considerably harder, when compared to predicting all dimensions at once. Thus, this can result in lower overall performance (see Appendix \ref{fig:singledomain-metaworld}). 
\end{itemize}

\begin{figure}
    \centering
    \includegraphics[width=0.49\textwidth]{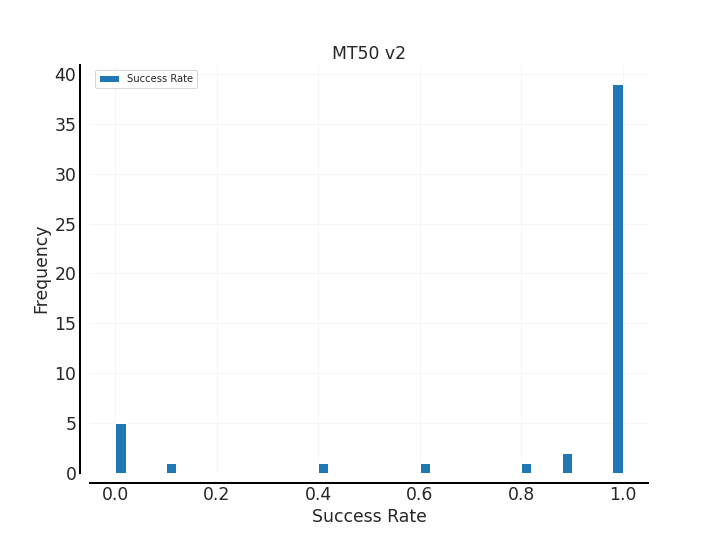}
    \caption{Histogram for success rates over all tasks. The majority of all MT50 tasks can be solved successfully at the end of training by single-task SAC, while on a minority of tasks the agents fails to achieve successful completion. }
    \label{fig:mt50_dist}
\end{figure}

\begin{table}[]
    \centering
    \caption{Performance scores for data collection on MT40.}
    \begin{tabular}{l c c c c}
    \toprule
                            \textbf{Task} & $|\mathcal{S}|$ & $|\mathcal{A}|$ & \textbf{Success Rate} &      \textbf{Reward} \\
    \midrule
                     assembly-v2 &  39 & 4 &  0.0 &     1206.9  \\
                   basketball-v2 &  39 & 4 &  0.9 &    1375.95  \\
                  bin-picking-v2 &  39 & 4 &  0.0 &     474.81  \\
                    box-close-v2 &  39 & 4 &  0.0 &     759.15  \\
         button-press-topdown-v2 &  39 & 4 &  1.0 &    1299.24  \\
    button-press-topdown-wall-v2 &  39 & 4 &  1.0 &    1296.16  \\
                 button-press-v2 &  39 & 4 &  1.0 &    1430.44  \\
            button-press-wall-v2 &  39 & 4 &  1.0 &    1508.16  \\
                coffee-button-v2 &  39 & 4 &  1.0 &    1499.17  \\
                  coffee-pull-v2 &  39 & 4 &  1.0 &    1313.88  \\
                  coffee-push-v2 &  39 & 4 &  0.6 &     508.14  \\
                    dial-turn-v2 &  39 & 4 &  0.8 &    1674.29  \\
                  disassemble-v2 &  39 & 4 &  1.0 &    1396.55  \\
                   door-close-v2 &  39 & 4 &  1.0 &     1535.4  \\
                    door-lock-v2 &  39 & 4 &  1.0 &    1712.65  \\
                    door-open-v2 &  39 & 4 &  1.0 &    1544.32  \\
                  door-unlock-v2 &  39 & 4 &  1.0 &    1733.64  \\
                 drawer-close-v2 &  39 & 4 &  1.0 &    1845.92  \\
                  drawer-open-v2 &  39 & 4 &  1.0 &    1710.65  \\
                  faucet-open-v2 &  39 & 4 &  0.9 &    1727.98  \\
                  hand-insert-v2 &  39 & 4 &  1.0 &    1607.17  \\
                 handle-press-v2 &  39 & 4 &  1.0 &    1854.79  \\
             handle-pull-side-v2 &  39 & 4 &  1.0 &    1613.72  \\
                  handle-pull-v2 &  39 & 4 &  1.0 &    1581.75  \\
                   lever-pull-v2 &  39 & 4 &  1.0 &    1449.05  \\
              peg-insert-side-v2 &  39 & 4 &  1.0 &    1545.19  \\
             pick-out-of-hole-v2 &  39 & 4 &  1.0 &    1435.64  \\
                   pick-place-v2 &  39 & 4 &  0.0 &       6.59  \\
              pick-place-wall-v2 &  39 & 4 &  0.1 &     702.59  \\
        plate-slide-back-side-v2 &  39 & 4 &  1.0 &    1766.24  \\
             plate-slide-back-v2 &  39 & 4 &  1.0 &    1773.56  \\
             plate-slide-side-v2 &  39 & 4 &  1.0 &    1663.35  \\
                  plate-slide-v2 &  39 & 4 &  1.0 &    1667.35  \\
                        reach-v2 &  39 & 4 &  1.0 &    1858.99  \\
                   reach-wall-v2 &  39 & 4 &  1.0 &    1831.14  \\
                       soccer-v2 &  39 & 4 &  0.4 &     445.84  \\
                   stick-push-v2 &  39 & 4 &  1.0 &    1470.71  \\
                   sweep-into-v2 &  39 & 4 &  1.0 &    1761.69  \\
                        sweep-v2 &  39 & 4 &  1.0 &    1458.35  \\
                  window-open-v2 &  39 & 4 &  1.0 &    1537.59  \\
                  \midrule
             Average &  - & - & 0.84 ± 0.34 & 1414.62 ± 439.39 \\
    \bottomrule
    \end{tabular}
    \label{tab:mt40}
\end{table}

\begin{table}[]
    \centering
    \caption{Performance scores for data collection on CW10.}
    \begin{tabular}{l c c c c}
    \toprule
                    \textbf{Task} & $|\mathcal{S}|$ & $|\mathcal{A}|$ & \textbf{Success Rate} &      \textbf{Reward} \\
    \midrule
         faucet-close-v2 &  39 & 4 &  1.0 &    1768.87 \\
               hammer-v2 &  39 & 4 &  1.0 &    1632.21 \\
    handle-press-side-v2 &  39 & 4 &  1.0 &    1842.17 \\
      peg-unplug-side-v2 &  39 & 4 &  1.0 &    1373.45 \\
            push-back-v2 &  39 & 4 &  1.0 &    1373.32 \\
                 push-v2 &  39 & 4 &  1.0 &    1672.88 \\
            push-wall-v2 &  39 & 4 &  1.0 &    1594.37 \\
          shelf-place-v2 &  39 & 4 &  1.0 &    1376.92 \\
           stick-pull-v2 &  39 & 4 &  1.0 &    1344.29 \\
         window-close-v2 &  39 & 4 &  1.0 &    1426.45 \\
    \midrule
                 Average &  39 & 4 &  1.0 ± 0.0 & 1540.49 ± 184.43 \\
    \bottomrule
    \end{tabular}
    \label{tab:cw10}
\end{table}

\begin{figure}
  \centering
    \subfigure{\includegraphics[width=0.19\textwidth]{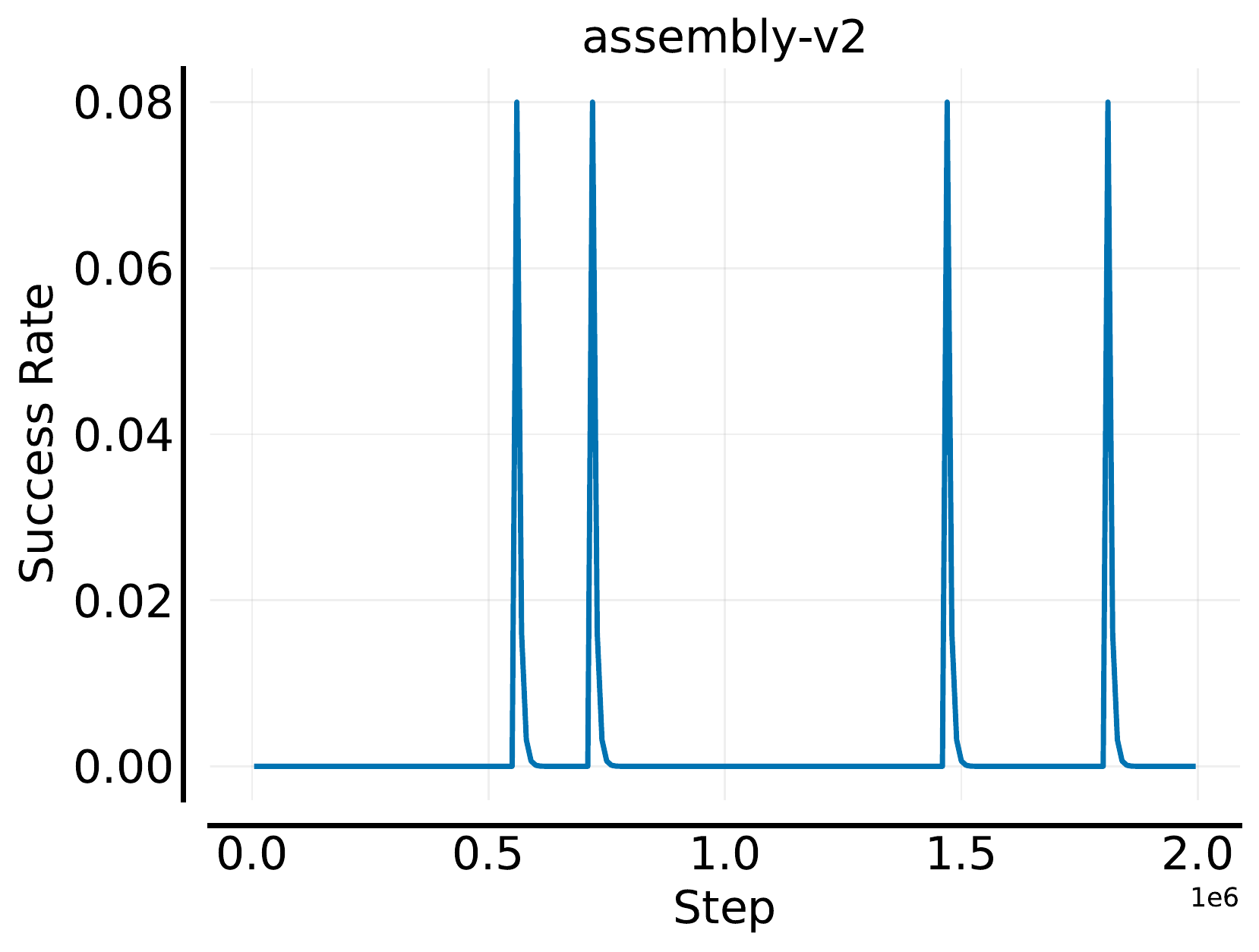}}
    \subfigure{\includegraphics[width=0.19\textwidth]{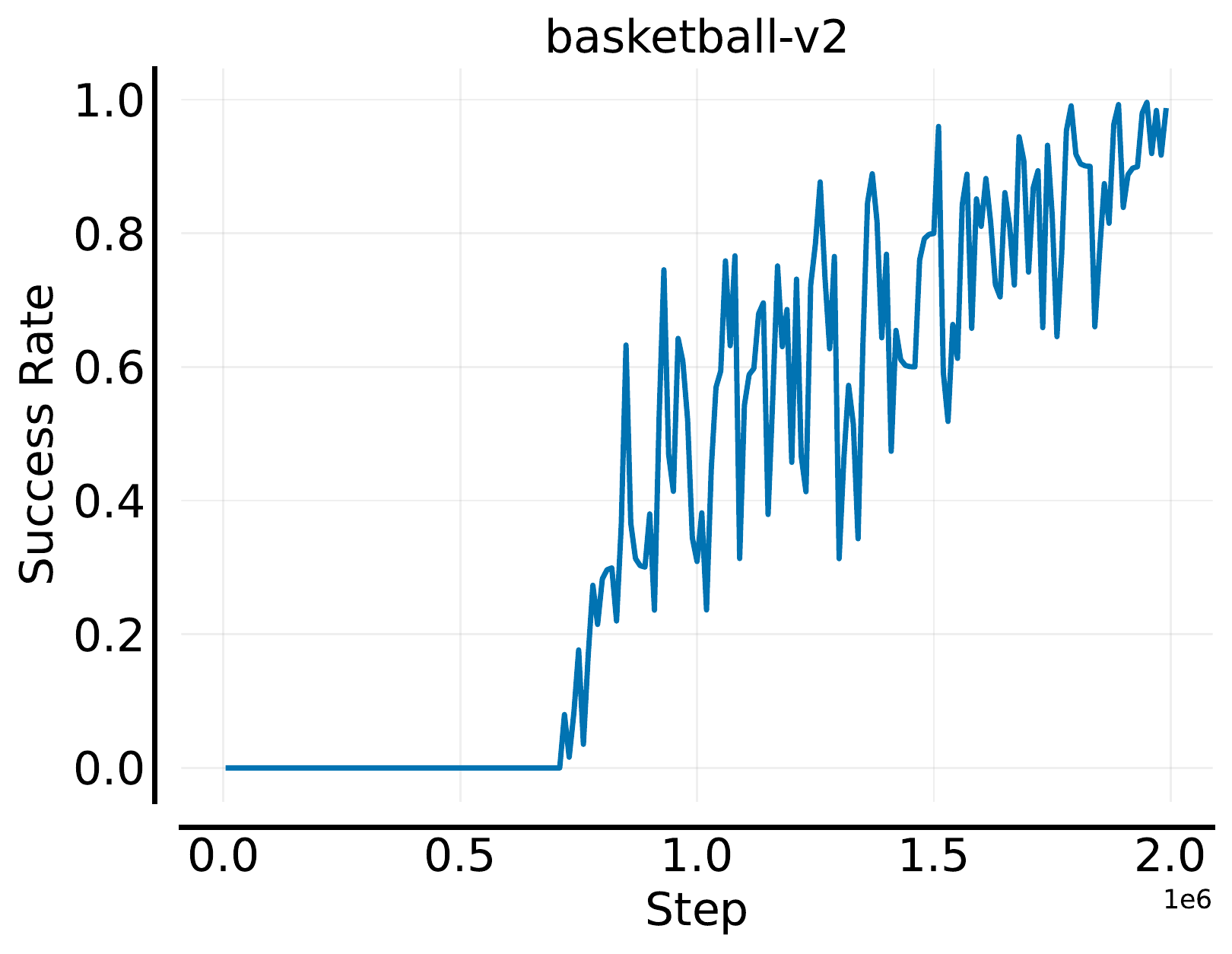}}
    \subfigure{\includegraphics[width=0.19\textwidth]{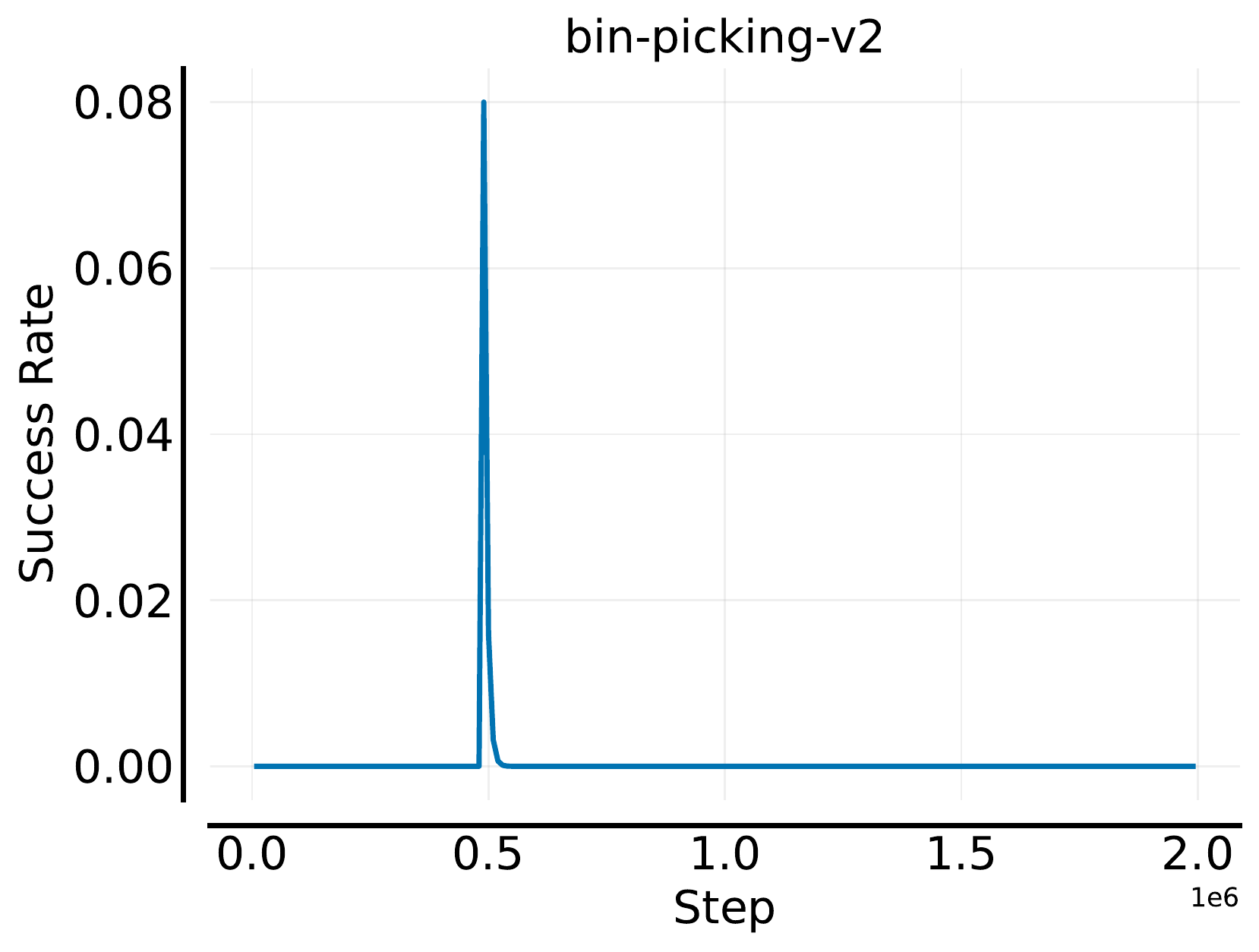}}
    \subfigure{\includegraphics[width=0.19\textwidth]{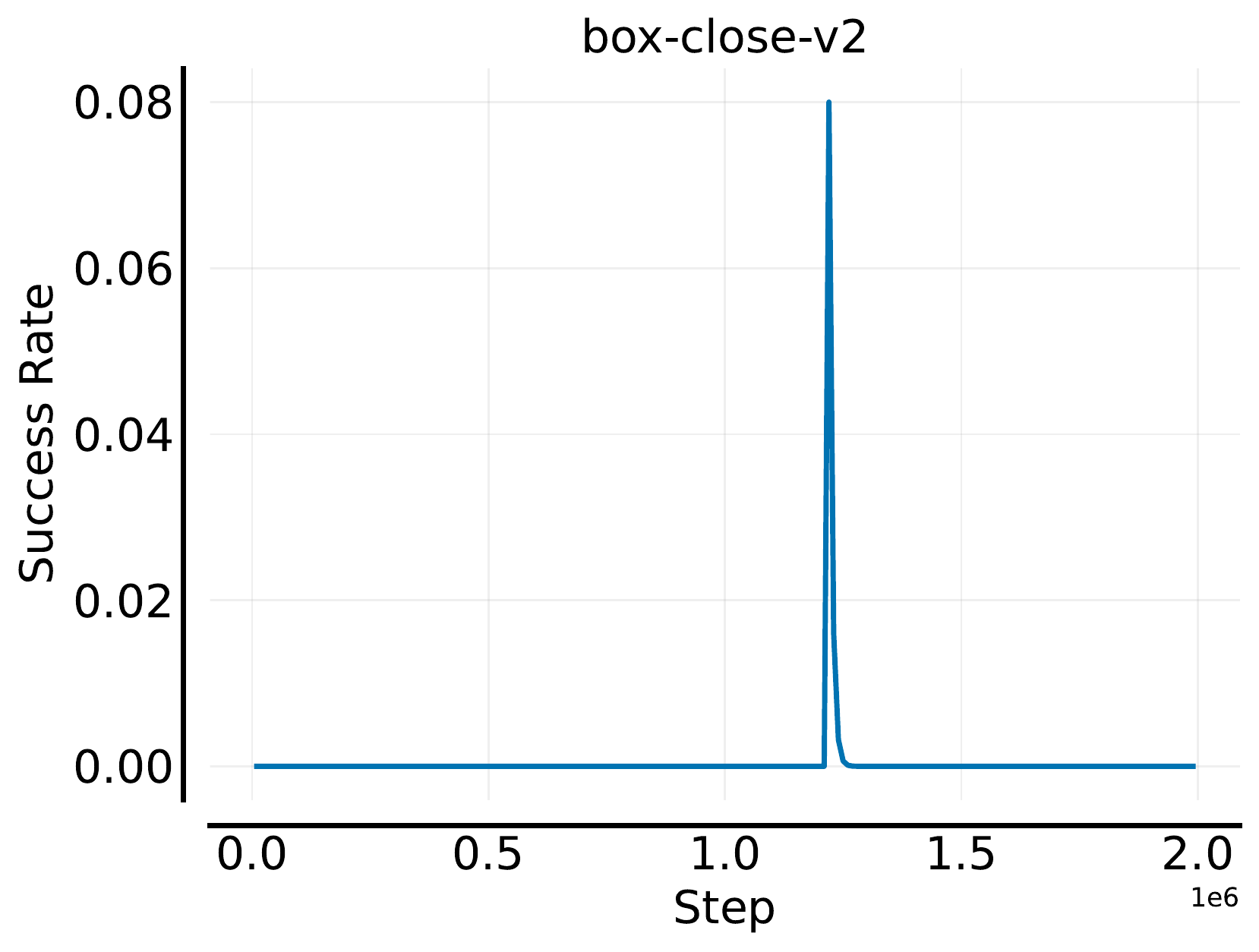}}  \subfigure{\includegraphics[width=0.19\textwidth]{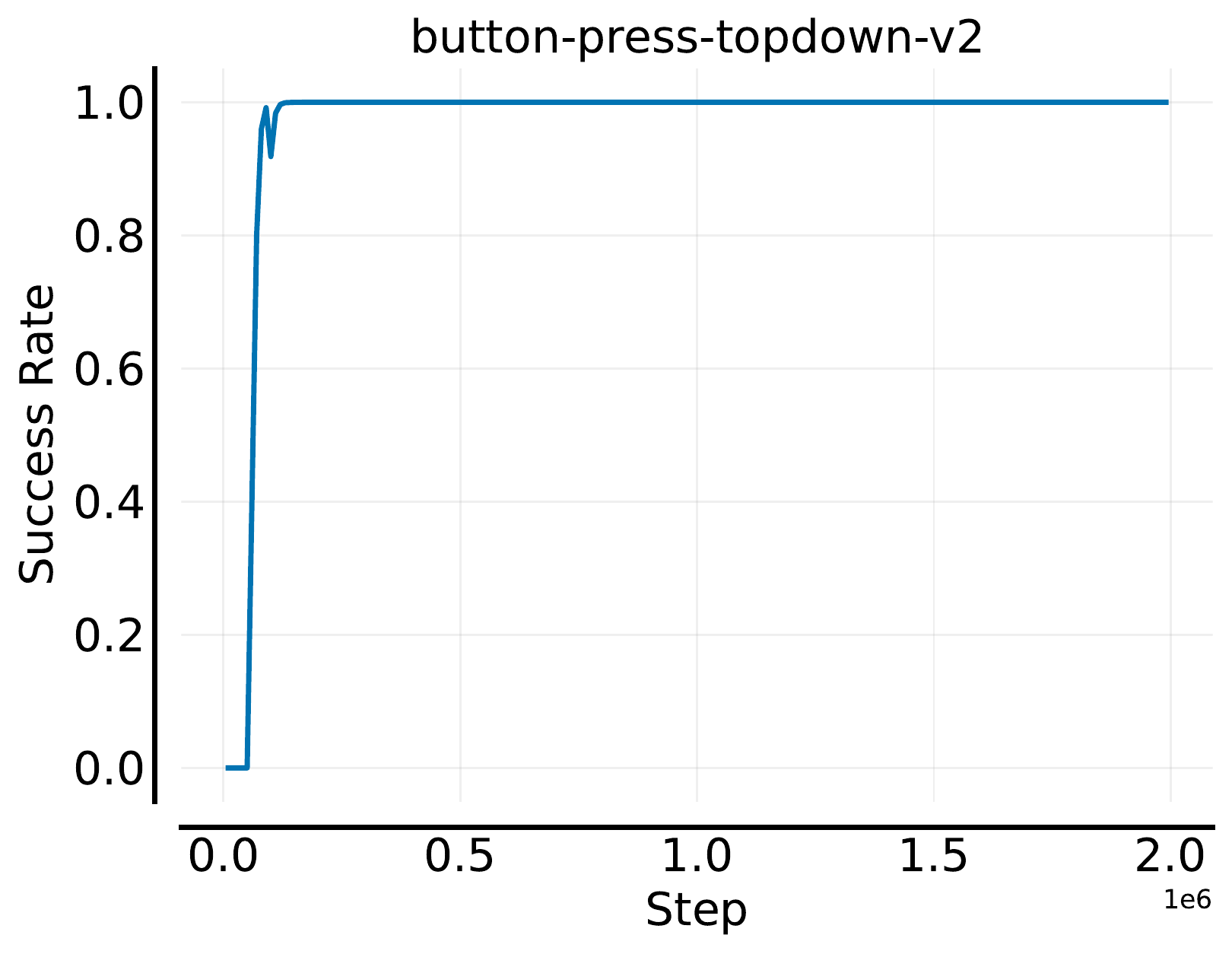}}
    
    \subfigure{\includegraphics[width=0.19\textwidth]{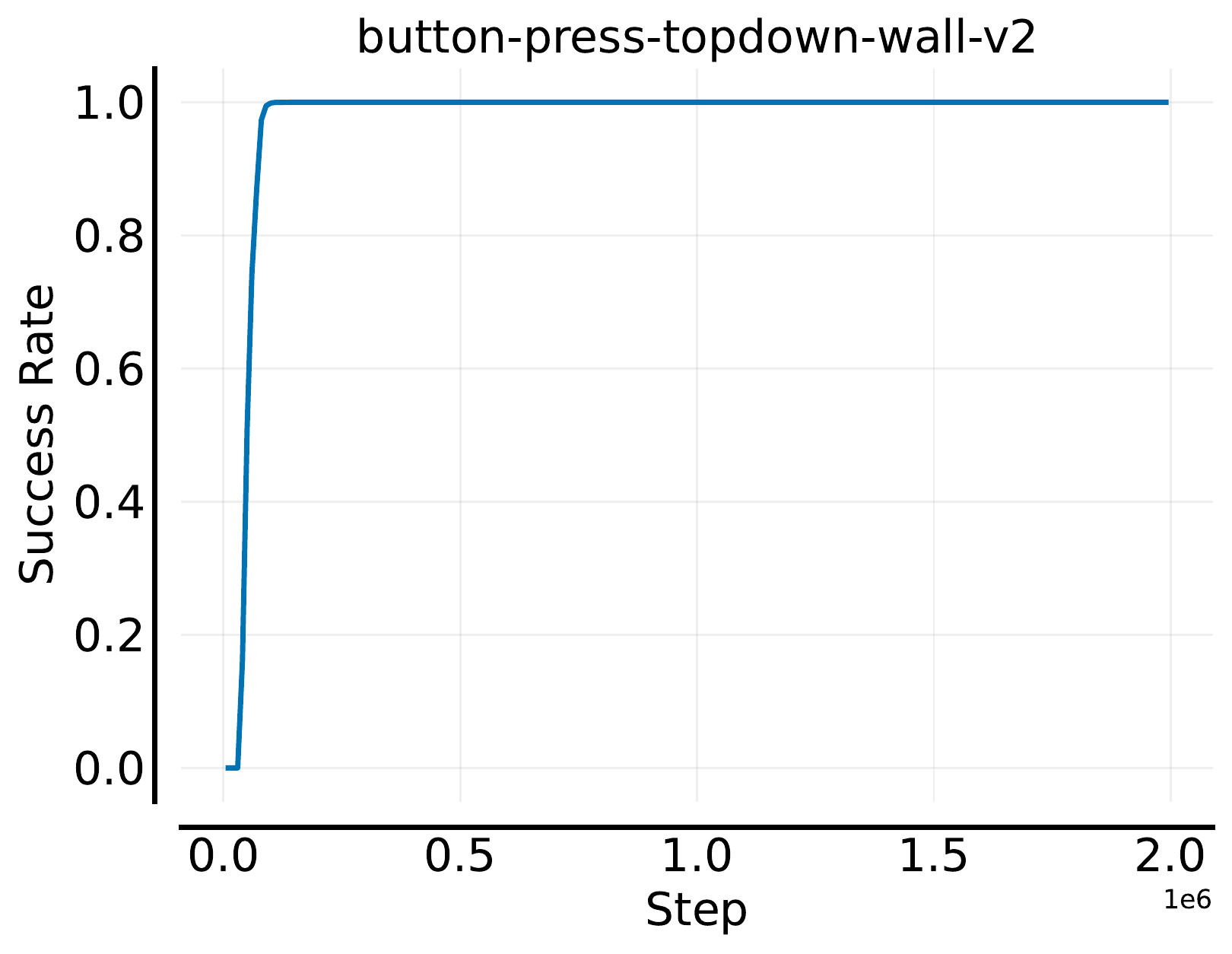}}
    \subfigure{\includegraphics[width=0.19\textwidth]{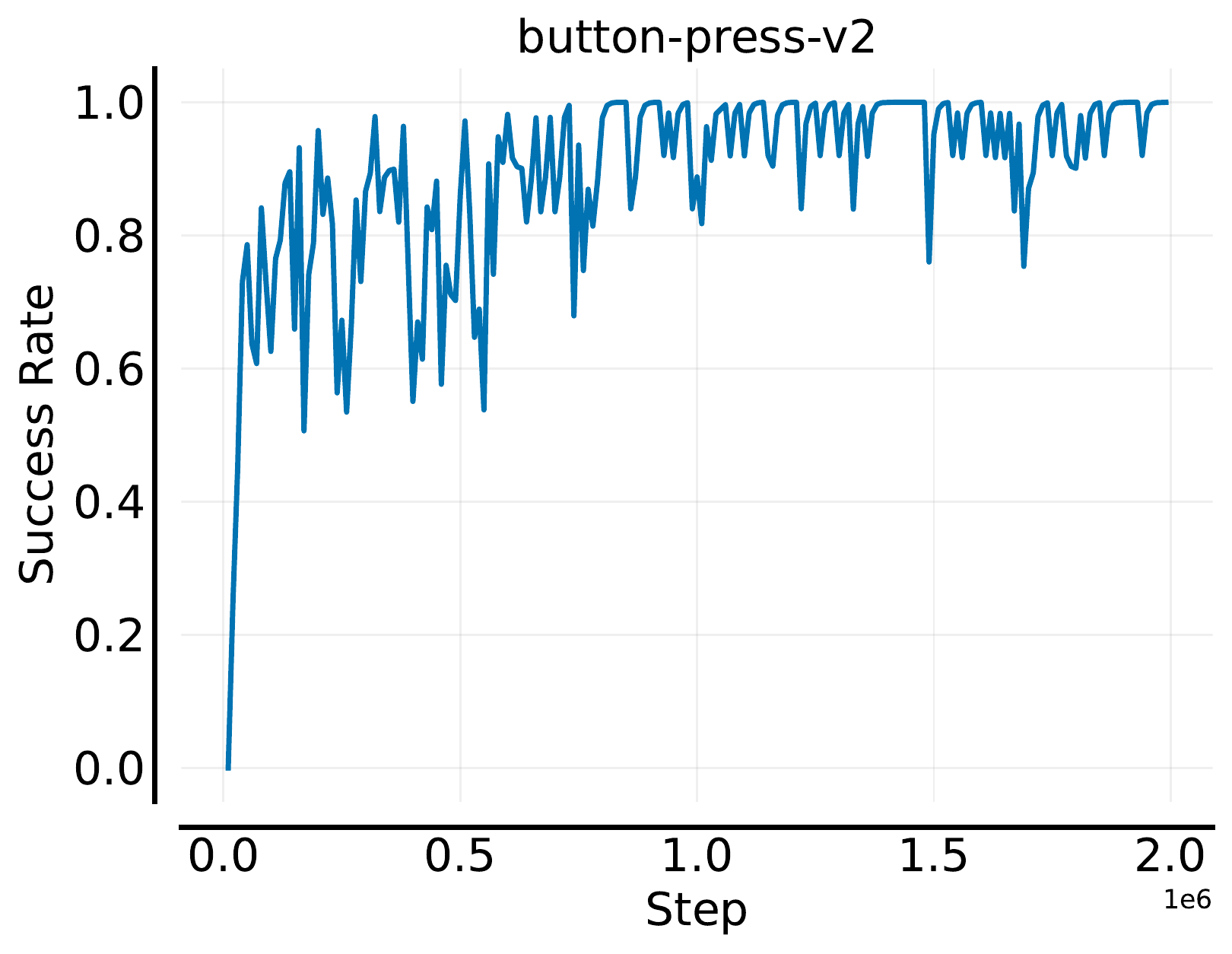}}
    \subfigure{\includegraphics[width=0.19\textwidth]{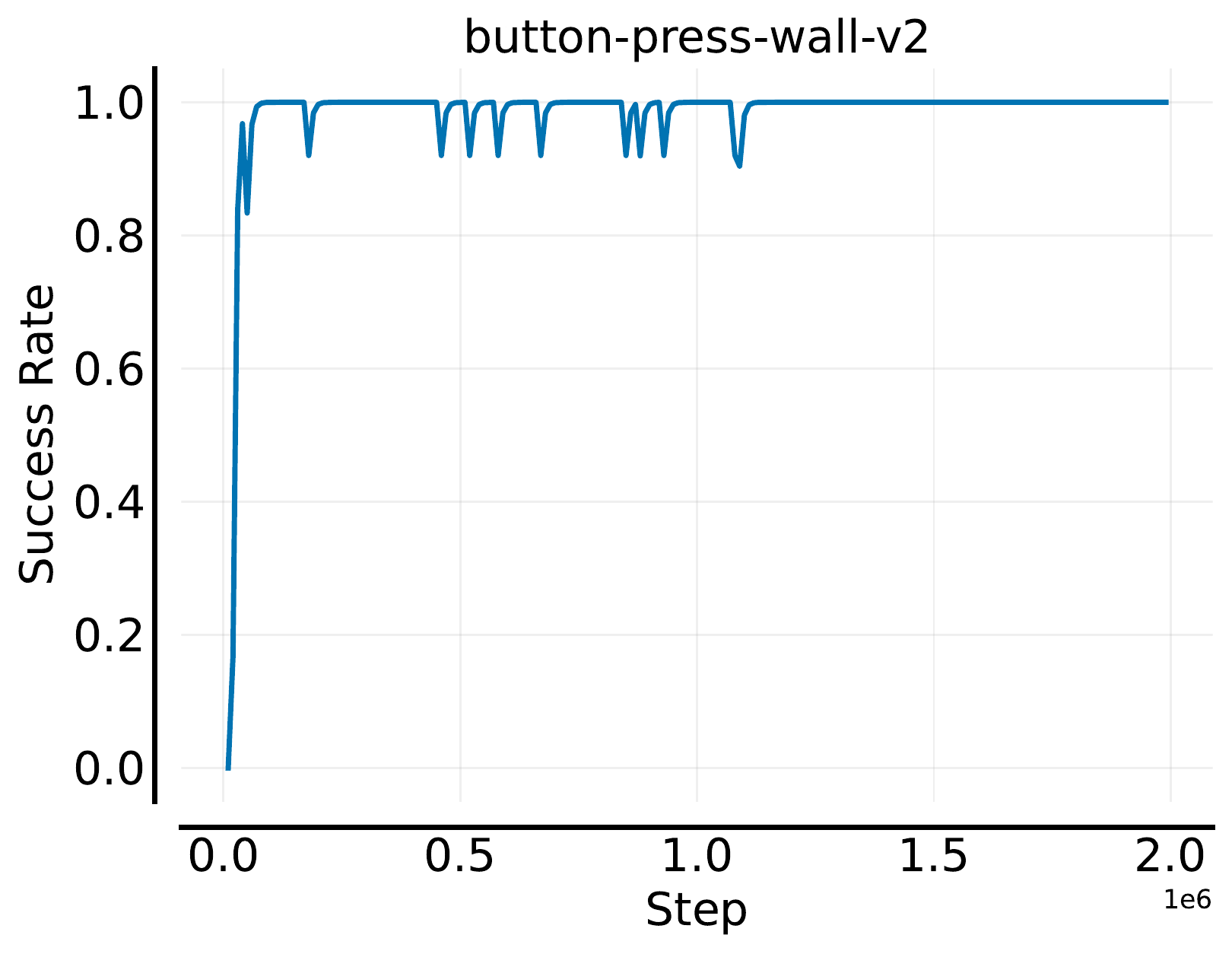}}
    \subfigure{\includegraphics[width=0.19\textwidth]{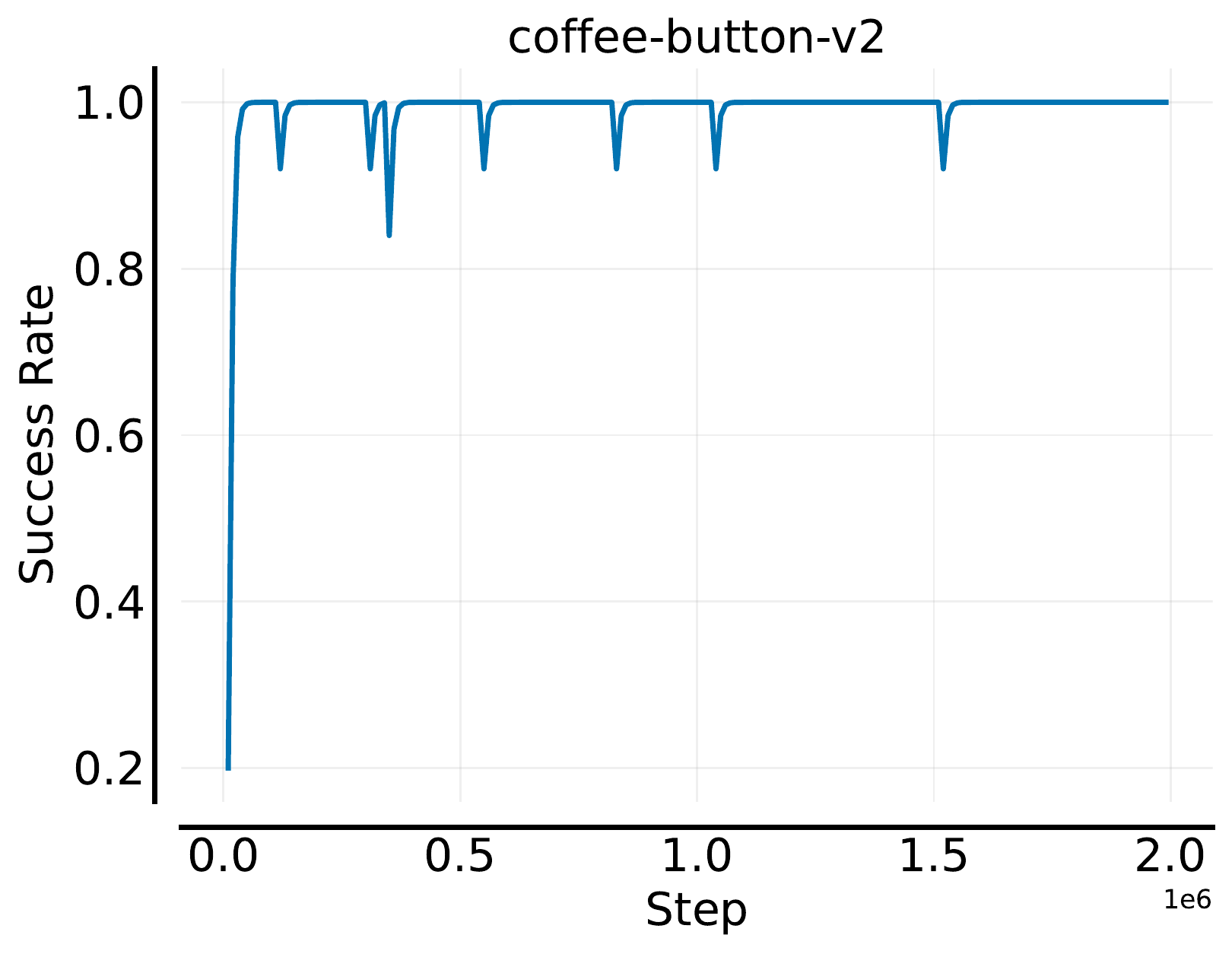}}
    \subfigure{\includegraphics[width=0.19\textwidth]{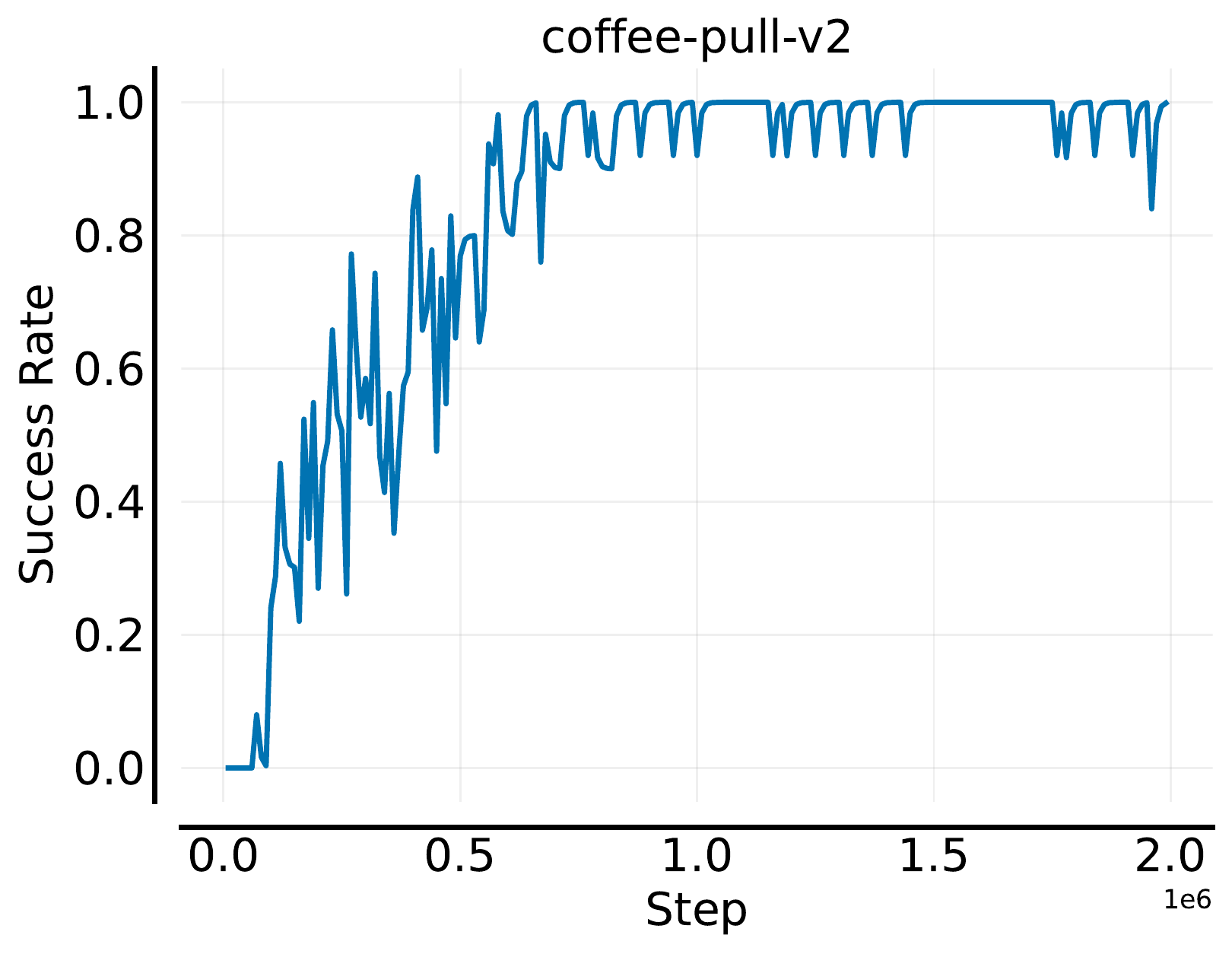}}
    
    \subfigure{\includegraphics[width=0.19\textwidth]{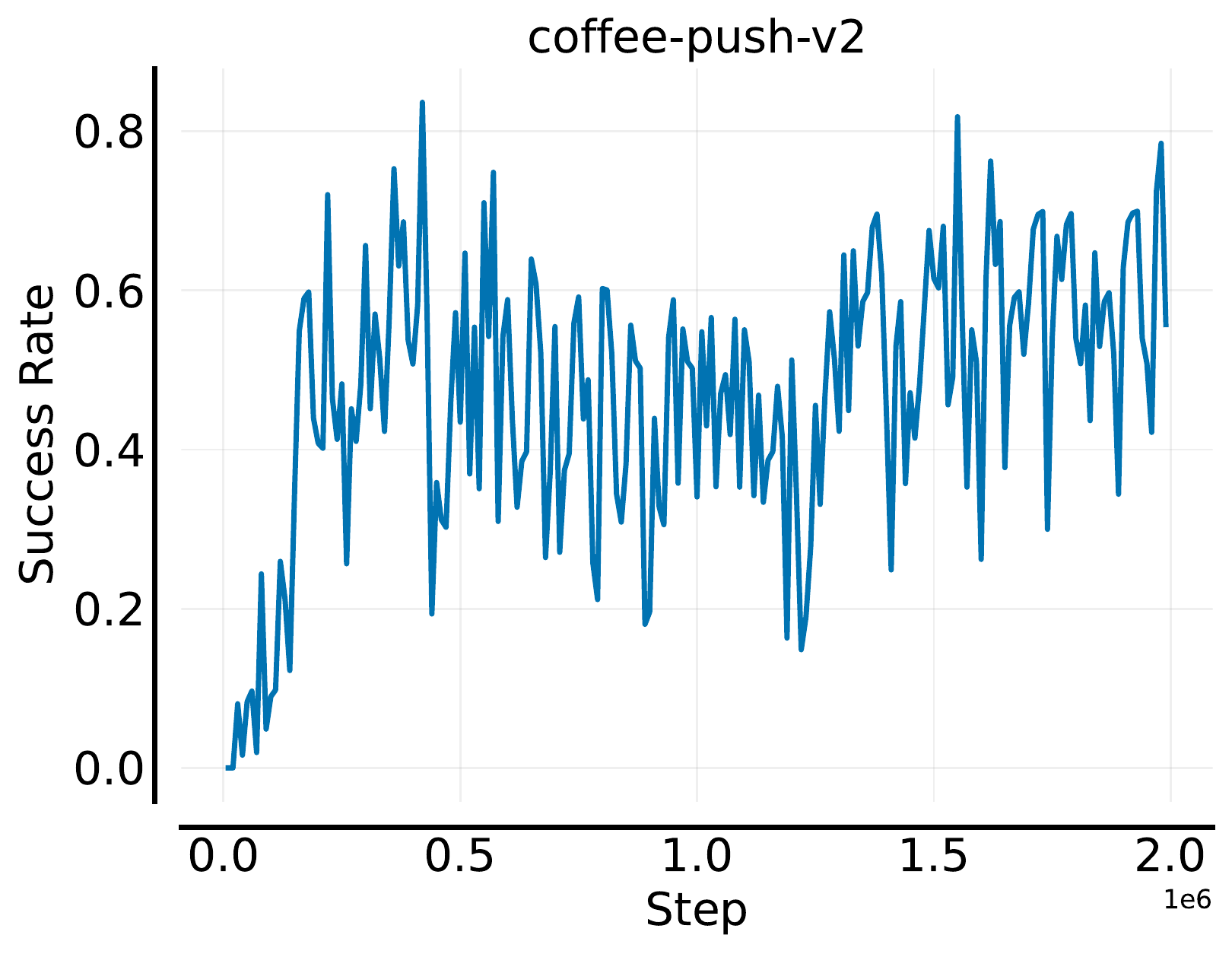}}
    \subfigure{\includegraphics[width=0.19\textwidth]{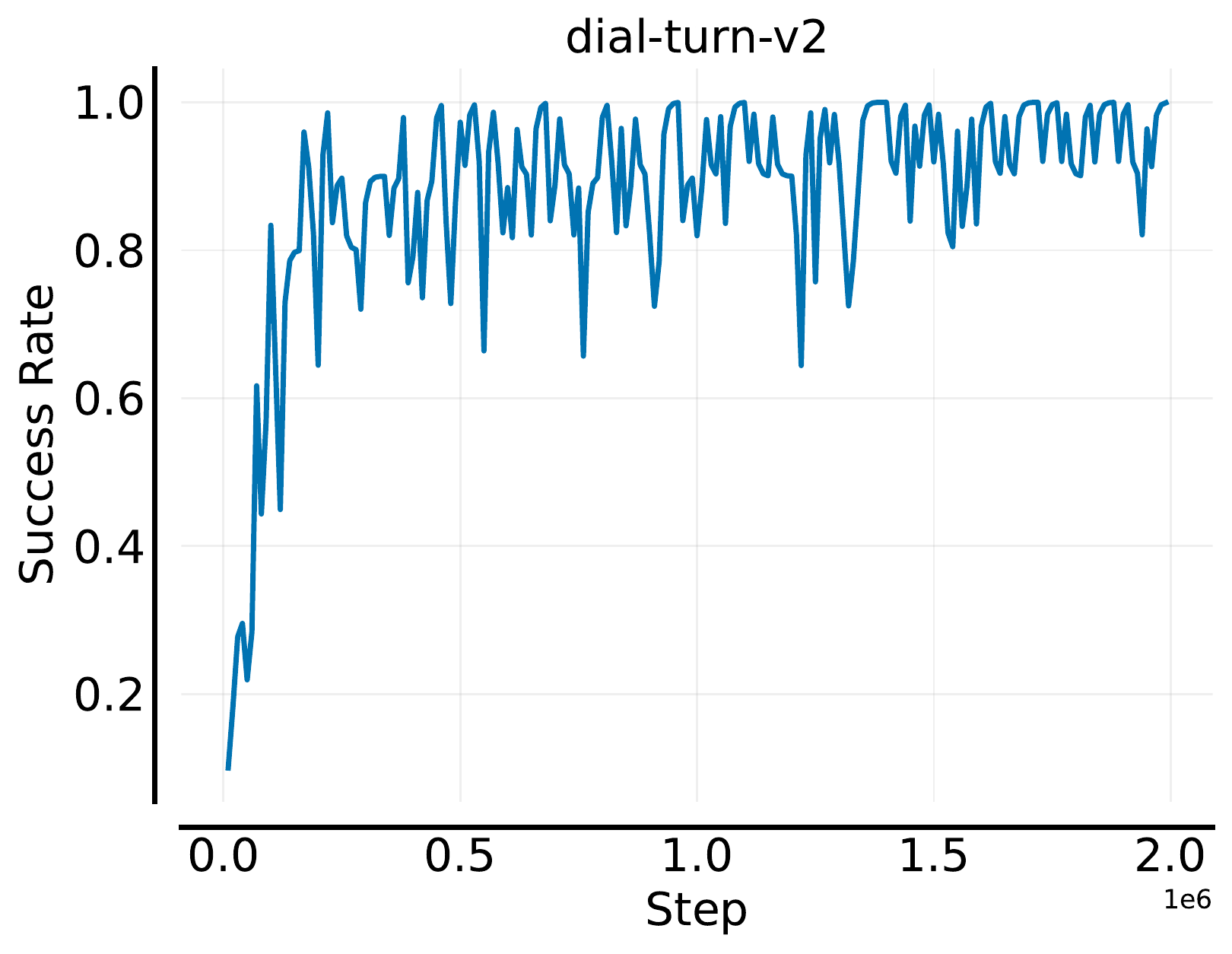}}
    \subfigure{\includegraphics[width=0.19\textwidth]{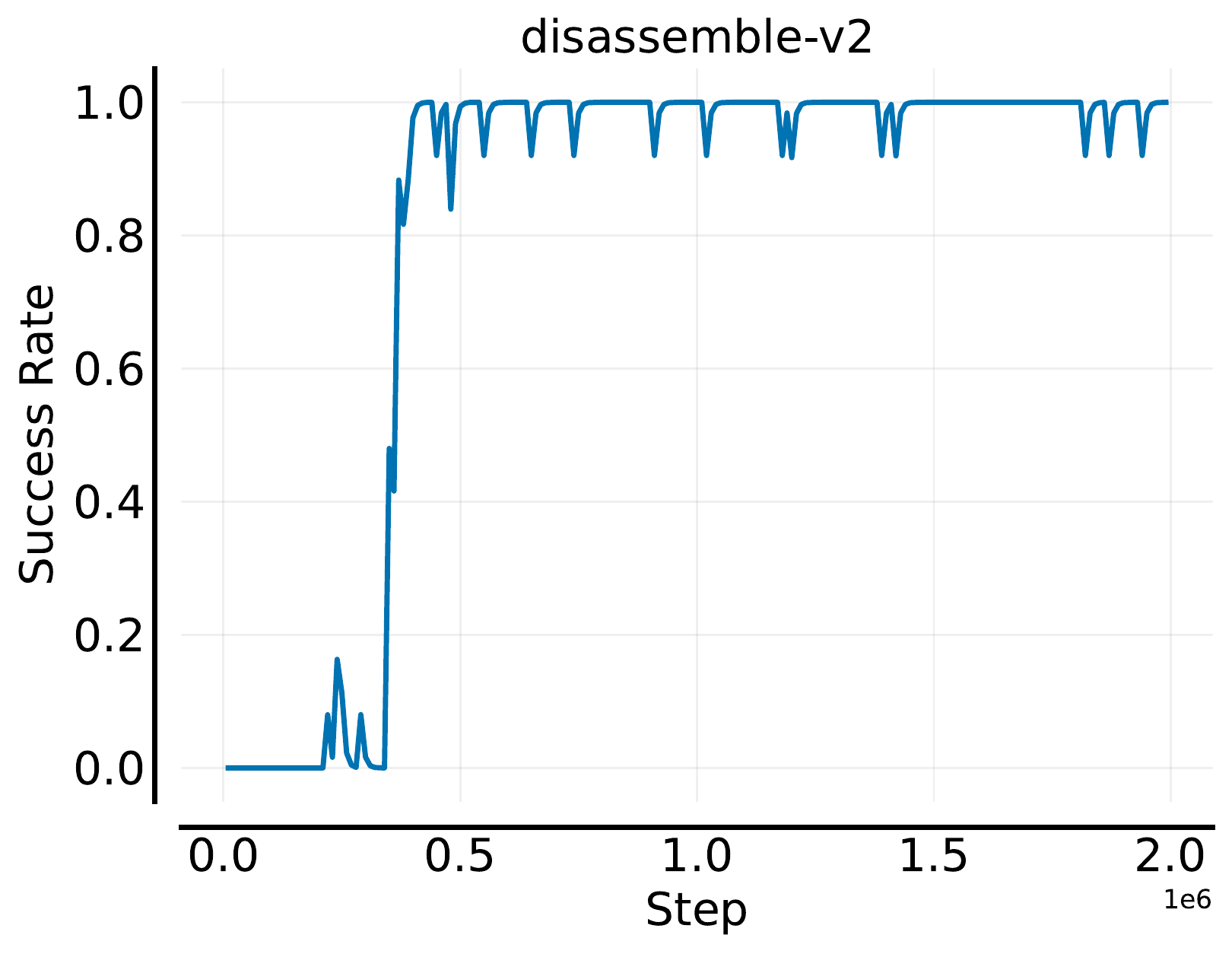}}
    \subfigure{\includegraphics[width=0.19\textwidth]{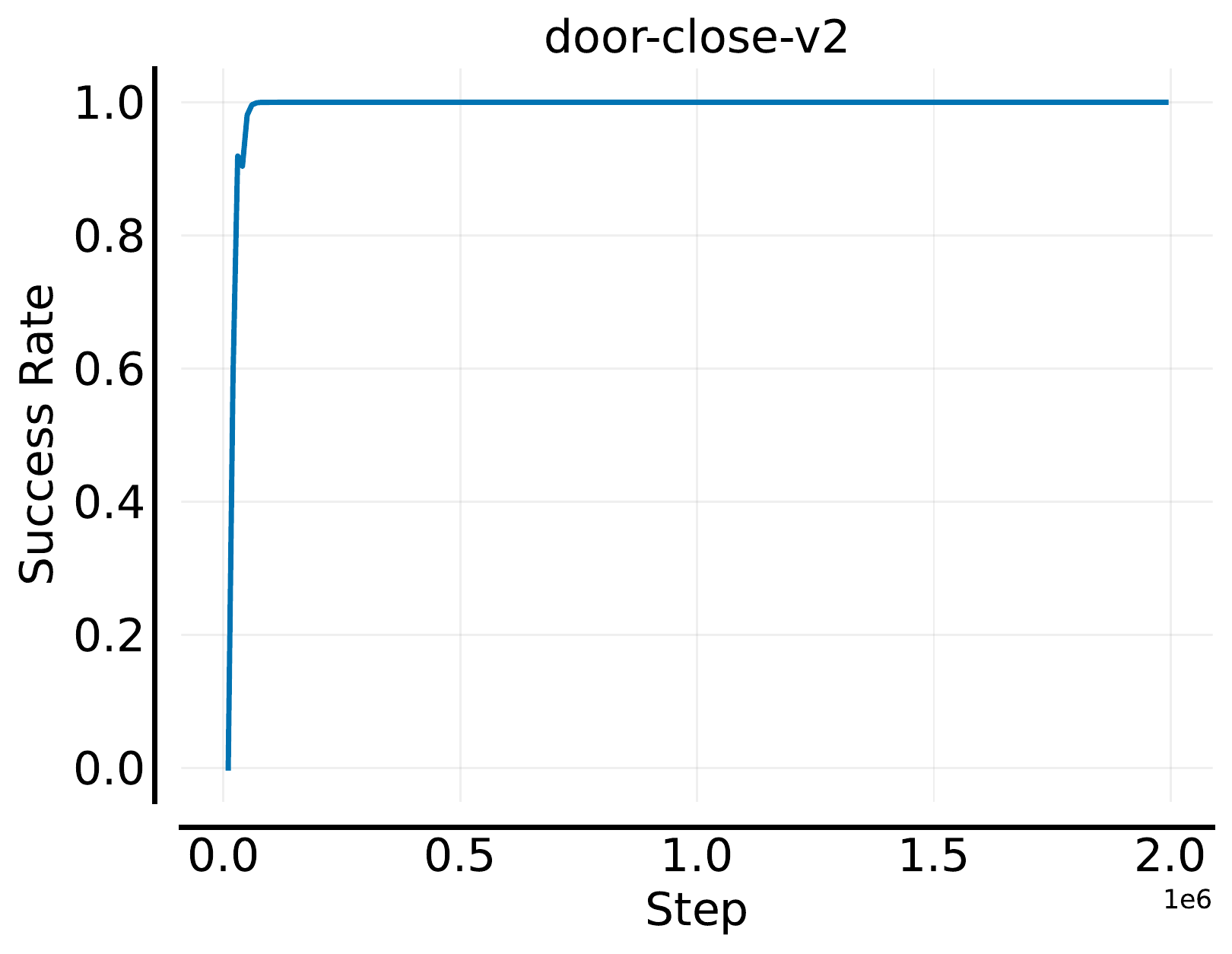}}
    \subfigure{\includegraphics[width=0.19\textwidth]{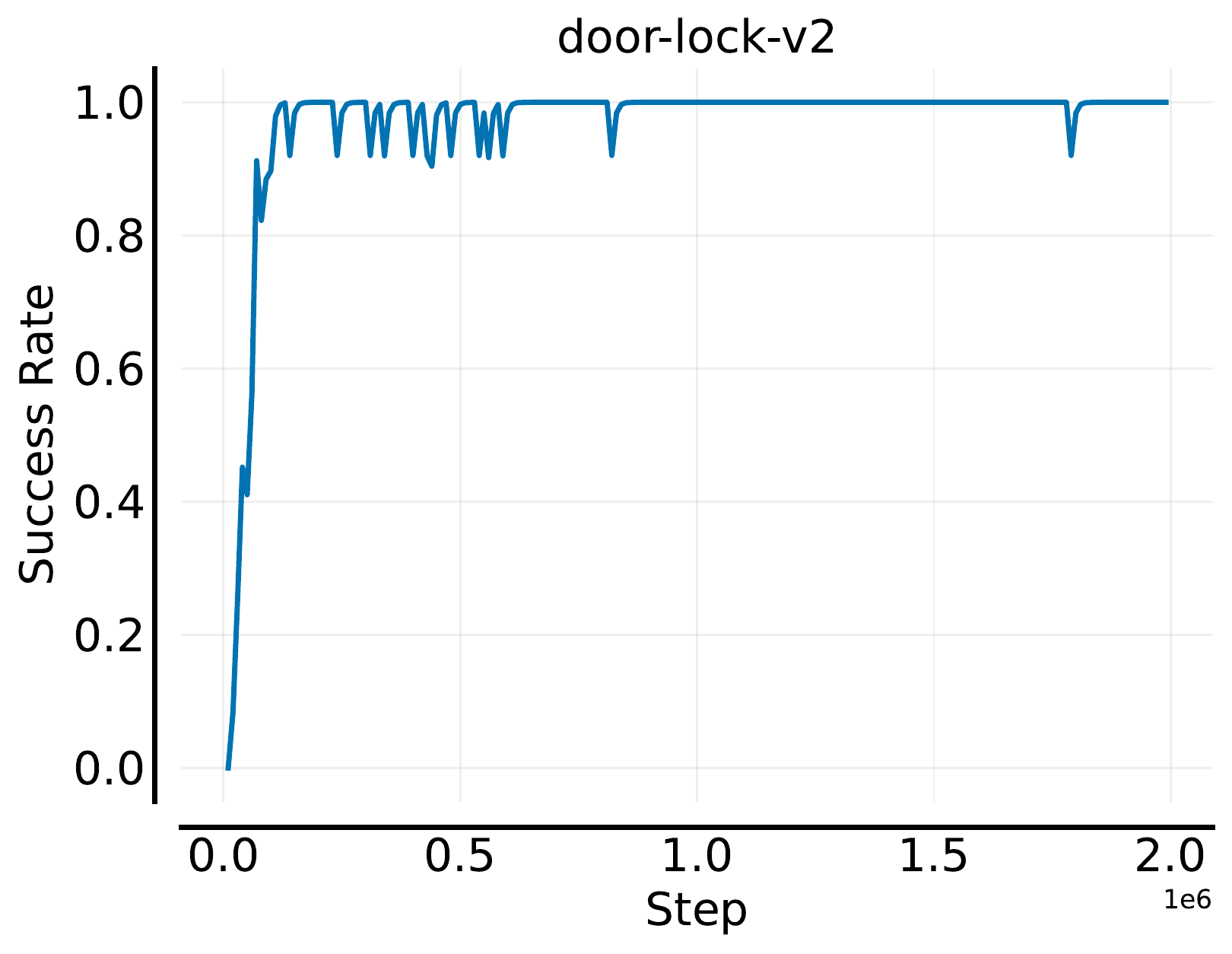}}
    
    \subfigure{\includegraphics[width=0.19\textwidth]{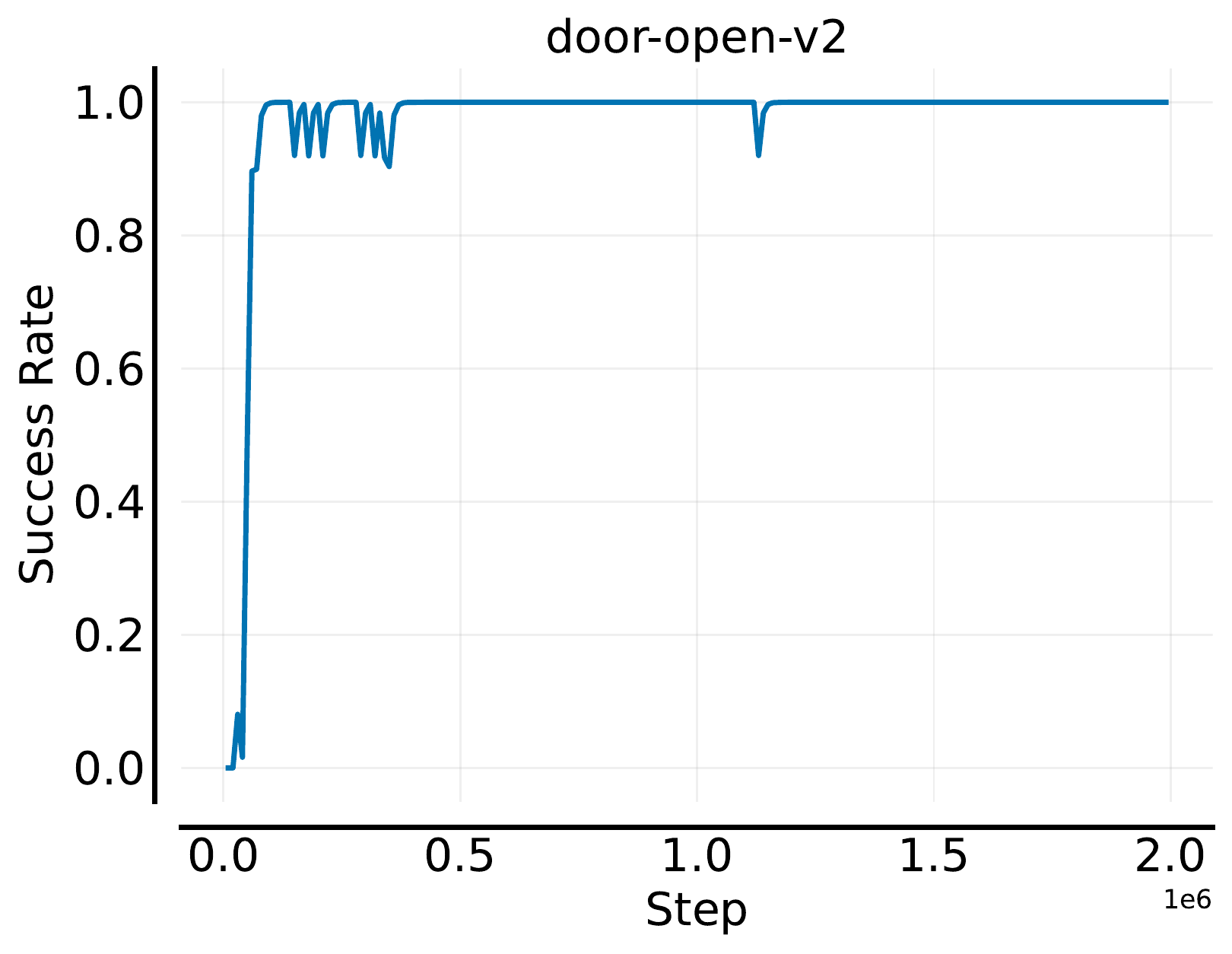}}
    \subfigure{\includegraphics[width=0.19\textwidth]{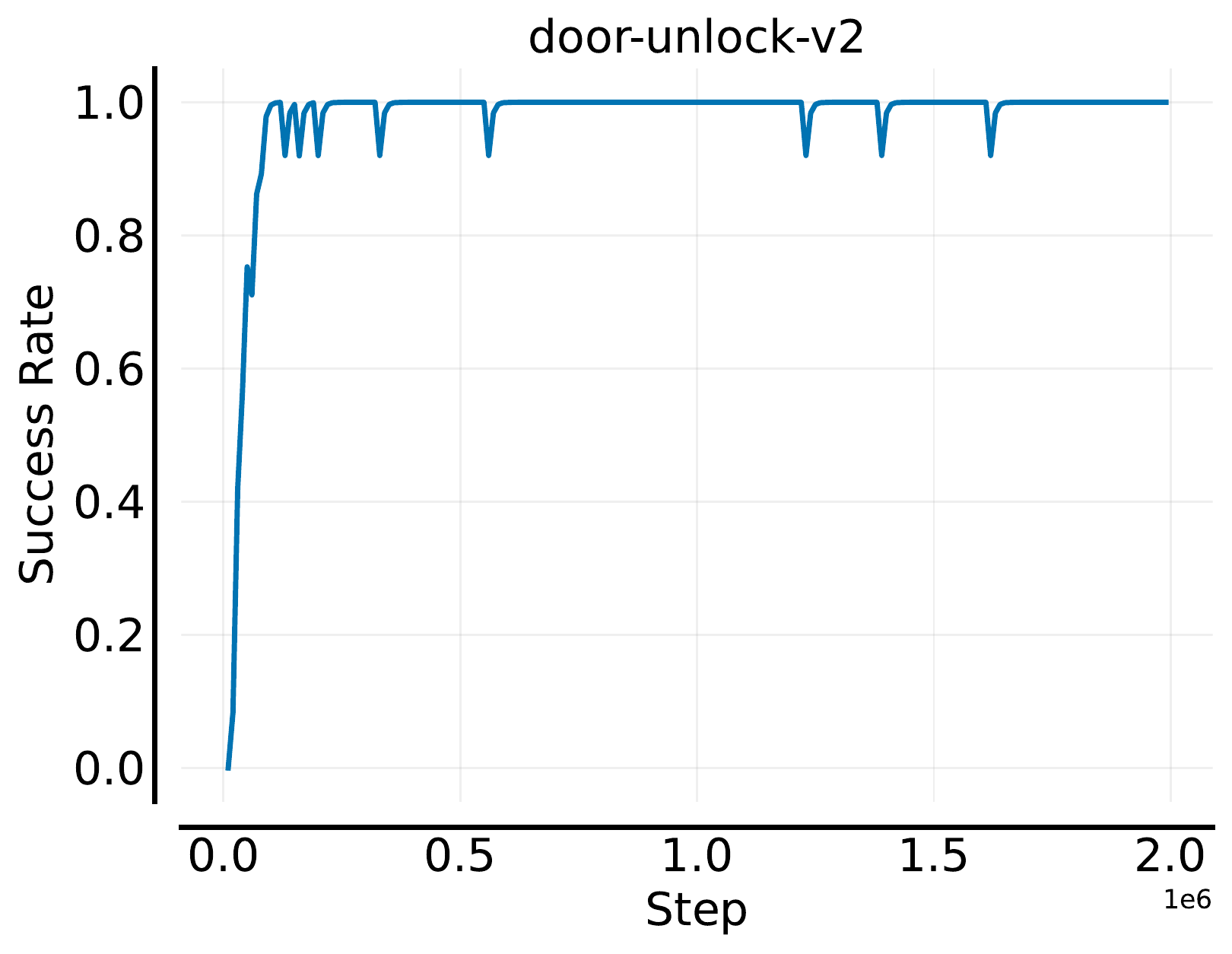}}
    \subfigure{\includegraphics[width=0.19\textwidth]{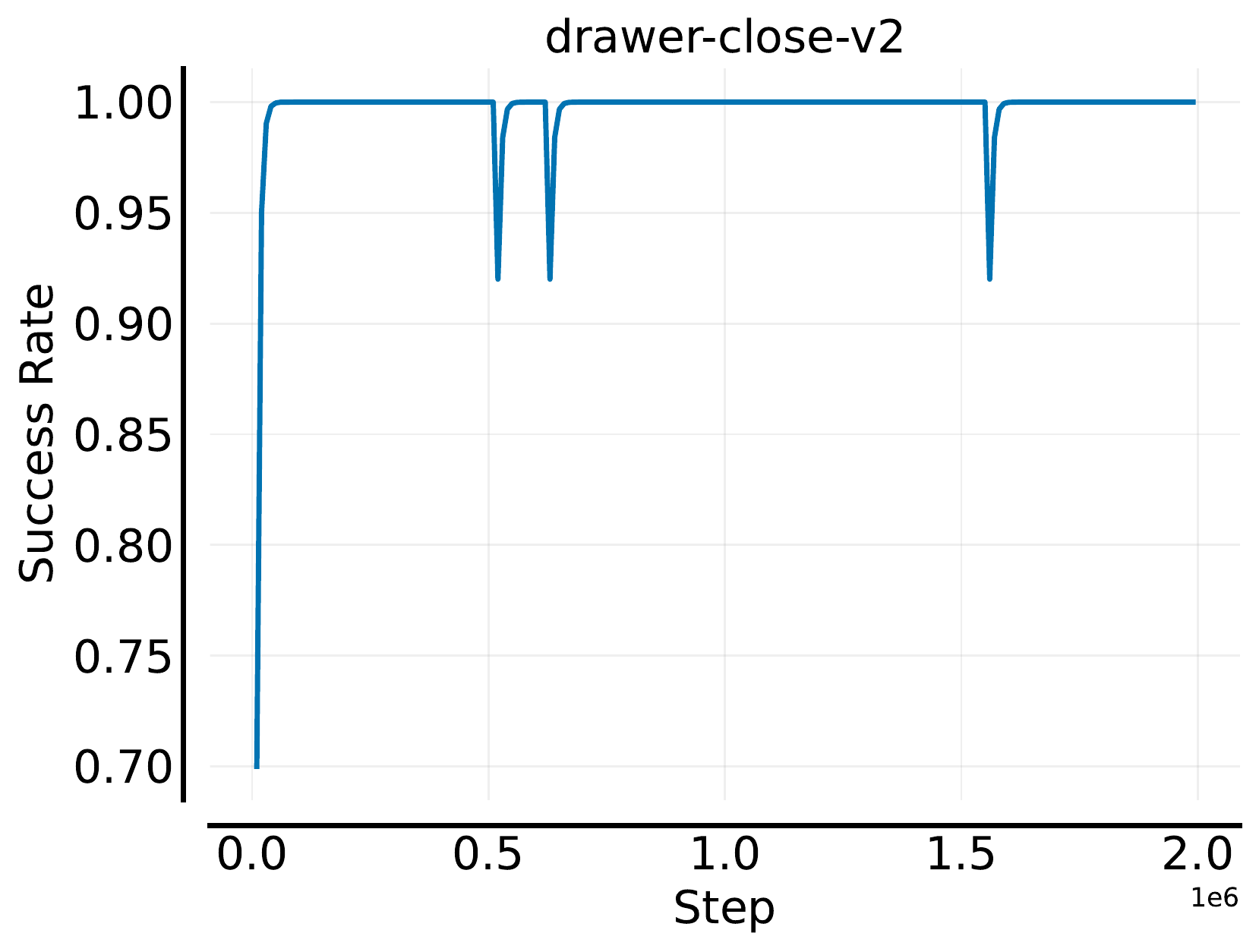}}
    \subfigure{\includegraphics[width=0.19\textwidth]{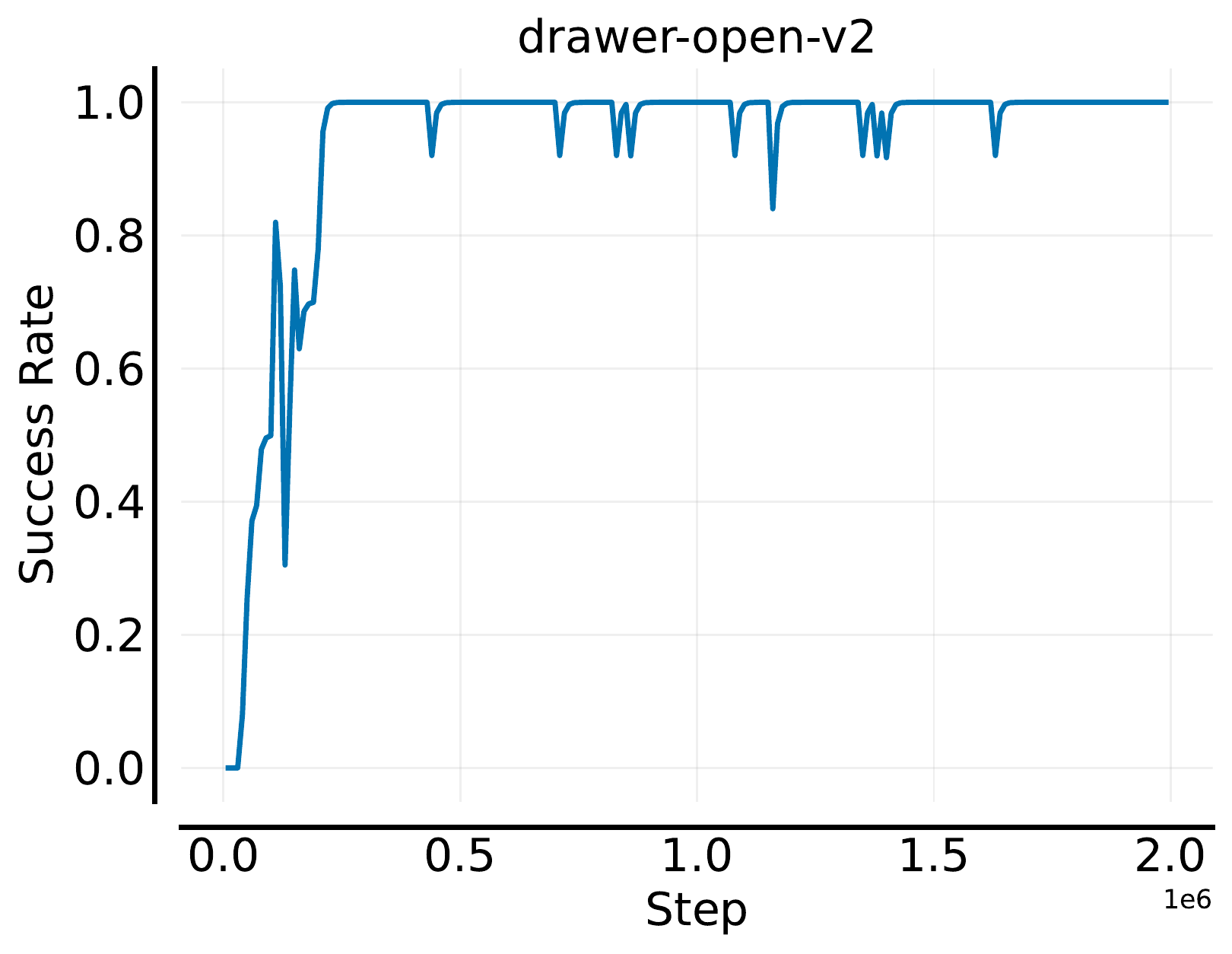}}
    \subfigure{\includegraphics[width=0.19\textwidth]{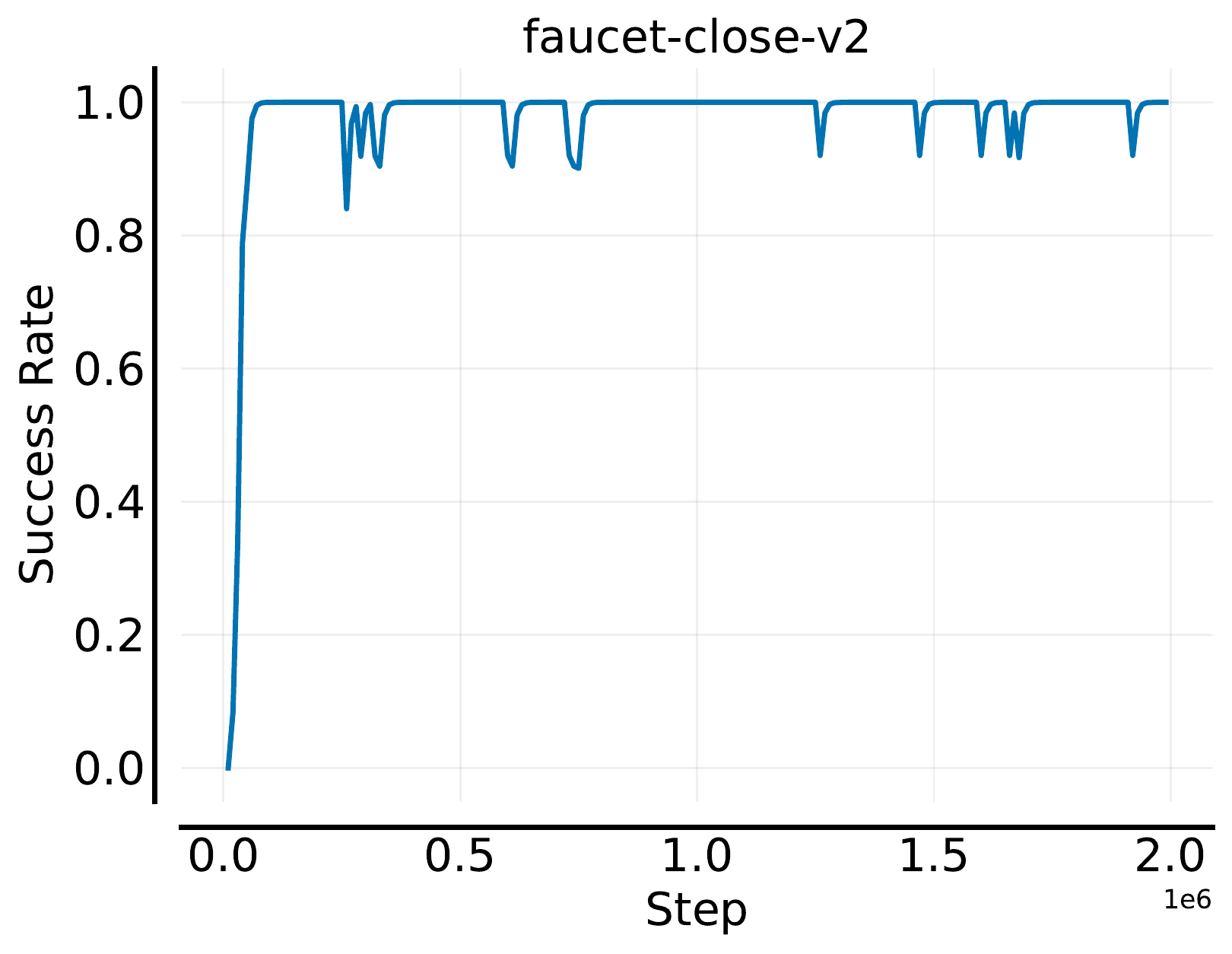}}
    
    \subfigure{\includegraphics[width=0.19\textwidth]{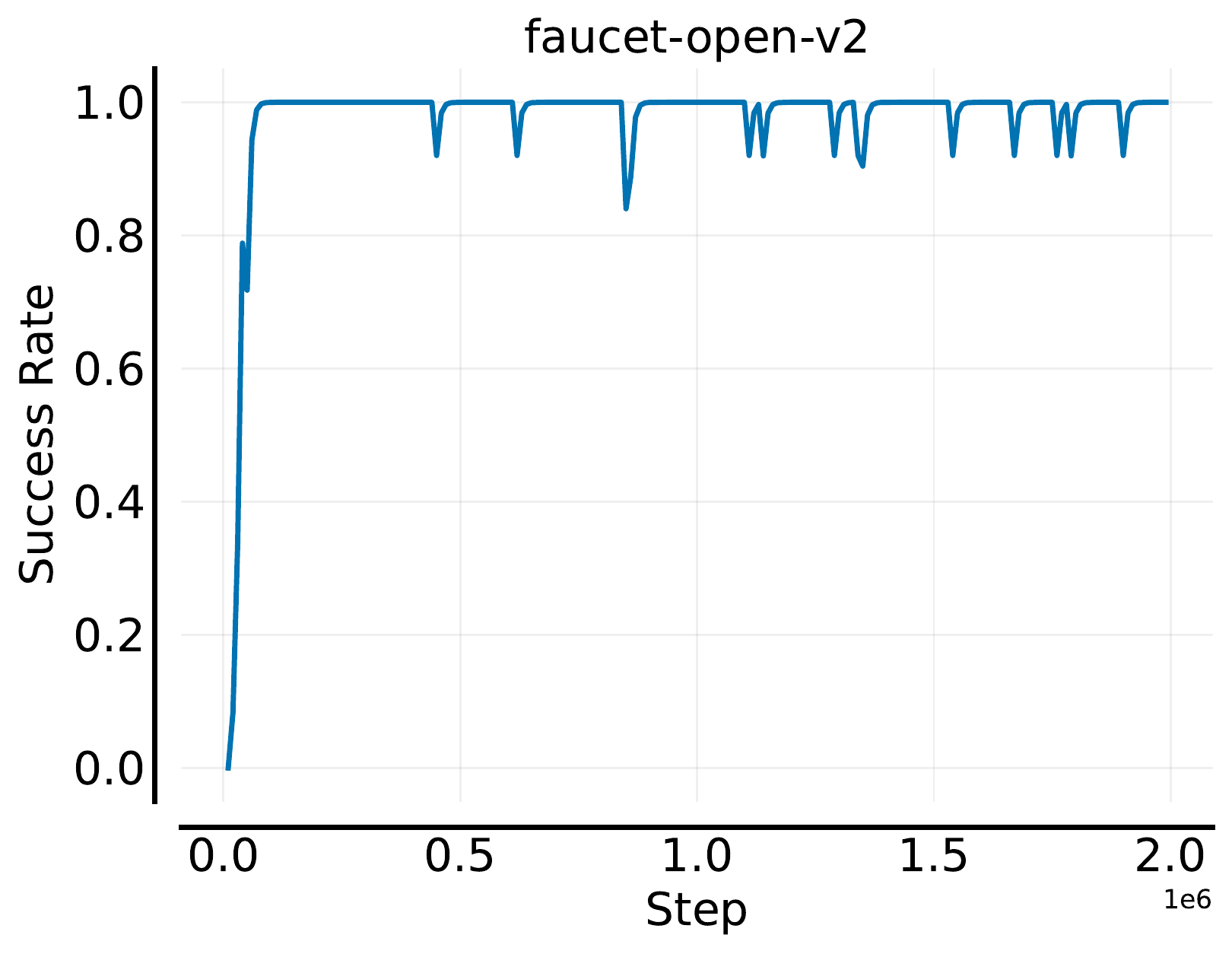}}
    \subfigure{\includegraphics[width=0.19\textwidth]{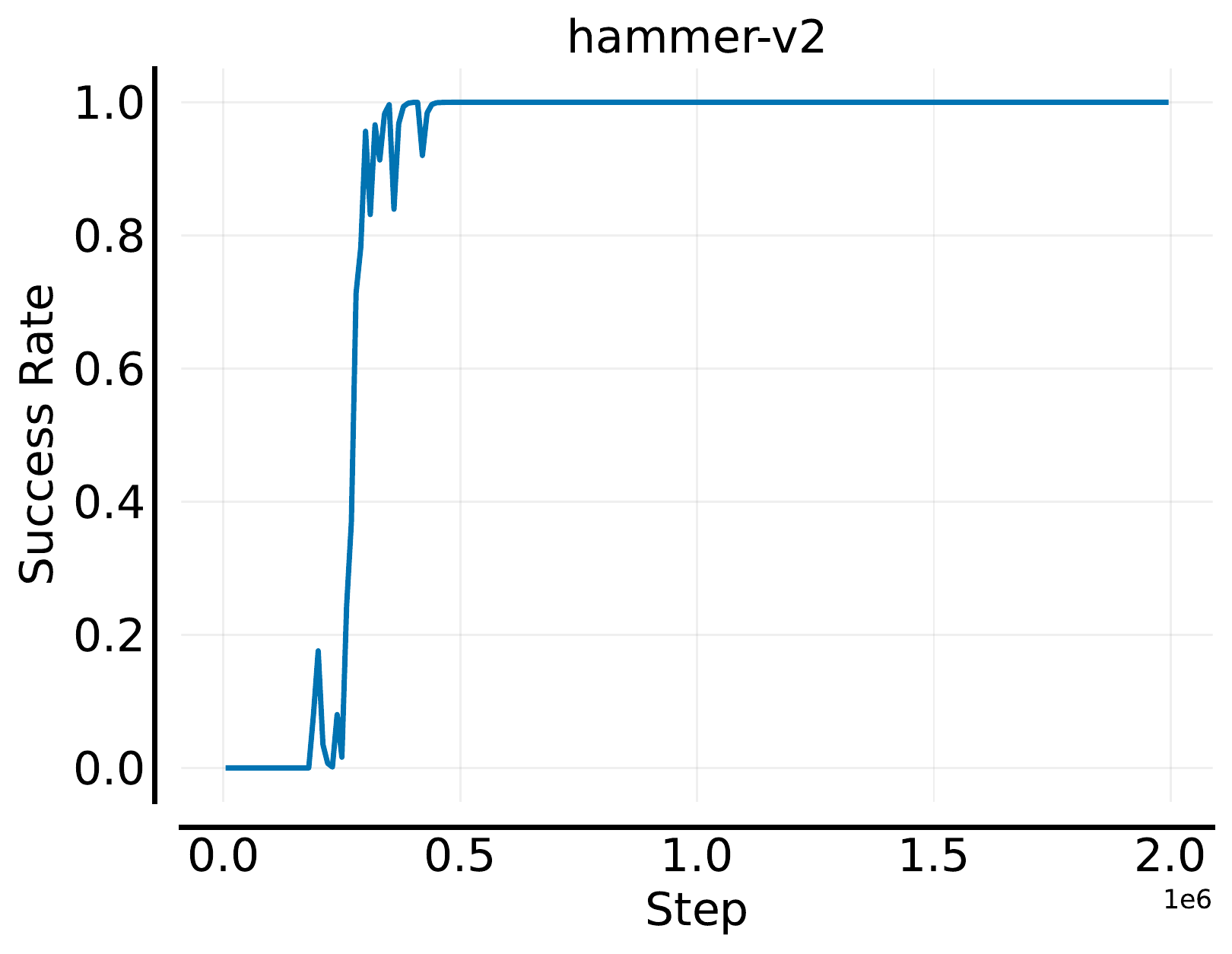}}
    \subfigure{\includegraphics[width=0.19\textwidth]{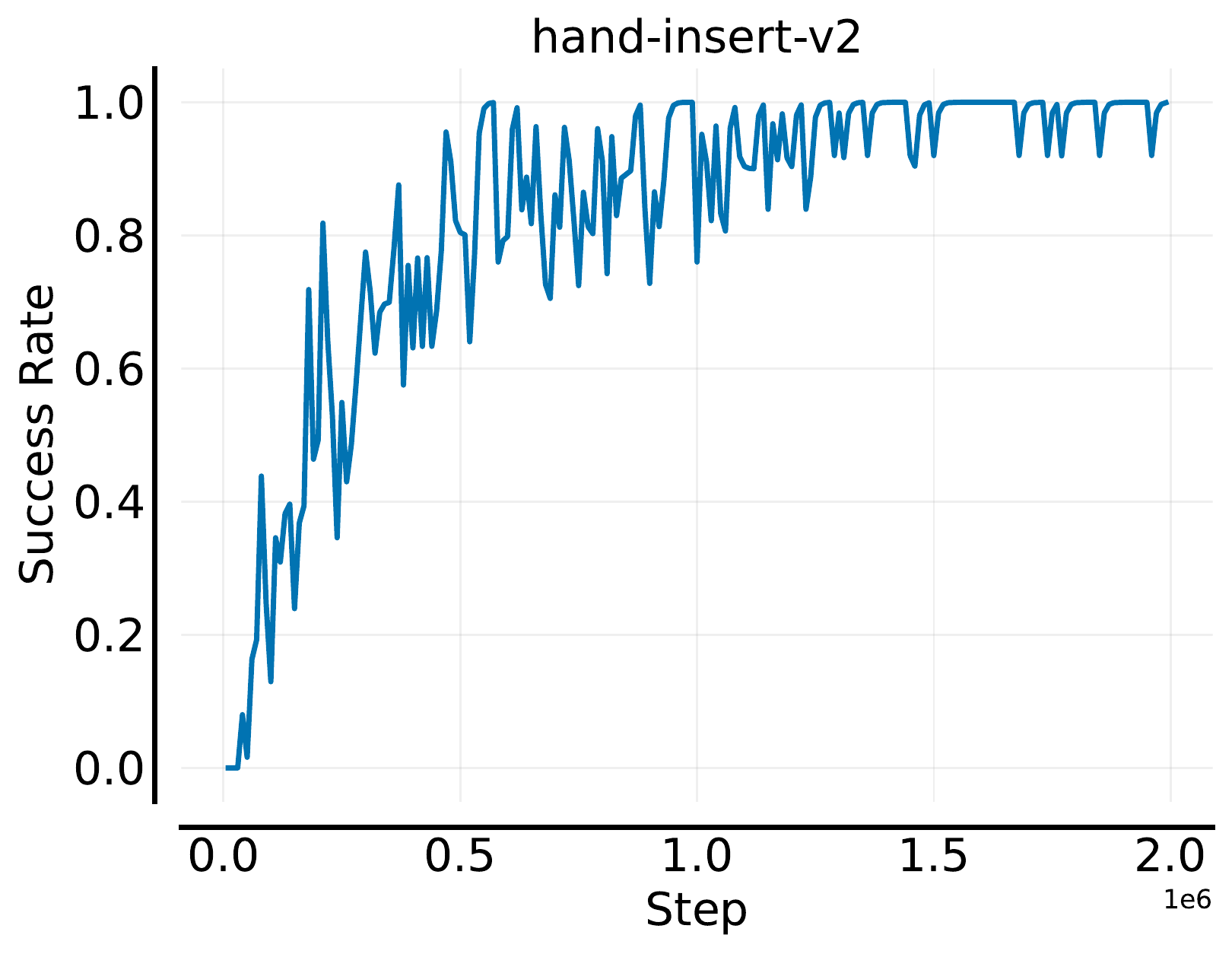}}
    \subfigure{\includegraphics[width=0.19\textwidth]{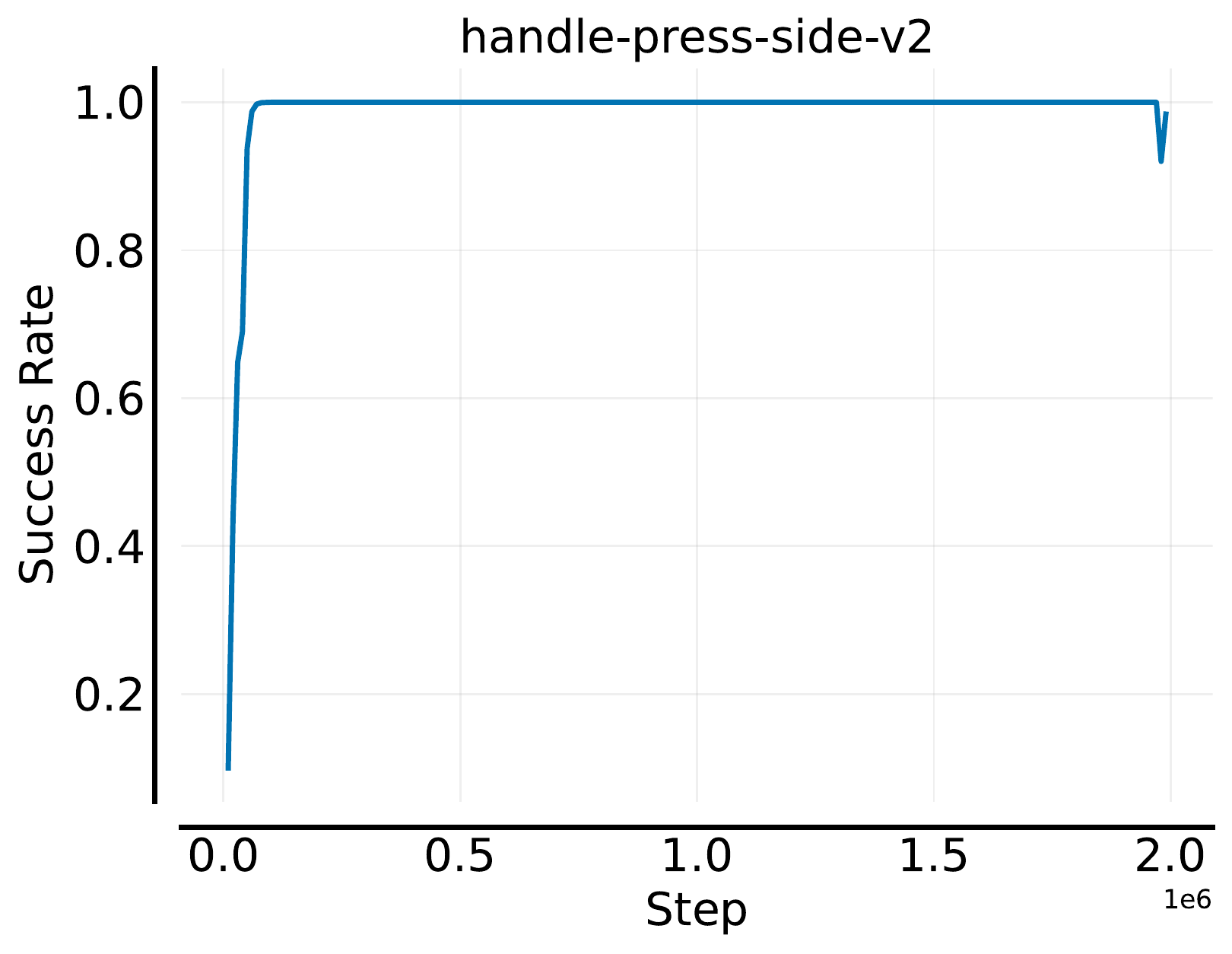}}
    \subfigure{\includegraphics[width=0.19\textwidth]{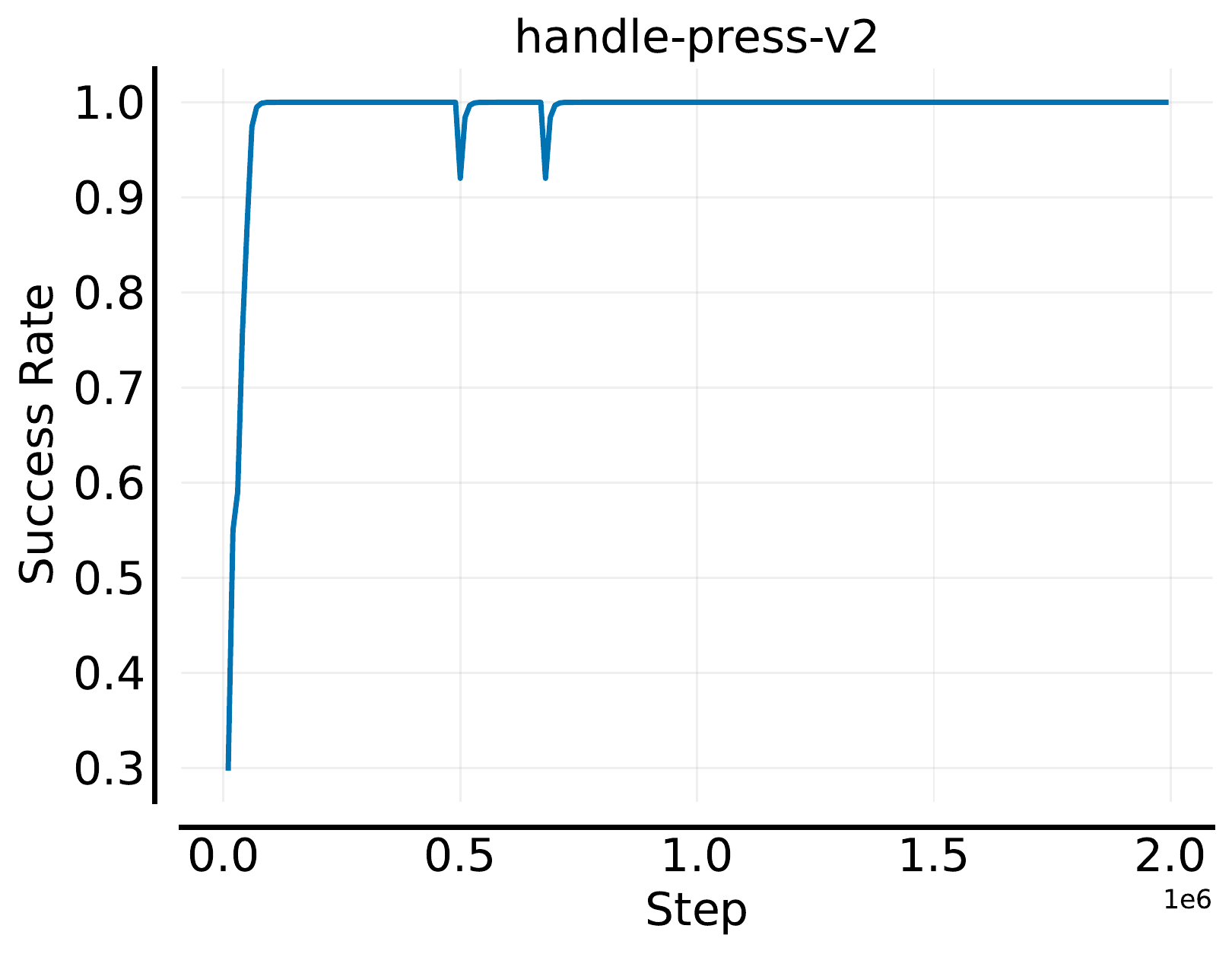}}
    
    \subfigure{\includegraphics[width=0.19\textwidth]{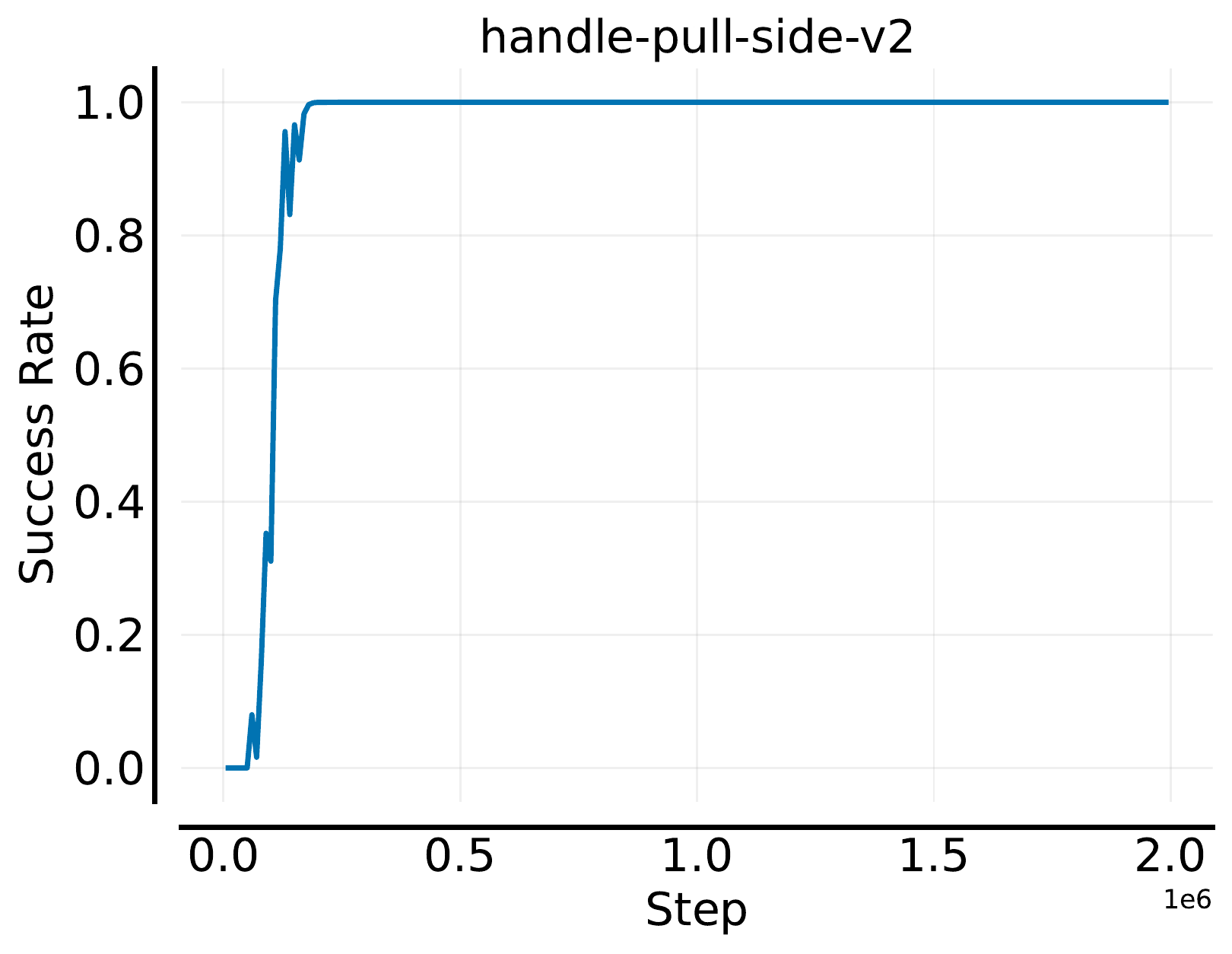}}
    \subfigure{\includegraphics[width=0.19\textwidth]{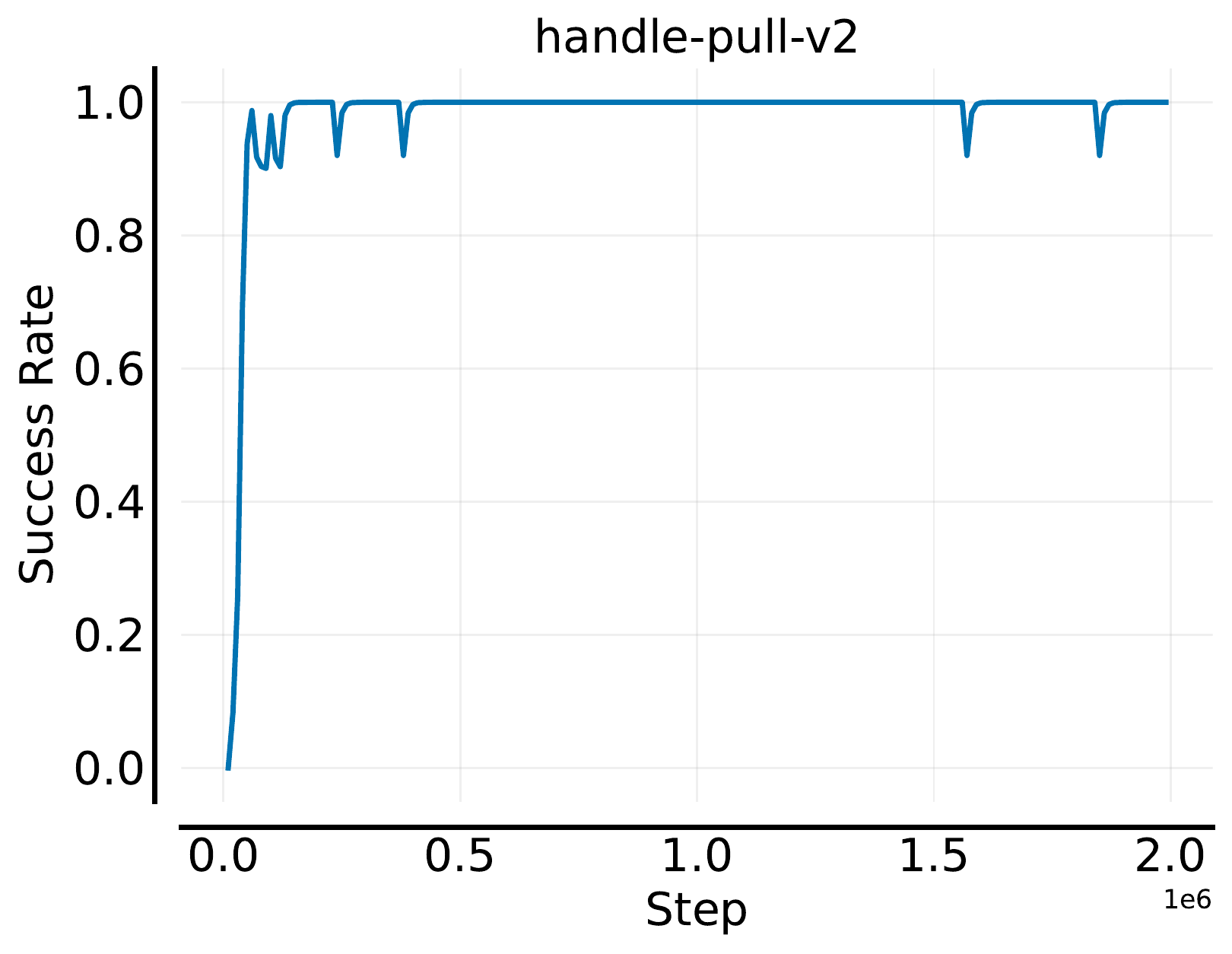}}
    \subfigure{\includegraphics[width=0.19\textwidth]{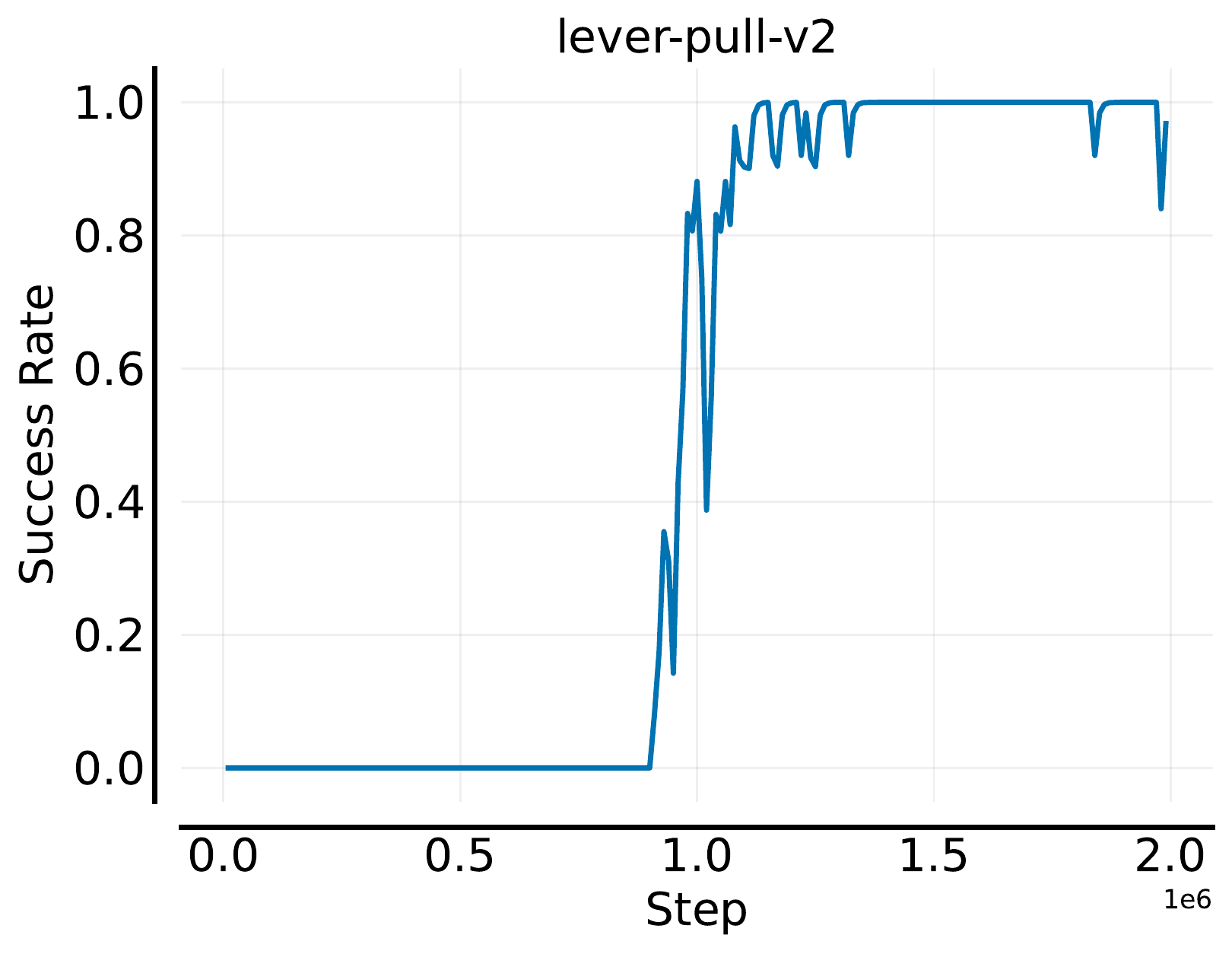}}
    \subfigure{\includegraphics[width=0.19\textwidth]{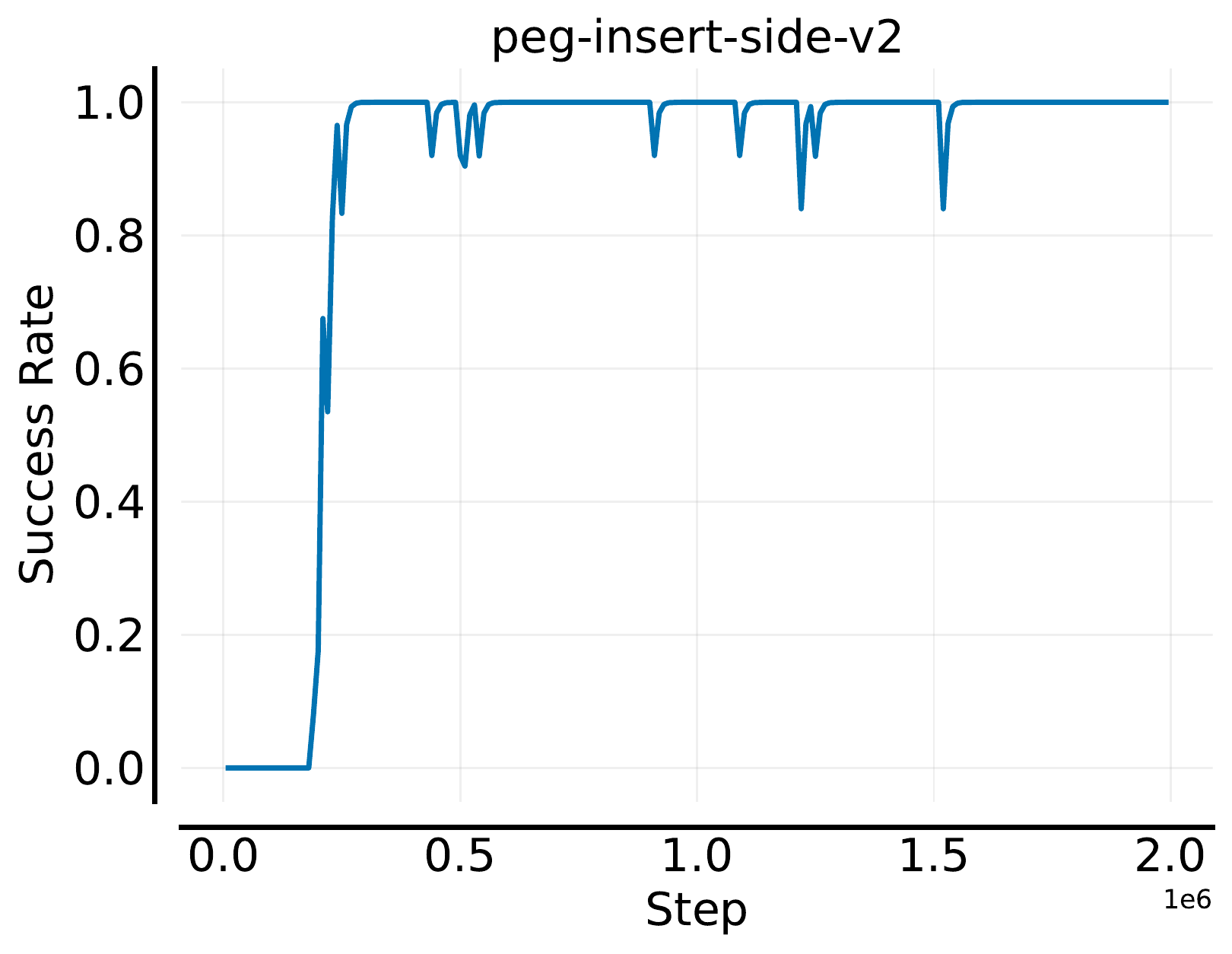}}
    \subfigure{\includegraphics[width=0.19\textwidth]{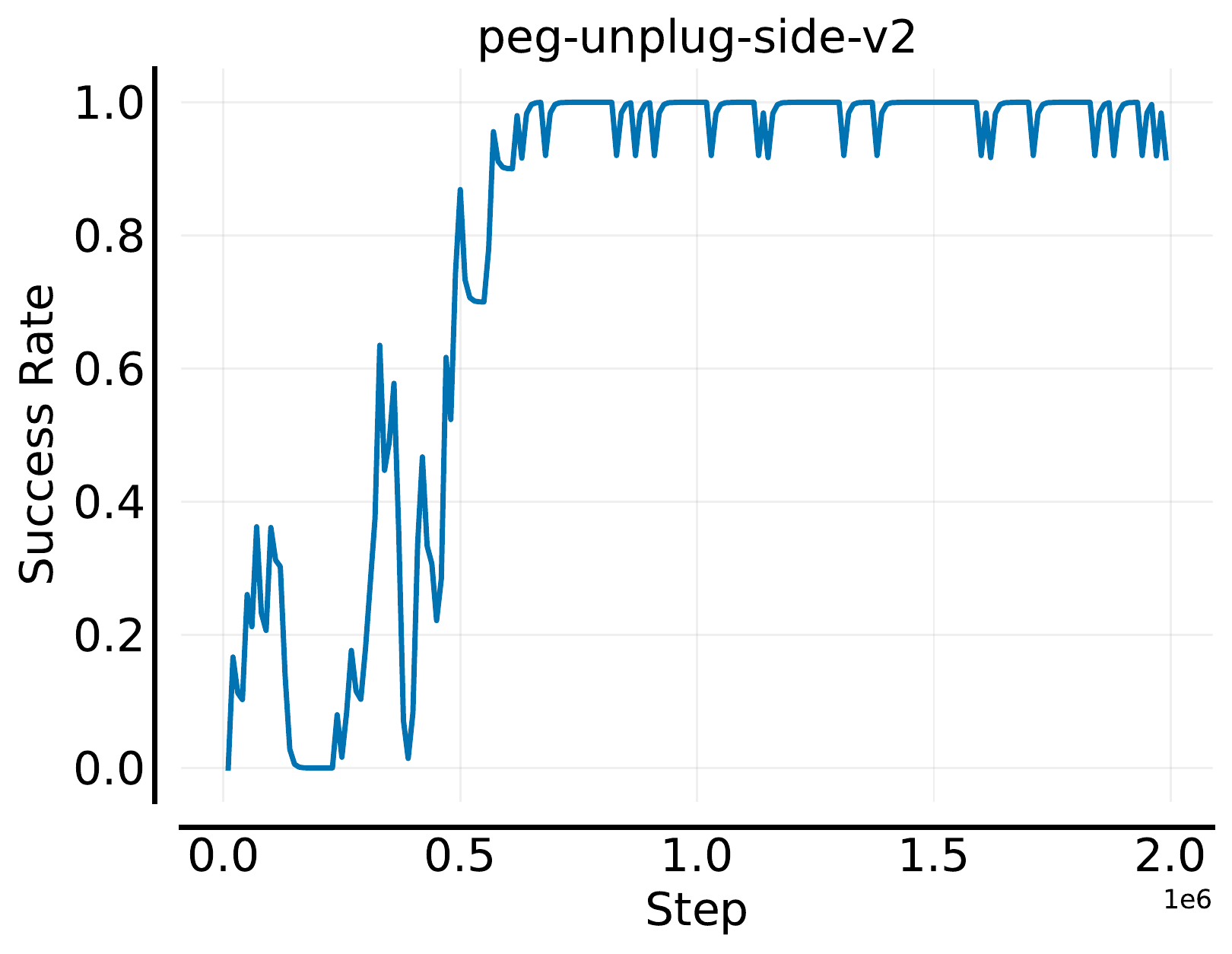}}
    
    \subfigure{\includegraphics[width=0.19\textwidth]{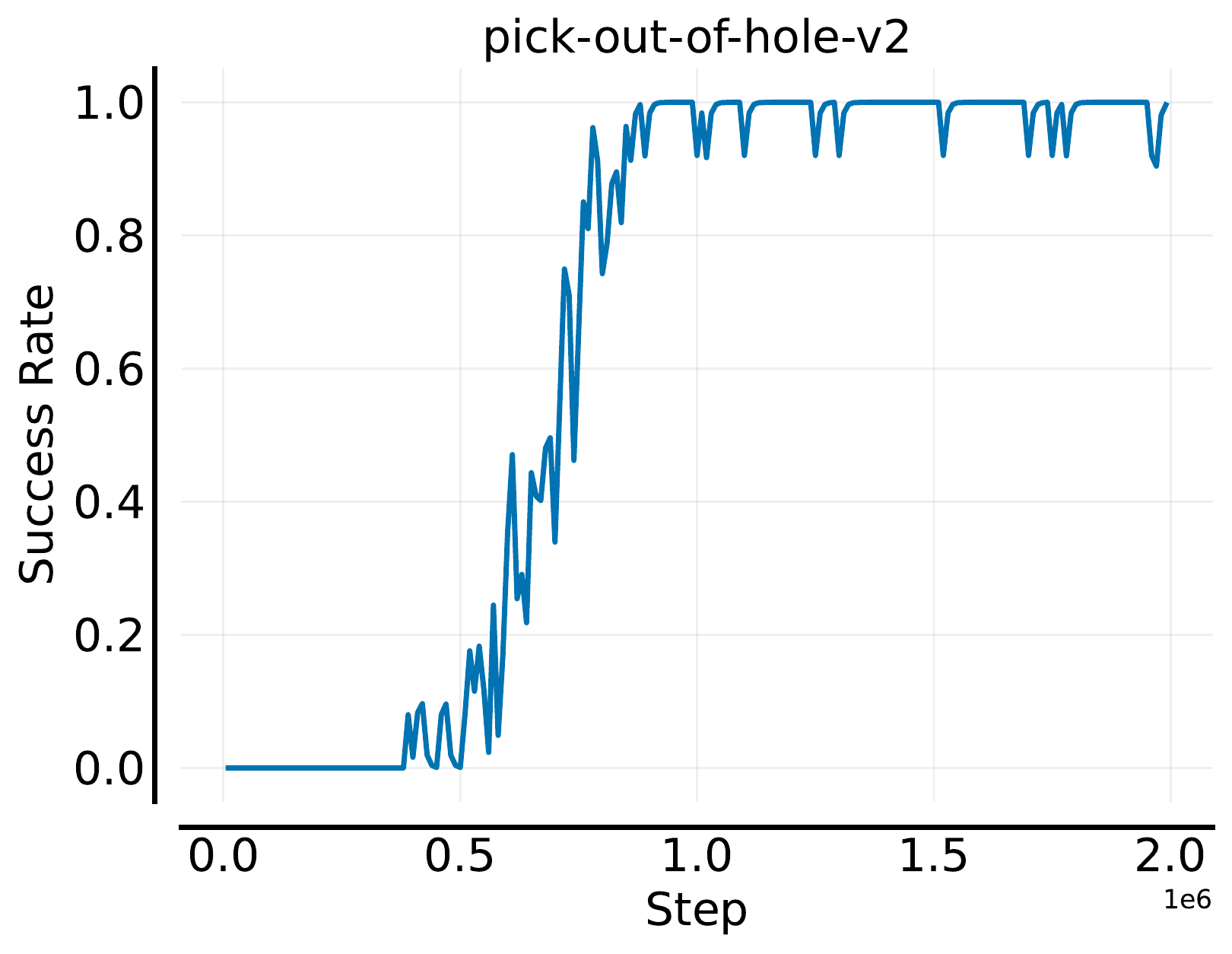}}
    \subfigure{\includegraphics[width=0.19\textwidth]{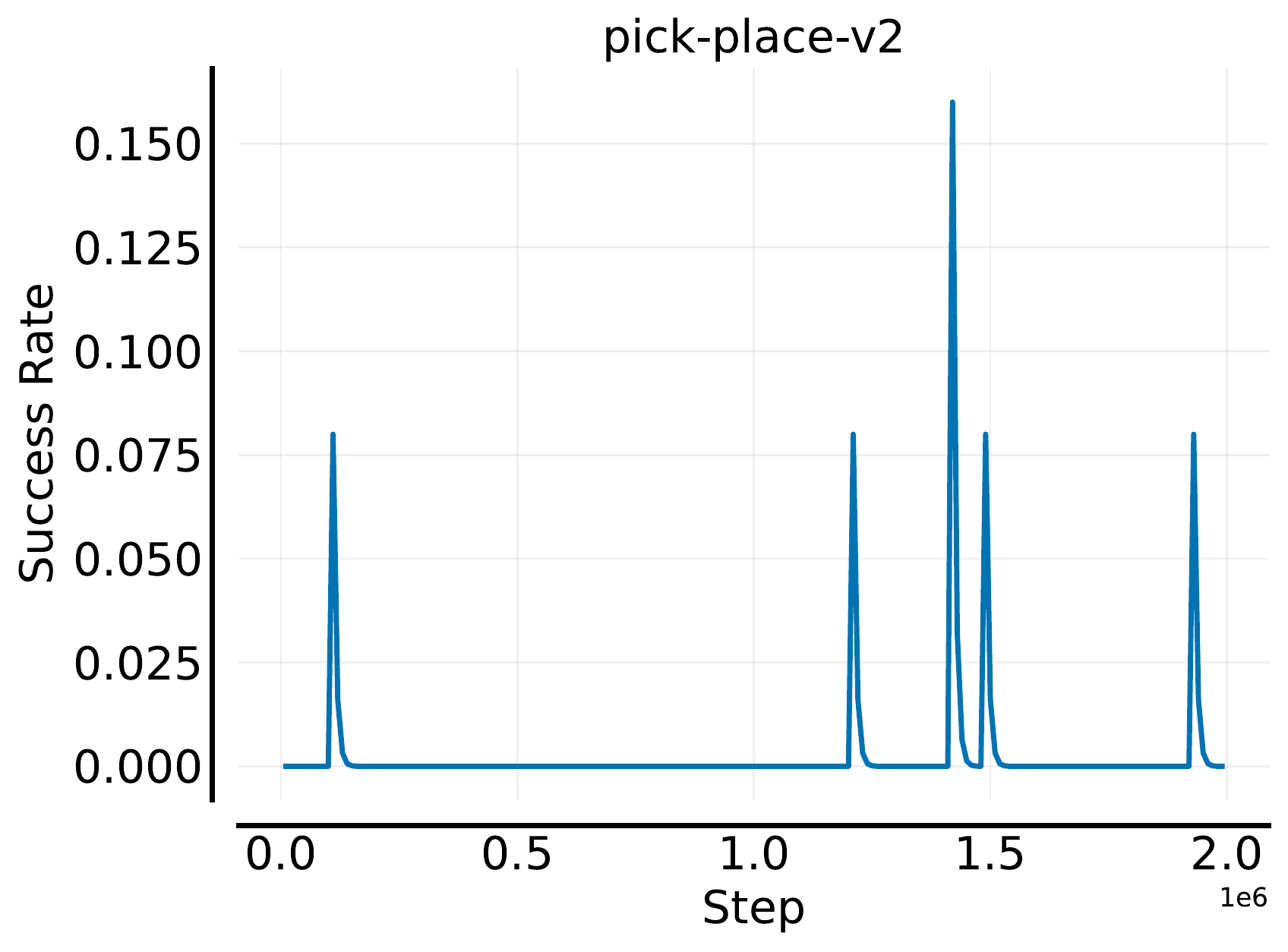}}
    \subfigure{\includegraphics[width=0.19\textwidth]{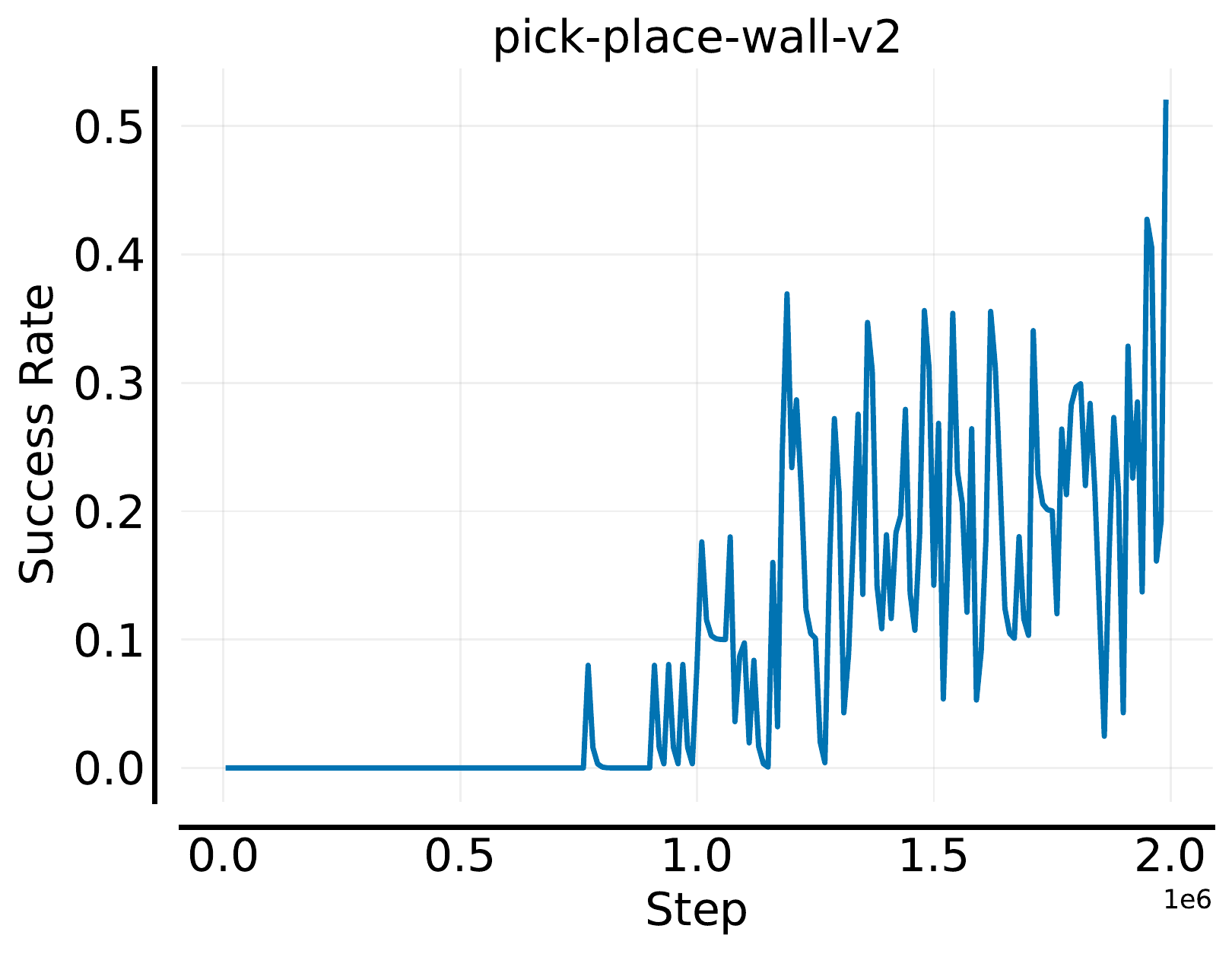}}
    \subfigure{\includegraphics[width=0.19\textwidth]{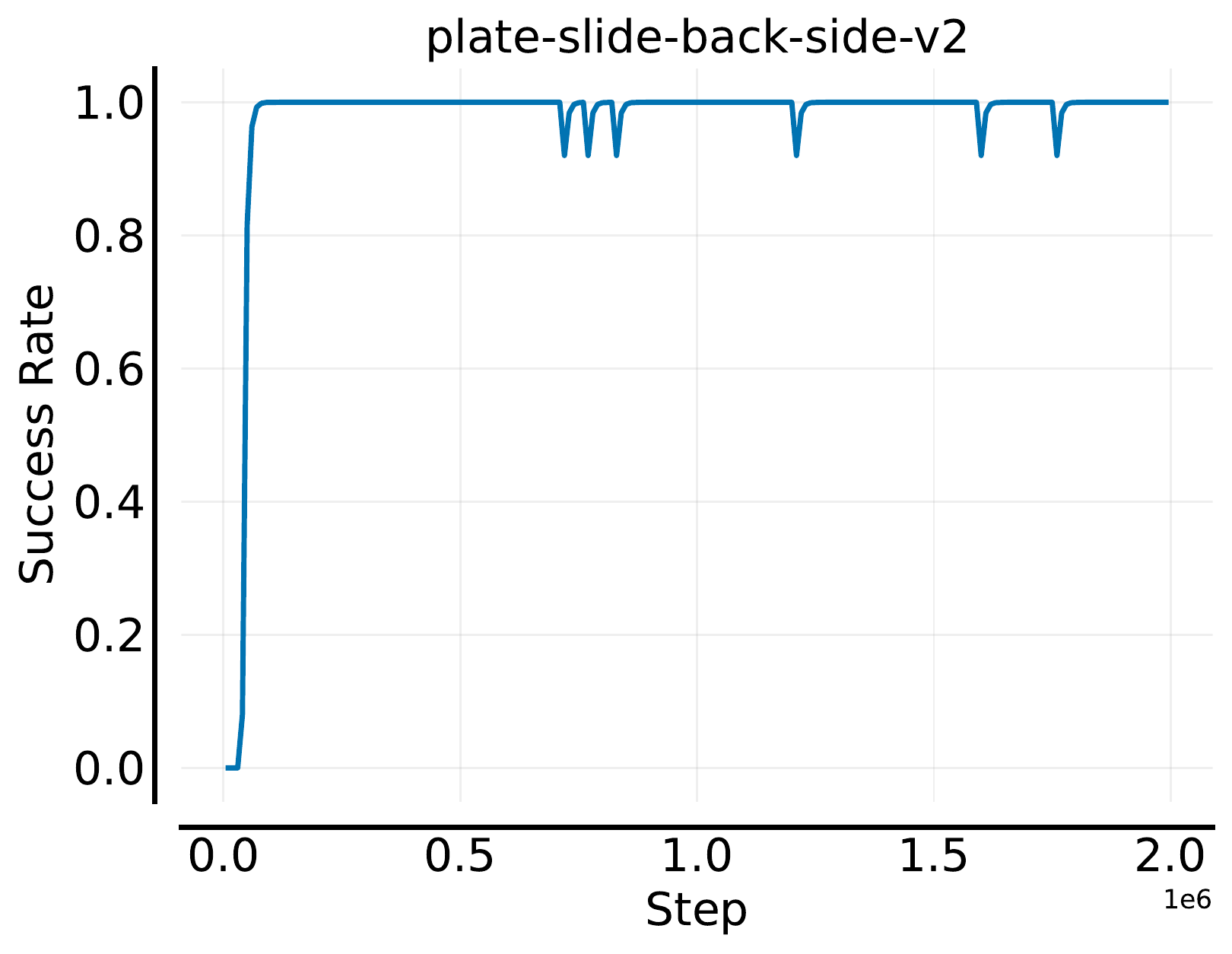}}
    \subfigure{\includegraphics[width=0.19\textwidth]{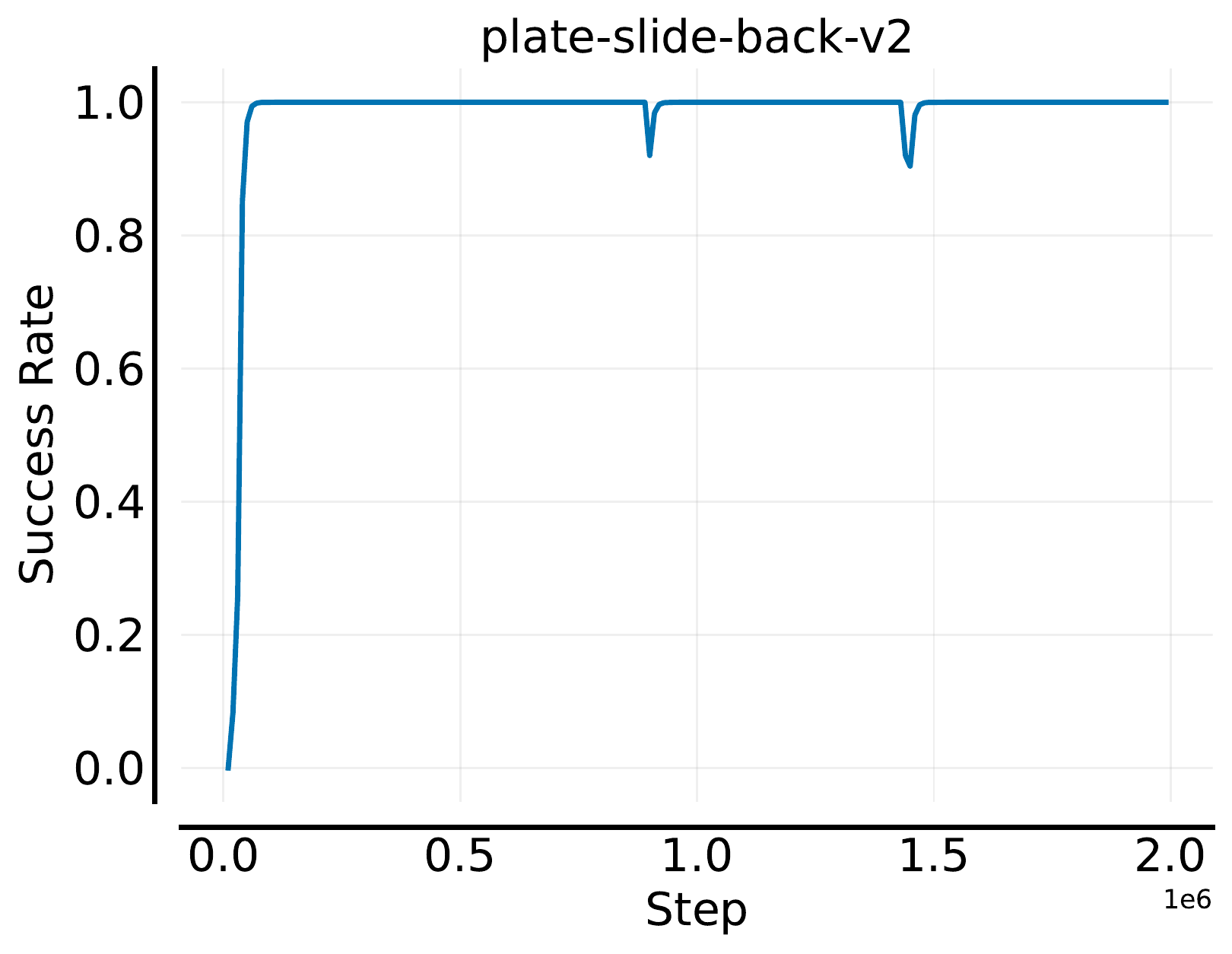}}
    
    \subfigure{\includegraphics[width=0.19\textwidth]{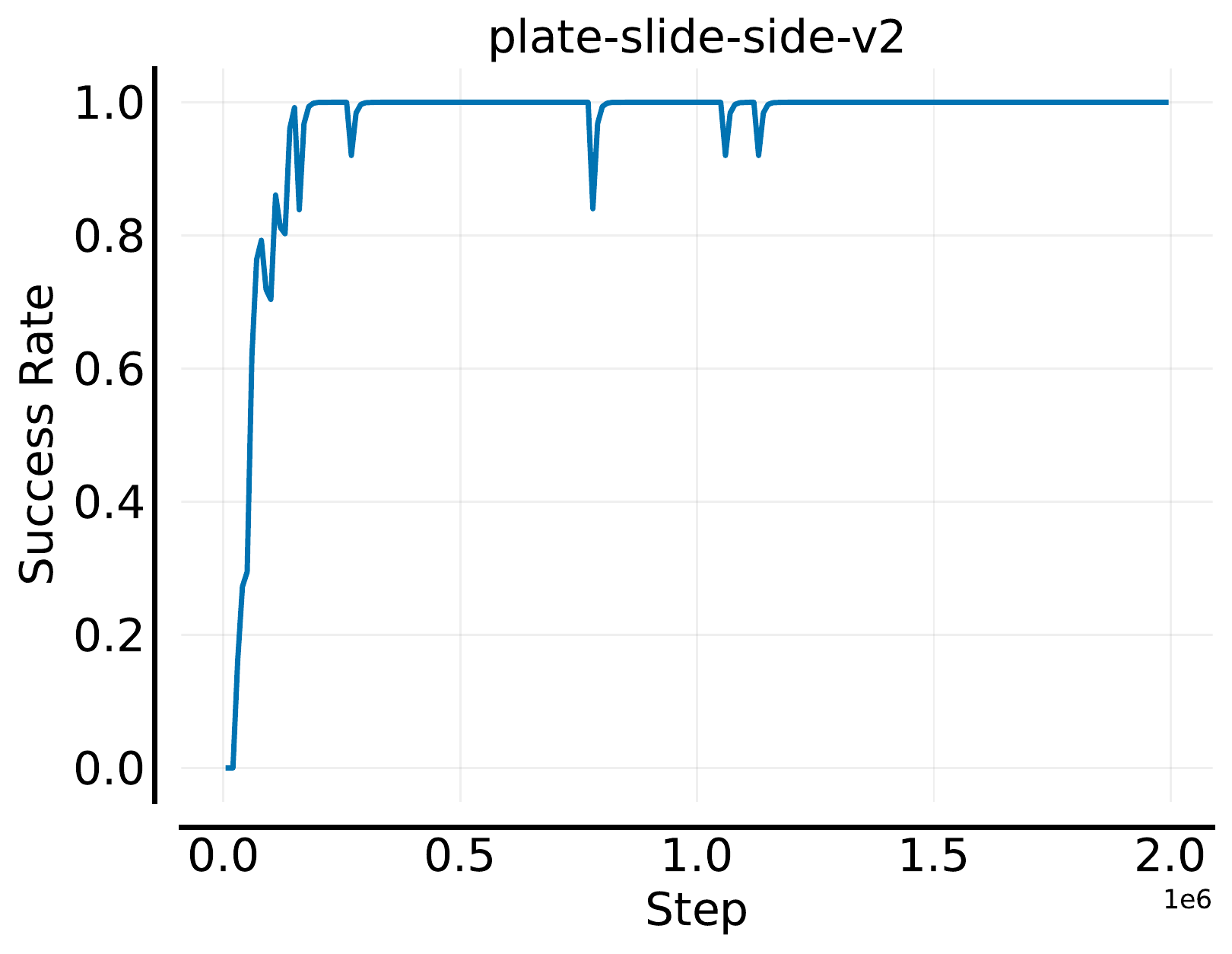}}
    \subfigure{\includegraphics[width=0.19\textwidth]{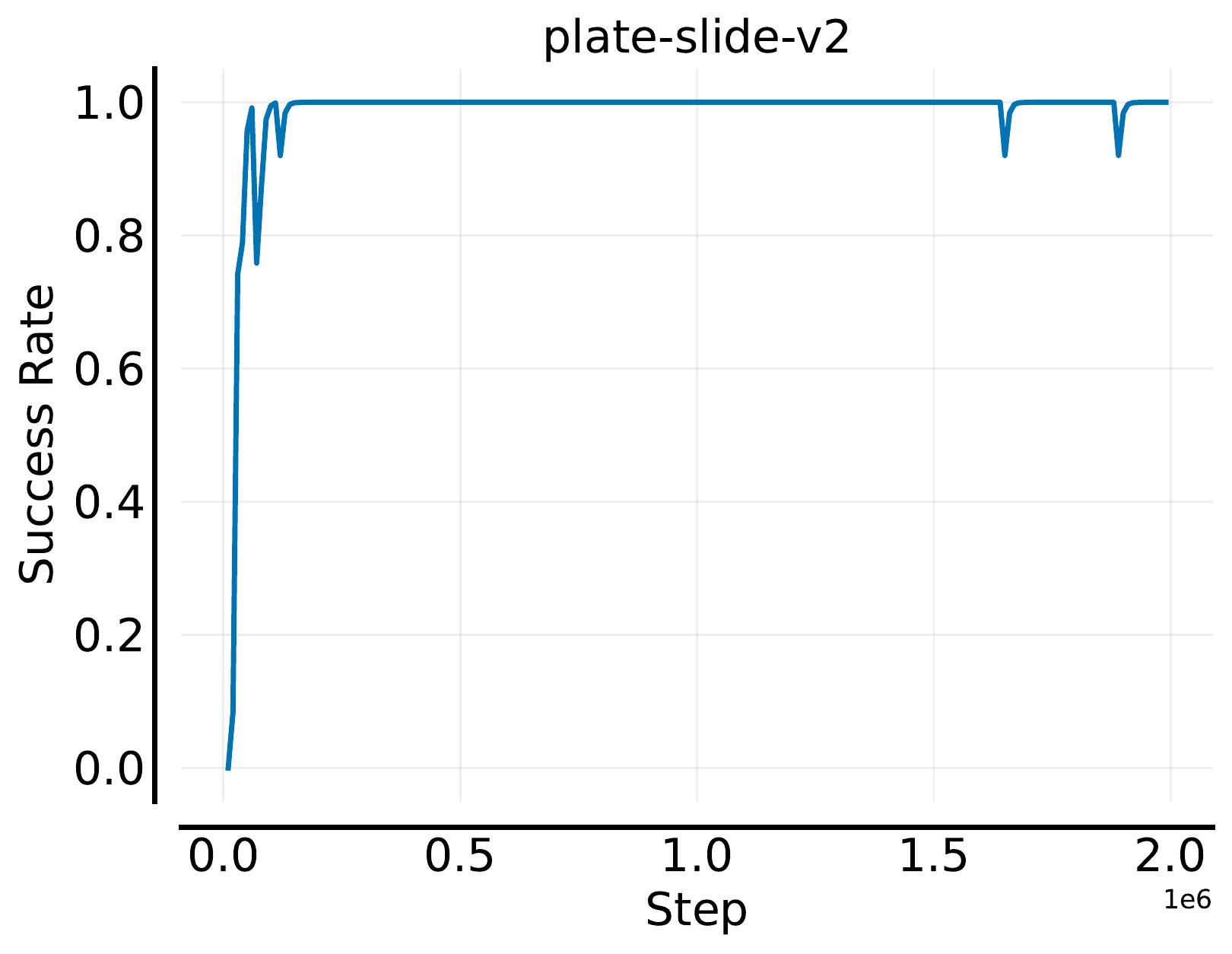}}
    \subfigure{\includegraphics[width=0.19\textwidth]{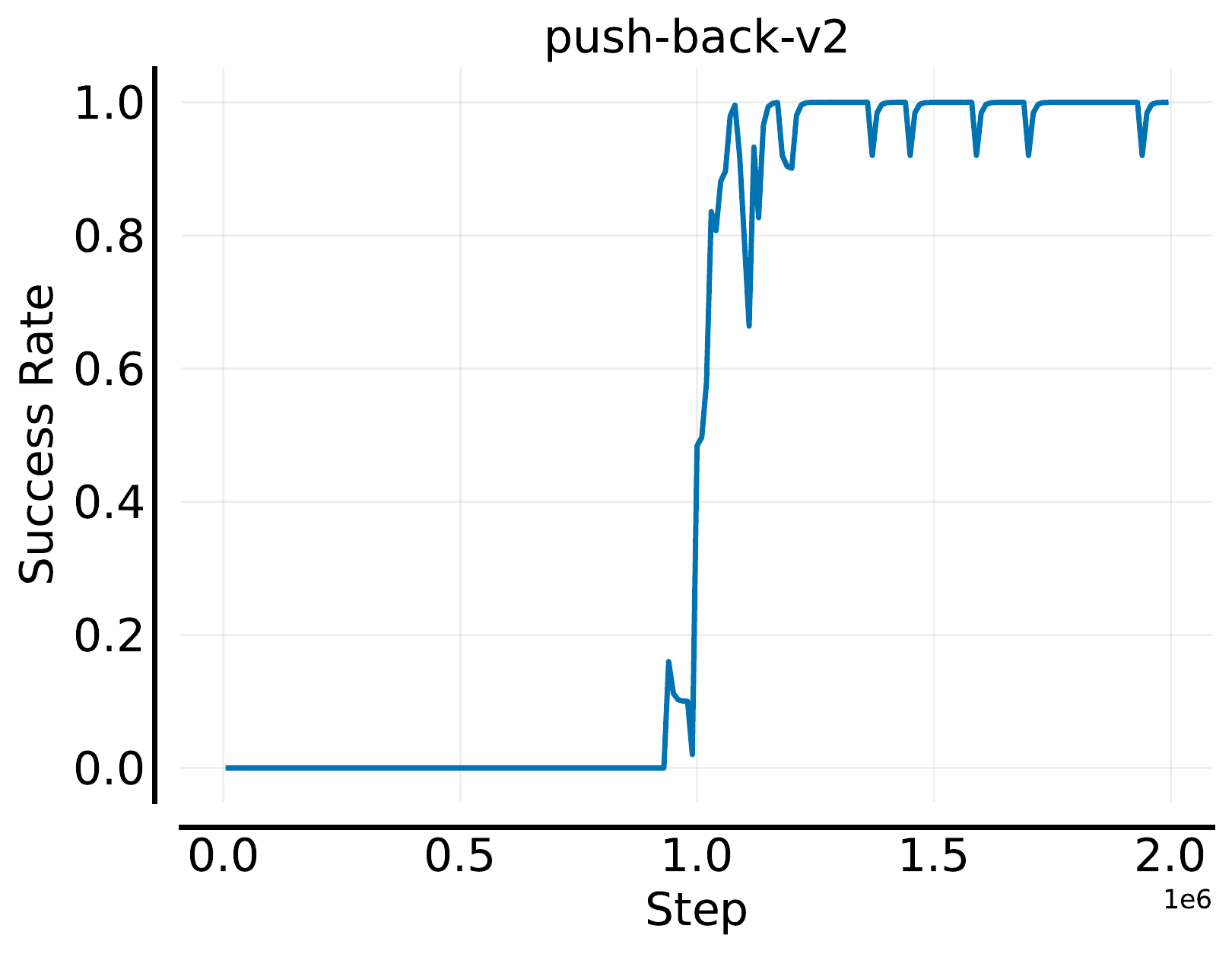}}
    \subfigure{\includegraphics[width=0.19\textwidth]{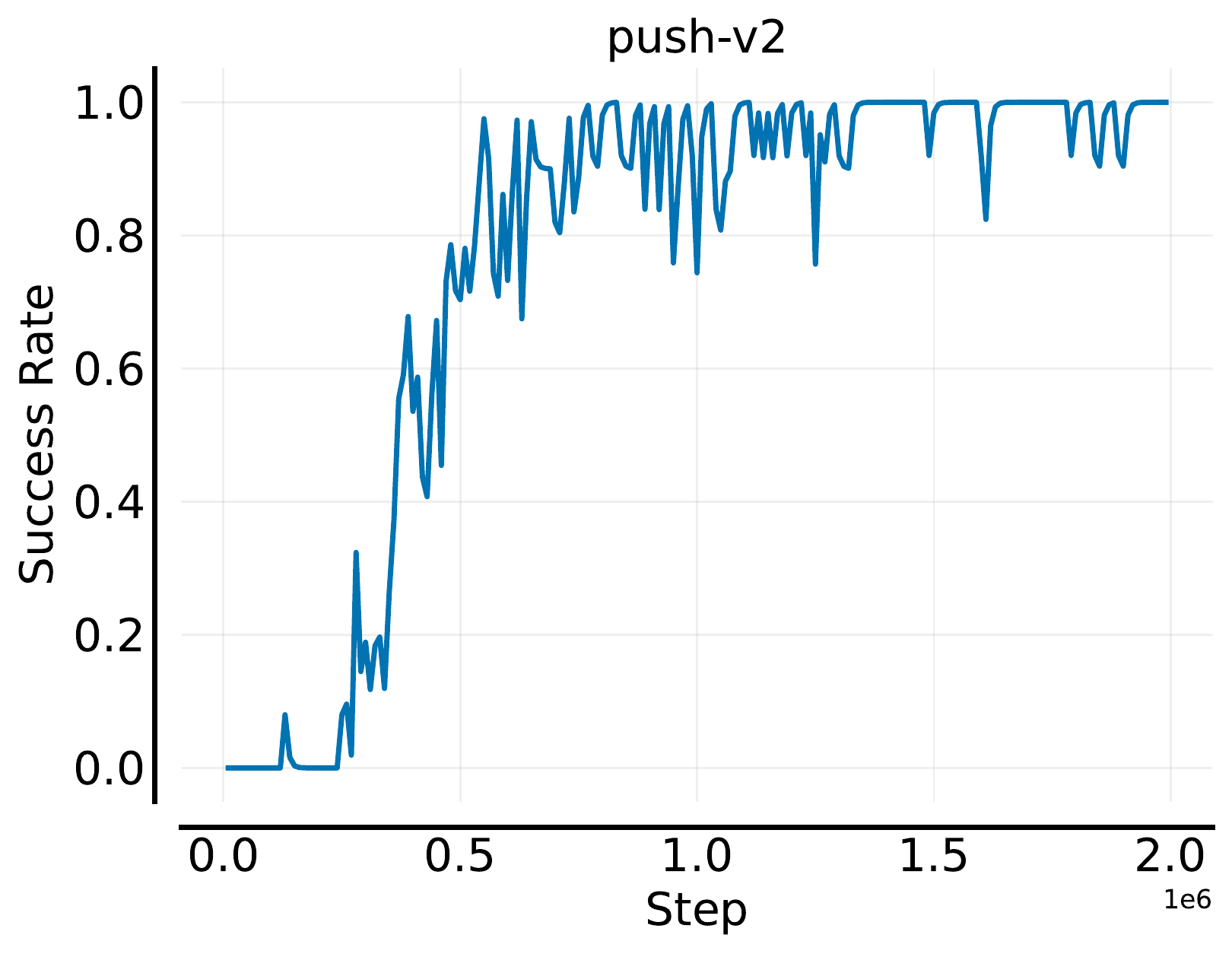}}
    \subfigure{\includegraphics[width=0.19\textwidth]{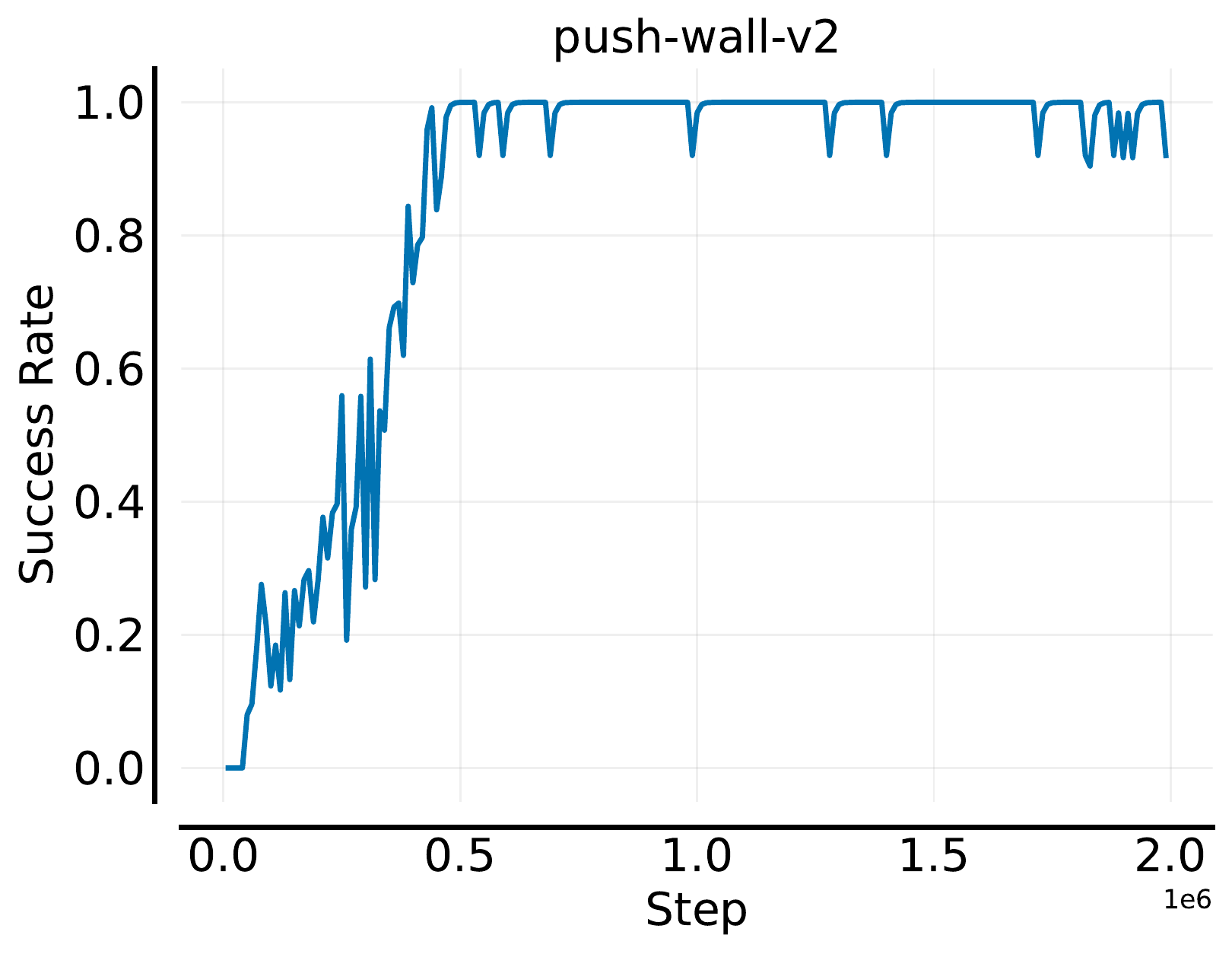}}
  \caption{Learning curves for data-collection runs on all MT40 tasks with SAC. We train for 2M environment interaction steps on each task and record the entire replay buffer.} 
  \label{fig:appendix-datacollection}
\end{figure}

\begin{figure}
  \centering
    \subfigure{\includegraphics[width=0.19\textwidth]{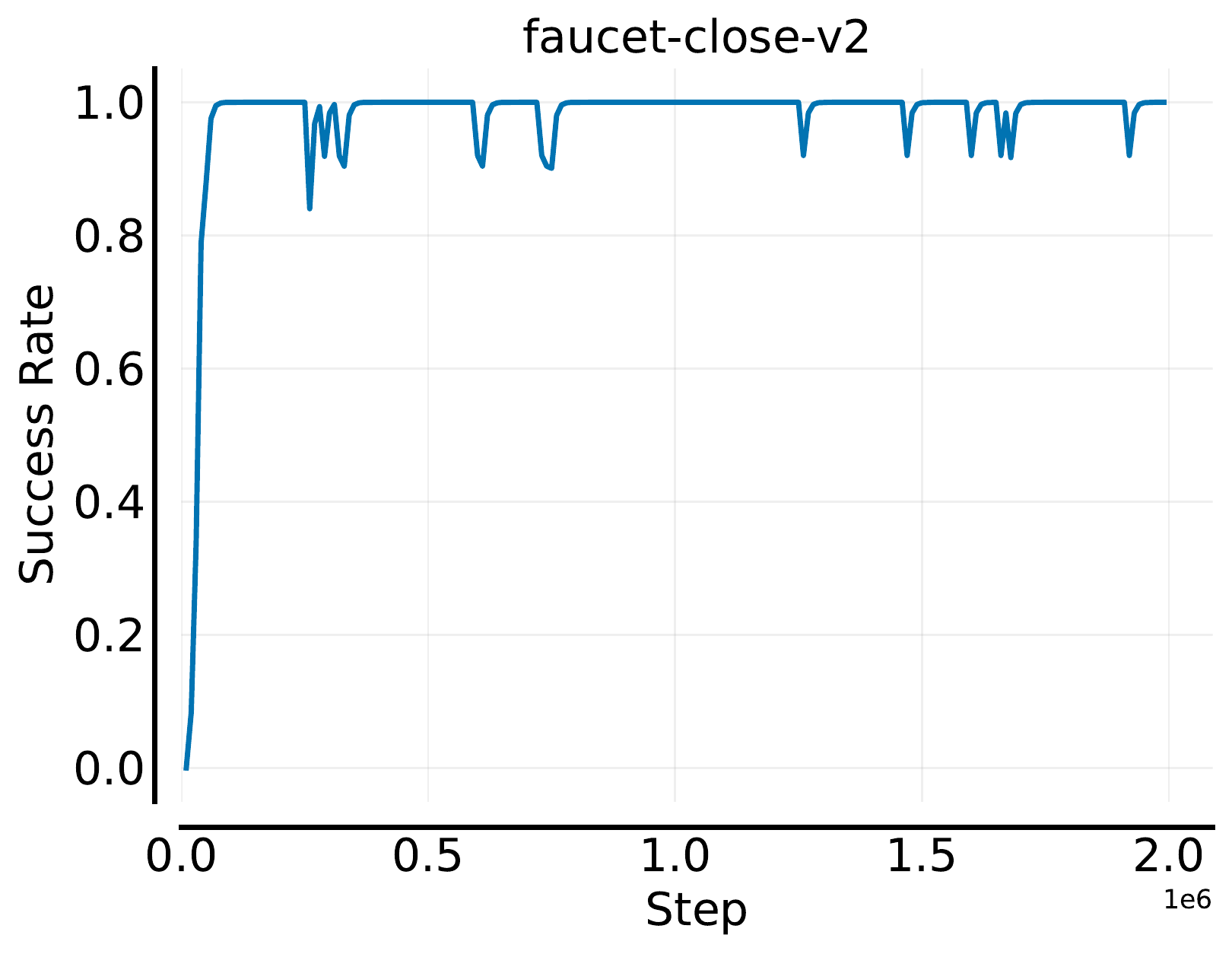}}
    \subfigure{\includegraphics[width=0.19\textwidth]{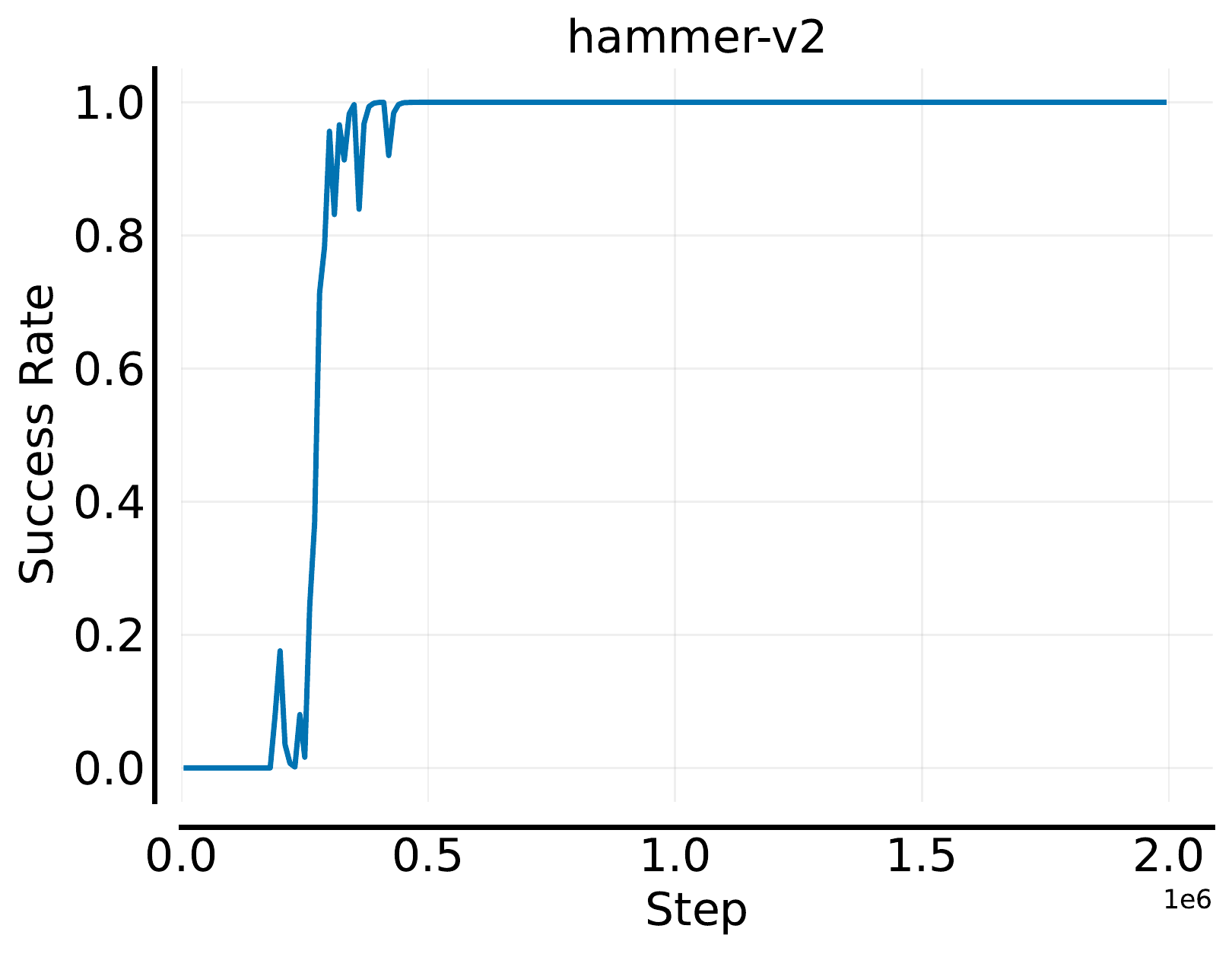}}
    \subfigure{\includegraphics[width=0.19\textwidth]{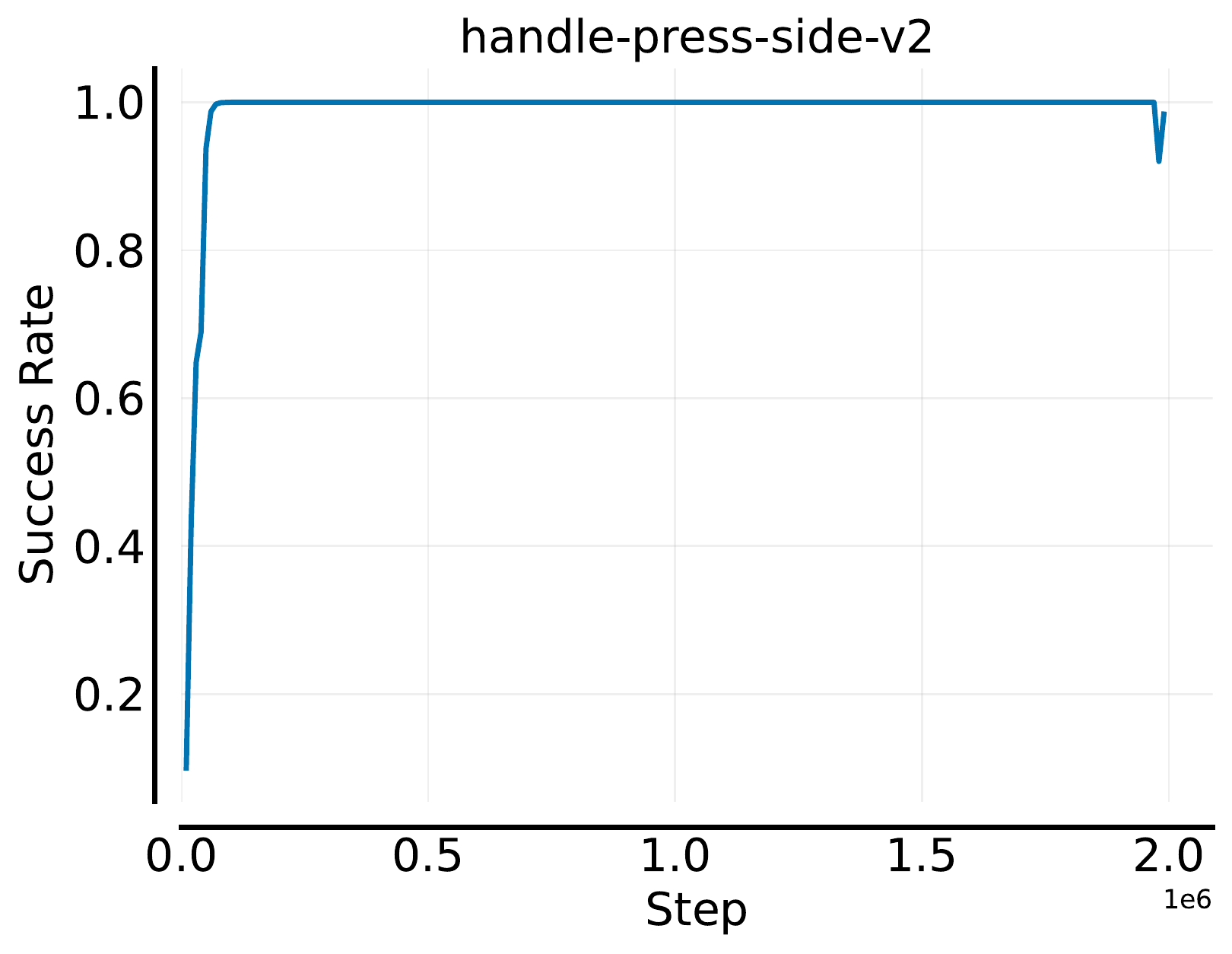}}
    \subfigure{\includegraphics[width=0.19\textwidth]{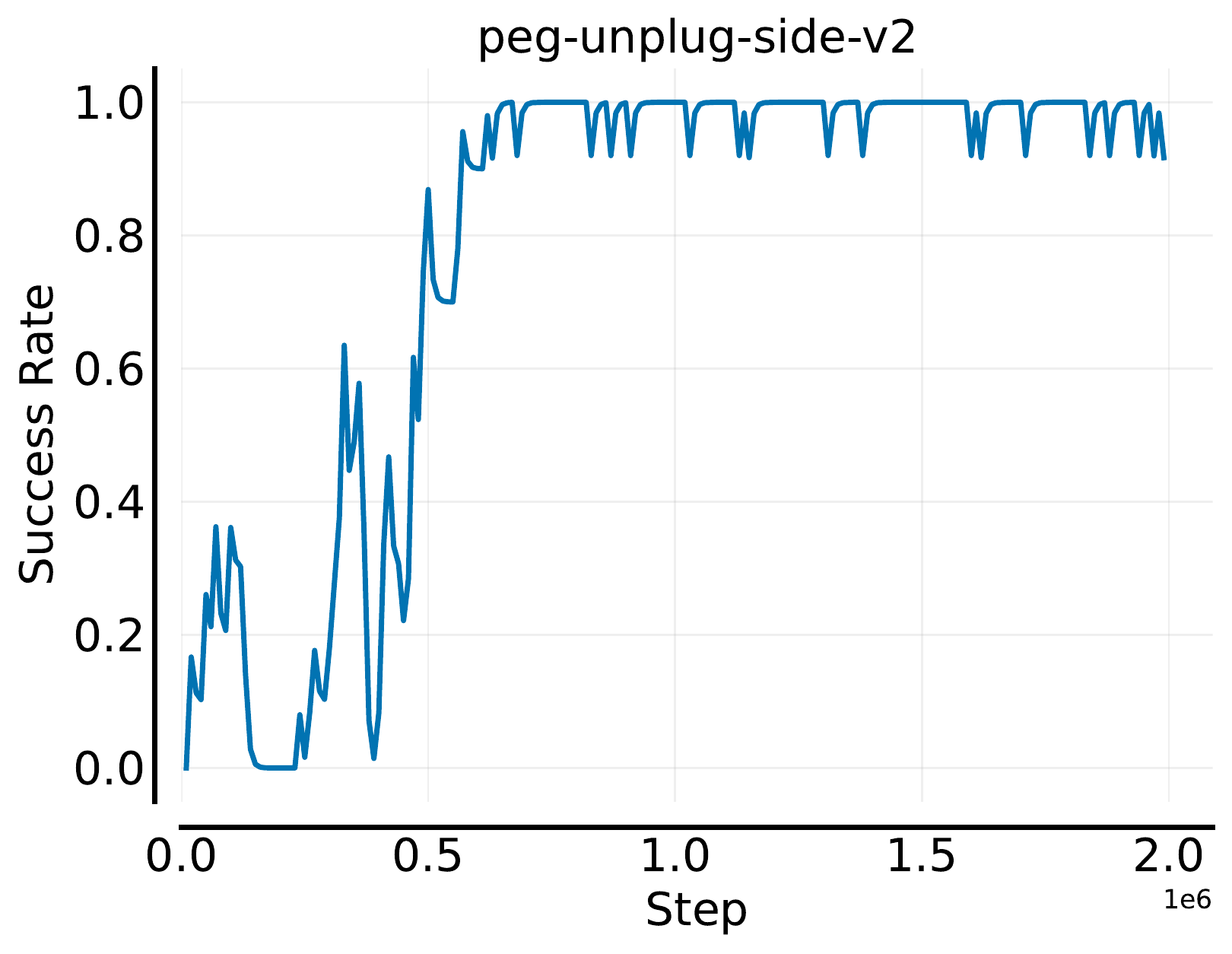}}
    \subfigure{\includegraphics[width=0.19\textwidth]{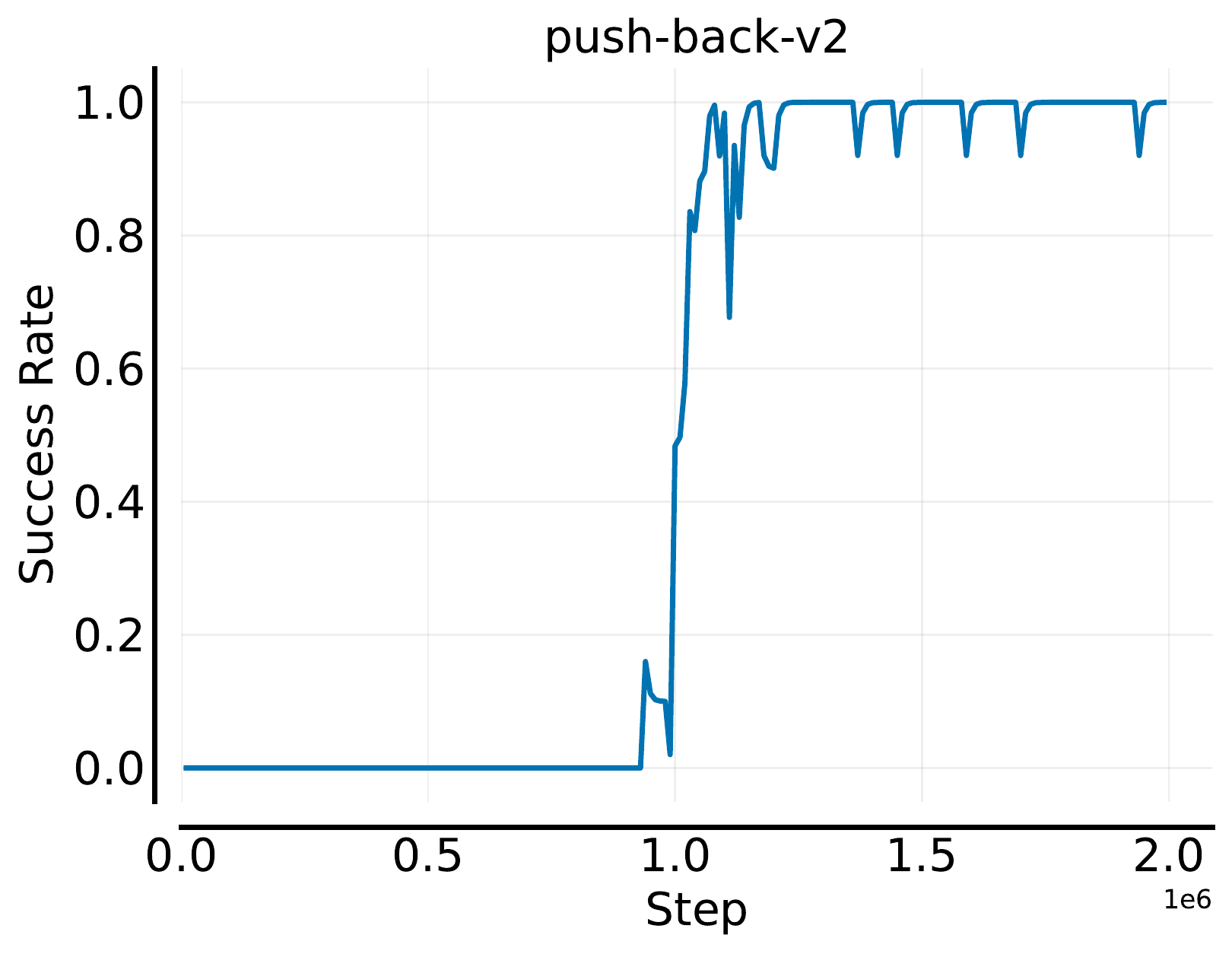}}

    \subfigure{\includegraphics[width=0.19\textwidth]{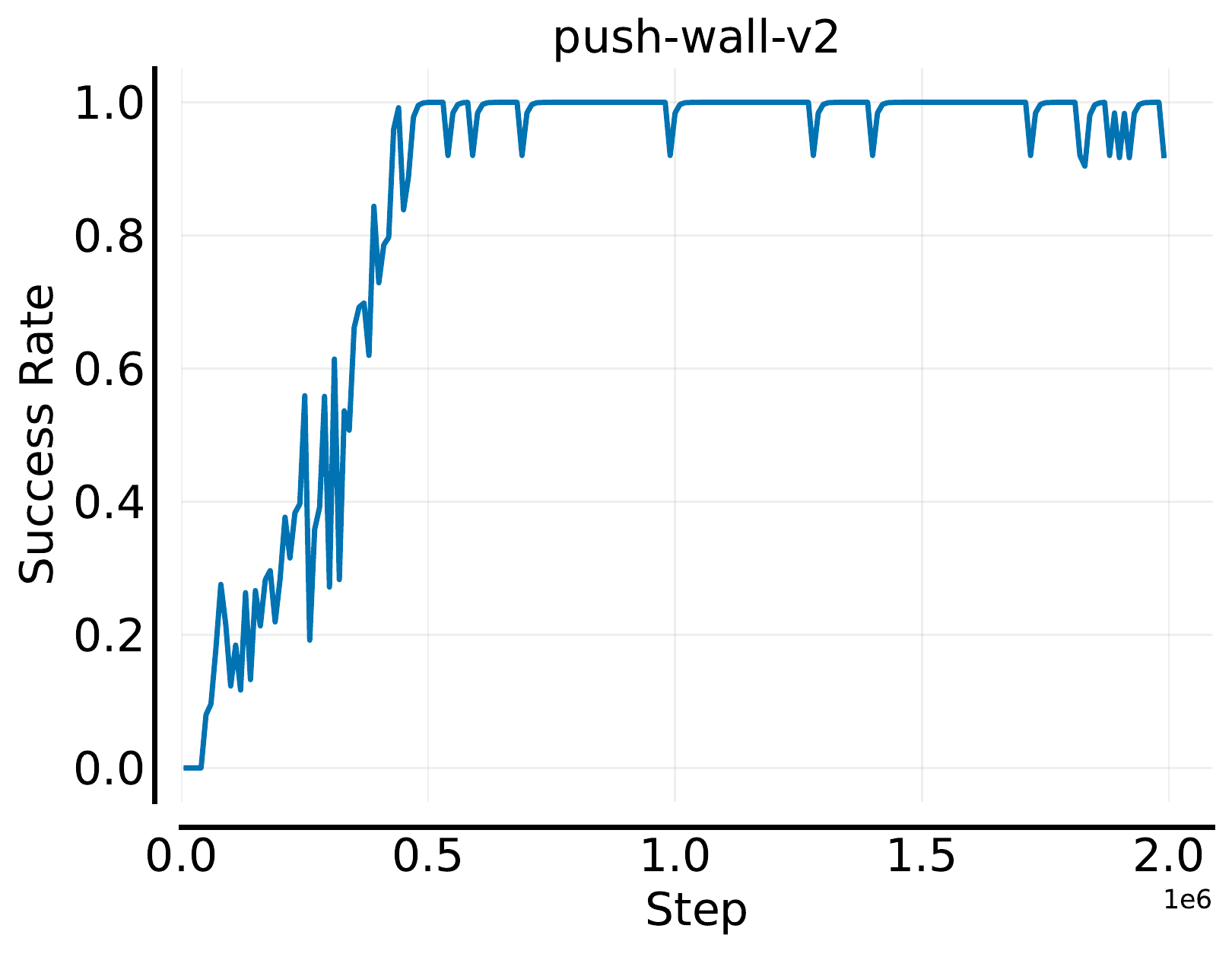}}
    \subfigure{\includegraphics[width=0.19\textwidth]{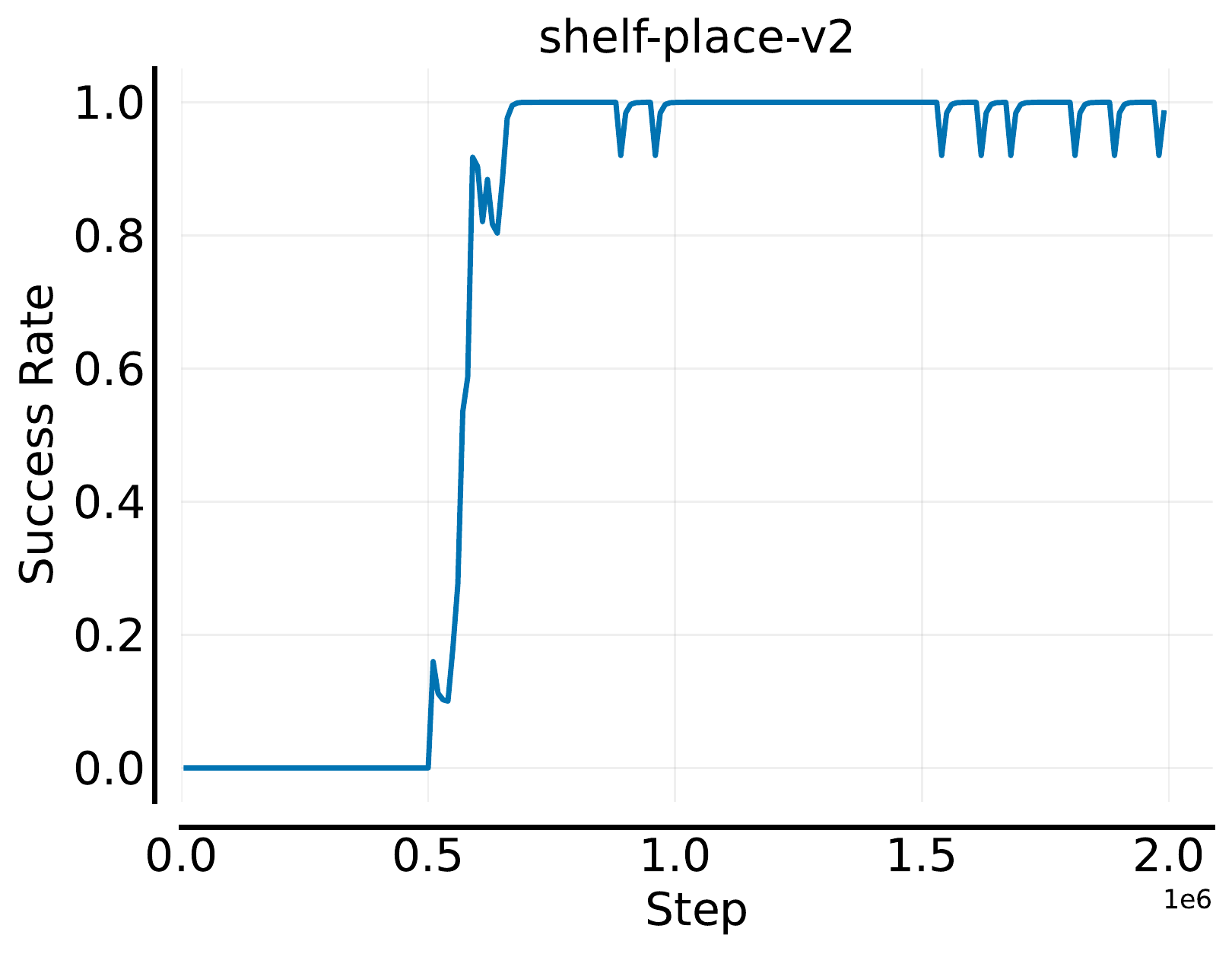}}
    \subfigure{\includegraphics[width=0.19\textwidth]{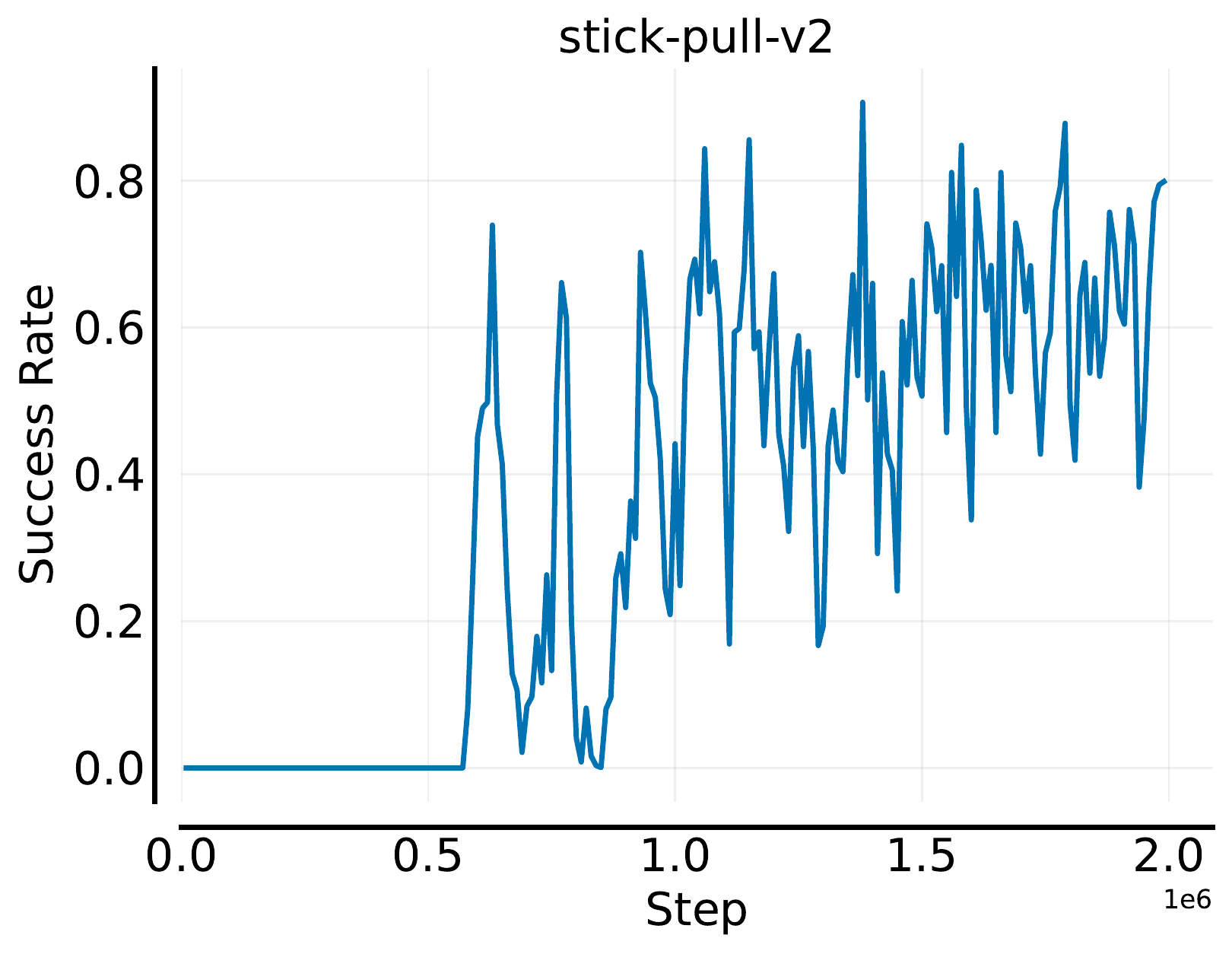}}
    \subfigure{\includegraphics[width=0.19\textwidth]{figures/neurips/data_collection/cw10/lc_rliable_faucet-close-v2_nolegend.pdf}}
    \subfigure{\includegraphics[width=0.19\textwidth]{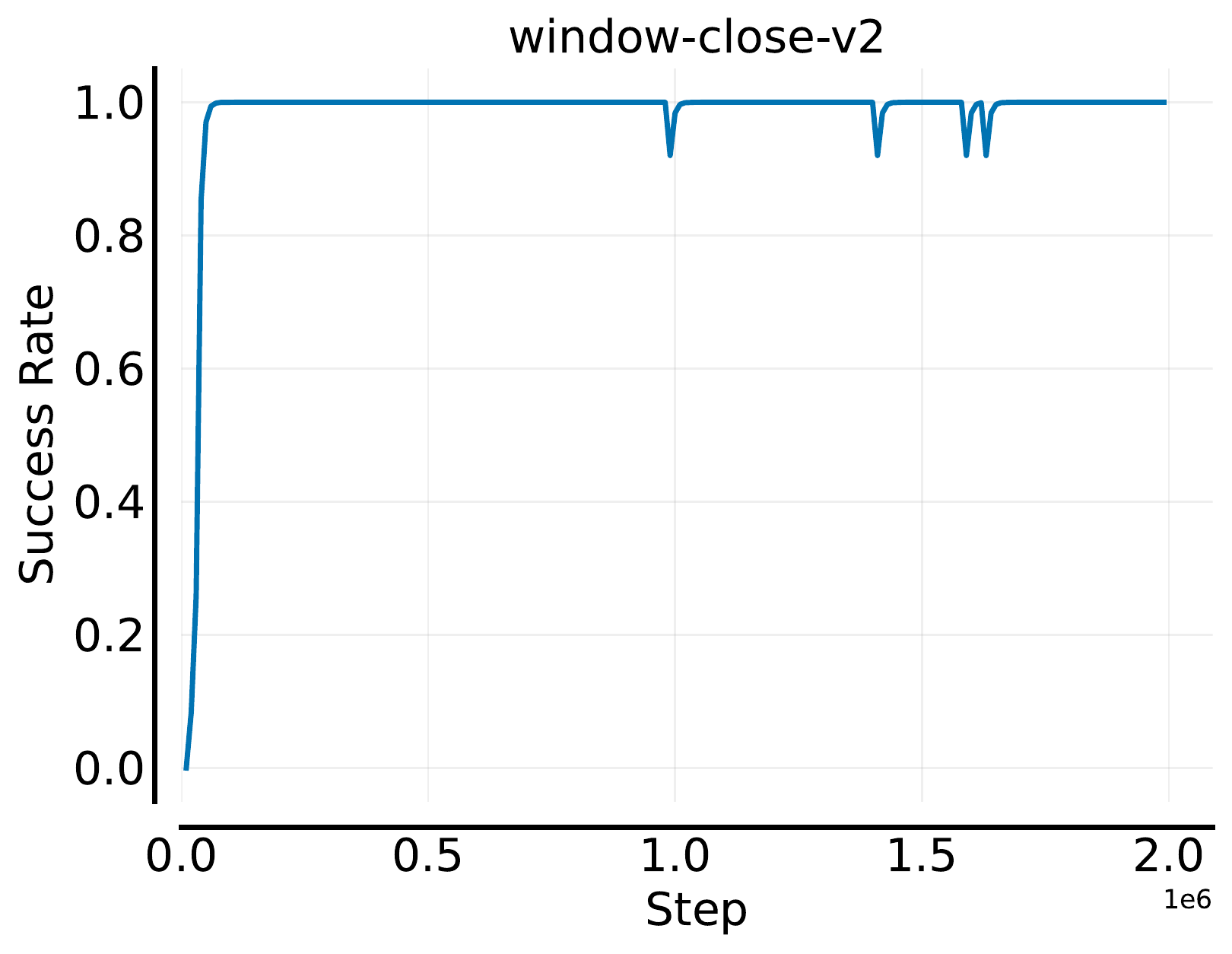}}
  \caption{Learning curves for data-collection runs on all CW10 tasks with SAC. We train for 2M environment interaction steps on each task and record the entire replay buffer.} 
  \label{fig:appendix-datacollection-cw10}
\end{figure}

\section{Dataset}\label{appendix:data}
To collect our dataset, we train expert agents via SAC \citep{Haarnoja:18}. 
We train a separate expert on each of the 66 considered tasks.

\subsection{Meta-World} \label{appendix:data-metaworld}
For each of the 50 Meta-World tasks, we train for 2M steps and record the entire replay buffer. Thus, the final Meta-World datasets contain 2M transitions (i.e., state-RTG-action-reward tuples). Each trajectory is 200 steps long, and each dataset consists of 10K trajectories. This amounts to 100M transitions in total for all 50 tasks, 80M and 20M for MT40 and CW10, respectively. 

We use the same network architecture as \citet{Wolczyk:21}, 4 linear layers with 256 neurons,  LayerNorm \citep{Ba:16} after the first layer, and LeakyReLU \citep{Maas:13} activations for both, actor and critic. We use $\alpha=0.01$ for the LeakyReLU instead of $\alpha=0.2$ used by \citet{Wolczyk:21}, since it lead to performance improvements.

After every 50 interaction steps, we perform 50 gradient steps using Adam \citep{Kingma:14}. We keep other parameters fixed at their default values in \texttt{stable-baselines3} \citep{Raffin:21}. This includes the learning rate of $3e^{-4}$, batch size of 256, discount factor 0.99, and automatic entropy tuning. The target networks are synced after every update step with an update coefficient of 0.005. We evaluate the current policy after every 10K interaction steps. Each evaluation run consists of 10 individual evaluation episodes, and scores are averaged over all 10 episodes. The success rates provide a measure of trajectory quality (TQ, \citealp{Schweighofer:22}) for all individual datasets. Additional measures such as state-action coverage (SACo) may also be computed.

In Tables \ref{tab:mt40} and \ref{tab:cw10}, we list the success rates and average rewards achieved by the expert agents on MT40 and CW10 tasks, respectively. In addition, we show the success rate distribution in Figure \ref{fig:mt50_dist}  and the learning curves on MT40 and CW10 in Figures \ref{fig:appendix-datacollection} and \ref{fig:appendix-datacollection-cw10}, respectively.
Across tasks, we observe notable differences in learning behaviour. The expert learns some tasks already after a few thousand interaction steps, while it requires more interaction steps for others. For some it does not exhibit any learning progress at all (e.g. \emph{assembly-v2}).  Overall, the expert achieves average success rates of 84\% and 100\% on MT40 and CW10, respectively.

\subsection{DMControl}
For each of the 16 considered DMControl tasks, we train for 1M steps. Therefore, the final DMControl datasets contain 1M transitions per task, amounting to 16M transitions in total (10M for DMC10, 6M for DMC6). All trajectories are of the same length and contain 1000 timesteps.

We utilize the same network architecture as for Meta-World (Section \ref{appendix:data-metaworld}), using 1024 neurons per layer instead of 256. Furthermore, we perform one gradient step per interaction step and keep all other hyper-parameters the same as on Meta-World.

\begin{figure}
  \centering
    \subfigure{\includegraphics[width=0.19\textwidth]{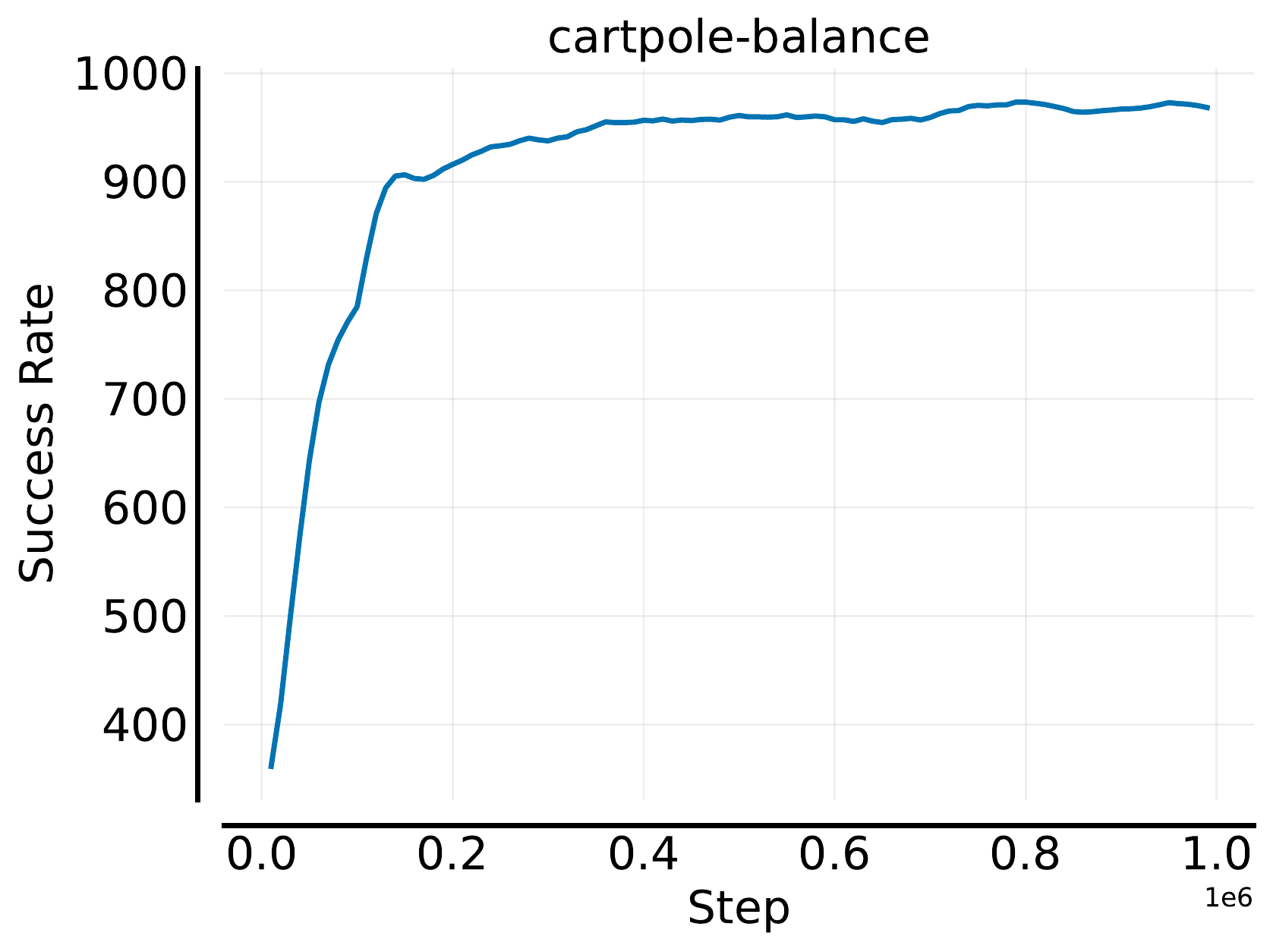}}
    \subfigure{\includegraphics[width=0.19\textwidth]{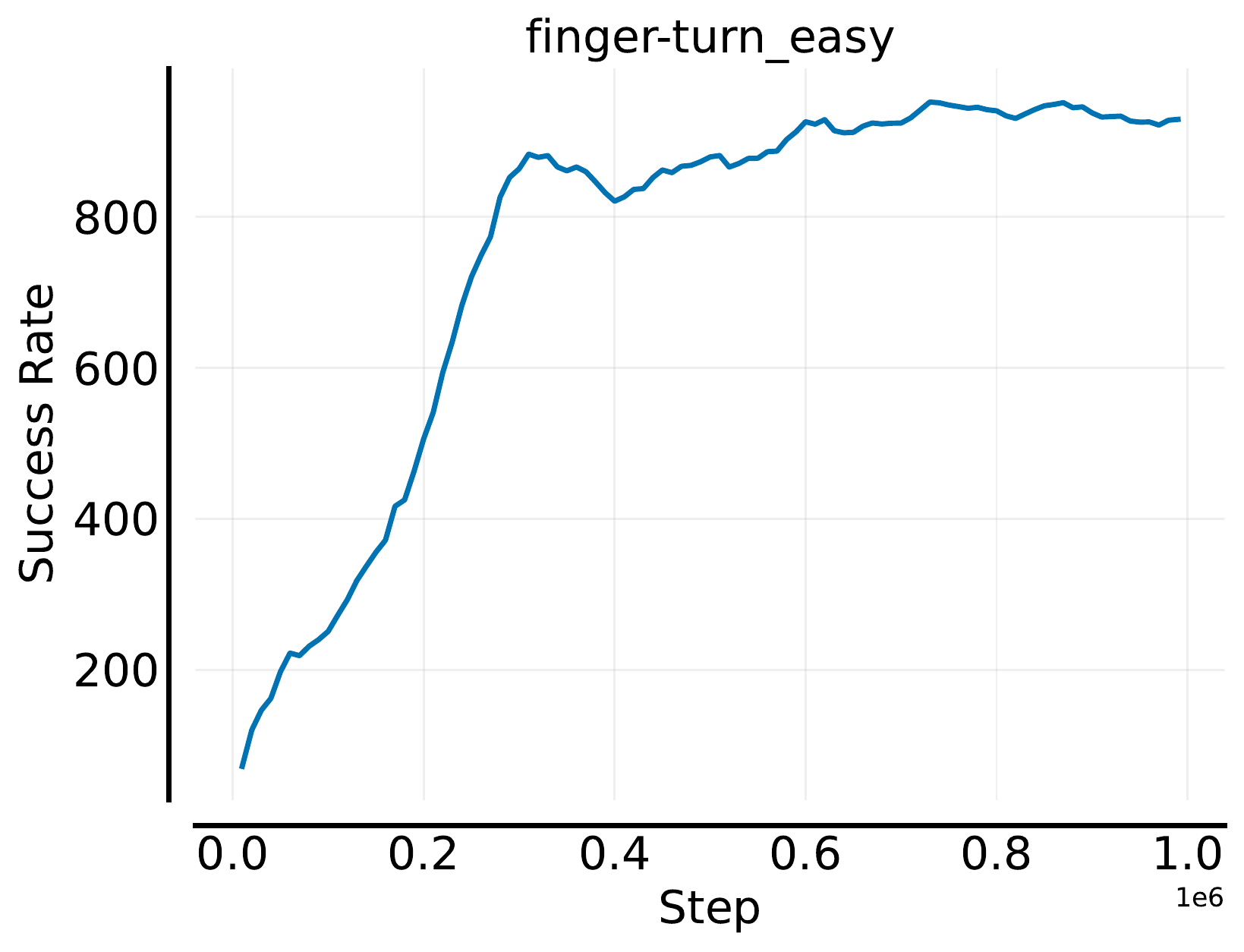}}
    \subfigure{\includegraphics[width=0.19\textwidth]{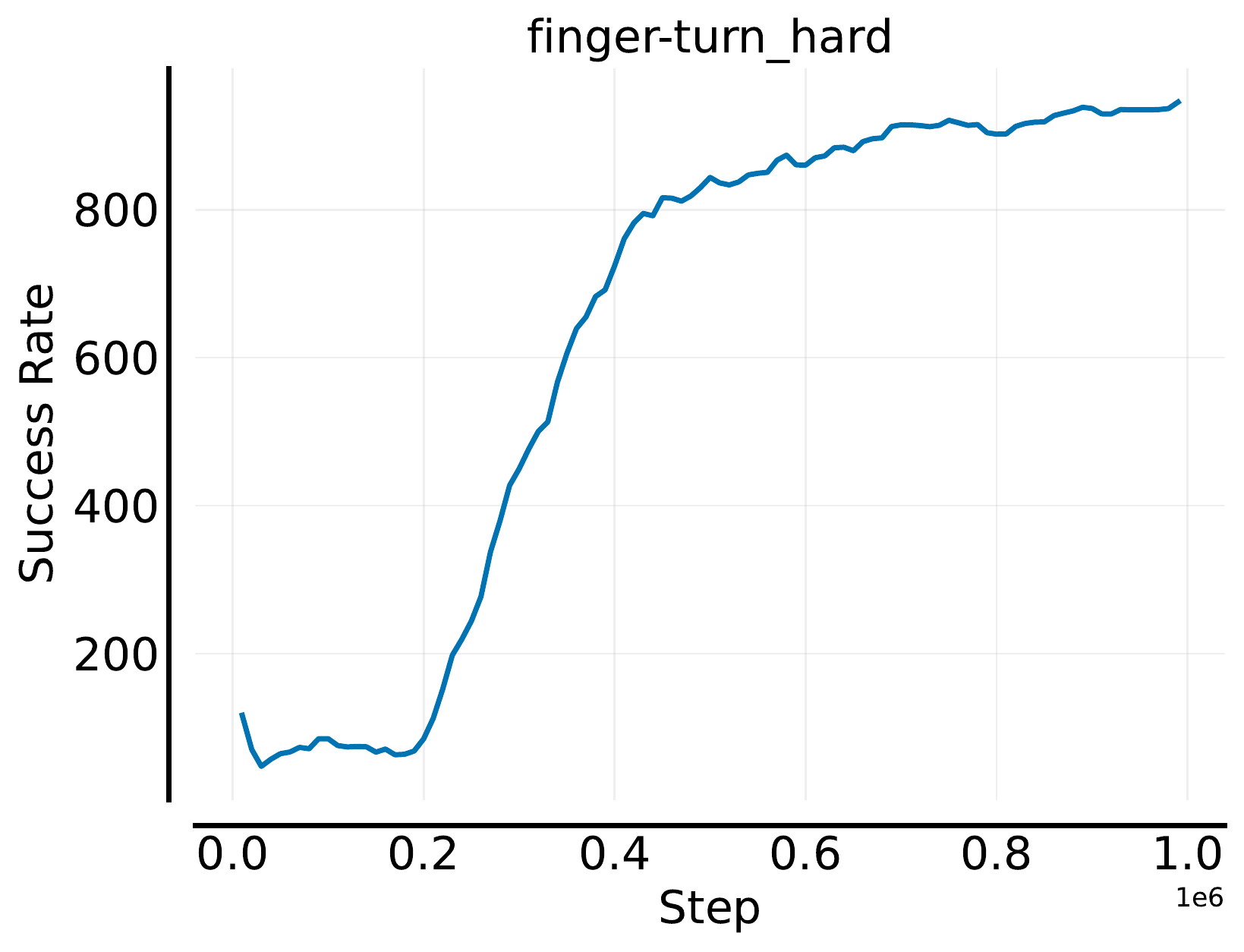}}
    \subfigure{\includegraphics[width=0.19\textwidth]{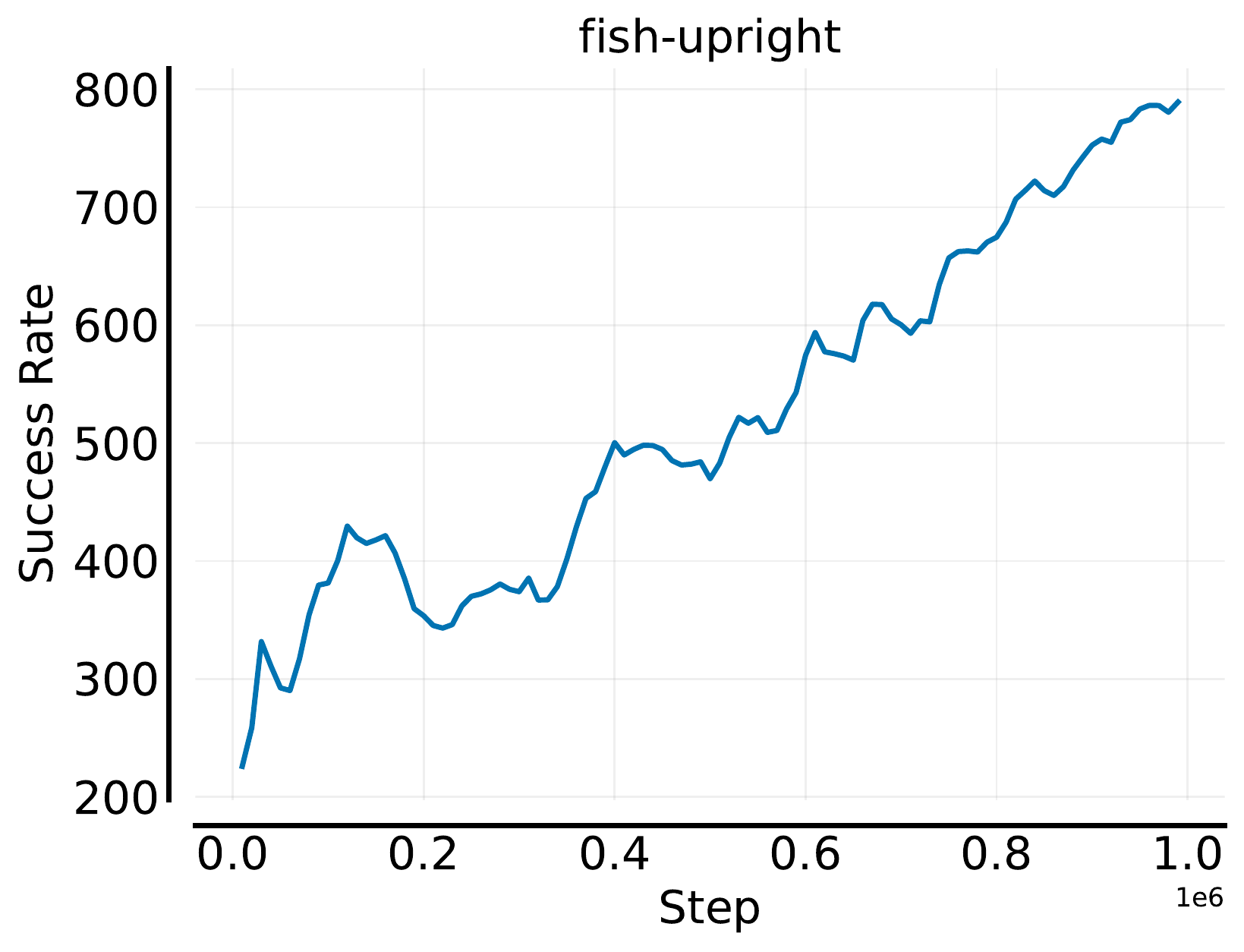}}
    \subfigure{\includegraphics[width=0.19\textwidth]{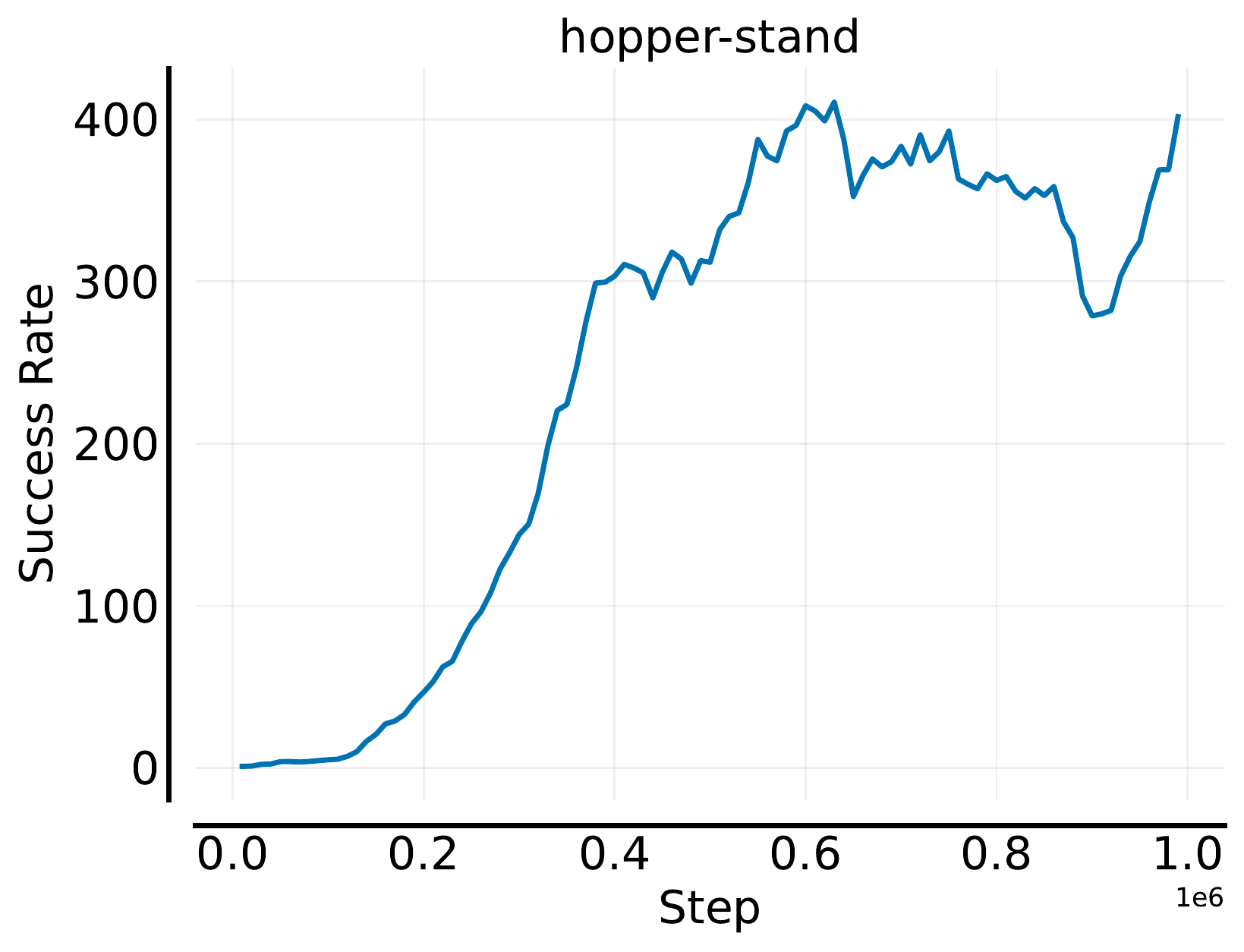}}
    
    \subfigure{\includegraphics[width=0.19\textwidth]{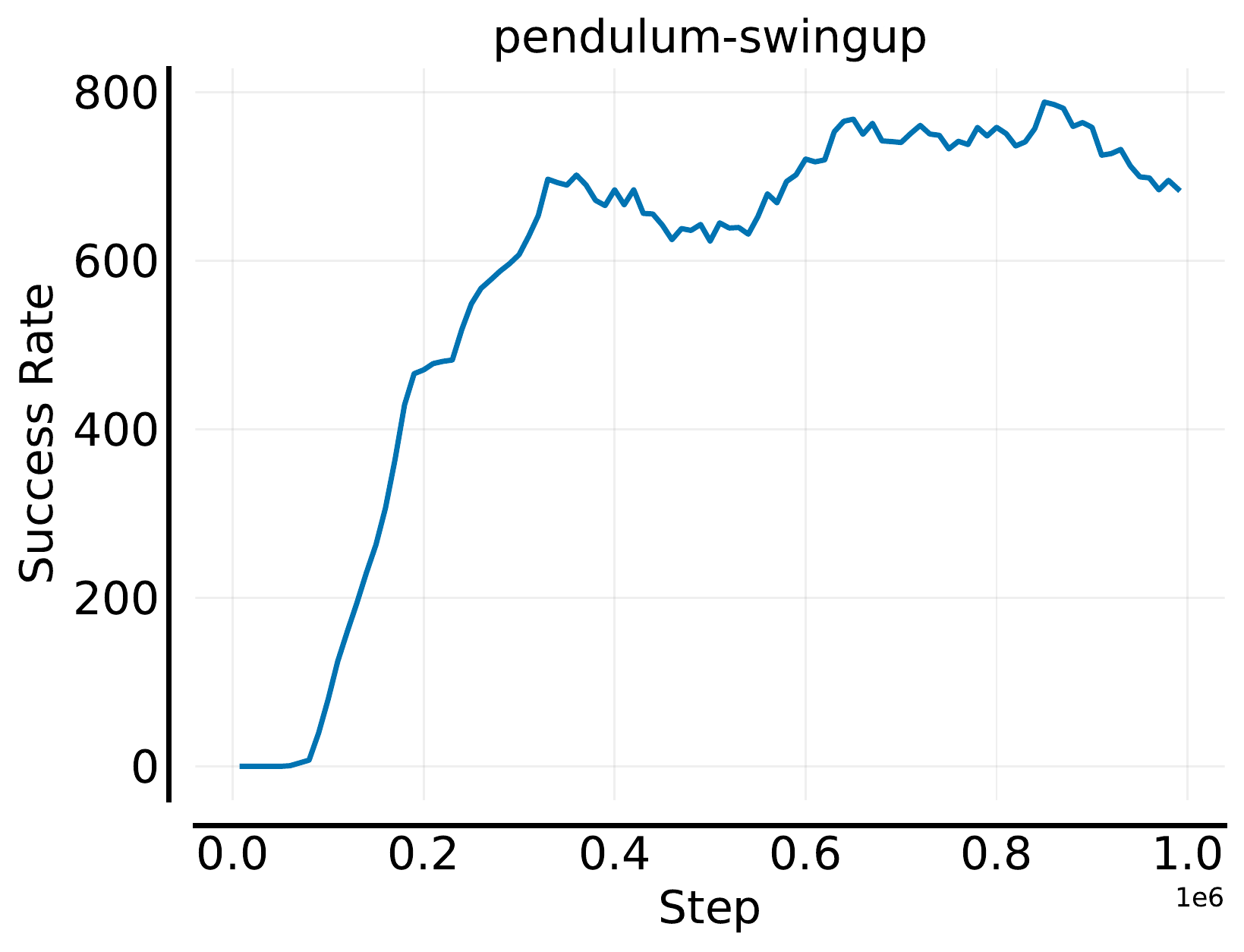}}
    \subfigure{\includegraphics[width=0.19\textwidth]{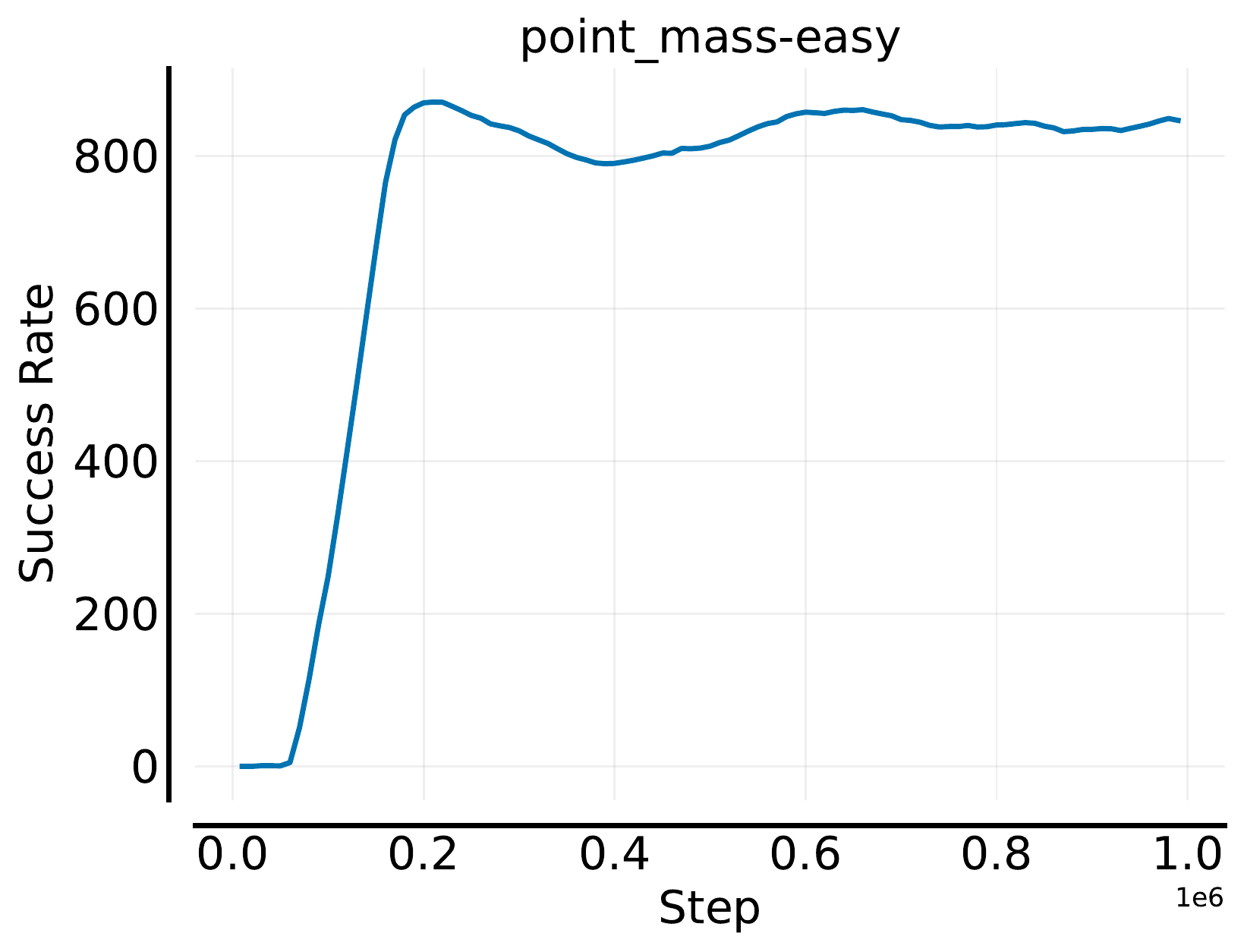}}
    \subfigure{\includegraphics[width=0.19\textwidth]{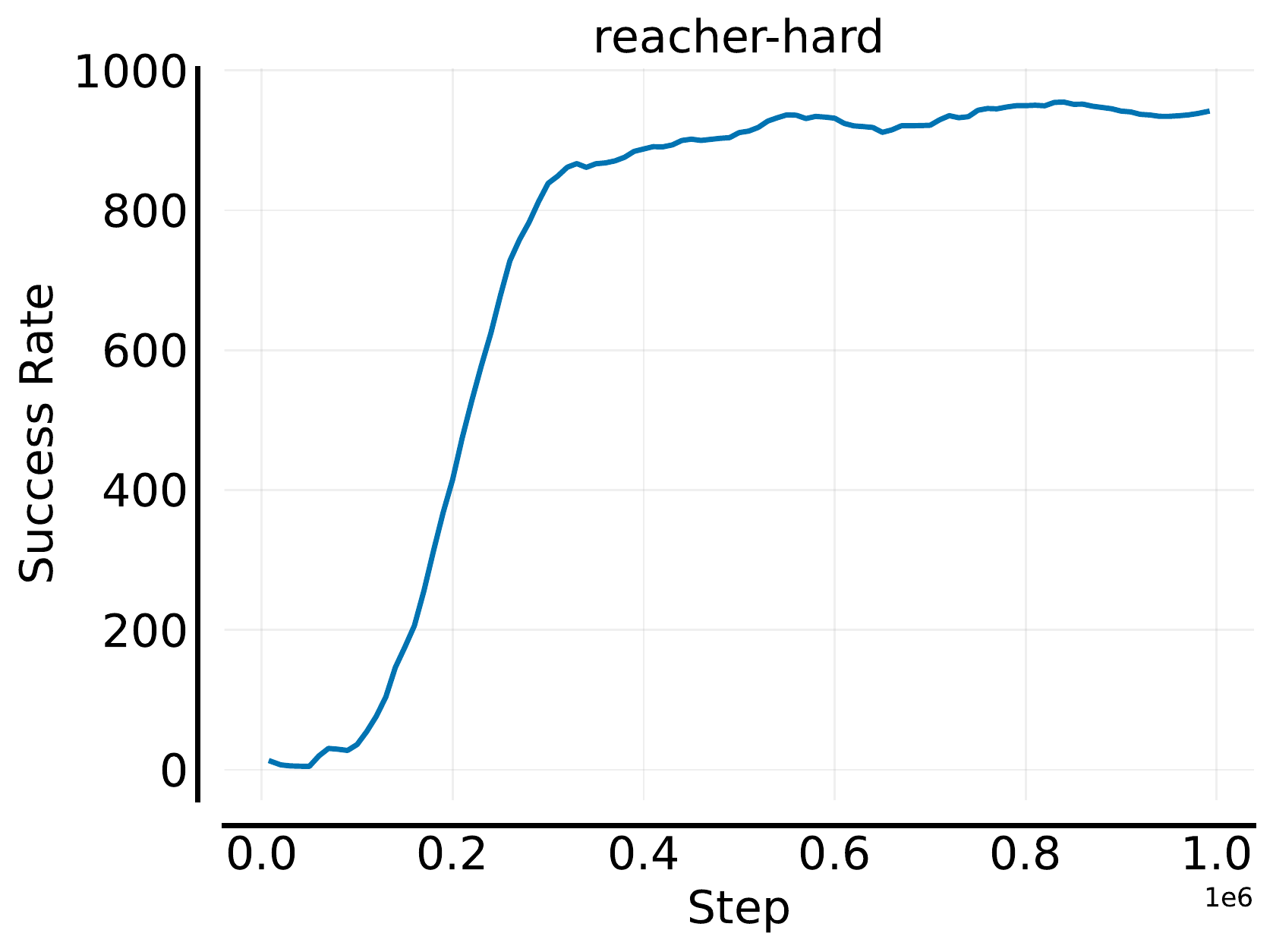}}
    \subfigure{\includegraphics[width=0.19\textwidth]{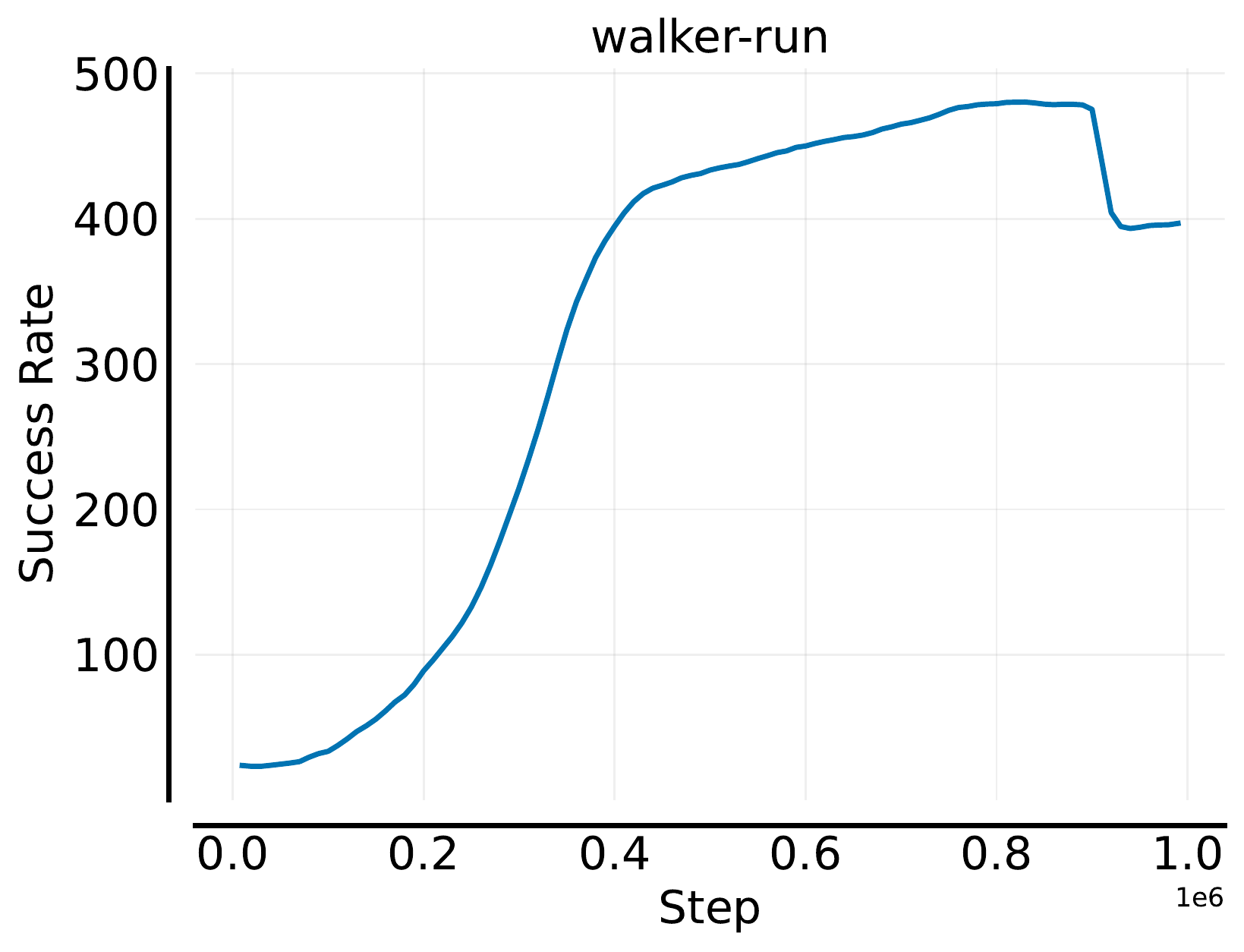}}
    \subfigure{\includegraphics[width=0.19\textwidth]{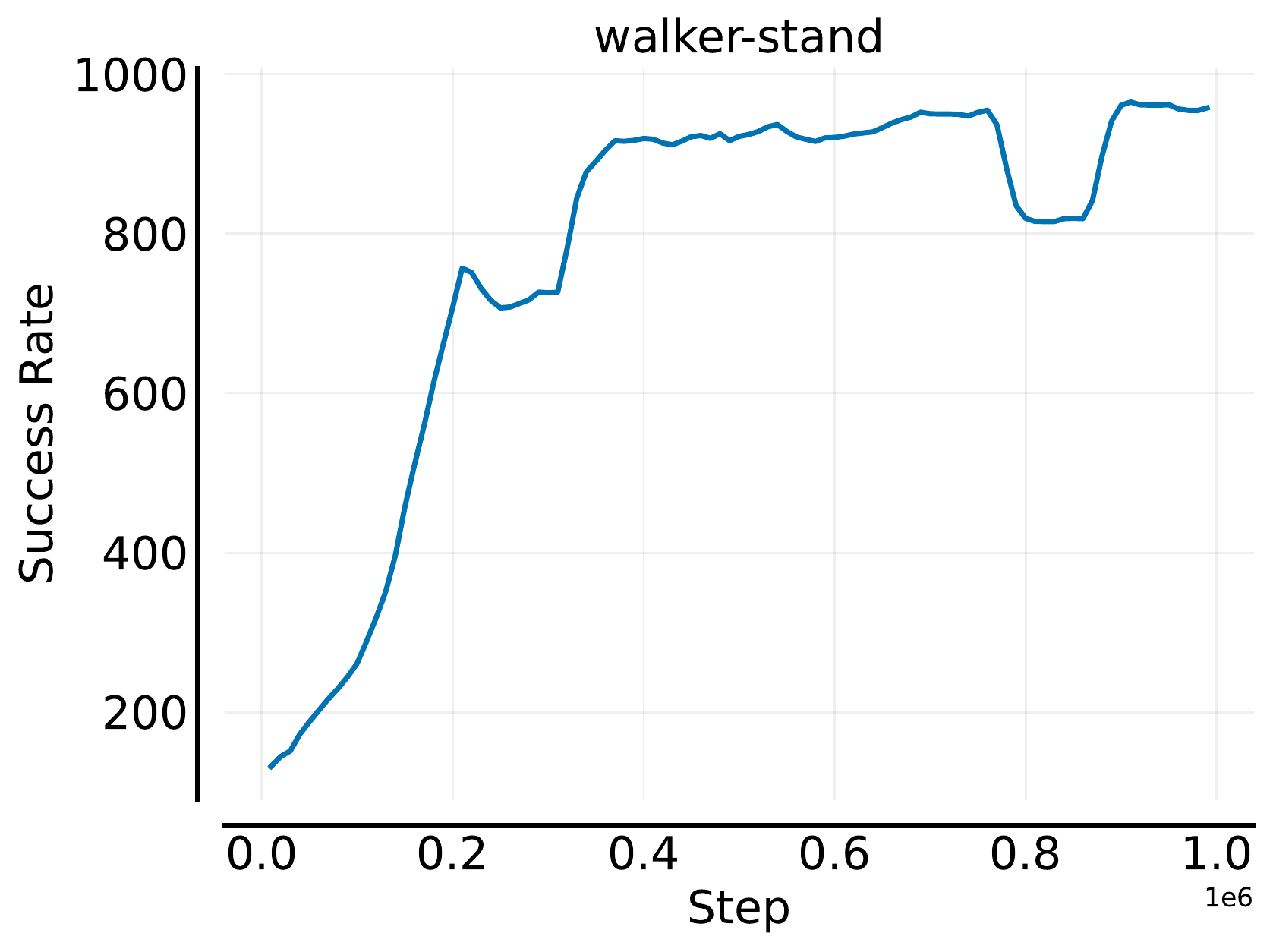}}
  \caption{Learning curves for data-collection runs on all DMC10 tasks with SAC. We train for 1M environment interaction steps on each task and record the entire replay buffer.} 
  \label{fig:appendix-datacollection-dmc10}
\end{figure}

We list the average rewards achieved by the expert agents on DMC10 and DMC6 tasks in Table \ref{tab:dmc10-dmc6}. The corresponding learning curves for all DMC10 tasks are available in Figure \ref{fig:appendix-datacollection-dmc10}. On both splits, the agent performs close to the best possible return of 1000, achieving a mean reward of 788 on DMC10 and 840 on DMC6.
\begin{table}[h]
\centering
    \caption{Performance scores for data collection on \textbf{(a)} DMC10 and \textbf{(b)} DMC6.}
    \subtable[DMC10]{
        \centering
        \begin{tabular}{l c c c}
        \toprule
        \textbf{Task} & $|\mathcal{S}|$ & $|\mathcal{A}|$ & \textbf{Reward} \\
        \midrule
        cartpole-balance &  5 & 1 &  967.49 \\
        finger-turn\_easy & 12 & 2 &   940.57 \\
        finger-turn\_hard & 12 & 2 &   940.15 \\
            fish-upright &  21 & 5 &  787.87 \\
            hopper-stand &  15 & 4 &  417.49 \\
        pendulum-swingup &  3 & 1 &  682.85 \\
         point\_mass-easy & 4 & 2 &   842.03 \\
            reacher-hard &  6 & 2 &  945.18 \\
              walker-run &  24 & 6 &  402.36 \\
            walker-stand &  24 & 6 &  957.61 \\
            \midrule
                Average & - & - & 788 ± 219 \\
        \bottomrule
        \end{tabular}
    }
    \subtable[DMC6]{
        \centering
        \begin{tabular}{l c c c}
            \toprule
            \textbf{Task} & $|\mathcal{S}|$ & $|\mathcal{A}|$ &  \textbf{Reward} \\
            \midrule
    ball\_in\_cup-catch & 8 & 2 &     971.1 \\
     cartpole-swingup &   5 & 1 &  823.48 \\
          cheetah-run &   17 & 6 &  411.76 \\
          finger-spin &   9 & 2 &  961.85 \\
         reacher-easy &   6 & 2 &  890.03 \\
          walker-walk &   24 & 6 &  928.84 \\
          \midrule
              Average & - & - & 840 ± 216 \\
            \bottomrule
            \end{tabular}
     }
    \label{tab:dmc10-dmc6}
\end{table}

\begin{figure}
  \centering
    \subfigure{\includegraphics[width=0.19\textwidth]{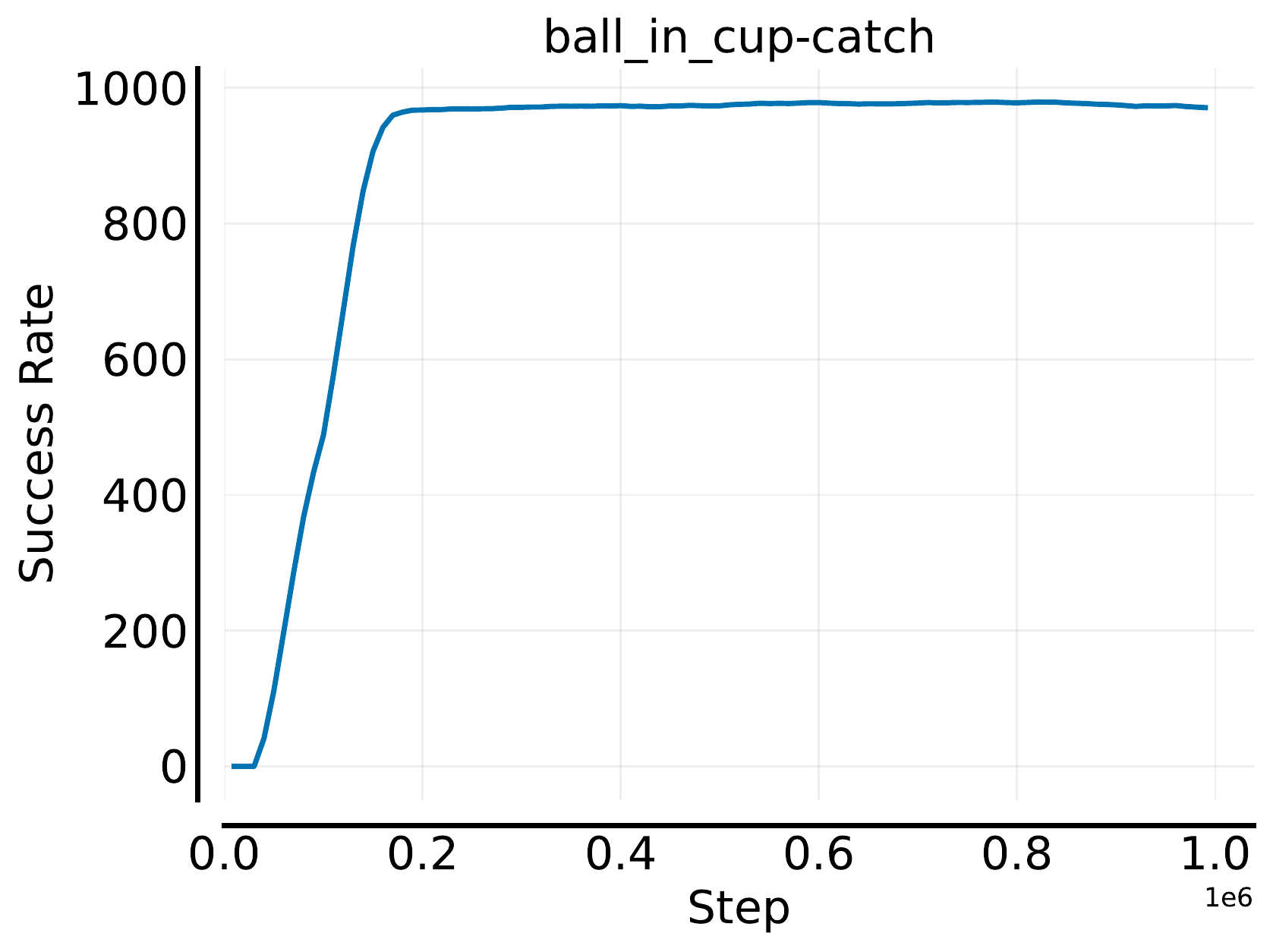}}
    \subfigure{\includegraphics[width=0.19\textwidth]{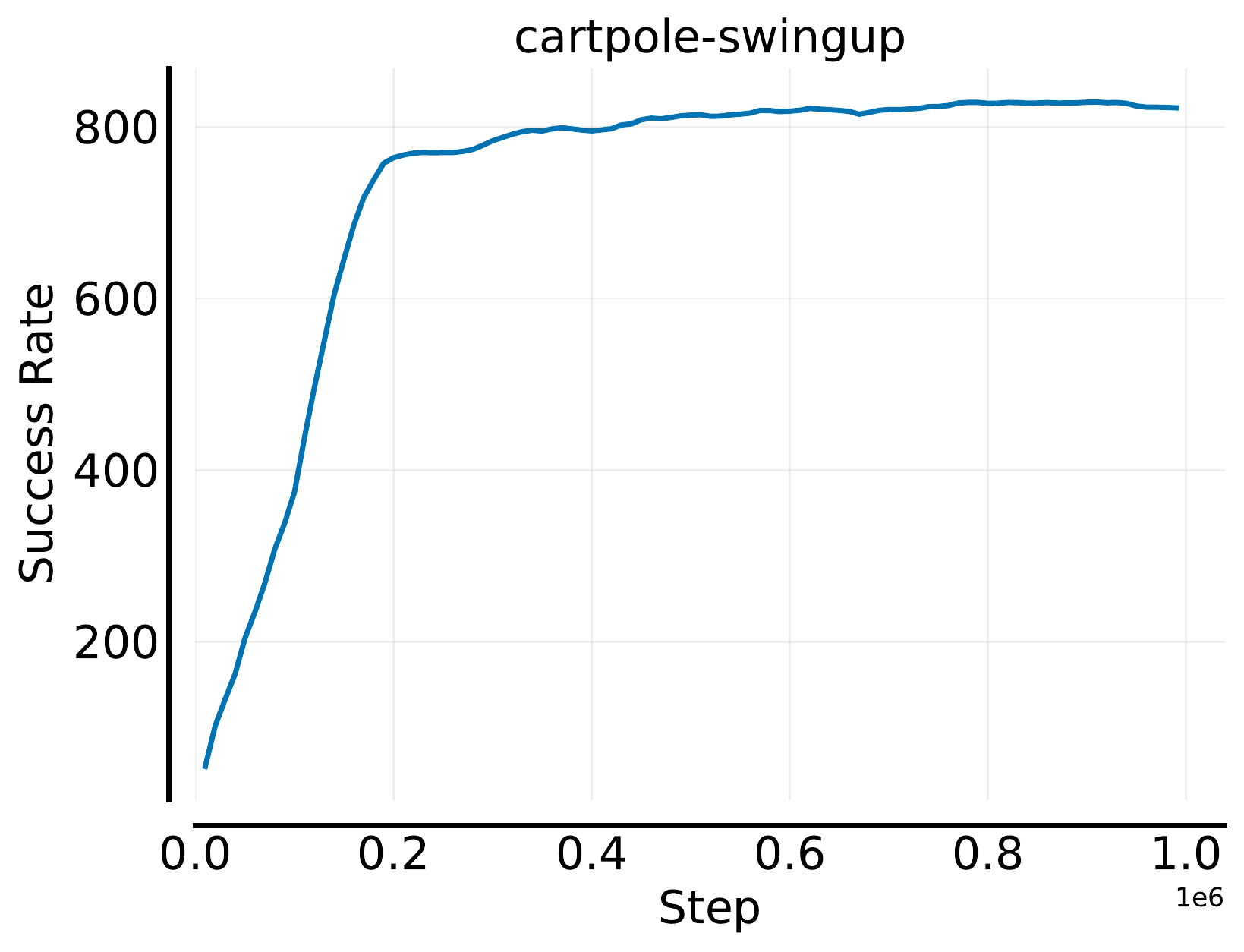}}
    \subfigure{\includegraphics[width=0.19\textwidth]{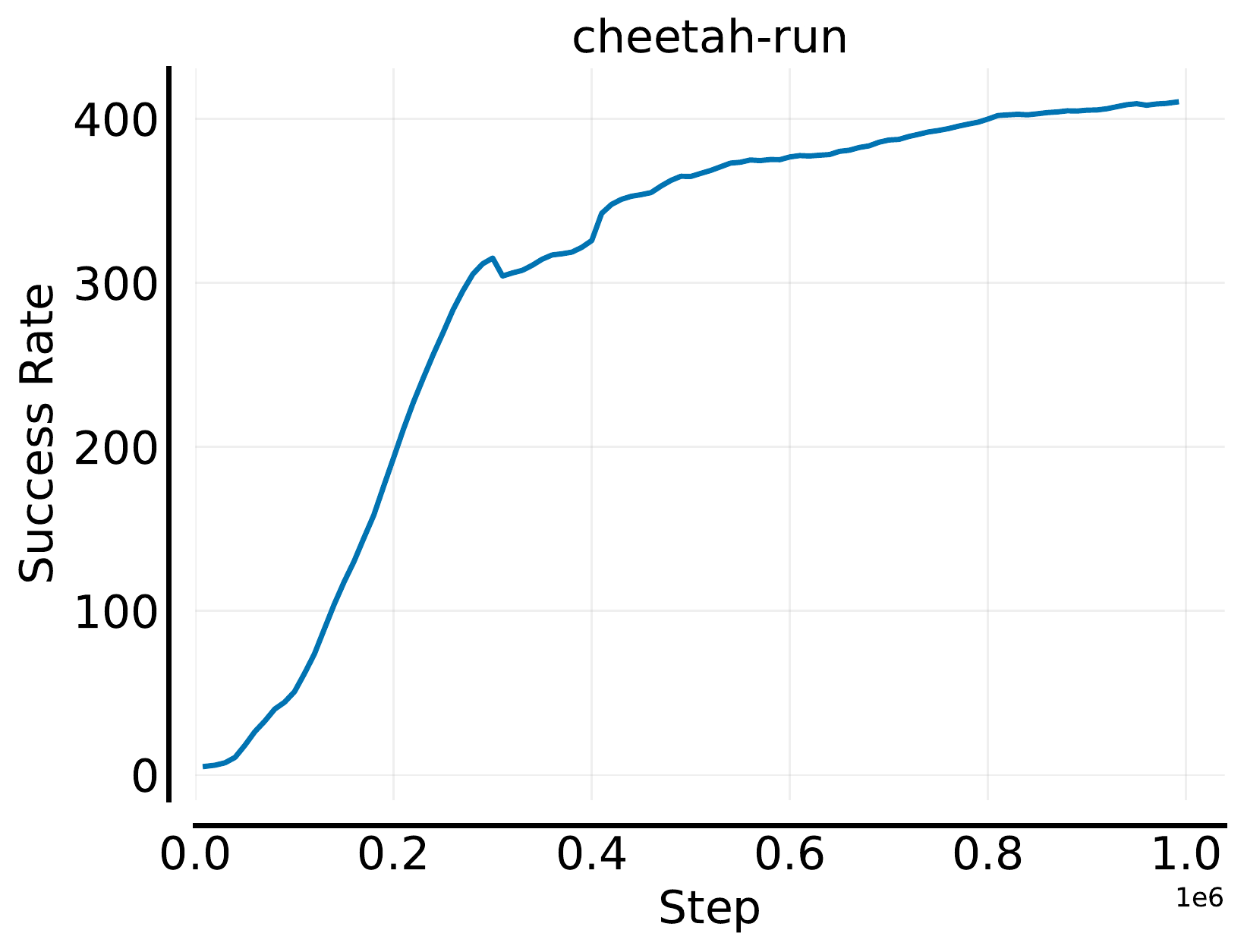}}

    \subfigure{\includegraphics[width=0.19\textwidth]{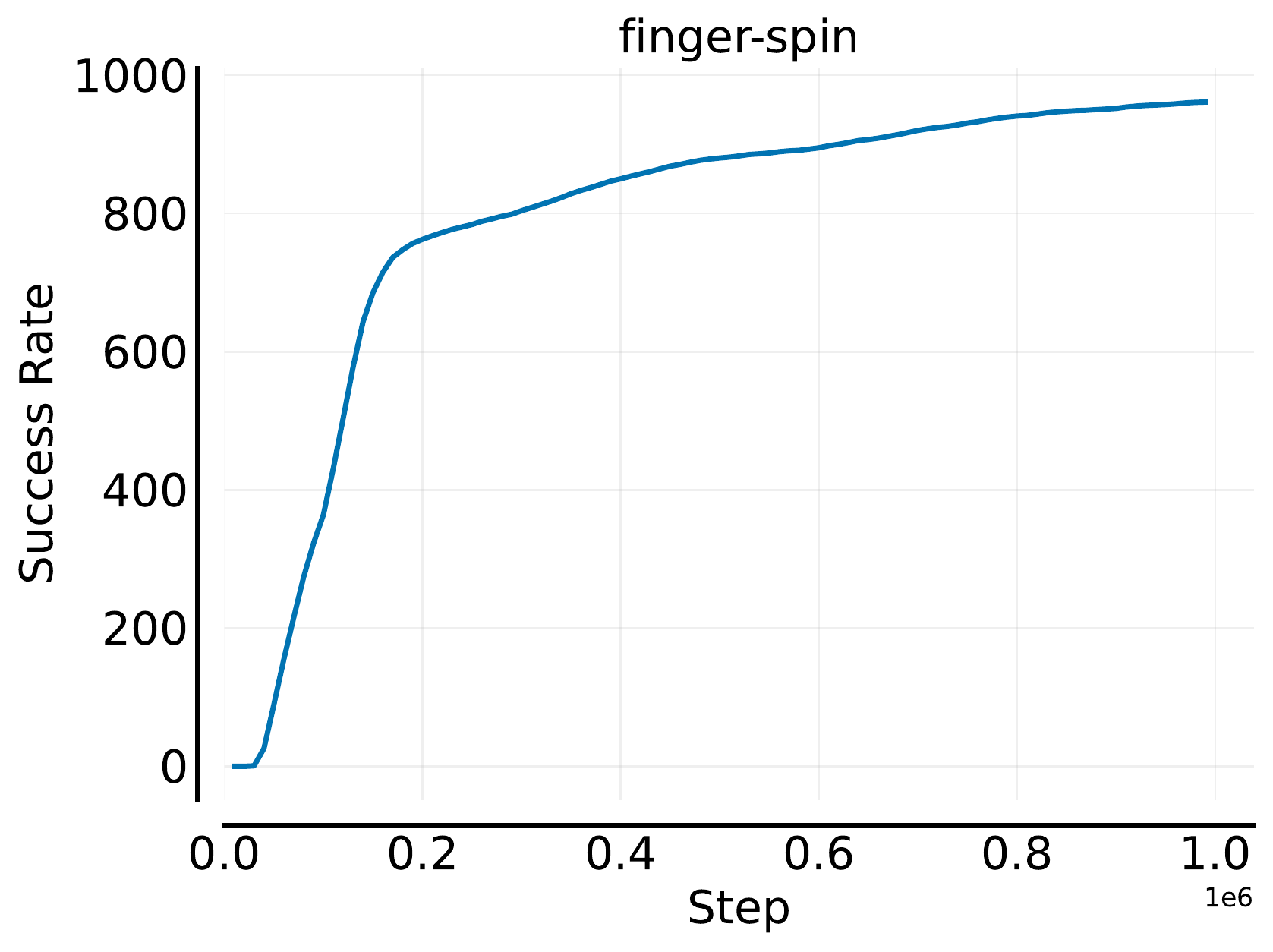}}
    \subfigure{\includegraphics[width=0.19\textwidth]{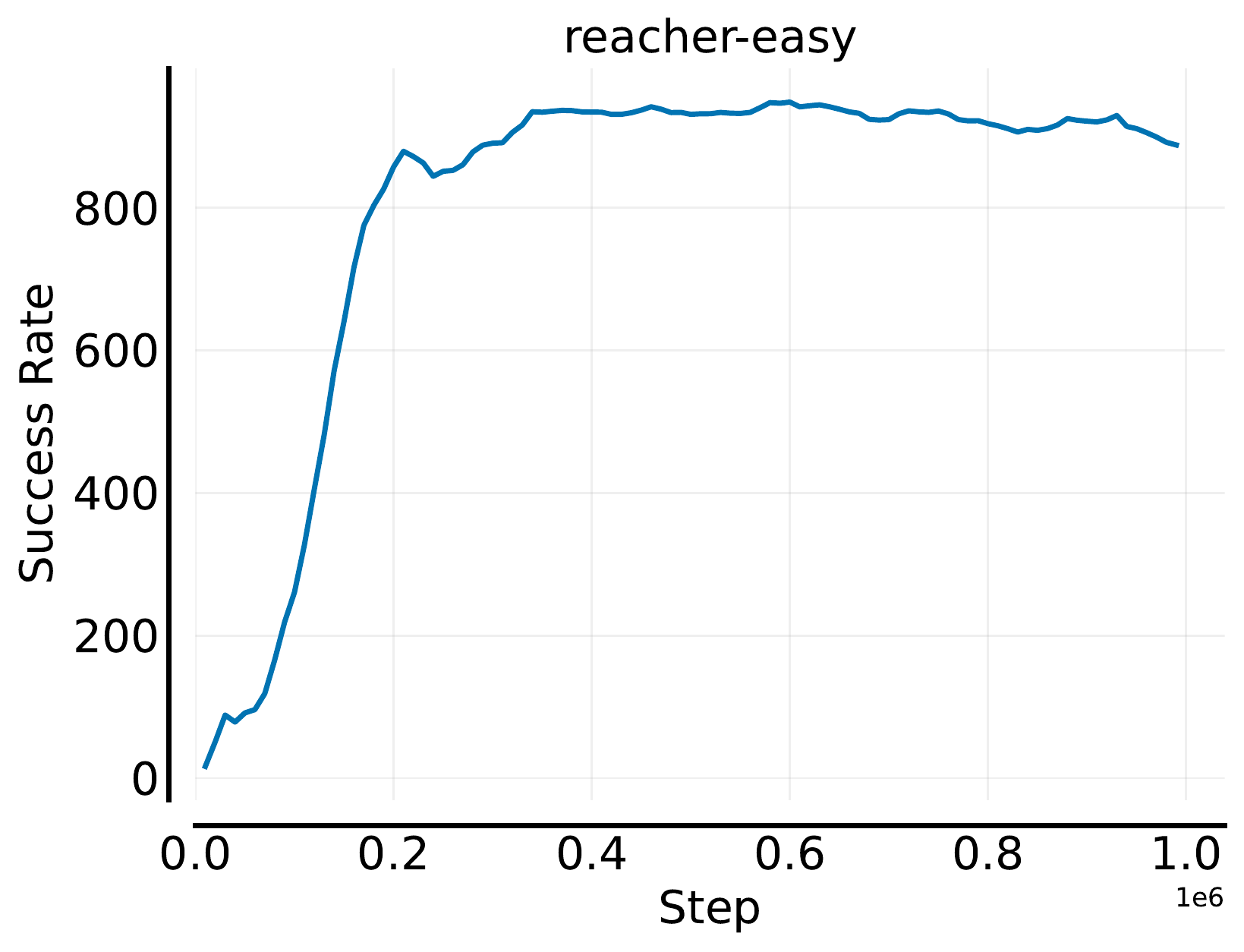}}
    \subfigure{\includegraphics[width=0.19\textwidth]{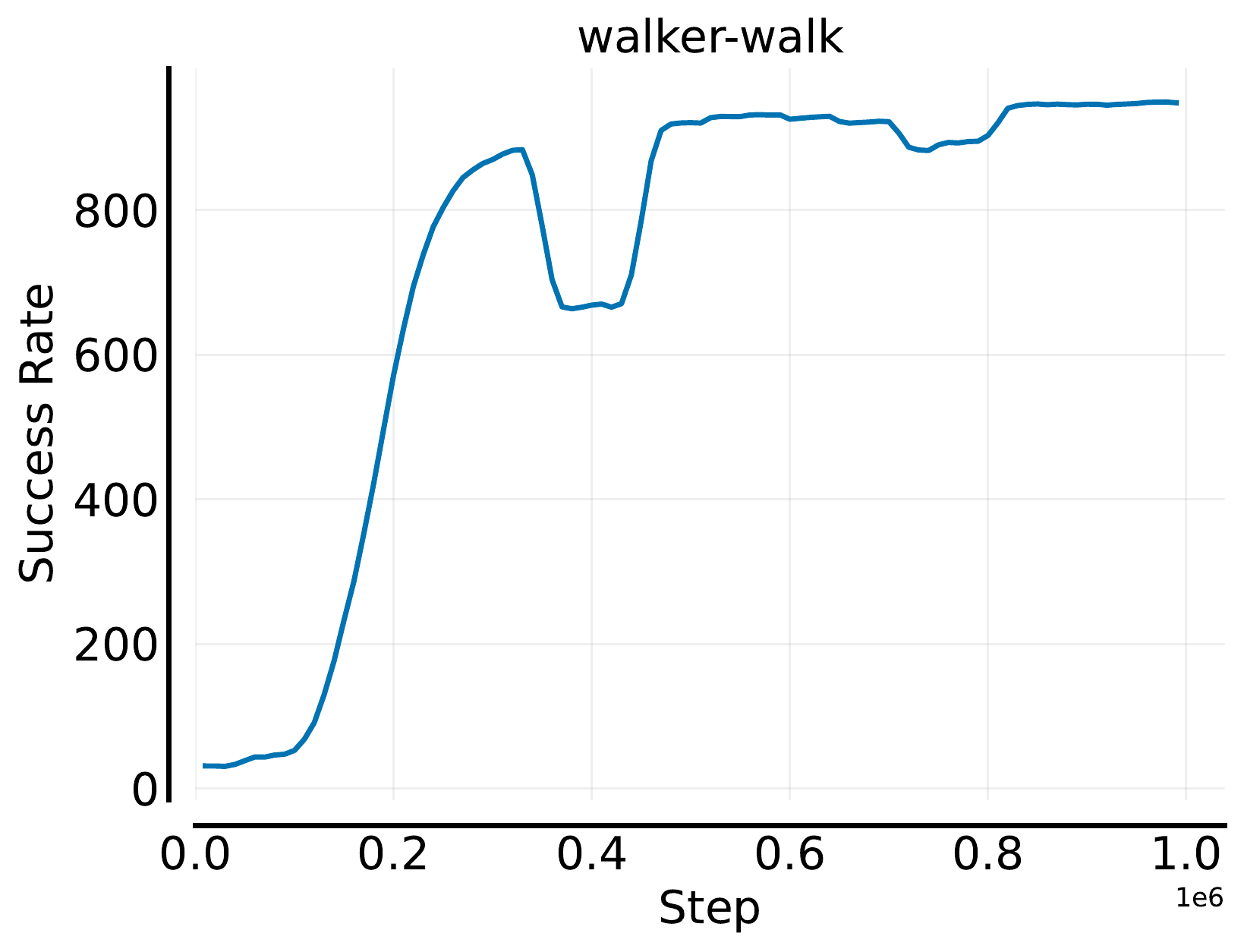}}
  \caption{Learning curves for data-collection runs on all DMC6 tasks with SAC. We train for 1M environment interaction steps on each task and record the entire replay buffer.} 
  \label{fig:appendix-datacollection-dmc6}
\end{figure}

\section{Pre-training}\label{appendix:pre-training}
We train our MDDT for a total of 1M update steps, with context length of 5 transitions (45 tokens). We use a learning rate of $1e^{-4}$ and 4000 linear warm-up steps, followed by a cosine decay to $1e^{-6}$. Furthermore, we use gradient clip of 0.25, weight decay of 0.01, dropout of 0.2, a batch size of 1024 sequences and train using the AdamW optimizer \citep{Loshchilov:18}. We base our implementation on the DT implementation in the \texttt{transformers} library \citep{Wolf:20}, and keep their default values for remaining parameters.

\begin{table}[]
    \centering
    \caption{Pre-training scores for multi-domain DT variants (40M-200M) trained on MT40 and DMC10. We vary the number of layers, the number of heads, and the embedding dimension. We report success rates for MT40, normalized scores for DMC10, and mean rewards across both domains.}
    \begin{tabular}{c c c c c  c r}
    \toprule
     \textbf{Layers} &  \textbf{Heads} &  \textbf{Embedding Dim} & \textbf{Params} & \textbf{MT40} & \textbf{DMC10} & \textbf{Mean Reward} \\
    \midrule
      6 &     12 &            768 &  40M  &   0.6 &   0.84 &  983.92 \\
      8 &     12 &            768 &  58M   &    0.6 &   0.93  &  1005.6  \\
      8 &     16 &           1024 &  102M   &   0.61 &   1.21  & 1055.75 \\
      10 &     20 &           1280 &  200M   &   0.6 &   1.33  &  1030.1\\
    \bottomrule
    \end{tabular}
    \label{tab:appendix-pretrain}
\end{table}

We use the same underlying GPT-2 like network architecture as \citet{Lee:22}. However, we do not use their proposed expert-action inference mechanism. Instead, we set the target return to the maximum return in the respective dataset and use a reward scale of 200 and 100 for Meta-World and DMControl, respectively. Furthermore, we use timestep-based positional embeddings, as described in Appendix \ref{appendix:mddt}.

\begin{figure}
    \centering
    \includegraphics[width=0.7\textwidth]{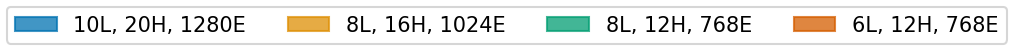}
    \subfigure[MT40]{
        \includegraphics[width=0.75\textwidth]{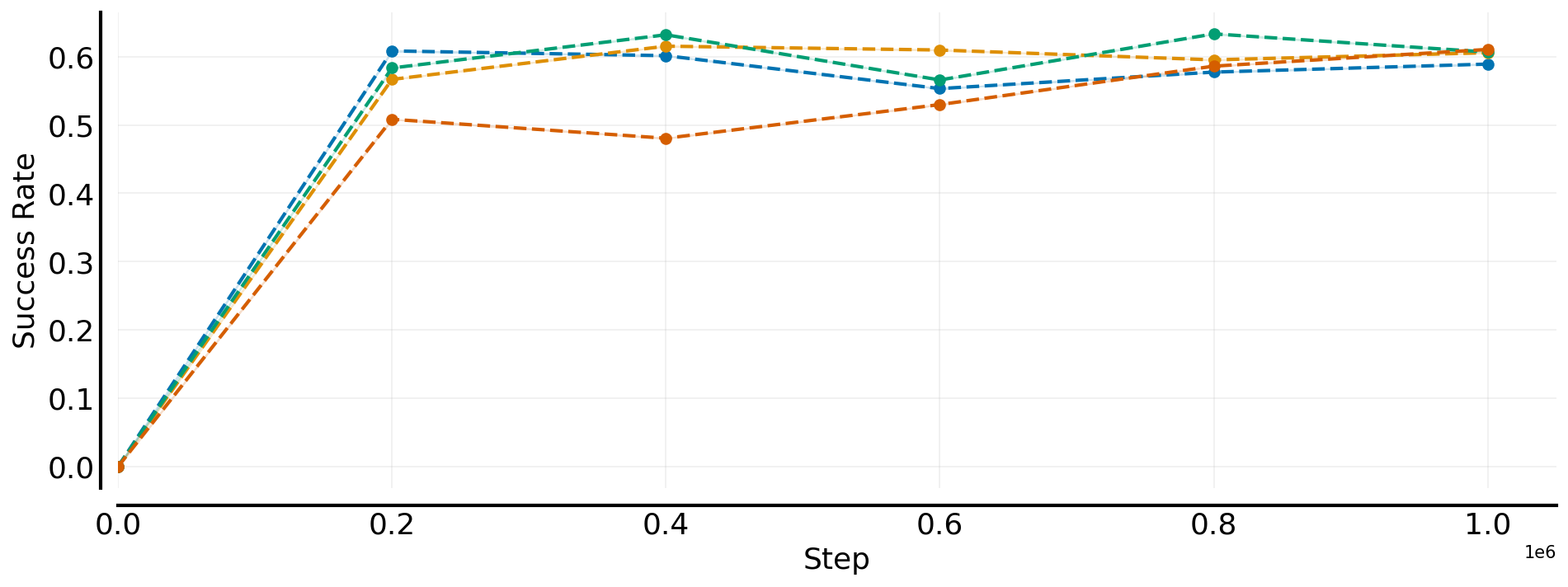}    
    }
    \subfigure[DMC10]{
        \includegraphics[width=0.75\textwidth]{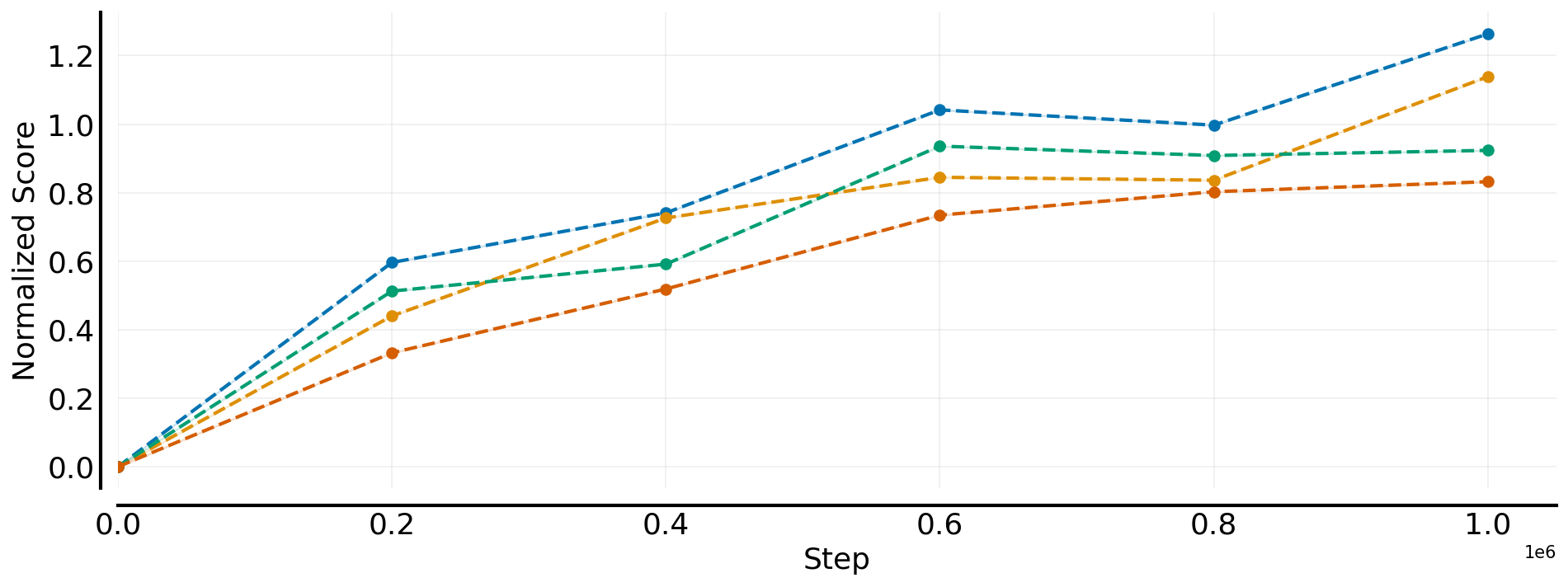}    
    }
    \caption{Learning curves for multi-domain pre-training runs on MT40 and DMC10 with different architectural choices of the Transformer network (L = \# of layers, H = \# of heads, E = embedding dimension). We train every model variant for 1M update steps with a batch size of 1024.}
    \label{fig:appendinx-pretrain-lc}
\end{figure}

In Table \ref{tab:appendix-pretrain} and Figure \ref{fig:appendinx-pretrain-lc}, we show the aggregate scores over all tasks at the end of training, and learning curves across different model sizes, respectively. %
We vary the number of layers, the number of heads, and the embedding dimension. We also experimented with higher context lengths, but only found slight performance gains at higher computational cost. Due to the excessive computational training cost, we only train one model per size. Generally, there is a trend that more complex models learn faster and attain higher performance on DMC10. 
However, for all our subsequent experiments, we used the model with 40M parameters, as it achieves a good trade-off between performance and training time. 
After pre-training, we evaluate this model across 3 environment seeds and attain an average success rate of 60\% ($\pm 0.5\%$) on MT40 and a normalized score of 94\% ($\pm 11\%$) on DMC10.  

\section{Task Separation Analysis}\label{appendix:task-sep}
To better understand what the DT learned, we visualize token embeddings using t-SNE \citep{Van:08}. First, we sample 256 state-RTG-action-reward sequences of length 5 (i.e., 45 tokens) for each task and record the hidden states of each (frozen) Transformer block. For each sequence, we then average the hidden states over the sequence, either across the entire sequence or individually per token-type (i.e., for states, RTGs, actions, rewards). 
We use the 40M pre-trained MDDT for our analysis. 
Thus, we end up with 256 768-dimensional vectors per task. We cluster all vectors using t-SNE projecting into a two-dimensional space with a perplexity of 30, 10K optimization steps, and the cosine distance. 

\begin{figure}
  \centering
  \subfigure{\includegraphics[width=1\textwidth]{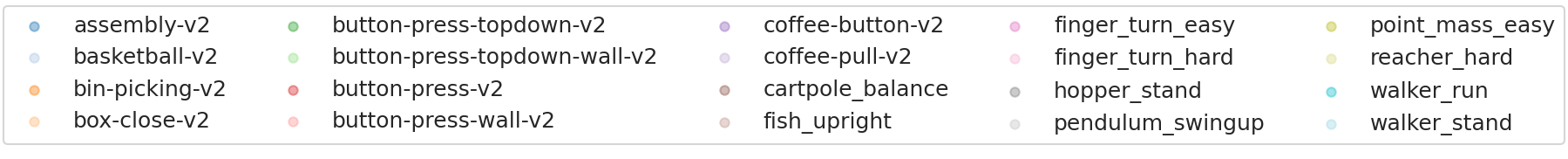}}
  \subfigure[States]{\includegraphics[width=0.4\textwidth]{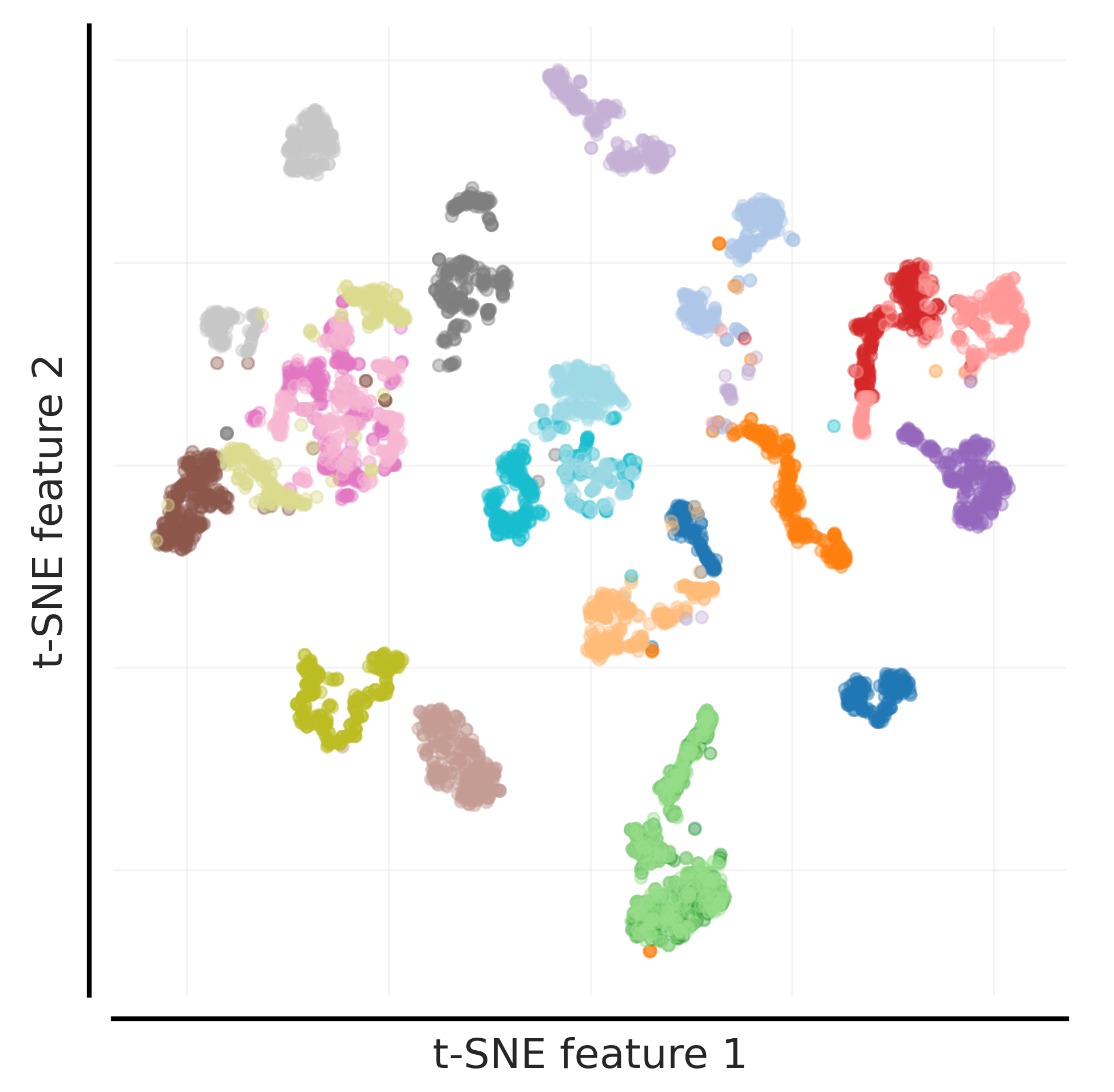}}
  ~
  \subfigure[Return-to-go]{\includegraphics[width=0.4\textwidth]{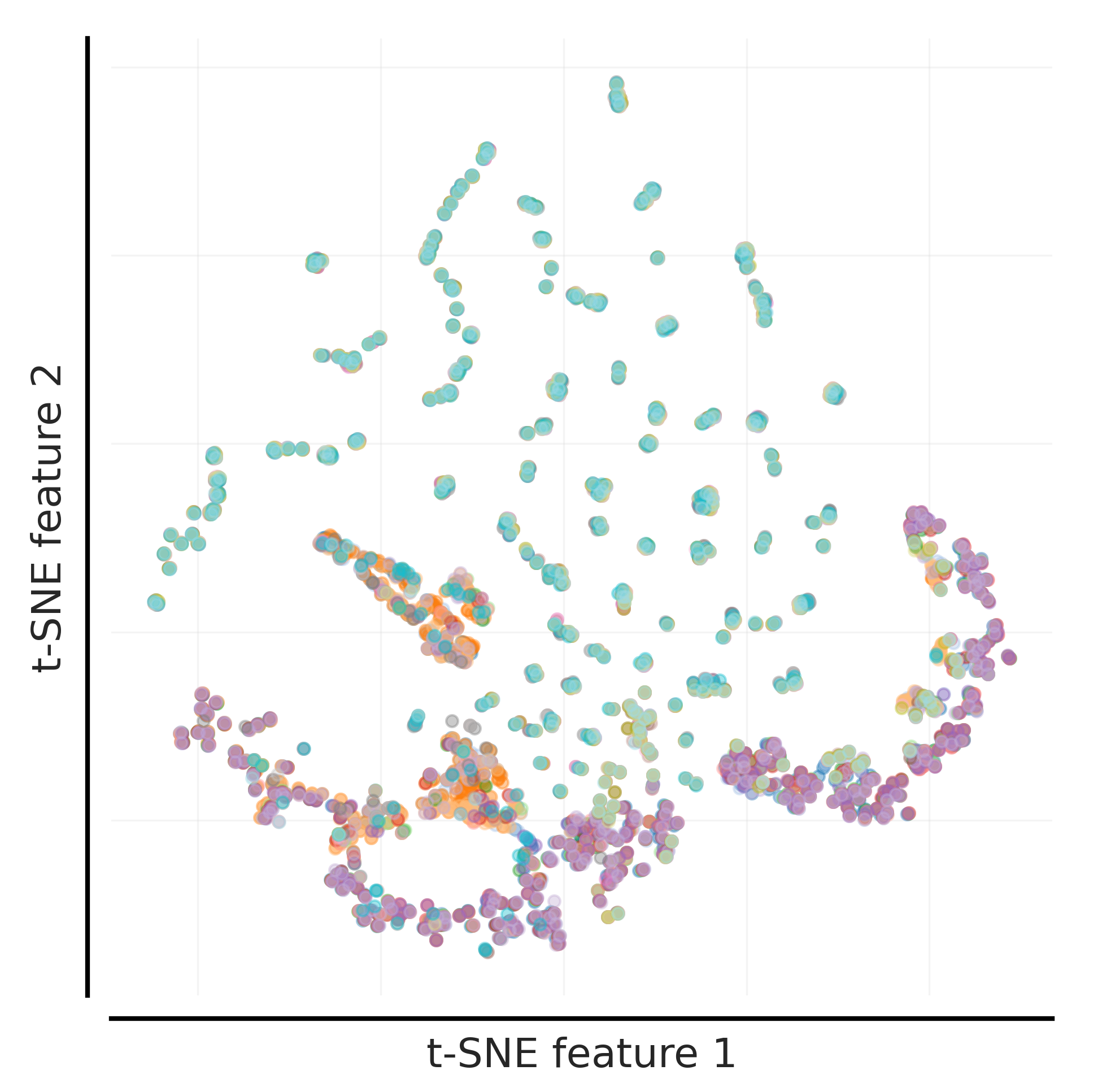}}

  \subfigure[Reward]{\includegraphics[width=0.4\textwidth]{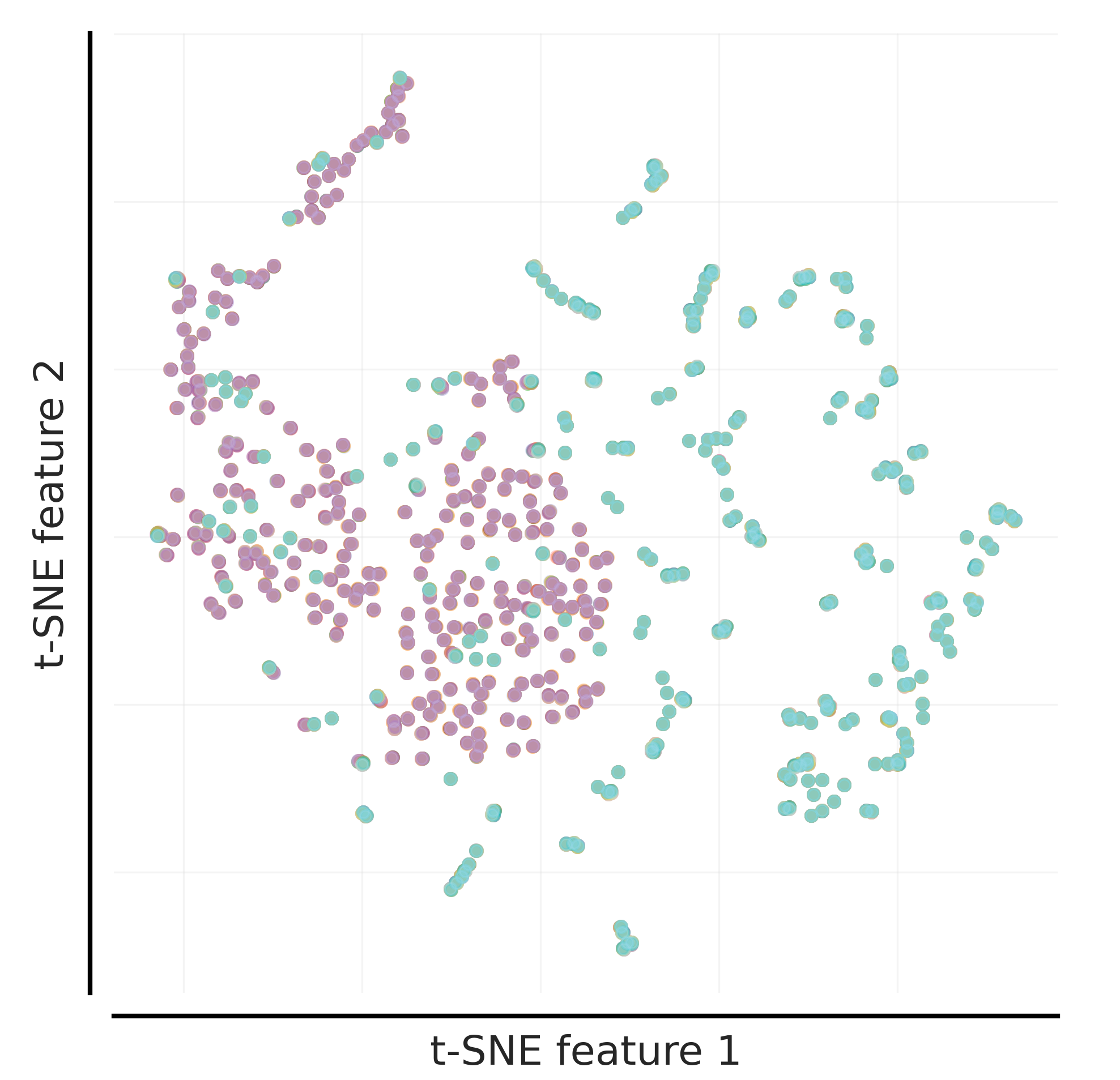}}
  ~
  \subfigure[Actions]{\includegraphics[width=0.41\textwidth]{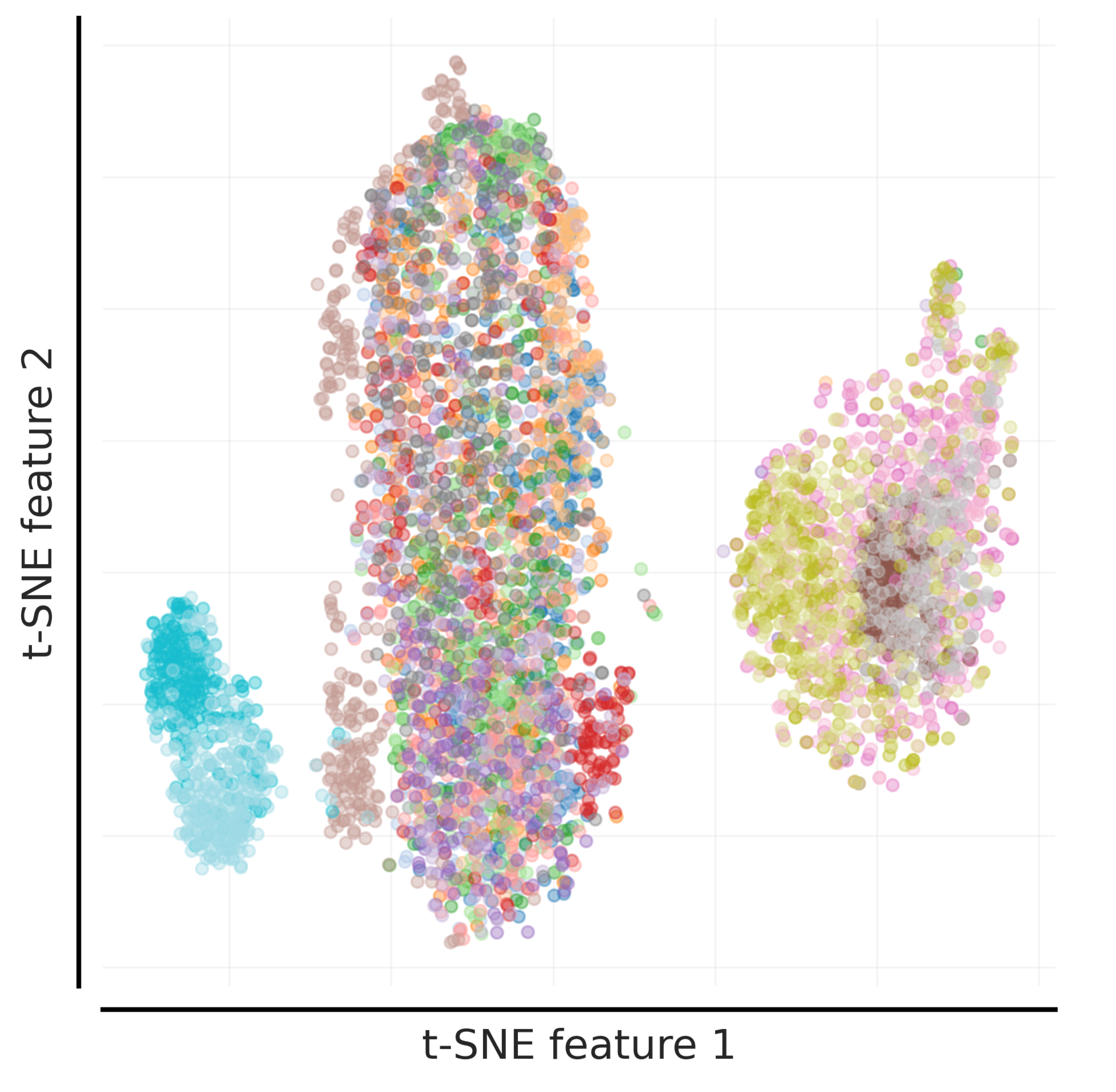}}

  \subfigure[All]{\includegraphics[width=0.4\textwidth]{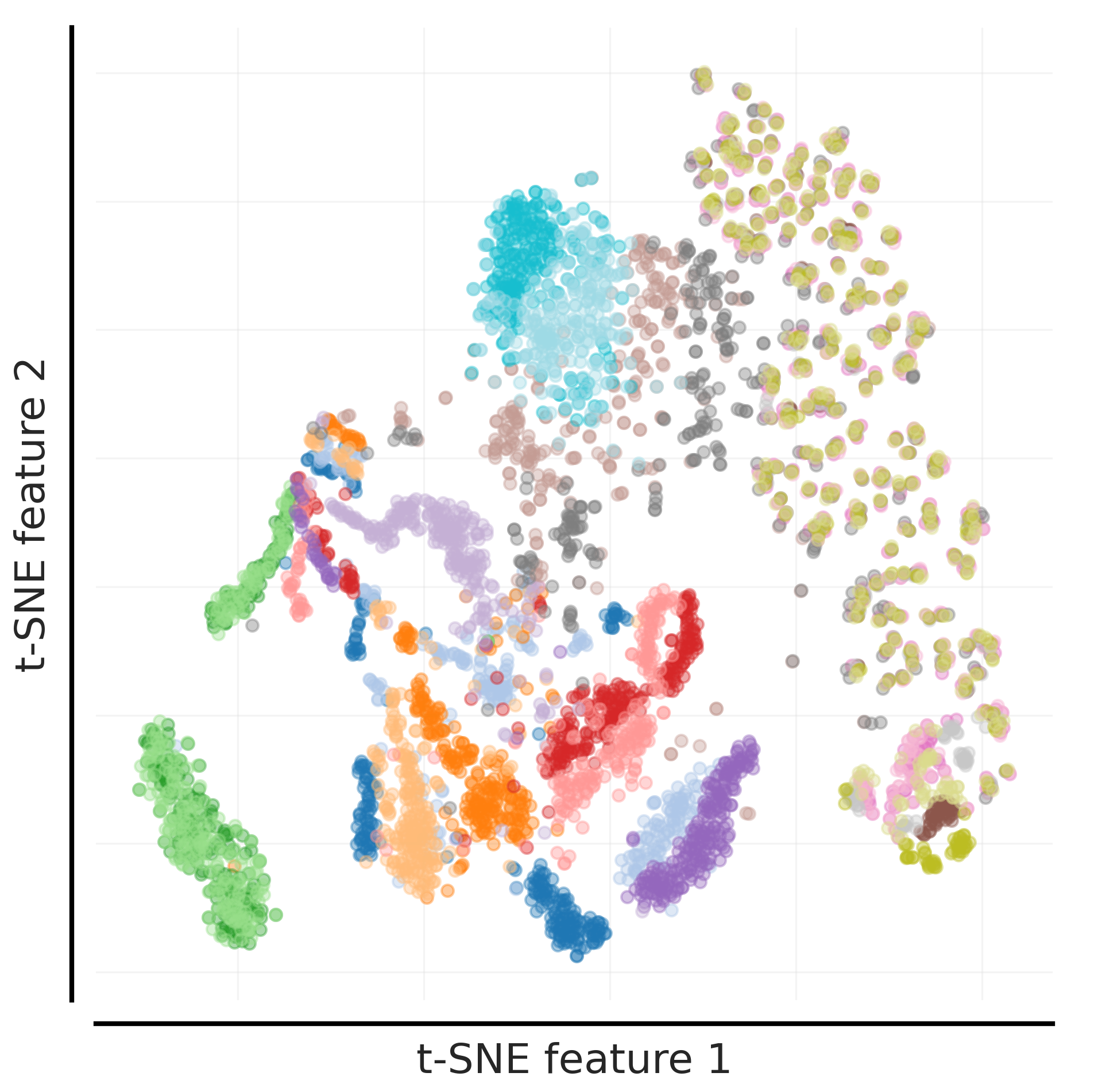}}
  \caption{t-SNE clustering for state, action, RTG, and reward embeddings individually, as well as for all token embeddings combined for the first ten tasks in MT40 and DMC10.} 
  \label{fig:appendix-tsne-tokens}
\end{figure}

In Figure \ref{fig:appendix-tsne-tokens}, we show the t-SNE visualizations of all tokens combined, as well as for state, action, RTG, and reward tokens individually on 10 tasks in MT40 and DMC10. For embedded state tokens, we observe good cluster separation. Also, for all tokens combined and actions some clusters can be observed, but no task-specific separation. For rewards and RTG, we do not observe distinct clusters. This is expected, since rewards and RTGs do not vary substantially among tasks.

\clearpage

\section{Single-Task Fine-Tuning}\label{appendix:adaptation}
We conduct a broad evaluation of different FT, PEFT and PBT methods to adapt the MDDT to each of the held-out fine-tuning tasks in a single-task setting (Section \ref{sec:exp-adaptation}). In this setting, we present results for fine-tuning on all CW10 and DMC6 tasks individually.

\begin{figure}[h]
     \centering
    \subfigure{\includegraphics[width=0.65\linewidth]{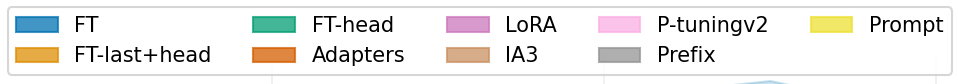}}\\[-1.5ex]
    \subfigure[CW10]{\includegraphics[width=0.75\linewidth]{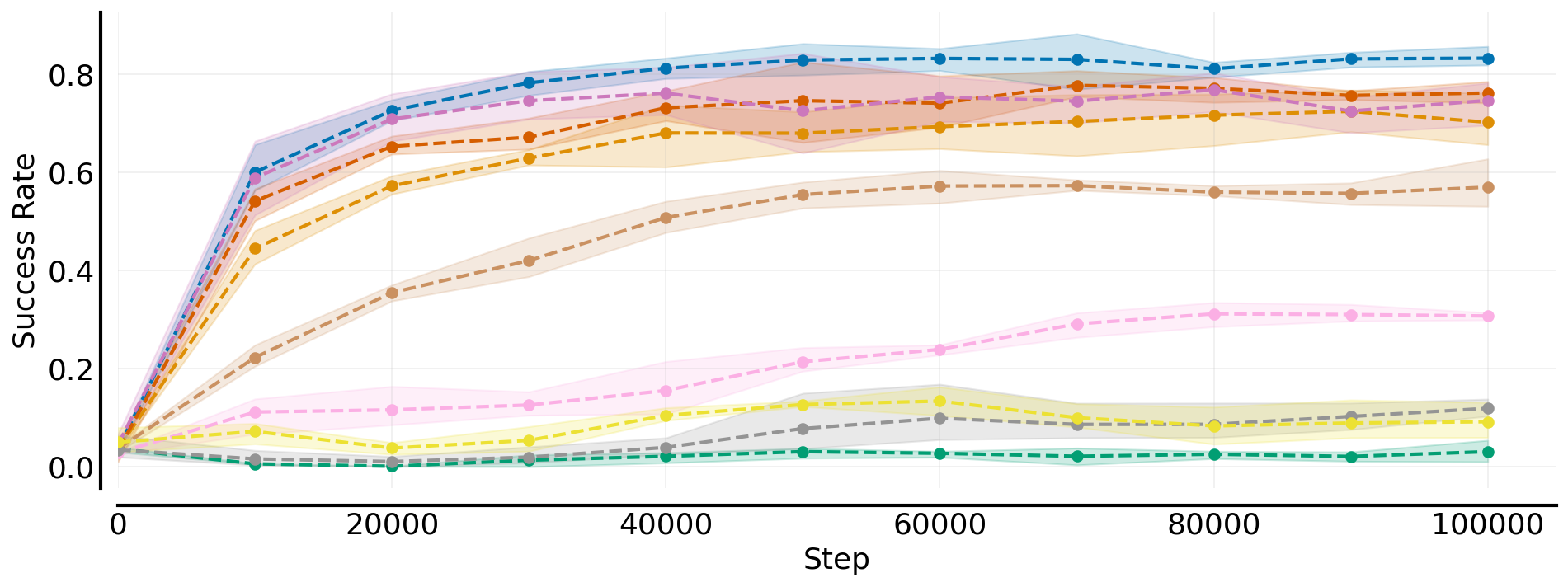}}
    \subfigure[DMC6]{\includegraphics[width=0.75\linewidth]{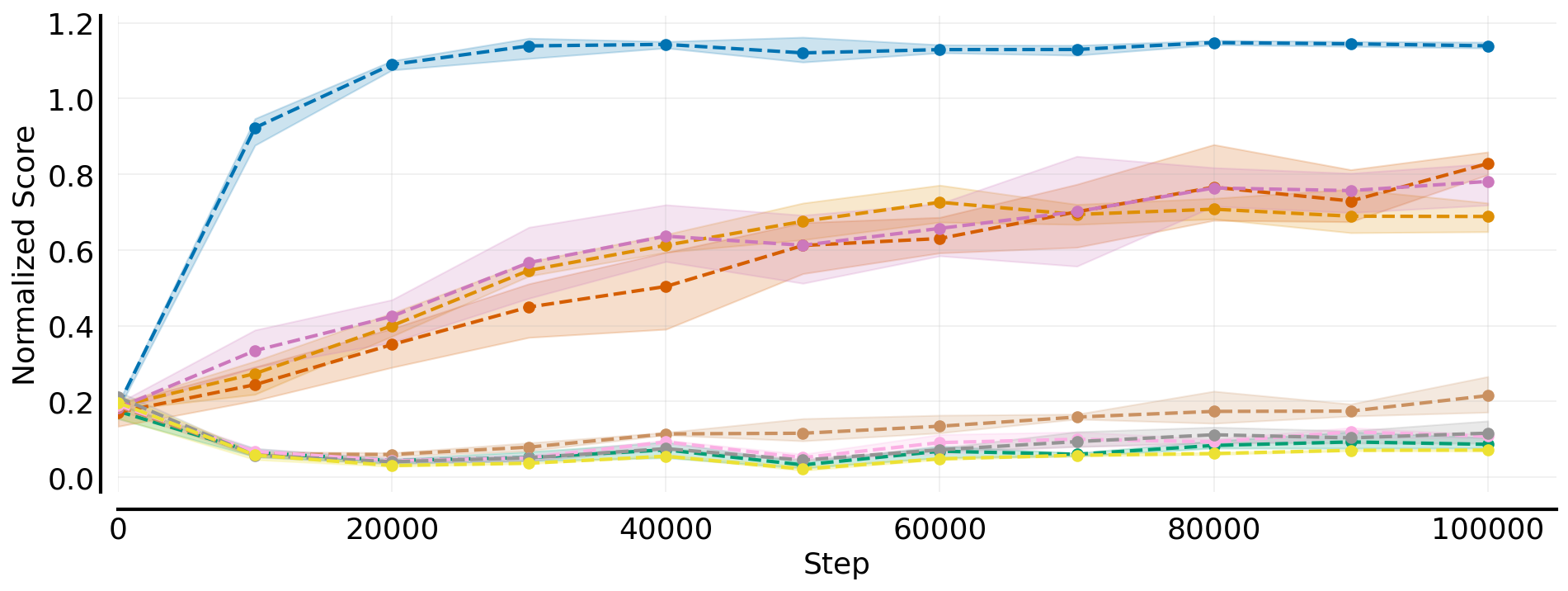}}
  \caption{Learning curves for single-task fine-tuning experiments on \textbf{(a)} CW10 and \textbf{(b)} DMC6. On every task, we train for 100K timesteps and consequently average the performances across all tasks in the data split. } 
    \label{fig:appendix-adapt-full-lc}
\end{figure}

\begin{figure}[h]
     \centering
     \includegraphics[width=0.47\linewidth]{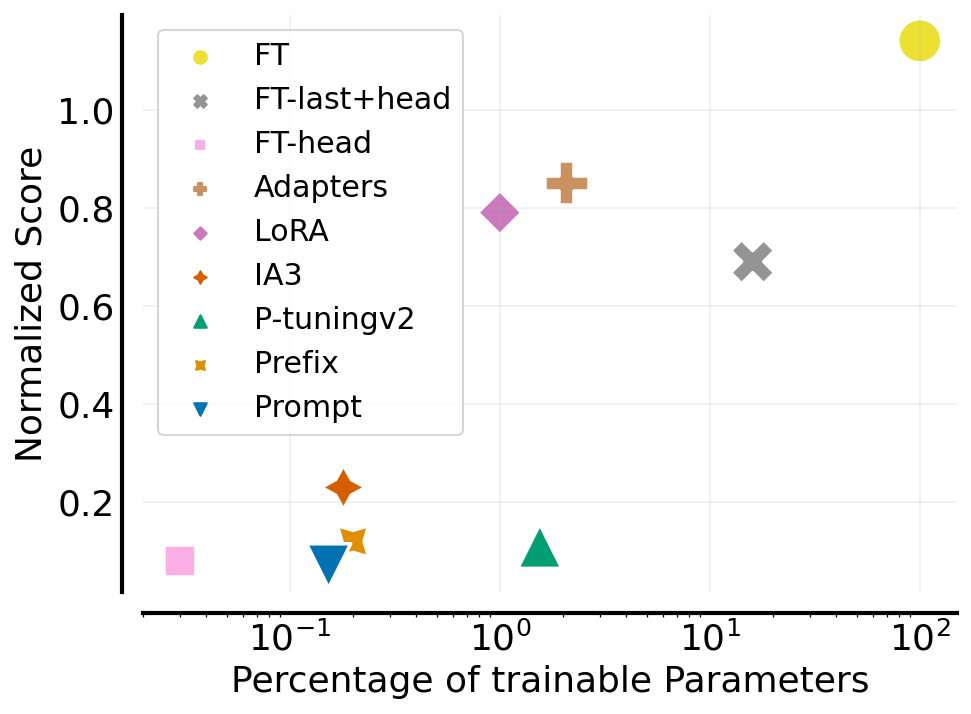}
    \caption{Normalized score vs. fraction of parameters trained for various fine-tuning techniques on single-task experiments for DMC6.}
    \label{fig:appendix-dcm6-perf-param}
\end{figure}

We compare 9 different methods on CW10 and DMC6, as listed in Section \ref{sec:exp-adaptation}. We train for 100K update steps on each task. Every 10K update steps, we evaluate the current model within the actual environment for 10 evaluation episodes and average performance across the evaluation episodes. The final performance scores are aggregated over all tasks in the task sequence. We show the aggregated task performances on CW10 and DMC6 in Table \ref{tab:appendix-adapt-full}. In addition, we provide the respective learning curves in Figure \ref{fig:appendix-adapt-full-lc}. %
In Figure \ref{fig:appendix-dcm6-perf-param} we compare the fraction of trainable parameters vs. attained performance for all considered fine-tuning techniques on DMC6.

For FT variations, Adapters, LoRA and (IA$\text{)}^3$, we use a learning rate of $1e^{-4}$. For all PBT methods, we increase the learning rate to $1e^{-3}$ which resulted in  better performance. For Adapters, we set a reduction factor of $16$ for the down-projection. For LoRA, we set the rank to 8 by default, but conduct an ablation on this choice in Section \ref{appendix:lora-rank}. For all prompt-tuning approaches, we use a prompt length of $25$ and a dropout rate of $0.2$. We found these values to work well in preliminary experiments. For LoRA, Adapters and (IA$\text{)}^3$ we do not use Dropout \citep{Srivastava:14}. 

Moreover, we add the performance scores for two multi-task baselines in Table \ref{tab:appendix-adapt-full}. While the first one is trained from scratch (FT-MT-scratch), the second is initialised from the pre-trained model (FT-MT-pre-trained). Both model variants are trained for 1M update steps on the multi-domain mix of CW10 and DMC6. The rationale for this comparison, is to investigate the effect of pre-training when learning new tasks. Indeed, we observe significantly higher fine-tuning scores on both domains for the pre-trained model.  

\begin{table}[]
    \centering
    \caption{Aggregate scores for all single-task fine-tuning experiments on CW10 and DMC6 datasets.}
    \begin{tabular}{l c c | c c}
    \toprule
    \multirow{2}{*}{\textbf{Method}} & \multicolumn{2}{c}{CW10} & \multicolumn{2}{c}{DMC6} \\
    &  \textbf{Success Rate} &  \textbf{Mean Reward} & \textbf{Norm. Score} & \textbf{Mean Reward} \\
    \midrule
    FT &  0.83 ± 0.21 & 1449.15 ± 329.56 &  1.14 ± 0.09 & 832.46 ± 198.22 \\
    FT-head+last &   0.7 ± 0.24 &  1232.9 ± 422.91 & 0.69 ± 0.37 &  547.76 ± 324.0 \\
    FT-head &  0.03 ± 0.11 &  127.23 ± 205.68 & 0.08 ± 0.09 &   81.23 ± 76.37 \\
    Adapters &  0.76 ± 0.23 & 1330.29 ± 346.19 & 0.85 ± 0.23 & 662.63 ± 255.08 \\
    LoRA &  0.75 ± 0.24 & 1319.89 ± 365.68 & 0.79 ± 0.26 &  616.8 ± 276.89 \\
    (IA$\text{)}^3$ &  0.64 ± 0.35 &  1131.7 ± 436.49 & 0.23 ± 0.26 & 195.95 ± 195.35 \\
    P-tuning v2 &  0.31 ± 0.39 &  556.97 ± 571.94 &  0.11 ± 0.11 &  102.33 ± 94.55 \\
    Prefix-tuning &  0.12 ± 0.25 &   282.6 ± 468.18 &  0.12 ± 0.14 & 112.24 ± 117.92 \\
    Prompt-tuning &  0.09 ± 0.23 &  178.25 ± 373.56 &  0.07 ± 0.08 &   76.27 ± 74.87 \\
    \midrule
    FT-MT-scratch  &  0.85 ± 0.07 & 1476.61 ± 395.13 &  0.92 ± 0.03 & 648.49 ± 354.7 \\
    FT-MT-pre-trained &  0.92 ± 0.02 & 1491.76 ± 332.0 &  1.04 ± 0.17 & 764.38 ± 230.28 \\
    \bottomrule
    \end{tabular}
    \label{tab:appendix-adapt-full}
\end{table}

\subsection{What rank is required in LoRA?} \label{appendix:lora-rank}
\begin{figure}[h]
     \centering
    \subfigure[CW10]{\includegraphics[width=0.45\linewidth]{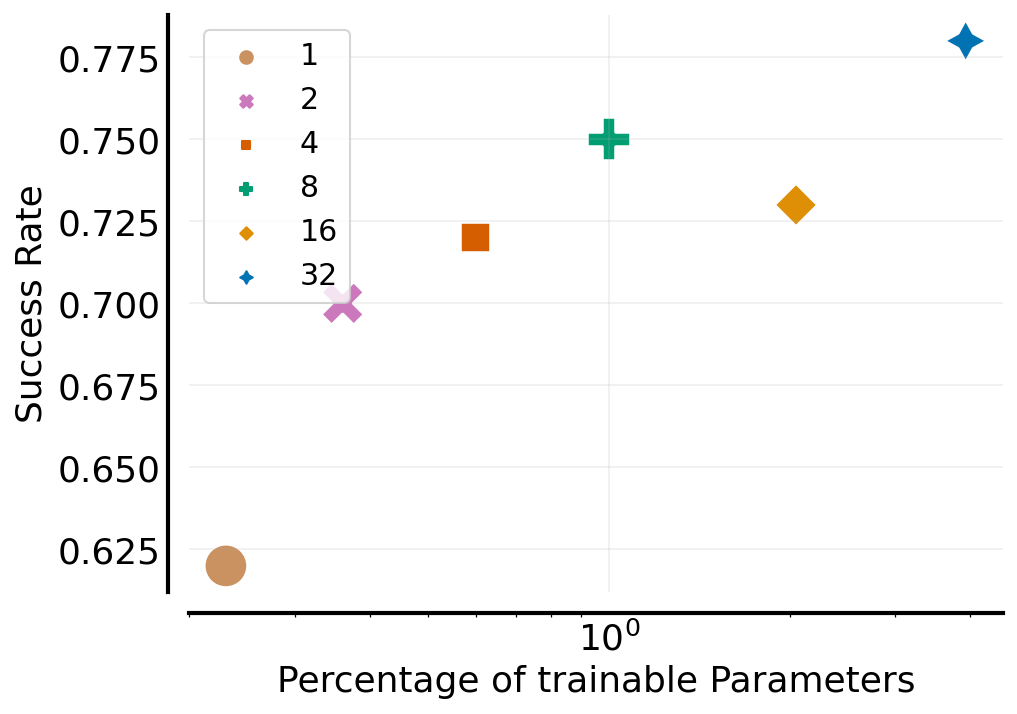}}
    \hspace{1cm}
    \subfigure[DMC6]{\includegraphics[width=0.45\linewidth]{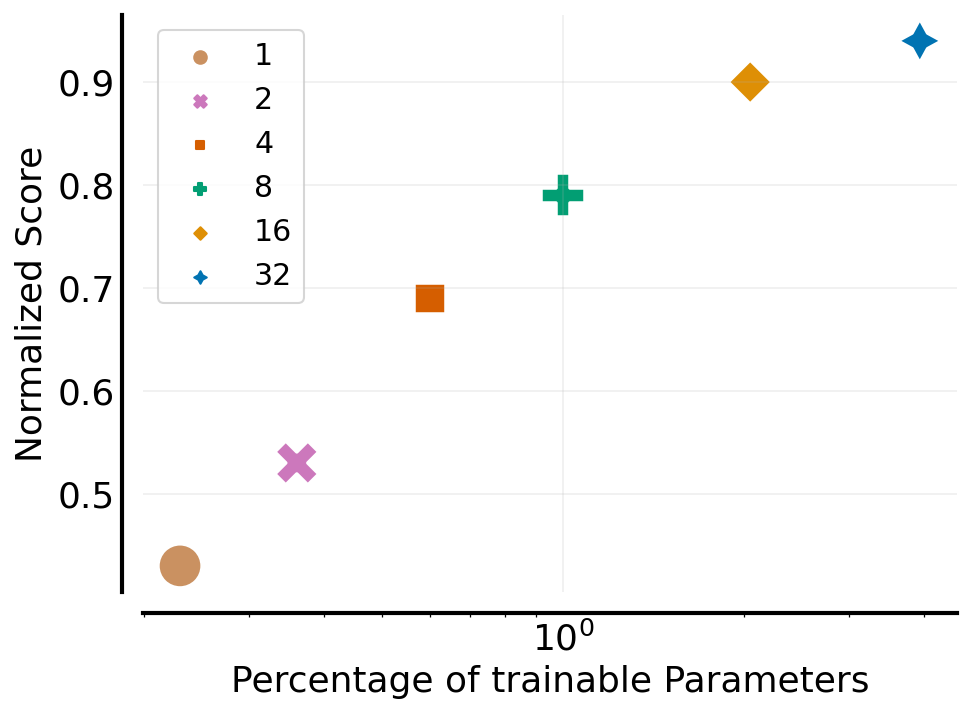}}
  \caption{Performance vs. fraction of parameters trained for different values for the rank in LoRA on \textbf{(a)} CW10 and \textbf{(b)} DMC6.} 
    \label{fig:appendix-lora-rank}
\end{figure}

To investigate the impact of the rank $r$ in LoRA, we performed a hyper-parameter search varying $r$ across the values $\{ 1,2,4,8,16,32\}$. We present the results of our experiments on CW10 and DMC6 in Figure \ref{fig:appendix-lora-rank}. Specifically, we plot the performance against the percentage of parameters trained. Note that the x-axis is on a log-scale. For all experiments presented in Sections \ref{sec:exp-adaptation} and \ref{sec:exp-cl}, we used a rank of $8$ for LoRA. As expected, increasing the rank can improve performance on both considered domains. However, this improvement comes at the expense of more trainable parameters.

\subsection{What matrices should be modulated using LoRA?} \label{appendix:lora-mod-targets}
Another important design choice when using LoRA is which of the weights of the pre-trained model to modulate.
To investigate this, we conduct an ablation study in which we vary the modulation targets. 
In principle, LoRA can be applied to any weight matrix in the pre-trained model. 
In our ablation study, we specifically focus on modulating the keys, values, and queries in the self-attention mechanism, as well as the position-wise feedforward layer.

\begin{figure}
    \centering
    \includegraphics[width=0.5\textwidth]{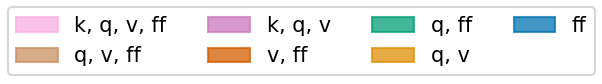}\\
    \includegraphics[width=0.45\textwidth]{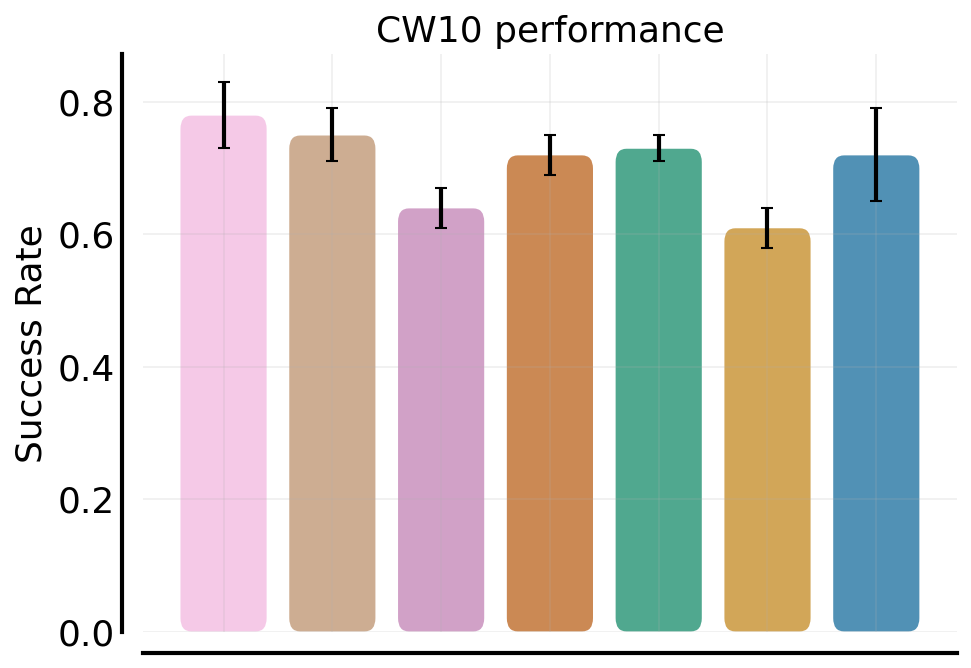}
    \caption{Modulation target ablation for L2M on CW10.}
    \label{fig:appendix-excludemodvecs}
\end{figure}

We present the results of our experiments in Figure \ref{fig:appendix-excludemodvecs}. We find that modulating the position-wise feedforward layer tends to be more important the modulating the self-attention mechanism. Overall, the best performance is obtained when modulating all considered targets. However, this comes at the cost of more parameters. 
Notably, these results narrow the performance gap to L2M-oracle.

\section{Continual Fine-Tuning}\label{appendix:cl}
Ultimately, our goal is to adapt the pre-trained model to multiple novel tasks, while alleviating forgetting of tasks learned during pre-training.
In this regard, we adapt the pre-trained model to all fine-tuning tasks in a sequential manner, as proposed by \citet{Wolczyk:21} for CW10.
Moreover, we evaluate forgetting by measuring the performance on tasks acquired during pre-training after fine-tuning. 
We compare 10 different methods in this setting, as listed in Section \ref{sec:exp-cl}.

We train each method for 100K steps per task in CW10, with a batch size of 256. After every 100K update steps, we switch to the next task in the sequence. After every 50K update steps, we evaluate the current model on all tasks in the task-sequence. For L2M, we use a pool size of 30. For all L2P-based approaches, we use a prompt size of 25 and a prompt pool of 100 by default. Following \citet{Wang:22}, we use the selection count  $n(\mathbf{k}_i)$ up to the previous task and $\lambda=0.5$ for regularization of L2M and all L2P-based approaches. 
For EWC, we tune the value of the penalty (see Section \ref{sec:appendix-ewc}). For L2M, we use learning rate of $5e^{-5}$ on CW10 and $1e^{-4}$ on DMC6. For all L2P variants, we use a learning rate of $1e^{-3}$.

We show the performance and forgetting scores for all considered methods in Table \ref{tab:crl_full}. All metrics are reported at the end of training. The forgetting scores are computed as defined by \citet{Wolczyk:21}.
In Table \ref{tab:crl_full}, we also include the multi-task performance scores on all CW10 tasks (same as in Appendix \ref{appendix:adaptation}). Multi-task FT represents the upper bound in terms of performance. Overall, L2M achieves the highest success rates/normalized scores and outperforms all other methods in both domains.

\begin{table}[]
    \centering
    \caption{Continual RL experiments on \textbf{(a)} CW10 and \textbf{(b)} DMC6.}
    \subtable[CW10]{
        \begin{tabular}{l c c c}
        \toprule
        \textbf{Method} & \textbf{Success Rate} &     \textbf{Mean Reward} & \textbf{Forgetting} \\
        \midrule
            FT                  &  0.09 ± 0.01 &  234.57 ± 414.86 &         0.75 ± 0.32 \\
            FT-last+head        &  0.12 ± 0.03 &  266.59 ± 406.42 &         0.65 ± 0.31 \\
            FT-head             &  0.07 ± 0.03 &  95.34 ± 127.2 &         0.08 ± 0.27 \\
            L2M-oracle          &  0.77 ± 0.03 &   1294.72 ± 385.32 &         0.05 ± 0.16 \\
            L2M                 &  0.65 ± 0.04 & 1136.17 ± 53.82 & 0.07 ± 0.08 \\
            L2P-Pv2             &    0.4 ± 0.02 &  707.84 ± 35.91 & 0.03 ± 0.06 \\
            L2P-PreT            &  0.34 ± 0.05 &  606.33 ± 61.97 & 0.01 ± 0.06 \\
            L2P-PT              &   0.23 ± 0.05 &  390.28 ± 64.87 & 0.11 ± 0.01 \\
            EWC                 &   0.17 ± 0.01 &  257.48 ± 18.18 &         0.78 ± 0.0 \\
            L2                  &    0.1 ± 0.0 &  237.07 ± 522.98 &           0.0 ± 0.0 \\
            \midrule
            FT-MT-scratch  &  0.85 ± 0.07 & 1476.61 ± 395.13  & -\\
            FT-MT-pre-trained &  0.92 ± 0.02 & 1491.76 ± 332.0 & - \\
        \bottomrule
        \end{tabular}
    }
    \subtable[DMC6]{
        \begin{tabular}{l c c c}
        \toprule
        \textbf{Method} & \textbf{Normalized Score} &     \textbf{Mean Reward} & \textbf{Forgetting} \\
        \midrule
        FT               &  0.27 ± 0.01 &   225.81 ± 10.2 &  0.83 ± 0.02 \\
        FT-last+head     &  0.25 ± 0.02 &  204.49 ± 22.07 &  0.51 ± 0.06 \\
        FT-head          &   0.06 ± 0.0 &    58.84 ± 0.64 &  -0.02 ± 0.0 \\
        L2M-oracle       &   0.7 ± 0.11 & 549.29 ± 254.69 &        -0.02 ± 0.27 \\
        L2M              &  0.56 ± 0.18 & 401.65 ± 122.75 &  0.11 ± 0.14 \\
        L2P-Pv2          &  0.32 ± 0.09 &  271.39 ± 72.49 &  0.02 ± 0.08 \\
        L2P-PreT         &  0.2 ± 0.04 &  182.33 ± 23.28 & -0.03 ± 0.01 \\
        L2P-PT           &  0.18 ± 0.03 &  158.09 ± 22.88 & -0.06 ± 0.02 \\
        EWC              &    0.3 ± 0.06 &   244.7 ± 54.12 &  0.49 ± 0.14 \\
        L2               &  0.2 ± 0.03 &   192.8 ± 23.86 & -0.01 ± 0.01 \\
        \midrule
        FT-MT-scratch   &  0.92 ± 0.03 & 648.49 ± 354.7 & - \\
        FT-MT-pre-trained &  1.04 ± 0.17 &  764.38 ± 230.28 & - \\
        \bottomrule
        \end{tabular}
    }
    \label{tab:crl_full}
\end{table}

\subsection{Query Representation Ablations}\label{appendix:queryrep}
For all L2M results, we use embedded state-tokens aggregated over the context as query to the modulation pool. It is possible to represent the query differently by using other tokens (see Figure \ref{fig:query-rep-token}), intermediate layers, or history lengths.  

\begin{wrapfigure}{r}{0.5\textwidth}
    \vspace{-0.28cm}
    \centering
    \includegraphics[width=0.45\textwidth]{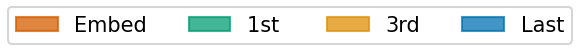}\\
    \includegraphics[width=0.45\textwidth]{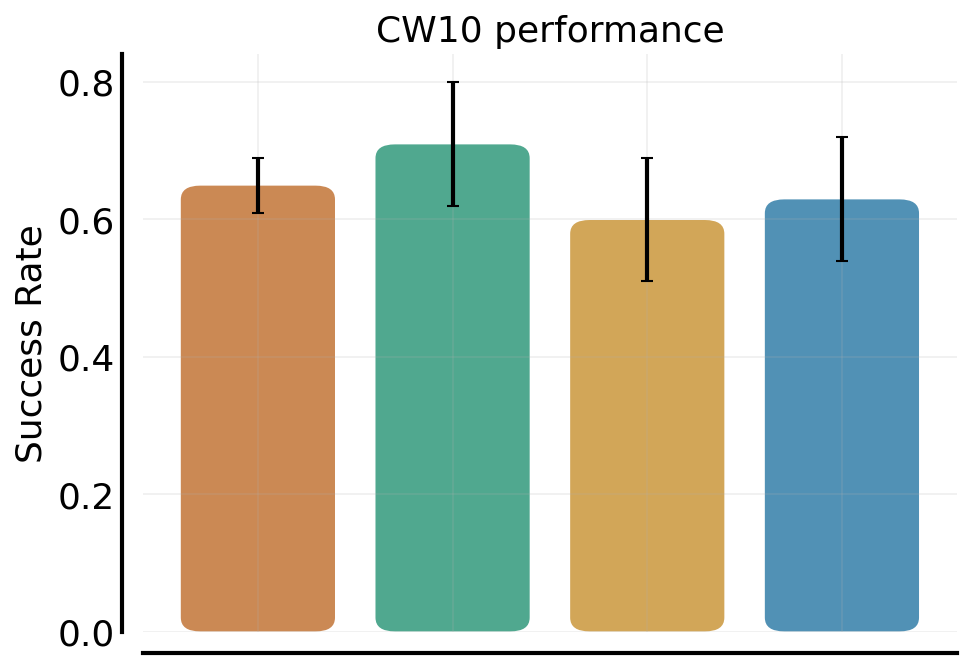}
    \caption{Layer representation ablation for L2M on CW10.}
    \label{fig:appendix-layers}
\end{wrapfigure}
\textbf{Layer Representation.}
Instead of using the representation of the embedding layer for the query, we can alternatively use the representations produced by intermediate Transformer layers. To this end, we conduct an ablation study in which we compare the representations extracted from the embedding layer against the representations from the first, middle (3rd block) and last Transformer block. We show the results of this experiment in Figure \ref{fig:appendix-layers}. Using the representations extracted from the middle or last Transformer blocks yields worse performance compared to the embedding layer. In contrast, the representation from the first Transformer block attains the highest scores overall, and narrows the gap to L2M-oracle. However, it is important to note that this performance gain comes at the cost of additional complexity since we need to propagate farther through the model than merely through the embedding layer. 

\begin{figure}
    \centering
    \includegraphics[width=0.45\textwidth]{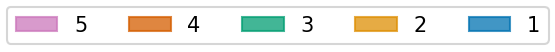}\\
    \includegraphics[width=0.45\textwidth]{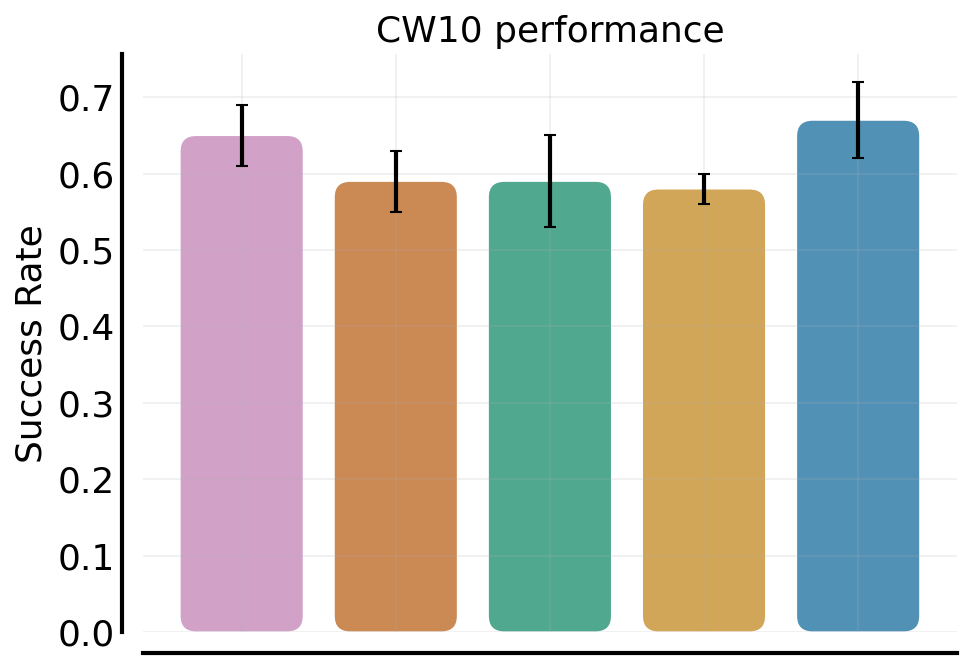}
    \caption{History length ablation for L2M on CW10.}
    \label{fig:appendix-histlen}
\end{figure}

\textbf{History Length.} To investigate the dependence of the query representation on the history length while keeping the context window as is, we conduct an ablation study. In Figure \ref{fig:appendix-histlen}, we report the results of this experiment. By default, we used the five last timesteps in the sequence to construct the query. However, we find that using only the most recent timestep performs similarly.

\begin{wrapfigure}{r}{0.5\textwidth}
    \vspace{-1.5cm}
    \centering
    \includegraphics[width=0.42\textwidth]{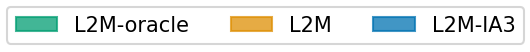}\\
    \includegraphics[width=0.45\textwidth]{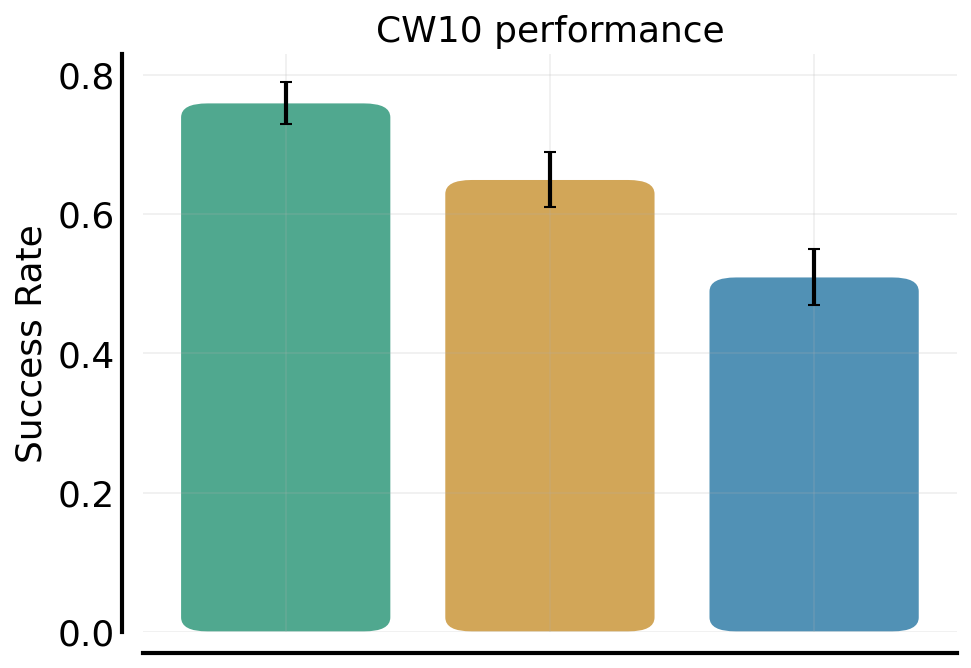}
    \caption{Alternative modulators ablation on CW10. We compare L2M to L2M-(IA$\text{)}^3$.}
    \label{fig:appendix-l2m-ia3}
\end{wrapfigure}
\subsection{Alternative Modulators}\label{appendix:altmod}
In L2M, we use LoRA by default. However, L2M can also be combined using other PEFT techniques%
Therefore, we conduct an ablation study in which we compare L2M against L2M in combination with (IA$\text{)}^3$ on CW10. Instead of the low-rank modulation matrices, (IA$\text{)}^3$ employs element-wise multiplication with learnable modulation vectors. We present the results in Figure \ref{fig:appendix-l2m-ia3}. 
While performance decreases with L2M-(IA$\text{)}^3$, it compares favourably in terms of parameter-efficiency. Depending on the task at hand, this may be preferred. 

\subsection{Single-domain DT on Meta-World}\label{sec:appendix-singledomain} 
\begin{figure}
     \centering
     \subfigure[Single-task FT on CW10]{\includegraphics[width=0.44\textwidth]{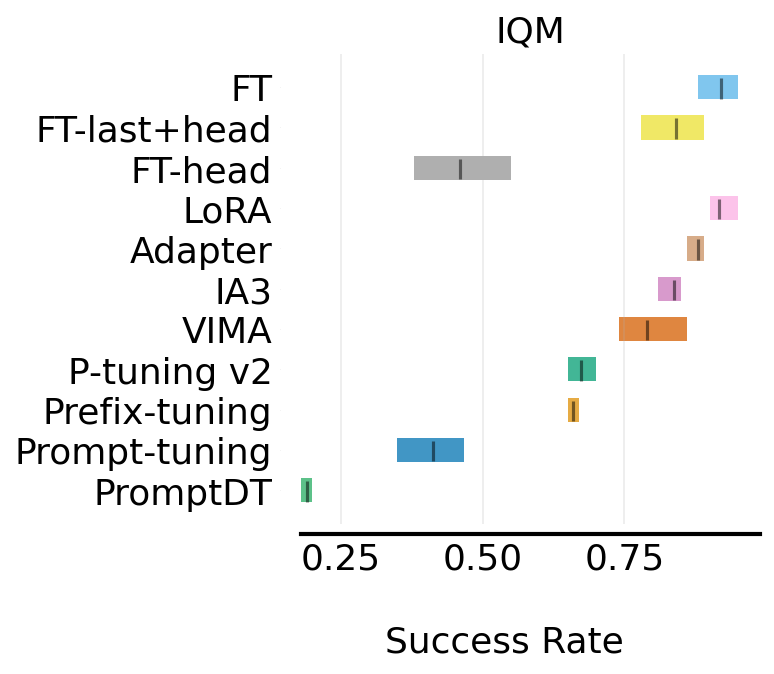}}
    \hspace{0.3cm}
     ~
     \subfigure[Continual FT on CW10]{\includegraphics[width=0.44\textwidth]{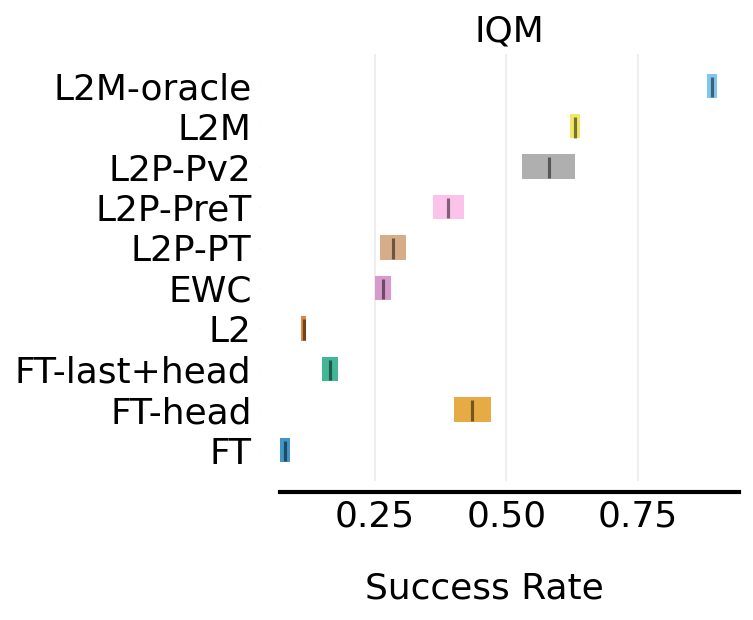}}
    \caption{IQM and 95\% CIs on CW10 for \textbf{(a)} single-task FT and \textbf{(b)} continual FT using a single-domain non-discretized DT pre-trained on MT40 with 13M parameters.}
    \label{fig:singledomain-metaworld}
\end{figure}

We pre-train a DT with 13M parameters on MT40 and compare to the same methods as in Sections \ref{sec:exp-adaptation} and \ref{sec:exp-cl}. In this setting, we additionally add two competitors, namely PromptDT \citep{Xu:22} and VIMA \citep{Jiang:22}. . We experimented with larger models, but found the selected model size to perform well. Our MDDT used a unified state space, a discretised action space and autoregressive action prediction (at inference time) to handle varying state and action spaces. In contrast, the single-domain DT does not require these mechanisms. Instead, the single-domain DT is trained to predict the continuous actions via the MSE loss. 
The results for the single-domain experiment are available in Figure \ref{fig:singledomain-metaworld}. 
The pre-trained model attains a success rate of 81\% on MT40. 
Overall, we find that the specialised single-domain model obtains considerably higher performance scores than the MDDT (see Sections \ref{sec:exp-adaptation} and \ref{sec:exp-cl}). 
However, these performance gains come at the cost of the loss of generality, as the specialised model can only handle the particular state/action space it was trained on.

\subsection{Cross-domain FT}\label{sec:appendix-crossdomain} 
We conduct another experiment, in which we pre-train a DT on Meta-World only (MT40) and then fine-tune it on DMControl (DMC6).
The pre-trained model relies on the unified state space and action discretisation. It has the same amount of parameters as the MDDT (40M) and attains an average success rate of 76\% on MT40 after pre-training.
Prior to fine-tuning, the model has never seen trajectories from DMControl. 

\begin{figure}[h]
     \centering
     \subfigure[Single-task FT on CW10]{\includegraphics[width=0.44\textwidth]{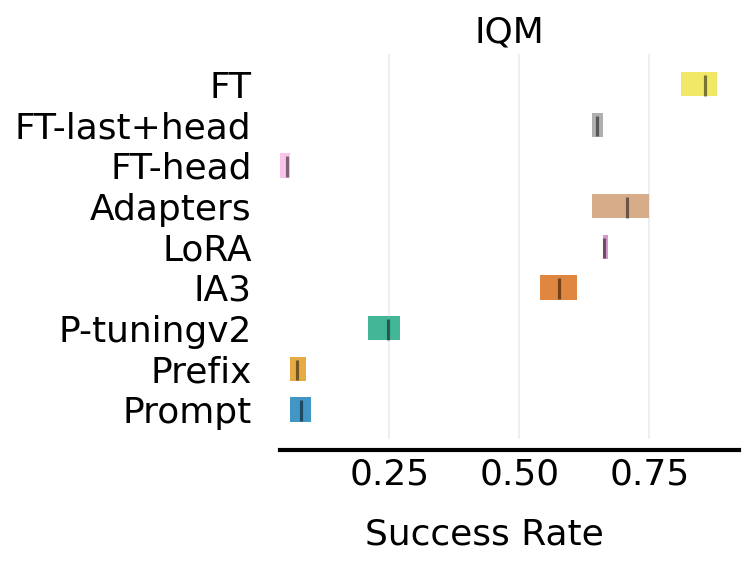}}
    \hspace{0.3cm}
     ~
     \subfigure[Single-task FT on DMC6]{\includegraphics[width=0.44\textwidth]{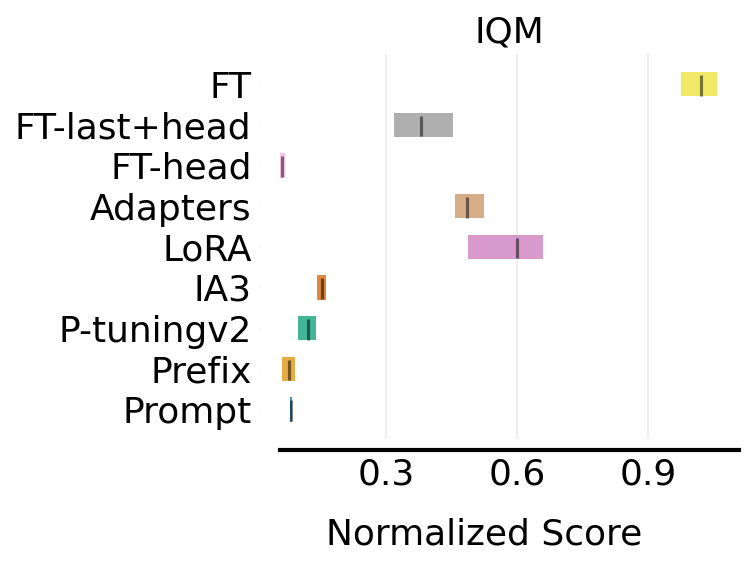}}
    \caption{IQM and 95\% CIs for single-task FT on \textbf{(a)} CW10 and \textbf{(b)} DMC6 using a single-domain discretized DT pre-trained on MT40 with 40M parameters.}
    \label{fig:crossdomainft}
\end{figure}

We show the results for different fine-tuning techniques in Figure \ref{fig:crossdomainft} 
We observe that the fine-tuning performance on DMC6 is considerably worse than for the MDDT (79\%). 
In addition, we also fine-tune the pre-trained single-domain model on the same domain (CW10). 
Interestingly, the final performance on CW10 is also lower compared to the MDDT model (75\%) that was pre-trained on both domains.
This experiment indicates, that multi-domain pre-training can, indeed, have a positive effect on the fine-tuning performance.

\subsection{EWC Hyper-Parameter Search}\label{sec:appendix-ewc}
We observed that EWC \citep{Kirkpatrick:17} performs worse than L2M in our experiments. 
By default, we used a regularization coefficient of $\lambda=10000$ for EWC, which performed best in preliminary experiments. 
Additionally, we compare different values for $\lambda=\{10, 400, 1e^3, 1e^4, 1e^5\}$ as used in EWC. %
We find that the optimal choice for $\lambda$ varies heavily between the two domains, but higher performance scores can be achieved by tuning $\lambda$.

\section{Hardware \& Training Times}
\textbf{Pre-training}. We run all our pre-training experiments on 4 NVIDIA A100 GPUs. Training times depend on the model size. 
Training the smallest model (40M parameters) takes roughly 1.5 days, while training the largest model (200M) takes roughly 3.5 days. To parallelize the computation across multiple GPUs/nodes, we leverage the distributed data parallel (DDP) feature provided by PyTorch. Throughout all our experiments, we use mixed precision training \citep{micikevicius2017mixed} as supported in PyTorch to speed up training time. 

\textbf{Fine-tuning}. For all our fine-tuning experiments, we use single GPU training on NVIDIA A100 or NVIDIA Titan V GPUs. Training times vary between the considered fine-tuning settings (single-tasks experiments, continual-learning experiments), domains (Meta-World, DMControl), and methods. For example, running a single seed for L2M on CW10 in the continual FT setting takes 1.25 days on one NVIDIA A100. Similarly, running a single seed for L2M on DMC6 takes roughly one day on one NVIDIA Titan V. 

\section{Potential Societal Impact}
Our work is purely academic. 
However, the presented ideas aim for more capable embodied agents. 
As such, there is potential for misuse, e.g., when used in real-world scenarios without proper verification or safeguards. 
While we do not expect to see the deployment of such agents for potentially malicious purposes in the foreseeable future, it is essential to ensure responsible use of this technology.
\end{document}